\documentclass[12pt, b5paper, oneside, english, final]{book}
\usepackage{./includes/sfahmi_thesis}
\usepackage[acronym, nonumberlist, nogroupskip, nopostdot, style=super]{glossaries} 
\newacronym{ati}{AtI}{Athletic Intelligence}
\newacronym{ci}{CI}{Cognitive Intelligence}

\newacronym{stance}{STANCE}{\textbf{S}oft \textbf{T}errain \textbf{A}daptation a\textbf{N}d \textbf{C}ompliance \textbf{E}stimation}

\newacronym{wbc}{WBC}{Whole-Body Control}
\newacronym{awbc}{c$^3$WBC}{Compliant Contact Consistent Whole-Body Control}
\newacronym{swbc}{sWBC}{Standard Whole-Body Control}
\newacronym{pwbc}{pWBC}{Passive Whole-Body Control}

\newacronym{c3}{\texttt{c}$^3$}{compliant contact consistent}
\newacronym{wbopt}{WBOpt}{Whole-Body Optimization}

\newacronym{ste}{TCE}{Terrain Compliance Estimator}
\newacronym{hc}{HC}{Hunt~and~Crossley's}
\newacronym{kv}{KV}{Kelvin-Voigt's}

\newacronym{mpc}{MPC}{Model Predictive Control}
\newacronym{qp}{QP}{Quadratic Program}

\newacronym{haa}{HAA}{Hip Adduction-Abduction}
\newacronym{hfe}{HFE}{Hip Flexion-Extension}
\newacronym{kfe}{KFE}{Knee Flexion-Extension}

\newacronym{lf}{LF}{Left-Front}
\newacronym{rf}{RF}{Right-Front}
\newacronym{lh}{LH}{Left-Hind}
\newacronym{rh}{RH}{Right-Hind}

\newacronym{hyq}{HyQ}{Hydraulically actuated Quadruped}

\newacronym{grfs}{GRFs}{Ground Reaction Forces}
\newcommand{\grfs}{\gls{grfs}~}

\newacronym{com}{CoM}{Center of Mass}
\newacronym{ode}{ODE}{Open Dynamics Engine}
\newacronym{zmp}{ZMP}{Zero Moment Point}
\newacronym{mae}{MAE}{Mean Absolute Tracking Error}
\newacronym{dofs}{DoFs}{Degrees of Freedom}

\newacronym{imu}{IMU}{Inertial Measurement Unit}

\newacronym{ekf}{EKF}{Extended Kalman Filter}
\newacronym{ukf}{UKF}{Unscented Kalman Filter}
\newacronym{xkf}{XKF}{eXogeneous Kalman Filter}
\newacronym{ltv}{LTV}{Linear Time-Varying}
\newacronym{ges}{GES}{Globally Exponentially Stable}
\newacronym{nlo}{NLO}{Non-Linear Observer}
\newacronym{pe}{PE}{Persistency of Excitation}
\newacronym{mcs}{MCS}{Motion Capture System}

\newacronym{ptal}{PTAL}{Proprioceptive Terrain-Aware Locomotion}
\newacronym{etal}{ETAL}{Exteroceptive Terrain-Aware Locomotion}

\usepackage[]{glossaries} 
\newacronym{tal}{TAL}{Terrain-Aware Locomotion}
\newacronym{to}{TO}{Trajectory Optimization}
\newacronym{drl}{RL}{Reinforcement Learning}
\newacronym{vital}{ViTAL}{Vision-Based Terrain-Aware Locomotion}
\newacronym{vfa}{VFA}{Vision-Based Foothold Adaptation}
\newacronym{vpa}{VPA}{Vision-Based Pose Adaptation}
\newacronym{fec}{FEC}{Foothold Evaluation Criteria}
\newacronym{cnn}{CNN}{Convolutional Neural Network} 
\newacronym{tr}{TR}{Terrain Roughness}
\newacronym{lc}{LC}{Leg Collision}
\newacronym{kfis}{KF}{Kinematic Feasibility}
\newacronym{fc}{FC}{Foot Trajectory Collision}
\newacronym{nsf}{$\mathrm{n_{sf}}$}{Number of Safe Footholds}
\newacronym{feasibles}{$\mathcal{F}$}{Set of Safe Footholds}
\newacronym{rcf}{RCF}{Reactive Controller Framework}
\newacronym{tbr}{TBR}{Terrain-Based Body Reference}
\newacronym{hyqreal}{HyQReal}{}

\makeglossaries
\newcommand{\pwbcPaper}{
\begin{tcolorbox}
\sffamily
\copyright~2019~IEEE. Reprinted, with permission.
S. Fahmi, C. Mastalli, M. Focchi and C. Semini, 
"Passive Whole-Body Control for Quadruped Robots: Experimental Validation Over Challenging Terrain,"
in IEEE Robotics and Automation Letters~(RA-L), vol.~4, no.~3, pp.~2553-2560, July~2019, 
\doi{10.1109/LRA.2019.2908502}.
\end{tcolorbox}
}

\newcommand{\stancePaper}{
\begin{tcolorbox}
\sffamily
\copyright~2020~IEEE. Reprinted, with permission.
S.~Fahmi, M.~Focchi, A.~Radulescu, G.~Fink, V.~Barasuol and C.~Semini, 
"STANCE: Locomotion Adaptation Over Soft Terrain," 
in IEEE Transactions on Robotics~(T-RO), 
vol.~36, no.~2, pp.~443-457, April~2020, 
\doi{10.1109/TRO.2019.2954670}.
\end{tcolorbox}
}

\newcommand{\lsensPaper}{
\begin{tcolorbox}
\sffamily
\copyright~2021~IEEE. Reprinted, with permission.
S.~Fahmi, G.~Fink and C.~Semini, 
"On State Estimation for Legged Locomotion over Soft Terrain," 
in IEEE Sensors Letters~(L-SENS), 
vol.~5, no.~1, pp.~1--4, January~2021, 
\doi{10.1109/LSENS.2021.3049954}.
\end{tcolorbox}
}

\newcommand{\vitalPaper}{
\begin{tcolorbox}
\sffamily
\copyright~2022~IEEE. Reprinted, with permission.
S.~Fahmi, V.~Barasuol, D.~Esteban, O.~Villarreal, and C.~Semini, 
"ViTAL: Vision-Based Terrain-Aware Locomotion for Legged Robots," 
in IEEE Transactions on Robotics~(T-RO), 
vol.~0, no.~0, pp.~1--20, November~2022, 
\doi{10.1109/TRO.2022.3222958}.
\end{tcolorbox}
}

\newcommand{\nmrm}[1]{{#1}}
\newcommand{\vc}[1]{#1} 
\newcommand{\mat}[1]{\ensuremath{\begin{bmatrix}#1\end{bmatrix}}}	

\newcommand{\mrm}[1]{\mathrm{#1}}
\newcommand{\Rnum}{\mathbb{R}} 
\newcommand{\grf}{F_{\mathrm{grf}}} 
\newcommand{\grfp}[1]{F_{\mathrm{grf,#1}}} 

\DeclareMathOperator{\Proj}{Proj}
\DeclareMathOperator{\sat}{sat}
\DeclareMathOperator{\vex}{vex}
\DeclareMathOperator{\SO}{SO}
\newcommand{\BF}{\mathcal{B}}
\newcommand{\NF}{\mathcal{N}}
\newcommand{\R}{\mathbb{R}}

\newcommand{\dtau}{\mathop{d\tau}}

\newcommand{\boldSubSec}[1]{\noindent\textbf{#1.}}

\newcommand{\appref}[1]{Section~\ref{#1}}
\newcommand{\criteria}{\gls{fec}\xspace}
\newcommand{\td}{touchdown\xspace}
\newcommand{\lo}{lift-off\xspace}
\newcommand{\nlo}{next lift-off\xspace}
\newcommand{\hipheight}{hip height\xspace}
\newcommand{\fectuple}{T} 
\newcommand{\fecout}{\mu_\mathrm{safe}} 
\newcommand{\hmap}{H} 
\newcommand{\hheight}{z_h} 
\newcommand{\bodyvel}{v_b} 
\newcommand{\gaitparams}{\alpha} 
\newcommand{\nominal}{p_n} 
\newcommand{\candidate}{p_c} 
\newcommand{\optimal}{p_*} 
\newcommand{\hmapvfa}{H_{\mathrm{vfa}}} 
\newcommand{\heurtuple}{\fectuple_{\mathrm{vfa}}} 
\newcommand{\g}{g(\heurtuple)} 
\newcommand{\ghat}{\hat{g}(\heurtuple)} 
\newcommand{\heurtuplevpa}{\fectuple_{\mathrm{vpa}}} 
\newcommand{\heurtuplevpaR}{\fectuple_{\mathrm{vpa},j}} 
\newcommand{\nsf}{\mathrm{n}_\mathrm{sf}}
\newcommand{\setfeasibles}{\mathcal{F}} 
\newcommand{\settuple}{\mathcal{T}} 
\newcommand{\setheights}{\mathcal{Z}} 
\newcommand{\elementfeasibles}{\mathrm{n}_{\mathrm{sf},i}}
\newcommand{\elementtuple}{T_i} 
\newcommand{\elementheights}{z_{h_i}} 
 
\newcommand{\hmapvpa}{H_{\mathrm{vpa}}} 
\newcommand{\hmapvpaR}{H_{\mathrm{vpa},j}} 
\newcommand{\gvpa}{g_{\mathrm{vpa}}(\heurtuplevpa)} 
\newcommand{\ghatvpa}{\hat{g}_{\mathrm{vpa}}(\heurtuplevpa)} 
\newcommand{\finitesetfeasibles}{\bar{\mathcal{F}}} 
\newcommand{\finitesettuple}{\bar{\mathcal{T}}} 
\newcommand{\finitesetheights}{\bar{\mathcal{Z}}} 
\newcommand{\hipheightsamples}{N_{z_h}}
\newcommand{\rbfn}{\hat{\mathcal{F}}} 
 
\graphicspath{
{chapters/figures/chapter1/}, {chapters/figures/chapter2/},
{chapters/figures/chapter3/}, {chapters/figures/chapter3/},
{chapters/figures/chapter4/}, {chapters/figures/chapter5/}}
\usepackage{wrapfig}

\begin{document}
\pagestyle{plain}
\glsresetall 
\thispagestyle{empty}
\begin{center}
\includegraphics[width=\textwidth]{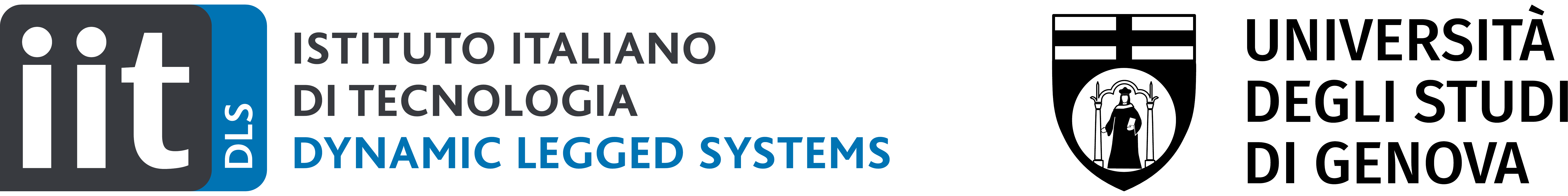}
\vfill
\Huge{\textbf{On Terrain-Aware Locomotion\\for Legged Robots}}
\vfill
\Large{\textbf{Shamel Fahmi}}
\vfill
\large
Istituto Italiano di Tecnologia, Italy
\\
Universit\`{a} degli Studi di Genova, Italy
\vfill
Thesis submitted for the degree of:\\
\textit{Doctor of Philosophy (PhD)}
\vfill
April, 2021
\end{center}
\normalsize
\newpage
\thispagestyle{empty}
\fontsize{10}{12}\selectfont
\noindent \textbf{Shamel Fahmi}

\noindent On Terrain-Aware Locomotion for Legged Robots

\noindent Doctor of Philosophy (PhD) in Bioengineering and Robotics

\noindent Curriculum: Advanced and Humanoid Robotics

\noindent Dynamic Legged Systems (DLS) lab, Italian Institute of Technology (IIT), Italy

\noindent \begin{center}\rule{200pt}{1pt}\end{center}

\noindent Tutors:

\noindent \textbf{Dr. Victor Barasuol}, \textbf{Dr. Michele Focchi}, 
and \textbf{Dr. Andreea Radulescu}

\noindent Dynamic Legged Systems (DLS) lab, Italian Institute of Technology (IIT), Italy

~

\noindent Principle Advisor:

\noindent \textbf{Dr. Claudio Semini}

\noindent Dynamic Legged Systems (DLS) lab, Italian Institute of Technology (IIT), Italy

\noindent \begin{center}\rule{200pt}{1pt}\end{center}

\noindent Annual Evaluation Committee:

\noindent \textbf{Prof. Roy Featherstone}

\noindent Advanced Robotics (ADVR) department, Italian Institute of Technology (IIT), Italy

~

\noindent \textbf{Dr. Enrico Mingo}

\noindent Humanoids \& Human Centered Mechatronics (HHCM) lab, 
Italian Institute of Technology~(IIT), Italy

~

\noindent \textbf{Dr. Geoff Fink}

\noindent Dynamic Legged Systems (DLS) lab, Italian Institute of Technology (IIT), Italy

\noindent \begin{center}\rule{200pt}{1pt}\end{center}

\noindent External Examination Committee:

\noindent \textbf{Prof. Auke Ijspeert}

\noindent Institute of Bioengineering, the Swiss Federal Institute of Technology at Lausanne~(EPFL), Switzerland

~

\noindent \textbf{Prof. Luis Sentis}

\noindent Department of Aerospace Engineering and Engineering Mechanics, University of Texas, USA

~

\noindent \textbf{Prof. Fabio Ruggiero}

\noindent Department of Electrical Engineering and Information Technology,
University of Naples Federico II, Italy 

\noindent \begin{center}\rule{200pt}{1pt}\end{center}

\noindent \copyright~2021 Shamel Fahmi. All rights reserved.
\normalsize
\newpage

\thispagestyle{empty}
~ \vfill
\begin{center}
\textit{To my parents Amal and Prof. Shamel,\\
to my grandmother Hagougi, \\
to my siblings Menna and Mohamed, \\
to our family's latest member Selim, \\
and to my partner Ilef.
}
\end{center}
~\vfill

\newpage
\thispagestyle{empty}
~\vfill
\begin{center}
\includegraphics[width=0.5\textwidth]{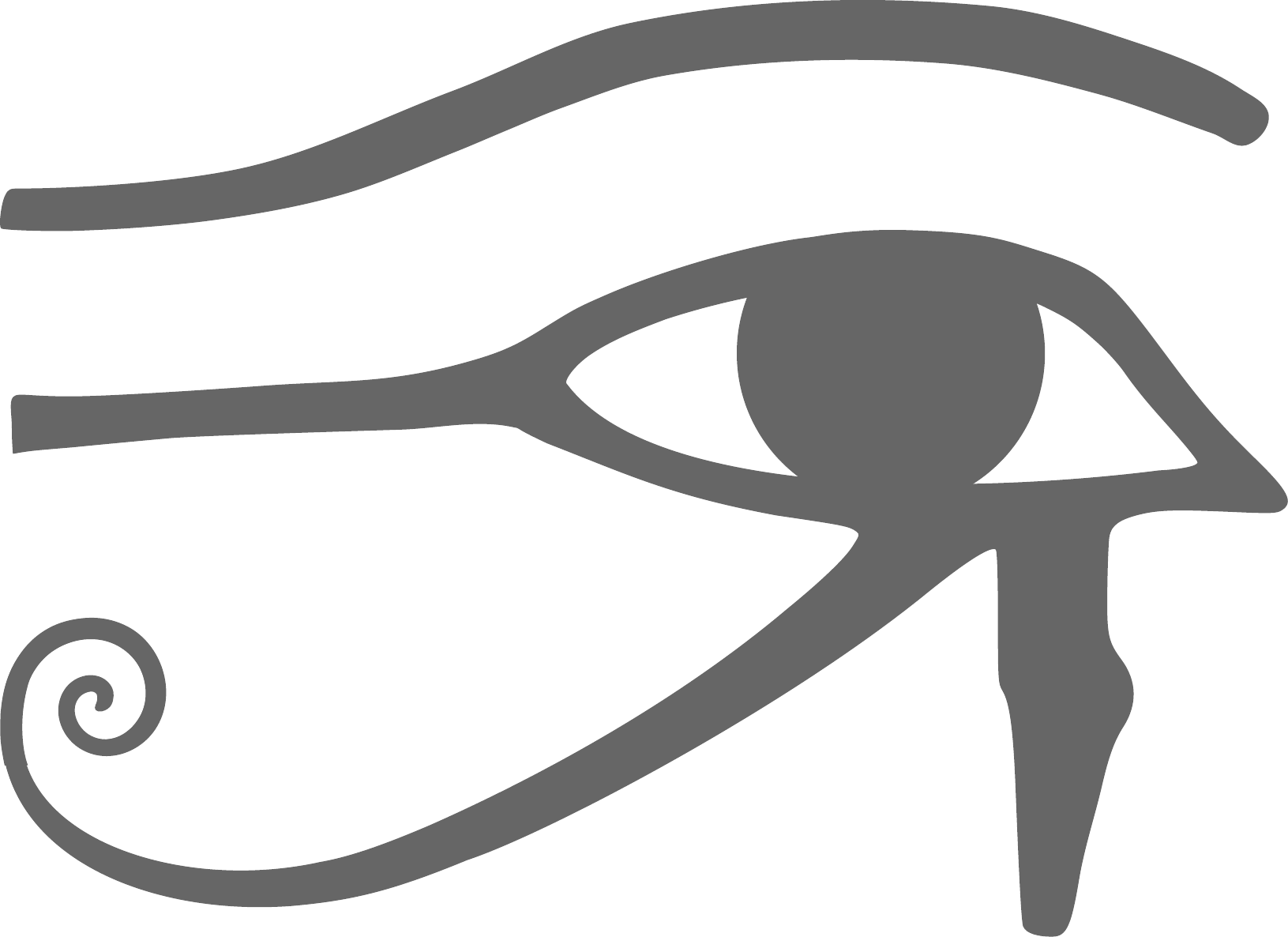}

~

\textit{
``What I hate is ignorance, smallness of imagination, 
the eye that sees no farther than its own lashes. 
All things are possible ... 
Who you are is limited only by who you think you are.''~-~ The Egyptian Book of the Dead.}
\end{center}
~\vfill

\newpage

\chapter*{Abstract}
\addcontentsline{toc}{chapter}{Abstract}
Legged robots are advancing towards being fully autonomous
as can be seen by the recent developments in academia and industry.  
To accomplish breakthroughs in dynamic whole-body locomotion, 
and to be robust while traversing unexplored complex environments, 
legged robots have to be \textit{terrain~aware}. 
	
\gls{tal} implies that the robot can 
perceive the terrain with its sensors, 
and can take decisions based on this information. 
The decisions can either be in planning, control, or in state estimation, 
and the terrain may vary in geometry or in its physical properties.  
\gls{tal} can be categorized into 
\gls{ptal}, which relies on the internal robot measurements to negotiate the terrain, 
and \gls{etal} that relies on the robot's vision to perceive the terrain. 
This thesis presents \gls{tal} strategies 
both from a proprioceptive and an exteroceptive perspective. 
The strategies are implemented at the level of
locomotion planning, control, and state estimation, 
and are using optimization and learning techniques. 
	
The first part of this thesis focuses on \gls{ptal} strategies that 
help the robot adapt to the terrain geometry and properties. 
At the \gls{wbc} level, achieving dynamic \gls{tal} requires reasoning about the robot
dynamics, actuation and kinematic limits as well as the terrain interaction. 
For that, we introduce a \textit{\gls{pwbc}} framework 
that allows the robot to stabilize and walk over challenging terrain while taking into account the 
terrain geometry (inclination) and friction properties. 
The~\gls{pwbc} relies on rigid contact assumptions which makes it suitable only for stiff terrain. 
As a consequence, 
we introduce~\textit{\gls{stance}}
which is a soft terrain adaptation algorithm that 
generalizes beyond rigid terrain.
\gls{stance} consists of 
a~\gls{awbc} that adapts the locomotion strategies
based on the terrain impedance,
and an online~\gls{ste} that senses and learns the terrain impedance properties
to provide it to the~\gls{awbc}. 
Additionally, we 
demonstrate the effects of 
terrains with different impedances 
on state estimation for legged robots.

The second part of the thesis focuses on \gls{etal} strategies 
that makes the robot aware of the terrain geometry 
using visual (exteroceptive) information. 
To do so, 
we present~\textit{\gls{vital}}
which is a locomotion planning strategy. 
\gls{vital} consists of
a~\gls{vpa} algorithm to plan the robot's body pose, 
and a~\gls{vfa} algorithm to select the robot's footholds. 
The~\gls{vfa} is an extension to the state of the art in foothold 
selection planning strategies.
Most importantly, the~\gls{vpa} algorithm 
introduces a different paradigm for vision-based pose adaptation.
\gls{vital} relies on a set of robot skills that 
characterizes the capabilities of the robot and its legs.
These skills are then learned 
via self-supervised learning using \glspl{cnn}.
The skills include (but are not limited to) the robot's ability to 
assess the terrain's geometry, avoid leg collisions, and to avoid reaching kinematic limits.
As a result, 
we contribute with
an online vision-based locomotion planning strategy that 
selects the footholds based on the robot capabilities, 
and the robot pose that maximizes the chances of the robot succeeding in reaching these footholds. 

Our strategies 
are based on optimization and learning methods, and
are extensively validated on the quadruped robots \acrshort{hyq} and \acrshort{hyqreal} in simulation and 
experiment. 
We show that with the help of these strategies, we can push dynamic legged robots one step closer 
towards being fully autonomous and terrain aware.  

\chapter*{Acknowledgments}
\addcontentsline{toc}{chapter}{Acknowledgments}

I would like to thank Claudio for giving me the chance to work at the DLS lab,
and for his continuous support during my PhD. 
Claudio creates an exceptional, motivating, and comfortable environment 
for everyone in the lab. 
He was always available when I needed him, and his suggestions were always critical.

During my PhD, 
I worked under the supervision of Andreea, Michele, and Victor
whom I truly enjoyed working with and learning from.
Andreea, 
thank you for being patient with me. I know I was hard to get along in the beginning. 
Thank you for covering up on how bad I am at making pretzels. 
Michele, thank you for saving me when Andreea left.
I truly enjoyed working with you, and writing down proofs on the white board. 
You always have this motivation and spark towards research that I am sure it will never fade. 
Victor, first of all, thank you for the barbecues; no one beats Victor when it comes to barbecues. 
Thank you for always pushing me to do my best. 
I will never forget the night we stayed late to submit a paper.
That night you stayed with us till we submitted, 
and you kept waking me up to support the team.

Working at the DLS lab is of one of the greatest experience that I have had. 
My experience would not have been the same without all the current and the past members of the 
DLS lab. 
Particularly,
I would like to thank Geoff whom I learned a lot from on the personal and professional level. 
Thanks for giving me your time albeit having loads of work on your shoulders, and
for teaching me that the world frame does not exist in reality. 
Thanks for being there when I wanted someone to talk to when I had a mental breakdown. 
Yet, none of that matters because you did not name your daughter Shamela!
Special thanks to Octavio who was my first close friend in Genoa. 
Thank you for being a part of my family, and for being my official interpreter.
Another special mention to Chundri. 
The lab has improved dramatically since Chundri stepped foot in the lab. 
I still owe him a Margarita.
I would also like to thank 
Angelo, 
Abdo, 
Letizia, 
Domingo, 
and Salvatore (the perfect Italian model). 

\newpage
Getting a PhD was a great achievement to me. 
But equally important was to get a work-life balance, and to live a healthy life. 
That would have never been possible without my friends. 
Particularly, Abril. 
Abril, thank you for being my coach at the gym and for taking it slow with me. 
Thank you for teaching me how to be fit, eat healthy, and enjoy my life outside work. 
You truly made me a different (better) person. 
Yet, none of that matters because you did not name your daughter Shamela!
I would also like to thank
Carlos G. and Marial (you still need to take me to that Jazz club!),
Andrea B. (I will not forgive you for leaving the house early without waking me up),
Romeo (what happened in Freiburg stays in Freiburg),
Eamon (you know we love you because of your awesome parents!), and
Fabrizia (for the paste di mandorla).
My thoughts are also dedicated to
Kristina (for the best birthday gift that I have ever had),
Juliet (my wing woman),
Lidia (for being in her top 10 list),
Maria (I am still available if you need a model), 
Sep (for his Christmas parties, and for never giving up on me going out for apperitivo),
Olmo (for his yearly New Year's Eve parties that never fails to impress), 
Francesca (you still did not take me out for pizza),
Mihail,
Anthony,
Dimitrious,
Mieke,
Edwin,
Aida, and~Wiebke.
I would also like to thank my friends
that albeit living far from each other, we're still close:
Anwar, Adel, Essam, Samer, Fay, Pavel, Ena, Eva, and Vassilina

I would have never reached this point 
without my parents, Amal and Shamel (yes, my dad is the true Shamel).
Thank you for your infinite love and support. 
Thank you for putting us first and for doing everything possible
to make us feel happy, supported, and appreciated. 
Thank you for believing in me, and for encouraging me to follow my dreams and passion. 
I would like to thank Menna, Mohamed, and our latest family member, Selim.
I live by your support and love. 

Last but not least, 
I would like to thank my creative, smart, and loving partner Ilef.
Thank you for drawing a smile on my face from the moment I saw you.
Thank you for your continuous love and support.

\vfill

\hfill Shamel Fahmi

\newpage
\chapter*{Preface}
\addcontentsline{toc}{chapter}{Preface}
\begin{itemize}
\item 
This doctoral thesis 
is building upon decades of research and development
in robotics, dynamics, controls, and machine learning. 
We expect that the reader has 
a basic knowledge about legged robotics 
before reading this thesis. 

\item 
This doctoral thesis is styled as a cumulative thesis. 
The main contents are
based on publications from peer-reviewed journals, 
with an exception of one chapter that contains yet unpublished material. 
\chapref{chap_intro}
gives a brief introduction to the problem and the state of the art, 
and it lists the contributions and the outline of the thesis. 
Chapters~\ref{chap_pwbc},~\ref{chap_stance},~and~\ref{chap_lsens} include the peer-reviewed publications
while \chapref{chap_vital} includes the yet to be published one. 
Finally, \chapref{chap_conc} concludes this thesis with 
a discussion and a summary of this work and its future directions.

\item
The articles 
that Chapters~\ref{chap_pwbc}-\ref{chap_vital} are based on
are my original work as a first author.
However, 
these articles are also the fruit of the effort of 
the supervisors and co-authors that assisted 
me during the period of my PhD. 
For this reason, 
I decided to use 
the active plural voice (we and our) 
instead of the singular voice (I and my) throughout the text.

\item \chapref{chap_pwbc} has been published in \cite{Fahmi2019}.
The concept and theory of this work has been developed by myself and M.~Focchi. 
The formulation and implementation has been developed by myself with the support of M.~Focchi. 
The experiments were conducted and analyzed by M.~Focchi with the support of C.~Mastalli and myself.
The manuscript was written by myself with the support of M.~Focchi and C.~Mastalli, 
and was reviewed by C.~Semini.

\item \chapref{chap_stance} has been published in \cite{Fahmi2020}.
The concept and theory of this work has been developed by myself and M.~Focchi. 
The formulation and implementation has been developed by myself with the support of M.~Focchi and A.~Radulescu. 
The experiments were conducted and analyzed by myself with the support of M.~Focchi and G.~Fink.
The manuscript was written by myself with the support of M. Focchi and G.~Fink, 
and it was reviewed by A.~Radulescu, V.~Barasuol, and C.~Semini.

\item \chapref{chap_lsens} has been published in \cite{Fahmi2021}.
The concept and theory of this work has been developed by myself with the support of G.~Fink. 
The formulation and implementation has been developed by myself with the support of G.~Fink. 
The experiments were conducted and analyzed by myself and G.~Fink.
The manuscript was written by myself and G.~Fink and was reviewed by C.~Semini.

\item Chapters~\ref{chap_vital}~and~\ref{chap_6} have been published in \cite{Fahmi2022}.
The concept and theory of this work has been developed by myself and V.~Barasuol. 
The formulation and implementation has been developed by myself with the support of
D.~Esteban, O.~Villarreal, and  V.~Barasuol. 
The experiments were conducted and analyzed by myself with the support of V.~Barasuol.
The manuscript was written by myself with the support of V.~Barasuol,
and it was reviewed by D.~Esteban, O.~Villarreal, and C.~Semini.
Note that, part of this chapter has been revised after the defense date.

\item The style of this dissertation has been adopted from A.~Winkler's dissertation \cite{Winkler2018}.
\end{itemize}

\clearpage

\addcontentsline{toc}{chapter}{Contents}
\tableofcontents
\clearpage

\addcontentsline{toc}{chapter}{\listfigurename}
\listoffigures 
\clearpage

\addcontentsline{toc}{chapter}{\listtablename}
\listoftables 
\clearpage


\addcontentsline{toc}{chapter}{Acronyms}
\printglossary[type=\acronymtype,title={Acronyms}]
\clearpage

\pagestyle{headings}

\glsresetall 
\chapter{Introduction}\label{chap_intro}
Marc Raibert gave a broad definition of intelligence in a recent talk, 
and divided it into: \gls{ci} and \gls{ati}\cite{Raibert2020}.  
\gls{ci}~allows us to make abstract plans, and to understand and solve broader problems. 
\gls{ati}~on~the~other hand, 
allows us to operate our bodies in such a way that we can 
balance, stand, walk, climb, etc. 
\gls{ati} also lets us do real-time perception so that we can interact with the world around us. 
Marc also noted that although not all of us are athletes, we still have a great amount of \gls{ati}
in us. 
This thesis is about \gls{ati} for legged robots;
it can perhaps be one step towards reaching animal-level~\gls{ati}.

\newpage
\section{Motivation}
Legged robots have been around for decades. 
Recently however, they have shown remarkable 
agile capabilities 
thanks to the research efforts of academia and industry. 
For this reason, legged robots 
are moving out of research labs
into the real world with the promise of being athletically intelligent. 
The promise is that legged robots are to aid humans in various applications. 
The applications include (but are not limited to)
warehouse logistics, 
inspection at industrial plants and construction sites, 
search and rescue, 
agriculture, 
package delivery, 
space exploration, 
etc. 
In all of these applications, 
there is perhaps one thing in common:
none of the terrains that the robots traverse are the same. 
In fact, these terrains are usually 
dynamic, unexplored, and uncertain. 
As a result, the core problem is that 
the terrain that robots traverse
introduces a large amount of uncertainty.
Therefore, for legged robots to 
achieve \gls{ati} and accomplish breakthroughs in dynamic whole-body locomotion, 
they have to be \textit{terrain~aware}.

\gls{tal} means that
the robot is able to \textbf{perceive} and \textbf{understand} the surrounding terrain, 
and is able to \textbf{take decisions} based on that.
In other words, 
the robot has to have a good knowledge of its surroundings
and use whatever sensors it has to perceive these surroundings and act upon them.
The terrain itself may vary in its geometry or in its physical properties, 
and the decisions can either be in planning, control, or in state estimation.
To~clarify, 
let us raise the following questions:

\begin{itemize}
\item Can the robot sense (see and feel) the world around it, and the terrain it is traversing?
\item Can the robot understand the differences between the geometrical and physical properties
of the terrain it is traversing?	
\item 
Can the robot plan its motion
based on its understanding of the terrain and its own limitations?
\item 
Can the robot quickly adapt this planned motion 
in case something goes wrong with it
(such as falling, slipping, external pushes, etc.)?
\end{itemize}
If the answer is yes, then the robot is terrain aware.

\begin{figure}[!t]
	\centering
	\includegraphics[width=0.95\textwidth]{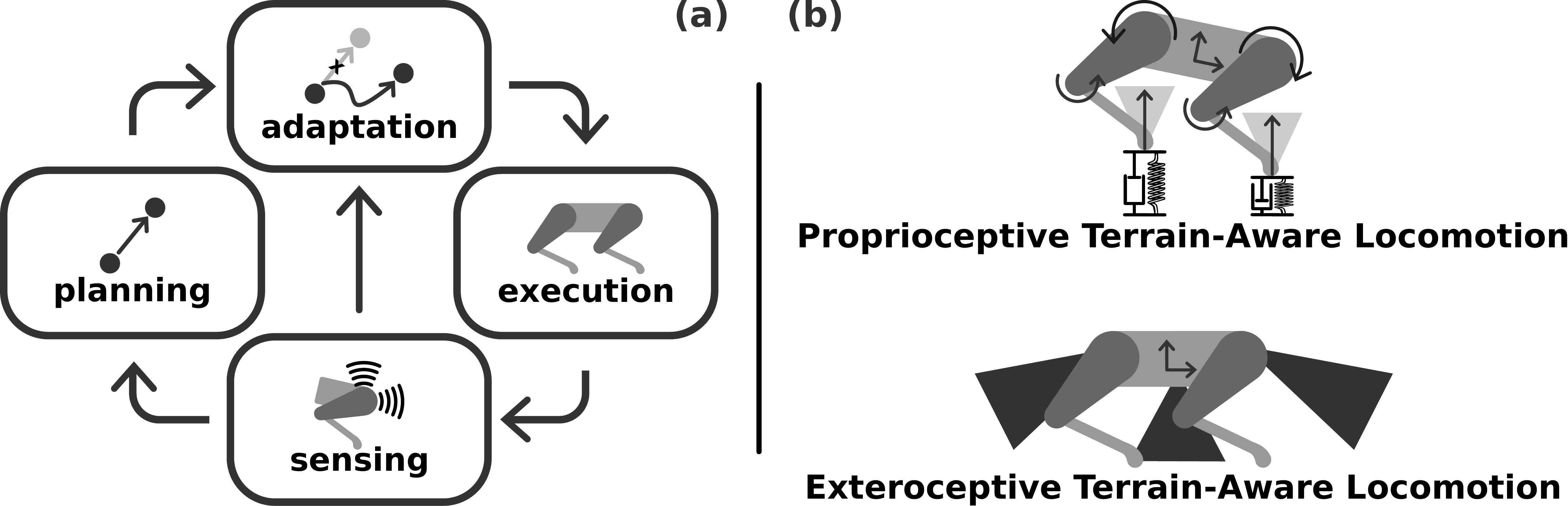}
	\caption
	[The Bigger Picture of Locomotion for Legged Robots.]
	{The Bigger Picture of Locomotion for Legged Robots.
	(a) an overview of the pipeline used in locomotion strategies for legged robots.
	(b) an illustration of the two categories of \acrfull{tal}.}
	\label{fig_chap0}
\end{figure}

\section{The Bigger Picture}\label{bigger_pic}
This section provides a broad overview
of the pipeline used in locomotion strategies for legged robots, 
and details on each block.
As shown in~\fref{fig_chap0}(a),
the pipeline has four main modules: sensing, planning, adaptation and execution. 

~

\textsf{\textbf{Sensing:}}~ 
This is the perception module
that encapsulates all of the robot's ability to perceive itself and its surroundings. 
For that, the robot relies on its onboard sensors 
to measure its base and joint states,
and build a map of its surroundings. 
These onboard sensors include 
IMUs, joint encoders, torque sensors, 
as well as its onboard LIDARs and cameras.

~

\textsf{\textbf{Planning:}}~
This is the trajectory generation module that plans the motion of the robot. 
The goal of the planning module is to
understand the perceived information from the sensing module,
and plan a motion for the robot to take accordingly.
This motion tends to be a long term horizon motion.

~

\textsf{\textbf{Adaptation:}}~
The adaption module acts as an intermediate module between the planning and execution modules. 
This module tends to be usually merged with the planning module. 
However, it is important to understand the core differences between both. 
The adaptation module has a re-planning nature.
This means that if the planned motion is not executed as programmed, 
or if something goes wrong during execution 
(such as if the robots falls, slips or gets disturbed or pushed),
the adaptation module can be able to sense this right away, 
and adapt the robot's motion accordingly. 

\textsf{\textbf{Execution:}}~
After perceiving the sensed information, and after planning and adapting the robot's motion, 
the robot finally gets to execute this motion at the joint level. 
Hence, the robot has to track its desired whole-body states while 
reasoning about its own dynamics and limits, and about the surrounding~terrain.

\vspace{-10pt}
\section{Terrain-Aware Locomotion (TAL)}\label{tal_intro}
\vspace{-10pt}
\gls{tal} can be categorize into \gls{ptal} and \gls{etal} as shown in~\fref{fig_chap0}(b).
\gls{ptal} relies on the internal robot measurements 
(mainly its whole body states) 
to acquire the terrain information that is surrounding the robot.
\gls{etal} relies on directly acquiring this information
using the robot's visual sensing. 

An early work on \gls{ptal} was on reflex actions that reactively
adapt the swinging legs trajectory to overcome obstacles
if a collision is detected~\cite{Focchi2013}.
Since proprioceptive sensors measures the internal robot states, 
detecting and localizing contacts on the robot is possible. 
For instance, some \gls{ptal} strategies rely on the 
joint position, velocity and/or torque measurements to detect and localize
contacts~\cite{Manuelli2016, Barasuol2019, Wang2020}, 
and to detect slippage~\cite{Focchi2018b, Nistico2022}. 
In addition to the terrain's geometry, 
\gls{ptal} strategies  are also used to infer and adapt to
the physical properties of the terrain.
For instance, several works have
adopted \gls{ptal} strategies in locomotion planning and control over different
terrain impedance parameters~\cite{Bosworth2016, Fahmi2020, Chatzinikolaidis2020}.
In these works, the robot was able to detect changes in the terrain impedance, 
and act upon it online. 

\gls{ptal} strategies are useful in many scenarios when visual feedback is denied 
(such as smoky areas, or areas with thick vegetation) or
when the terrain map is unreliable. 
However, based on their proprioceptive nature, 
the actions from \gls{ptal} strategies are limited 
to corrective actions because
predicting future robot-terrain interactions
using only the robot's internal states is insufficient. 
This means that \gls{ptal} strategies do not act on what is ahead of the robot. 
Hence, an action has to happen first before triggering a reactive strategy; 
the foot has to collide before triggering a step reflex, 
or touch the terrain before inferring its physical properties.

Unlike \gls{ptal}, \gls{etal} relies mainly on visual information. 
This gives \gls{etal} strategies the advantage of looking ahead of the robot.
One famous \gls{etal} strategy is in selecting 
the best footholds based 
on the terrain information and the capabilities of the legs. 
This is often referred to as foothold selection~\cite{Kolter2008, Kalakrishnan2009, Belter2011, Barasuol2015}. 
Apart from foothold selection 
and similar to \gls{ptal}, 
\gls{etal} strategies have also been used to infer the terrain properties
from images using deep learning~\cite{Filitchkin2012, Wellhausen2019, Ahmadi2020}.

\section{Contributions}\label{thesis_contribution}
This thesis summaries our work done on \gls{tal} for legged robots. 
Our work includes strategies implemented for both \gls{ptal} and \gls{etal},
and is applied at the levels of planning, control, and state estimation. 
This thesis is divided into two parts. 
The first part focuses on \gls{ptal} 
and the second part focuses on \gls{etal}. 
This thesis is based on four main articles
papers~\cite{Fahmi2019,Fahmi2020, Fahmi2021, Fahmi2022}.
Each article is self-contained and is included in a stand-alone chapter. 
The remainder of this section summarizes the motivation and contributions of each of these articles. 

\subsection*{C1: Passive Whole-Body Control for Quadruped Robots}
\pwbcPaper

To achieve \gls{ati} as explained earlier in this chapter,  
the locomotion strategy should be able to reason about
the robot's capabilities, and to be terrain aware.
Thus, as a first step, the first paper contributes to 
\gls{ptal} strategies by presenting a \gls{wbc} framework.

This paper presents a \gls{pwbc} framework for quadruped robots, 
and focuses on the experimental validation. 
The \gls{pwbc} is aware of the terrain geometry and friction properties. 
Additionally, 
the \gls{pwbc} 
achieves dynamic locomotion while compliantly balancing the robot's trunk. %
To do so, 
we formulate the motion tracking as a \gls{qp} that takes into account
the full robot rigid body dynamics, the actuation limits, the joint limits 
and the contact interaction.
To be terrain aware, we encode the terrain geometry (inclination), 
and frictional properties in the \gls{qp} formulation. 
To maintain contact consistency with the rigid terrain, 
we also encode the rigid contact interaction in the \gls{qp} formulation. 

To validate the approach used in this paper, 
we analyze the \gls{pwbc}'s robustness against
inaccurate terrain friction properties,
and the robot's ability to adapt to any sudden change in the actuation limits. 
We also present extensive experimental trials on the \acrshort{hyq} robot, 
and 
validate the \gls{ptal} capabilities of the \gls{pwbc} under various terrain conditions and gaits. 
The paper also includes extensive implementation details gained
from the experience with the real platform.

\subsection*{C2: STANCE: Locomotion Adaptation over Soft Terrain}
\stancePaper

\begin{remark}
This work has been selected as a finalist for the IEEE RAS Italian Chapter Young Author Best Paper Award 2020, and for the IEEE RAS Technical Committee on Model-Based Optimization for Robotics Best Paper Award 2020.
\end{remark}

The previous paper presented a \gls{pwbc} framework that was \textit{rigid} contact consistent.
In other words, the \gls{pwbc} was terrain aware with respect to rigid terrain. 
In fact, most of \gls{wbc} frameworks fail to generalize beyond rigid terrains.  
To be terrain aware, the robot should be able to adapt to terrains 
with different impedances.
For that, we focused on extending the \gls{ptal} capabilities of 
the previously presented \gls{pwbc}, 
and adapting it to multiple terrains with different impedances (such as soft terrain).
We study compliant terrain since it is an unsolved issue for legged locomotion. 
Legged locomotion over soft terrain is difficult 
because of the presence of unmodeled contact dynamics 
that standard \glspl{wbc} do not account for.
This introduces uncertainty in locomotion and affects the stability and performance of the system. 

Therefore, 
this paper proposes a novel soft terrain adaptation algorithm called 
\gls{stance}.
From its name, 
\gls{stance} consists of 
a \gls{awbc} that is aware of the terrain impedance,
and an online \gls{ste} that senses and estimates the terrain impedance. 
The \gls{awbc} exploits the knowledge of the terrain to generate an optimal solution that is contact consistent. This terrain knowledge is provided to the \gls{awbc} by the \gls{ste}.

In this paper, 
we show that  \gls{stance}  can adapt online to any type of terrain compliance (stiff or soft). 
To do so, 
we evaluated \gls{stance} 
both in simulation and experiment on \acrshort{hyq}, 
and we compared it with the state of the art \gls{pwbc} from the previous paper.
We demonstrated the capabilities of \gls{stance} with multiple terrains of different compliances,
with aggressive maneuvers, different forward velocities, 
and external disturbances. \gls{stance} allowed \acrshort{hyq} to adapt online to terrains with different compliances (rigid and soft) without pre-tuning. 
\acrshort{hyq} was able to  successfully deal with the transition between different terrains and showed the ability to  differentiate between compliances~under~each~foot.

\subsection*{C3: State Estimation for Legged Locomotion over Soft Terrain}
\lsensPaper

The previous \gls{stance} paper presented a \gls{ptal} strategy to adapt to soft terrain. 
One of the limitations of that paper was in state estimation for legged robots over soft terrain. 
This is a limitation because  
most of the work done on state estimation for legged robots 
is designed for rigid contacts, 
and does not take into account the physical parameters of the terrain.
Thus,
this paper is a step towards extending
the \gls{ptal} capabilities of legged robots to state estimation.
In detail, 
this paper
answers the following questions: 
how and why does soft terrain affect state estimation for legged robots? 
To do so, 
we utilize a state estimator that fuses IMU measurements with leg odometry
that is designed with rigid contact assumptions.
We experimentally validate the state estimator with \acrshort{hyq} trotting
over both soft and rigid terrain. 
Then, we demonstrate that soft terrain negatively affects state estimation for legged robots, 
and that the state estimates have a noticeable drift over soft terrain compared to rigid terrain.

\subsection*{C4: ViTAL: Vision-Based Terrain-Aware Locomotion}
\vitalPaper
Unlike the previous contributions that were 
\gls{ptal} strategies, the second part 
of this thesis (and the fourth contribution)
is an \gls{etal} strategy. 
This work focuses particularly on 
vision-based planning strategies 
that decouple locomotion planning into foothold selection and pose adaptation.
Despite the work done for foothold selection, pose adaptation strategies lag behind. 
The core problem of the current pose adaptation strategies is that 
they focus on finding \textit{one optimal} solution based on \textit{given} selected footholds. 
This is a problem because
there are no guarantees on what would happen 
if the selected footholds are not reached, or if the robot gets disturbed. 
If any of these cases happen, 
the robot may end up in a pose 
that makes the feet reach kinematic limits, or collide with the terrain. 
This would in turn compromise the robot's performance and safety. 
To solve this problem,
we should not find body poses that are optimal with respect to a given foothold, 
but rather find body poses that maximize the chances of reaching safe footholds.

With this in mind, we present a locomotion planning strategy called \gls{vital}.
\gls{vital} consists of 
a pose adaptation algorithm called \gls{vpa}, 
and a foothold selection algorithm called \gls{vfa}. 
The \gls{vfa} is an extension of state of the art foothold selection strategies.
The \gls{vpa} introduces a different paradigm for pose adaptation strategies. 
The \gls{vpa} 
is a pose adaptation algorithm that 
finds the body pose that maximizes the number of safe footholds
based on a set of skills. 
The skills represent the capabilities of the robot and its legs
including the ability to assess the terrain's geometry, avoid leg collisions, and to avoid reaching kinematic limits during the swing and stance phases.
These skills are then learned via self-supervised learning using \glspl{cnn}. 
Therefore, \gls{vital}
is an online strategy that simultaneously plans the robot's body pose and footholds 
based on the robot capabilities.

To validate~\gls{vital}, we use the \acrshort{hyq} and \acrshort{hyqreal} robots.
Thanks to~\gls{vital}, our robots are able to 
climb various obstacles including stairs and gaps at different speeds. 
We also compare the \gls{vpa} with a baseline strategy
that selects the robot pose based on given selected footholds, 
and show that it is indeed not robust enough to select robot poses based only on given footholds.


\newpage~\thispagestyle{plain} \newpage

\part{Proprioceptive Terrain-Aware Locomotion}
\glsresetall \chapter[Passive Whole-Body Control for Quadruped Robots]{\vspace{-5pt}Passive Whole-Body Control\\ for Quadruped Robots}\label{chap_pwbc}

\pwbcPaper

\boldSubSec{Abstract}
We present experimental results using a passive whole-body control approach for 
quadruped robots
that achieves \textit{dynamic} locomotion while compliantly balancing the robot's trunk.
We formulate the motion tracking as a Quadratic Program (QP) that takes into account
the full robot rigid body dynamics, the actuation limits, the joint limits 
and the contact interaction. 
We analyze the controller's robustness against
inaccurate friction coefficient estimates and unstable footholds, 
as well as its capability to redistribute the load  
as a consequence of enforcing actuation limits.
Additionally, we present practical implementation details gained
from the experience with the real platform.
Extensive experimental trials on the \unit[90]{kg} Hydraulically actuated Quadruped (HyQ)
robot validate the
capabilities of this controller under various terrain conditions and gaits.
The proposed approach is superior for accurate 
execution of highly dynamic motions with respect 
to the current state of the art.

\boldSubSec{Accompanying Video} \href{https://youtu.be/Lg3V_juoE1w}{\texttt{https://youtu.be/Lg3V\_juoE1w}}

\section{Introduction}\label{sec:introduction}
Achieving dynamic locomotion 
requires reasoning about the robot's dynamics, actuation limits and interaction with the environment 
while traversing challenging terrain (such as rough
or sloped terrain).
Optimization-based techniques can be exploited to  attain these objectives
in locomotion planning and control of legged robots.
For instance, one approach is to use non-linear \gls{mpc} 
while taking into consideration the full dynamics of the robot. 
Yet, it is often challenging to meet real-time
requirements because the solver can  get stuck in local minima, 
unless  proper warm-starting  is used \cite{Erez2013}.
Thus, 
current research often relies on low dimensional models or constraint relaxation
approaches to meet such requirements (e.g. \cite{Kuindersma2014}). 
Other approaches rely on decoupling the motion planning  from the  motion control
\cite{Farshidian2017b,Cabezas2018,Bellicoso2018}. 
Along this line, an optimization-based  motion planner 
could rely on low dimensional models to compute 
\gls{com} trajectories and footholds while a locomotion controller  tracks these
trajectories. 

Many recent contributions in locomotion control have been proposed in the
literature that were successfully tested on bipeds and quadrupeds
(e.g. \cite{Herzog2016,Koolen2016,Henze2016,Farshidian2017a,Bellicoso2018,Kim2018}). Some of them
are based on  quasi-static
assumptions or lower dimensional models  \cite{Stephens2010,Ott2011,Focchi2017}.
This often limits the dynamic locomotion capabilities of the robot \cite{Herzog2016}. 
Consequently, another approach, that is preferable for dynamic motion, is based on
\gls{wbc}. \gls{wbc} facilitates such decoupling between the motion planning and control in such a way that it is easy 
to accomplish multiple tasks while respecting the robot's behavior \cite{Farshidian2017a}.
These tasks might include motion tasks for the robot's end effectors 
(legs and feet) \cite{Henze2016,Farshidian2017a}, but also could be 
utilized for contacts anywhere on the robot's body \cite{Henze2017} or for a 
cooperative manipulation task between robots \cite{Bouyarmane2018}.
\gls{wbc} casts the locomotion controller 
as an \textit{optimization} problem,  in which, by incorporating the full dynamics of the legged robot, 
all of its \gls{dofs} are exploited in order to spread the desired motion tasks globally to all the joints.
This allows us to reason about multiple tasks  and solve them in an optimization fashion while respecting the 
full system dynamics and the actuation and interaction constraints. 
\gls{wbc} relies on the fact that robot dynamics 
and constraints could be formulated, at each loop, as linear constraints with a convex cost function (i.e., a \gls{qp})  \cite{Kuindersma2014}.
This allows us to solve the optimization problem in real-time.

Passivity theory is proven to guarantee a certain degree of robustness during interaction with the environment 
\cite{Stramigioli2015}. For that reason, such tool is commonly exploited in the design of locomotion controllers to 
ensure a 
passive contact interaction.  
Passivity based \gls{wbc} in humanoids was introduced first by \cite{Hyon2007} to effectively balance 
the robot when experiencing contacts. By providing compliant tracking and gravity compensation, the humanoid was 
able 
to adapt to unknown disturbances. The same approach was further extended first  by \cite{Ott2011} and later by 
\cite{Henze2016}. 
The former extended \cite{Hyon2007}  to posture control, 
while the  latter analyzed the passivity of a humanoid 
robot in multi-contact scenarios (by exploiting  the 
similarity with PD+ control \cite{Ortega2013}).
\begin{sidewaysfigure}
\centering
\includegraphics[width=1.0\textwidth]{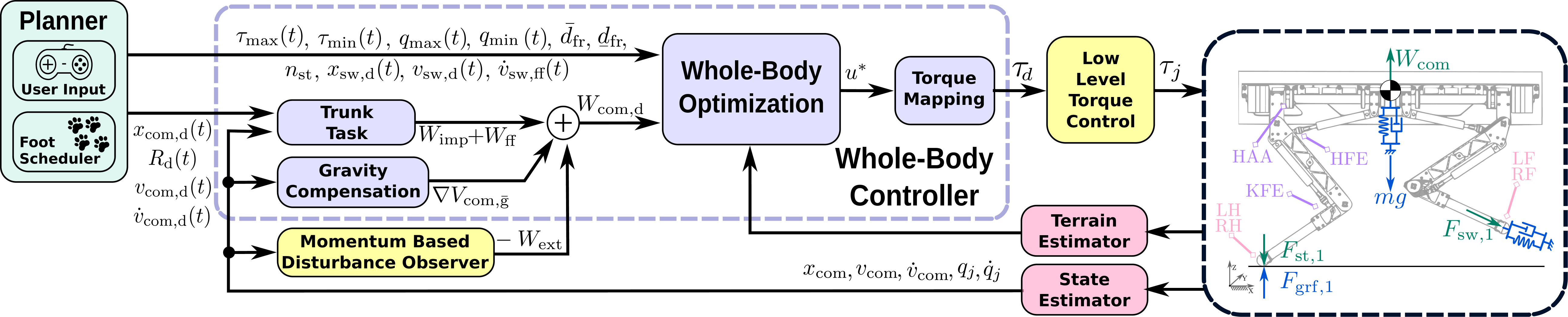}
\caption
[Overview of the \acrshort{wbc} as part of our locomotion framework.]
{Overview of the whole-body controller as part of our
locomotion framework.  The dashed black box (cartouche) presents an overview of the robot's joints, 
feet and generated wrenches and forces.
LH, LF, RH and RF are Left-Hind, Left-Front, Right-Hind and Righ-Front legs, respectively. HAA, HFE and KFE are 
Hip Abduction/Adduction, Hip Flexion/Extension and Knee Flexion/Extension, respectively. 
}
\label{fig:blockDiagram}
\end{sidewaysfigure}

In our previous work \cite{Focchi2017}, 
the locomotion controller was
designed for \textit{quasi-static} motions using only the robot's centroidal dynamics. 
Under that assumption, we noticed that 
during dynamic motions, the effect of the leg dynamics  no longer negligible; and thus, it
becomes necessary to abandon the quasi-static assumption to achieve  good tracking.
Second, since the robot is constantly interacting  with the
environment (especially during walking and running), it is crucial to ensure a
compliant and passive interaction.
For these reasons, in this paper, 
we improve our previous work \cite{Focchi2017} by implementing a passivity based \gls{wbc} 
that incorporates the full robot dynamics  and interacts  compliantly with the 
environment, while satisfying the kinematic and torque limits.
Our \gls{wbc} implementation is capable of achieving \textit{faster} dynamic motions 
than our  previous work.
We also integrate terrain mapping and state estimation on-board 
and present some practical implementation details  gained from the experience with the real platform. 

\textit{Contributions:} 
In this paper, we mainly present \textit{experimental} contributions in which we 
demonstrate the effectiveness of the controller both in simulation and experiments on \gls{hyq}. 
Compared to previous work on passivity-based \gls{wbc} \cite{Henze2016,Ott2011},  
in which experiments were conducted on the robot while standing (not walking or running),
we tested our controller on \gls{hyq} during crawling and trotting. 
Similar to the recent successful work of  \cite{Bellicoso2018} and \cite{DiCarlo2018}
 in  quadrupedal 
locomotion over rough terrain, we used similar terrain templates to present experiments of our passive \gls{wbc} on 
\gls{hyq} using multiple gaits over slopes and rough terrain of different heights.

The rest of this paper is structured as follows:
%
In \sref{sec:wholeBodyController} we present the detailed formulation and design of our \gls{wbc}
followed by its passivity analysis in \sref{sec:Passivity}. 
\sref{sec:Implementations} presents further crucial implementation details. 
Finally we present our simulation and experimental results in \sref{sec:expResults} followed by our conclusions in 
\sref{sec:conclusion}. 
\section{Whole-Body Controller (WBC)}
\label{sec:wholeBodyController}
In this section we present and formulate our \gls{wbc}. 
Figure \ref{fig:blockDiagram} depicts the main components of our locomotion framework. 
Given high-level user velocity commands, the planner generates a reference motion
online \cite{Focchi2018} or offline \cite{Cabezas2018}, and provides it to the 
\gls{wbc}.
Such references include the desired trajectories for \gls{com}, trunk orientation
and  swing legs.
The \textit{state estimator} supplies the controller with an estimate of the actual state of the robot,
by fusing  leg odometry, inertial sensing, visual odometry and LIDAR
while, the 
\textit{terrain estimator}, provides an estimate of the  terrain inclination (i.e. surface normal).
Finally, there is a momentum-based observer that estimates external 
disturbances \cite{Focchi2018} and a lower-level torque controller.
The goal of the designed \gls{wbc} is to keep the quadruped robot balanced (during running, walking or 
standing) while interacting passively with the environment.
The motion tasks of a quadruped robot can be categorized into 
a \textit{trunk task} and a \textit{swing task}. 
The trunk task regulates the position  of the \gls{com} and the  orientation of the trunk\footnote{Since \gls{hyq} is 
not equipped with arms, it suffices for us to control the trunk orientation instead of 
the whole robot angular momentum.}  and is achieved by 
implementing a Cartesian-based impedance controller with a feed-forward term\footnote{This is similar to a PD+ 
controller \cite{Ortega2013}.}.
The swing task regulates the swing foot trajectory in order to place 
it in the desired location while achieving enough clearance from the terrain.
Similar to the trunk task, the swing task is achieved by implementing a Cartesian-based impedance controller with a 
feed-forward term. 
The \gls{wbc} realizes these tasks by computing the optimal generalized accelerations and contact forces 
\cite{Herzog2016} via \gls{qp} and mapping them to the desired 
joint torques while taking into 
account the full dynamics of the robot, the properties of the terrain (\textit{friction constraints}), 
the unilaterality of the contacts (e.g. the legs can only push and not pull) (\textit{unilateral constraints}), 
and the actuator's \textit{torque/kinematic limits}. 
The desired torques,  will be sent to the 
lower-level (torque) controller.

\subsection{Robot Model} 

For a legged robot with $n$ \gls{dofs} and $c$  feet,
the forward kinematics of each foot is defined by $n_{\nmrm{a}}$  coordinates\footnote{Without the loss of 
generality, we consider a quadruped robot with $n=12$ \gls{dofs} with point feet, where $c = 4$ and 
$n_{\nmrm{a}} = 3$.}.
The total dimension of the feet operational space is $n_{\nmrm{f}} = n_{\nmrm{a}} 
c$. This can be separated into stance  ($n_{\mrm{st}} = n_{\nmrm{a}} c_{\mrm{st}}$) and swing feet 
($n_{\mrm{sw}} = n_{\nmrm{a}} c_{\mrm{sw}}$).
Since we are interested in regulating the position of 
the \gls{com},  we formulate the dynamics 
 in terms of the \gls{com}, using its
velocity rather than the base velocity\footnote{In this coordinate 
system, the inertia matrix is block diagonal \cite{Ott2011}. For the detailed 
implementation of the dynamics using the base velocity, see \cite{Hyon2007}.}
\cite{Ott2011}. Assuming that all the external forces are exerted on the \textit{stance feet}, we write the equation of motion that 
describes the full dynamics of the robot as:
\begin{equation}
\hspace{-0.5em}
\underbrace{\mat{M_{\mrm{com}} &	\vc{0}_{6\times n} \\
		\vc{0}_{n\times 6} & \bar{\vc{M}}_j}}_{\vc{M}(\vc{q})} \!\!
\underbrace{\mat{\dot{\vc{v}}_\mrm{com} \\ \ddot{\vc{q}}_j}}_{\ddot{\vc{q}}}\! + \! 
\underbrace{\mat{h_{com} \\ \bar{\vc{h}}_j}}_{h} \! = \!
\mat{ \vc{0}_{6\times n} \\ \vc{\tau_j}} +
\underbrace{\mat{\vc{J}_\mrm{st,com}^T \\ \vc{J}_\mrm{st,j}^T}}_{\vc{J}_\mrm{st}(\vc{q})^T}\vc{\grf}
\hspace{-0.5em}
\label{eq:full_dynamicsCOM}
\end{equation}
where the first 6 rows  represent the (un-actuated) floating 
base part and the remaining $n$ rows represent the actuated part.
$\vc{q}  \in SE(3) \times \Rnum^n$ represents the pose of the whole floating-base system while
$\vc{\dot{q}}=\mat{v_\mrm{com}^T & \vc{\dot{q}}_j^T}^T \in \Rnum^{6+n}$  and  $\ddot{\vc{q}} = [\dot{v}_{\mrm{com}}^T \ 
\ddot{q}_j^T]^T\in\Rnum^{6+n}$  are the vectors 
of generalized velocities and accelerations, respectively.
$v_\mrm{com} = [\dot{x}_\mrm{com}^T \  \vc{\omega}_b^T]^T$ $\in \Rnum^{6}$ 
and $\dot{v}_\mrm{com} = [\ddot{x}_\mrm{com}^T \  \dot{\vc{\omega}}_b^T]^T$ $\in \Rnum^{6}$
are the spatial velocity and acceleration of the floating-base expressed at the \gls{com}.
$\vc{M}(q) \in \Rnum^{(6+n) \times (6+n)}$ is the  inertia matrix,
where $M_\mrm{com}(q)$ $\in \Rnum^{6\times6}$ is the 
composite rigid body inertia matrix of the robot expressed at the \gls{com}.
$\vc{h}\in \Rnum^{6+n}$ is the force vector that accounts for Coriolis,
centrifugal, and gravitational forces
\footnote{Note that $h_{com} = -mg + 
v_{com} \times^*M_{com} v_{com}$ 
according to the spatial algebra notation, where $m$ is the total robot mass.}.
$\vc{\tau}\in\Rnum^n$ are the actuated joint torques while
$\vc{\grf} \in\Rnum^{n_\mrm{st}}$ is the vector of \gls{grfs} (contact forces).
%
%
In this context, the floating base Jacobian $J$ $\in\Rnum^{n_\mrm{f} \times(6+n)}$
is separated into swing Jacobian $J_{\mrm{sw}}$  $\in\Rnum^{n_{\mathrm{sw}} \times (6+n)}$ and stance 
Jacobian  $J_{\mrm{st}}$   $\in\Rnum^{n_{\mathrm{st}} \times (6+n)}$ which could be further expanded into 
$J_{\mathrm{st,com}}$  $\in\Rnum^{n_{\mathrm{st}} \times 6}$, $J_{\mathrm{st,j}}$  $\in\Rnum^{n_\mathrm{st} \times n}$, 
$J_{\mathrm{sw,com}}$ $\in\Rnum^{n_\mathrm{sw} \times 6}$ and  $J_{\mathrm{sw,j}}$ $\in\Rnum^{n_\mathrm{sw} \times 
n}$.  
%
%
%
%
The  operator $\bar{[\cdot]}$ denotes  the
matrices/vectors recomputed after the coordinate transform to the \gls{com} \cite{Hyon2007}. 
Following the sign convention in \fref{fig:blockDiagram}, recalling the first 6 rows in 
\eref{eq:full_dynamicsCOM}, and  by defining the gravito-inertial  \textit{\gls{com} wrench} as $W_\mrm{com} = M_\mrm{com}
\dot{v}_\mrm{com} +  h_{com} $ $\in \Rnum^6$, we can write the \textit{floating-base dynamics} as:
\begin{eqnarray}
	W_\mrm{com}  =  J_\mrm{st,com}^T \grf 
\label{eqn4}
\end{eqnarray}
such that  $J_{\mrm{st,com}}^T$ maps $\grf$ to the \gls{com} wrench space.

	The feet velocities $v=[v_{\mrm{st}}^T \ v_{\mrm{sw}}^T]^T$ $\in\Rnum^{n_\mrm{f}}$ could be separated into stance $v_{\mrm{st}}$ 
	$\in\Rnum^{n_{\mrm{st}}}$ and swing $v_{\mrm{sw}}$ $\in\Rnum^{n_{\mrm{sw}}}$ feet velocities.
	The mapping between $v$ and the generalized velocities $\dot{q}$ is: 
	\begin{subequations}
		\begin{eqnarray}
		v &=& J \dot{q} 
		\label{pass1} \\
		v  =\begin{bmatrix} J_{\mrm{com}} & J_{\nmrm{j}} \end{bmatrix}\begin{bmatrix} v_\mrm{com}  \\ 
		\dot{q}_\nmrm{j} \end{bmatrix}
		&=&  J_{\mrm{com}}v_\mrm{com} +  J_{\nmrm{j}} \dot{q}_\nmrm{j}
		\label{eq_pass4}
		\end{eqnarray} 
	\end{subequations}
such that  $J_{\mathrm{com}}$  $\in\Rnum^{n_{\mathrm{f}} \times 6}$  and $J_{\nmrm{j}}$ $\in\Rnum^{n_{\mathrm{f}} 
\times n}$. Similar to the feet velocities, we split the feet force vector $F = [F_{\mrm{st}}^T \ 
F_{\mrm{sw}}^T]^T$ 
$\in\Rnum^{n_\mrm{f}}$, into  $F_{\mrm{st}}$ $\in\Rnum^{n_\mrm{st}}$ and $F_{\mrm{sw}}$  $\in\Rnum^{n_\mrm{sw}}$.
\begin{assumption}
	The robot is walking over rigid terrain in which the 
	stance feet do 
	not move (i.e., $v_{\mrm{st}}= 
	\dot{v}_{\mrm{st}} = 0$).
	\label{ass1}
\end{assumption}

\subsection{Trunk and Swing Leg Control Tasks}
\label{secTrunkTask}
To compliantly achieve a desired motion of the \textit{trunk}, 
we define the desired wrench at the CoM  $W_{\mrm{com,d}}$ 
using the following: 
1) a Cartesian impedance at the \gls{com} $W_\mrm{imp}$ that is represented by
a stiffness term  ($\nabla V_{\mrm{com,K}} = K_{\mrm{com}}\Delta x_{\mrm{com}}$)  with positive definite stiffness 
matrix $K_{\mrm{com}}\in R^{6\times6}$
and a damping term ($D_{\mrm{com}} \Delta v_{\mrm{com}}$) with positive definite damping matrix $D_{\mrm{com}} \in R^{6\times6}$,
2) a virtual gravitational potential gradient to render  gravity compensation $(\nabla V_{\mrm{com,\bar{g}}} =  
mg )$\footnote{$\nabla V_{[.]}$ denotes the gradient of a potential function $V_{[.]}$. For more information 
regarding the Cartesian stiffness and gravitational potentials, see
\cite{Henze2016}.},
3) a feedforward term to improve tracking ($W_\mrm{ff} = M_{\mrm{com}} \dot{v}_\mrm{com,d}$) and a compensation term for external 
disturbances  
$-W_\mrm{ext}$ 
\cite{Focchi2018}:
\begin{subequations}
	\begin{eqnarray}
	W_{\mrm{com,d}} &=& 	W_\mrm{imp} + \nabla V_{\mrm{com,\bar{g}}} + W_\mrm{ff}   - W_\mrm{ext}
	\label{desiredCoMwrench}\\
	W_\mrm{imp} &=& \nabla V_{\mrm{com,K}}   + D_{\mrm{com}} \Delta v_{\mrm{com}}  
	\end{eqnarray}
\end{subequations}
such that $\Delta x_{\mrm{com}} = x_{\mrm{com,d}} - x_{\mrm{com}} $, $\Delta v_{\mrm{com}} = v_{\mrm{com,d}} - 
v_{\mrm{com}}$ are the tracking 
errors  $\in \Rnum^6$ of the position and velocity, respectively.

Similarly, the tracking of the swing task is obtained by the virtual force $F_{\mrm{sw,d}} \in \Rnum^{n_{\mrm{sw}}}$.
This is generated by 
1) a Cartesian impedance at the swing foot that is represented by
a stiffness term  ($\nabla V_{\mrm{sw}} = K_{\mrm{sw}}\Delta x_{\mrm{sw}}$)  with positive definite stiffness 
matrix  $K_{\mrm{sw}} \in R^{n_{\mrm{sw}} \times n_{\mrm{sw}}}$
and a damping term ($D_{\mrm{sw}} \Delta v_{\mrm{sw}}$) with positive definite damping matrix $D_{\mrm{sw}} \in 
R^{n_{\mrm{sw}} \times n_{\mrm{sw}}}$, and 2) a feedforward term to improve tracking ($F_{\mrm{sw,ff}} = 
M_{\mrm{sw}}\dot{v}_{\mrm{sw,ff}}$):
\begin{eqnarray}
	F_{\mrm{sw,d}} &=&  \nabla V_{\mrm{sw}} + D_{\mrm{sw}} \Delta v_{\mrm{sw}} + F_{\mrm{sw,ff}} 
\label{eq:swingTask}
\end{eqnarray}
such that $\Delta x_{\mrm{sw}} = x_{\mrm{sw,d}} - x_{\mrm{sw}} $ and $\Delta v_{\mrm{sw}} = 
v_{\mrm{sw,d}} - 
v_{\mrm{sw}}$ are the tracking errors of the swing feet positions and velocities 
respectively. 
Alternatively, it is possible to write this task at the acceleration level, 
with the difference that the gains $\mrm{K}_{\mathrm{sw}}$ and $\mrm{D}_{\mathrm{sw}}$ have no physical meaning:
\begin{eqnarray}
	\dot{\vc{v}}_{\mathrm{sw,d}} = \dot{\vc{v}}_{\mathrm{sw,ff}} +
								\mrm{K}_{\mathrm{sw}} \Delta x_{\mrm{sw}} +
								\mrm{D}_{\mathrm{sw}} \Delta v_{\mrm{sw}} 
\label{eq:swingTaskAcc}
\end{eqnarray}
\subsection{Optimization}
\label{sec:optimization}
To fulfill the motion tasks in \sref{secTrunkTask} and to distribute 
the load on the stance feet, while respecting the mentioned constraints,
we formulate the \gls{qp}: 
\begin{subequations}
	\begin{eqnarray} 
	& \underset{\vc{u}=[\ddot{\vc{q}}^T \hspace{0.2cm} \vc{\grf}^T]^T}{\min} 	\Vert W_\mrm{com} - W_\mrm{com,d}  
	\Vert^2_Q+ 
	\Vert\vc{u}\Vert^2_{R} \label{cost}\\
  &	\hspace{-2cm}  \text{s.t.} \hspace{0.5cm} \quad \vc{Au} = \vc{b} \label{eqCon}\\
	 &\hspace{-0.6cm} \underline{\vc{d}} < \vc{Cu} < \bar{\vc{d}} \label{ineqCon}
\end{eqnarray}
\label{eq:qp_problem}
\end{subequations}
such that our decision variables  $\vc{u}=[\ddot{\vc{q}}^T \  \vc{\grf}^T]^T \in\Rnum^{6+n+n_{\mrm{st}}}$ 
are the generalized 
accelerations $\ddot{\vc{q}}$ and the contact forces 
$\grf$. 
The cost function \eref{cost} is designed to minimize the \textit{trunk task} and to regularize the solution.
The equality constraints \eref{eqCon} 
encode dynamic consistency, stance constraints and swing tasks. 
The inequality constraints \eref{ineqCon} 
encode
friction constraints, joint kinematic and torque limits. 
All constraints 
are stacked
in the matrix $\vc{A}^T=\mat{\vc{A}_{p}^T & \vc{A}_{\mrm{st}}^T & \vc{A}_{\mrm{sw}}^T}$ and
$\vc{C}^T=\mat{\vc{C}_{\mrm{fr}}^T & \vc{C}_j^T &\vc{C}^T_{\tau}}$ and detailed in
the following sections.
	
\subsubsection{Cost}
The first term of the cost in \eref{cost}
represents the \textit{tracking} error 
between the actual $W_\mrm{com}$  and  the desired $W_\mrm{com,d}$ \gls{com} wrenches from 
\eref{eqn4} and \eref{desiredCoMwrench} respectively.
Since  $W_\mrm{com}$ 
is not a decision
variable, we compute it from the contact forces (see \eref{eqn4}) and
re-write $\|W_{\mrm{com}}-W_{\mrm{com,d}}\|_{\vc{Q}}^2$
in the form of $\|\vc{Gu}-\vc{g}_0\|_{\vc{Q}}^2$ with:
%
	\begin{eqnarray}
G  = \mat{0_{6 \times (6+n)} & J^T_{\mrm{st,com}}}, \quad g_0 =  W_{\mrm{com,d}}
\end{eqnarray}

\subsubsection{Physical consistency}
\label{sec:physical_contraints}
To enforce physical  consistency between $\grf$ and $\ddot{q}$, we impose the dynamics of the unactuated part of the 
robot
(the trunk dynamics in \eref{eqn4}) as an equality constraint:
\begin{equation}
\vc{A}_p = \mat{\vc{M}_{\mrm{com}}  &0_{6 \times n} & -J^T_{\mrm{st,com}}},\quad
\vc{b}_p = - h_{com}
\label{eq:physicalEq}
\end{equation}
%
%
%
%

\subsubsection{Stance condition}
\label{sec:stance_contraints}
We can encode the stance feet constraints by re-writing them at the acceleration level in order to be compatible with 
the decision variables. Since $v_{\mrm{st}} = J_{\mrm{st}} \dot{q}$, differentiating once in time, yields to
$\dot{v}_{\mrm{st}} = \vc{J}_{\mrm{st}}\ddot{\vc{q}}+\dot{\vc{J}}_{\mrm{st}}\dot{\vc{q}}$. 
Recalling \assref{ass1} yields $ \vc{J}_{\mrm{st}}\ddot{\vc{q}}+\dot{\vc{J}}_{\mrm{st}}\dot{\vc{q}} = 0$ which is 
encoded as:
%
\begin{equation}
\vc{A}_{\mrm{st}} = \mat{\vc{J}_{\mrm{st}}  & \vc{0}_{n_{\mrm{st}} \times n_{\mrm{st}} }}, \quad
\vc{b}_{\mrm{st}} = -\dot{\vc{J}}_{\mrm{st}}\dot{\vc{q}}
\label{eq:stanceEq}
\end{equation}
such that $\dot{\vc{J}}_{\mrm{st}}$ is the time derivative of $J_{\mrm{st}}$.
For numerical precision, we
compute the product $\dot{\vc{J}}_{\mrm{st}}\dot{\vc{q}}$ using spatial algebra.
\subsubsection{Swing task}
\label{sec:swingTask}
Similar to \sref{sec:stance_contraints}, we can encode the swing task directly as an
\textit{equality constraint}, i.e. by enforcing the swing feet to
follow a desired swing acceleration
$\dot{\vc{v}}_{\mrm{sw}}(\vc{q})=\dot{v}_{\mrm{sw,d}}\in\Rnum^{n_{\mrm{sw}}}$ 
yielding:
\begin{equation}
	\vc{J}_{\mrm{sw}}\ddot{\vc{q}} + \dot{\vc{J}}_{\mrm{sw}}\dot{\vc{q}} =
	\dot{\vc{v}}_{{\mrm{sw,d}}}
\end{equation}
that in matrix form becomes\footnote{Alternatively, it is possible to write the
swing task at the joint space rather than in the operational space by changing
the matrix $\vc{A}_{\mrm{sw}}, \vc{b}_{\mrm{sw}}.$}:
\begin{equation}
	\vc{A}_{\mrm{sw}} = \mat{\vc{J}_{\mrm{sw}}  & \vc{0}_{n_{\mrm{sw}}\times n_{\mrm{st}}}}, \quad
	\vc{b}_{\mrm{sw}} = \dot{\vc{v}}_{{\mrm{sw,d}}} -\dot{\vc{J}}_{\mrm{sw}}\dot{\vc{q}}
\label{eq:swingEq}
\end{equation}
%
%
%
%
where we computed $\dot{\vc{v}}_{{\mrm{sw,d}}}$  as in \eqref{eq:swingTaskAcc}.
Note that this implementation is analogous to the trunk task in \sref{secTrunkTask}. 
The difference is that this implementation is at the 
acceleration level while the other is at the force level.
However, without any loss of generality, 
the formulation  \eqref{eq:swingTask} could also be used. 
In Section \ref{sec:slacks}
we incorporate slacks in the optimization to allow temporary violation of the swing
tasks (e.g. useful when the kinematic limits are reached).
\subsubsection{Friction cone constraints}
\label{sec:friction_constraints}
To avoid slippage and obtain a smooth
loading/unloading of the legs, we incorporate friction/unilaterality constraints. For that, we ensure that the contact forces lie
inside the friction cones and their normal components stay within some
user-defined values (i.e. maximum and minimum force magnitudes). We approximate
the friction cones with square pyramids to express them with linear constraints.
The fact that the ground contacts are unilateral, 
can be naturally encoded by setting an  ``almost-zero''
lower bound on the normal component, 
while the upper bound allows us to regulate
the amount of ``loading'' for each leg. We define the friction inequality constraints as:
\begin{align}
\underline{\vc{d}}_{\mrm{fr}} < \vc{C}_{\mrm{fr}}\vc{u} < \bar{\vc{d}}_{\mrm{fr}}, \quad
\vc{C}_{\mrm{fr}} =
\mat{\vc{0}_{p\times(6+n)} & \vc{F}_{\mrm{fr}}}
\label{eq:frictionIneq}
\end{align}
with:
\begin{equation}
\vc{F}_{\mrm{fr}} = \mat{\vc{F}_0 & \dots & \vc{0} \\
	\vdots & \ddots & \vdots \\
	\vc{0} & \dots & \vc{F}_c},  \ \ \
\underline{\vc{d}}_{\mrm{fr}} = \mat{\underline{\vc{f}}_0 \\
	\vdots \\
	\underline{\vc{f}}_c}, \ \ \
\bar{\vc{d}}_{\mrm{fr}} = \mat{\bar{\vc{f}}_0 \\
	\vdots \\
	\bar{\vc{f}}_c}
\end{equation}
where $\vc{F}_{\mrm{fr}}\in\Rnum^{p\times n_\mrm{st}}$ is a block diagonal 
matrix that encodes the friction cone
boundaries and select the normals, for each stance leg and
$\underline{\vc{d}}_{\mrm{fr}},\bar{\vc{d}}_{\mrm{fr}}\in\Rnum^{p}$ are the lower/upper
bounds respectively. For the detailed  implementation of the friction constraints refer to \cite{Focchi2017}.

\subsubsection{Torque limits}
\label{sec:torque_constraints}
We notice that the torques be obtained from the decision
variables since they can be expressed as a bi-linear function
of $\ddot{q}_j$ and $\grf$. 
Therefore, the constraint on the joint torques (i.e., the actuation limits $\vc{\tau}_{\min} < \vc{\tau}_j <
\vc{\tau}_{\max}$) can be encoded by exploiting the actuated part of the dynamics
\eref{eq:full_dynamicsCOM}:
\begin{eqnarray}
	\underline{\vc{d}}_{\tau} < \vc{C}_{\tau}\vc{u} < \bar{\vc{d}}_{\tau},\quad
	\vc{C}_{\tau} = \mat{0_{n \times 6} & \vc{\bar{M}}_j & - \vc{J}_{\mrm{st,j}}^T}\\
	\underline{\vc{d}}_{\vc{\tau}} = - \bar{\vc{h}}_j + \vc{\tau}_{\min}(\vc{q}_j), \quad 
	\bar{\vc{d}}_{\vc{\tau}} = - \bar{\vc{h}}_j +  \vc{\tau}_{\max}(\vc{q}_j) \nonumber
	\label{eq:torqueIneq}
\end{eqnarray}

where $\vc{\tau}_{\min}(\vc{q}_j), \vc{\tau}_{\max}(\vc{q}_j)\in\Rnum^n$ are
the lower/upper bounds on the torques. In the case of our quadruped robot, 
these bounds must be recomputed at each
control loop because they  depend on the joint positions. This is due to
the presence of linkages on the sagittal joints (HFE and KFE), 
that set a joint-dependent profile on the maximum torque
 (non-linear in the joint range).

\subsubsection{Joint kinematic limits}
\label{sec:kinematic_constraints}
We enforce joint kinematic constraints as function of the joint accelerations
(i.e. $\ddot{\vc{q}}_{j_{\min}} < \ddot{\vc{q}}_j < \ddot{\vc{q}}_{j_{\max}}$).
We select them via the matrix $\vc{C}_j$:
\begin{eqnarray}
	&\underline{\vc{d}}_j < \vc{C}_j \vc{u} < \bar{\vc{d}}_j,\quad
	{C}_j = \mat{\vc{0}_{n\times 6}  &  \vc{I}_{n\times n}  &  \vc{0}_{n
		\times n_{\mrm{st}}}}\\
	& \underline{\vc{d}}_j =  \ddot{\vc{q}}_{j_{\min}}(\vc{q}_j), \quad
	\quad \bar{\vc{d}}_j = \ddot{\vc{q}}_{j_{\max}}(\vc{q}_j)
	\label{eq:jointKinIneq} 
\end{eqnarray}
such that $\ddot{\vc{q}}_{j_{\min}}(\vc{q}_j)$  and $\ddot{\vc{q}}_{j_{\max}}(\vc{q}_j)$ are the upper/lower bounds on 
accelerations. These bounds 
should be recomputed at each control
loop. They are set  in order to make the joint to reach the end-stop at a zero
velocity in a time interval $\Delta t = 10dt$, where $dt$ is the loop
duration. For instance, if the joint is at a distance $ 
\vc{q}_{j_{\max}}-\vc{q}_j$ from the end-stop with a velocity $\dot{\vc{q}}_j$, the
deceleration to cover this distance in a time interval  $\Delta t$, and approach
the end-stop with zero velocity, will be:
\begin{equation}
\ddot{\vc{q}}_{j_{\min,\max}} = -\frac{2}{\Delta t^2}(\vc{q}_{j_{\min,\max}} -
\vc{q}_j - \Delta t\, \dot{\vc{q}}_j)
\label{eq:accellBounds}
\end{equation}

\subsection{Torque computation}
\label{secTorqueComp}
We map the optimal solution  $\vc{u}^* = \mat{\ddot{\vc{q}}^* & 
\vc{\grf}^*}$ obtained by solving \eref{eq:qp_problem}, 
into desired  joint torques
$\vc{\tau}_{d}^{*} \in\mathbb{R}^n$ using the actuated part of the 
dynamics equation of the robot as:
\begin{equation}
\vc{\tau}_{d}^{*} = \bar{M}_j\ddot{\vc{q}_j}^* + \bar{\vc{h}}_j -
\vc{J}_{\mrm{st,j}}^T\vc{\grf}^*
\label{eq:torques}
\end{equation}
\section{Passivity Analysis}  
\label{sec:Passivity}    
The overall system consists of the \gls{wbc}, the robot and 
the environment.
This system is said to be \textit{passive} if all these components, and their interconnections are passive 
\cite{Stramigioli2015}.
If the robot and the environment are passive, and the controller is proven to be passive, then the overall 
system is passive \cite{Schaft}. 
A system (with input $u$ and output $y$) is said to be passive if there exists a storage function $S$ that is bounded 
from below and its derivative $\dot{S}$ is less than or equal to its supply rate ($s=y^Tu$). 
In this context, we define the total energy stored in the controller to be the candidate storage function for the 
controller $S=V$.
The rest of this section is devoted to analyze the passivity of the overall system. 
\begin{assumption}
	 A feasible solution exists for the \gls{qp} in \eref{eq:qp_problem} 
	 in which the motion 
	 tasks are achieved. Moreover, we do not consider the feed-forward terms in \eref{desiredCoMwrench} and \eref{eq:swingTask}  leaving 
	 this to future developments.   
	 \label{ass2}
\end{assumption} 

We start by defining the velocity error at the joints and at the stance feet to be $\Delta \dot{q}_j= 
\dot{q}_{j,d} - \dot{q}_\nmrm{j}$, and $\Delta v_{\mrm{st}} = v_{\mrm{st,d}}-v_{\mrm{st}}$, respectively. 
We also define the  desired feet forces $F_\mrm{d}= [F_{\mrm{st,d}}^T \ F_{\mrm{sw,d}}^T]^T$ such that, 
by following the sign convention in \fref{fig:blockDiagram}, 
the mapping between $F_\mrm{d}$ and $W_\mrm{com,d}$
is expressed as%
\footnote{Since we are analyzing the passivity of the controller,
we are interested in the forces exerted by the robot on the environment rather
than the forces exerted by the environment on the robot. Hence the mapping in \eref{sign} is  
negative.}:
\begin{equation}
W_\mrm{com,d} =  - J_\mrm{com}^T F_\mrm{d} \label{sign},  \end{equation}
while mapping between $F_d$ and $\tau_j$   is expressed as\footnote{Assuming a perfect low level torque control 
tracking (i.e., $\tau_\mrm{d} = \tau$).}
\begin{eqnarray}
\tau &=&  J_{\nmrm{j}}^T  F_d \label{torquemapp} .\label{pass6}
\end{eqnarray}
By defining $\nabla V_\mrm{com} = \nabla V_\mrm{com,K} + \nabla V_\mrm{com,\bar{g}}$ and recalling \eref{sign}, we 
rewrite  
\eref{desiredCoMwrench} under  \assref{ass2} as: 
\begin{subequations}
	\begin{eqnarray}
	\nabla V_{\mrm{com}} &=& 	W_{\mrm{com,d}}  - D_{\mrm{com}} \Delta v_{\mrm{com}} \label{pass9b}\\
	&=& - J_{\mrm{com}}^{T} F_\mrm{d} - D_{\mrm{com}}  \Delta v_{\mrm{com}} \label{pass9c}.
	\end{eqnarray}
\end{subequations}
\subsection{Analysis}
The overall energy in the whole-body controller is the one stored in the virtual impedance 
at the CoM and the potential energy due to gravity compensation ($V_{\mrm{com}}$),
and the energy stored in the virtual impedances at the swing feet ($ V_{\mrm{sw}}$)\footnote{In this analysis we use the formulation \eref{eq:swingTask}.}: 
\begin{equation}
V =  V_{\mrm{com}} + V_{\mrm{sw}} .
\end{equation}
The time derivatives are\footnote{The time derivative of an arbitrary storage function $\dot{V}(\Delta x(t))$ that is a 
function of $\Delta x(t)$ could be written as $\dot{V} = \frac{d}{dt} \Delta x^T(t) \cdot \frac{\partial}{ \partial 
\Delta x(t)} V$ that is written for short as $\dot{V} = \Delta v^T \nabla V $.}:
\begin{equation}
\dot{V} = \dot{V}_{\mrm{com}} + \dot{V}_{\mrm{sw}} = \Delta v_\mrm{com}^T  \nabla V_{\mrm{com}}  + \Delta 
v_\mrm{sw}^T \nabla V_{\mrm{sw}}
\label{vdot}.
\end{equation}
Recalling \eref{eq:swingTask} and \eref{pass9c}, \eref{vdot} yields:
\begin{eqnarray}
\dot{V} &=& \Delta v_\mrm{com}^T (-J_{\mrm{com}}^{T} F_\mrm{d} - D_\mrm{com} \Delta v_\mrm{com}) + 
\nonumber\\
&~& \Delta v_\mrm{sw}^T (F_\mrm{sw,d} - D_\mrm{sw} \Delta v_\mrm{sw}) ) .
\end{eqnarray}
We regroup $\dot{V}$ in terms of the non-damping terms $\dot{V}_1$ and damping terms  $\dot{V}_2$ yielding:
\begin{subequations}
	\begin{eqnarray}
	\dot{V_1} &=& - \Delta v_\mrm{com}^TJ_{\mrm{com}}^{T} F_\mrm{d}  + \Delta v_\mrm{sw}^T 
	F_\mrm{sw,d} \label{pass12aa} \\
	\dot{V_2} &=& - \Delta v_\mrm{com}^T D_\mrm{com} \Delta v_\mrm{com} - \Delta v_\mrm{sw}^T 
	D_\mrm{sw} \Delta v_\mrm{sw} \label{pass12aaa}.
	\end{eqnarray}
\end{subequations}
We rewrite \eref{eq_pass4} in terms of $\Delta v_\mrm{com}$, $\Delta v$ and $\Delta \dot{q}_\mrm{j}$ as:
\begin{subequations}
	\begin{eqnarray}
	\Delta	v &=&  J_{\mrm{com}} \Delta v_{\mrm{com}} +   J_{\mrm{j}} \Delta \dot{q}_\mrm{j}  \\
	\Delta v_{\mrm{com}}^T  J_{\mrm{com}}^T &=& - \Delta \dot{q}_\mrm{j}^T J_{\mrm{j}}^T + \Delta v^T.
	\label{pass14c}
	\end{eqnarray}
\end{subequations}
Plugging, \eref{pass14c} in \eref{pass12aa} yields\footnote{From the definition of  $v$ and $F_d$, we get $v^TF_d = 
v_{\mrm{st}}^T F_{\mrm{st,d}} +  v_{\mrm{sw}}^T 
	F_{\mrm{sw,d}} $.}:
\begin{subequations}
	\begin{eqnarray}
\dot{V_1} &=&  \Delta \dot{q}_\mrm{j}^T J_{\mrm{j}}^T F_\mrm{d} - \Delta v^T F_\mrm{d}  + \Delta 
v_\mrm{sw}^T F_\mrm{sw,d} \\
&=& \Delta \dot{q}_\mrm{j}^T J_{\mrm{j}}^T F_\mrm{d}   - \Delta v_\mrm{st}^T F_\mrm{st,d} \label{pass15a} .
\end{eqnarray}
\end{subequations}
Plugging \eref{pass6} and into \eref{pass15a} yields%
:
\begin{eqnarray}
\dot{V_1} &=&  \Delta \dot{q}_\mrm{j}^T \tau  - \Delta v_\mrm{st}^T F_\mrm{st,d} \label{pass16a} .
\end{eqnarray}
Under  \assref{ass1}, \eref{pass16a} yields:
\begin{eqnarray}
\dot{V_1} &=&   \Delta \dot{q}_\mrm{j}^T \tau \label{pass17a} .
\end{eqnarray}
Thus, $\dot{V}$ could be rewritten as:
\begin{eqnarray}
\dot{V} =  \Delta \dot{q}_\mrm{j}^T \tau - \Delta v_{\mrm{com}}^T D_{\mrm{com}}\Delta v_{\mrm{com}}    - \Delta 
v_{\mrm{sw}} ^T D_{\mrm{sw}}\Delta v_{\mrm{sw}}.
\label{pass92}
\end{eqnarray}

\subsection{Proof}
Under \assref{ass2}, the designed \gls{wbc} is an impedance control with 
gravity compensation 
that, similar to a PD+ \cite{Ortega2013}, defines a map of $( \dot{q}_\mrm{j} -
\dot{q}_\mrm{j,d}) \mapsto - \tau$\footnote{Note that  $ \dot{q}_\mrm{j} - \dot{q}_\mrm{j,d}  = -\Delta \dot{q}_\mrm{j}$.
	Thus, the controller with the map $( \dot{q}_\mrm{j} - \dot{q}_\mrm{j,d}) \mapsto - \tau $ 
	has a supply rate of $ \Delta \dot{q}_\mrm{j}^T \tau$.}. 
This controller is passive if $V$ is bounded from below and $\dot{V}\leq   \Delta \dot{q}_\mrm{j}^T \tau$. 
Since $V$ consists of positive definite potentials that resemble Cartesian 
stiffnesses at the CoM and the swing leg(s), and under the assumption
 the gravitational potential is bounded from below (see \cite{Ott2008}), 
 $V$ is also bounded from below. Additionally, recalling \eref{pass92} 
 proves that the controller is indeed passive; thus, the overall system is passive. 
\section{Implementation Details}
\label{sec:Implementations}
This section  describes some pragmatic details 
that we found crucial in the implementation on the real platform.

\subsection{Stance task}
\label{sec:stanceAccel}
Uncertainties in estimating the terrain's normal
direction and friction coefficient could result in slippage. This can lead 
to considerable  motion of the stance feet with  possible loss of stability.
To avoid this, a joint impedance feedback loop could be run in parallel to the \gls{wbc},
at the price of losing the capability of optimizing the \gls{grfs}.
A cleaner solution is to incorporate, in the optimization,   Cartesian impedances specifically designed
to keep the relative distance among the stance feet constant (we denote it \textit{stance task}).

\textit{The stance task} has an influence \textit{only} when there is 
an anomalous motion in the stance feet, retaining
the possibility to freely optimize for \gls{grfs} in normal situations.
This can be achieved by re-formulating the stance
condition in \eref{eq:stanceEq} as a desired 
stance feet acceleration   $\dot{v}_{\mrm{st,d}}$ as:
\begin{equation}
\dot{v}_{\mrm{st,d}}  =   K_{\mrm{st}}(\hat{x}_{\mrm{st}} - x_{\mrm{st}}) - D_\mrm{st}v_{\mrm{st}}
\label{eq:stanceAccelerations}.
\end{equation} 

This term is added to $b_{st}$ (in \eref{eq:stanceEq}) as $b_{st} = -\dot{J}_{st}\dot{q} + 
\ddot{x}^d_{f_{st}} $ 
%
%
where  $\hat{x}_{\mrm{st}}$ is a sample of the foot 
position at the touchdown (expressed in the world frame).

\subsection{Constraint Softening}
\label{sec:slacks}
Adding  slack variables to an optimization problem
is commonly done to avoid infeasible solutions,
by allowing a certain degree of constraint violation. 
Infeasibility can occur when hard constraints are conflicting with each other,
which can be the case in our \gls{qp}. 
Hence, some of the equalities/inequalities in \eref{eq:qp_problem}
 should be relaxed if they are in conflict.

We decided not to introduce slacks in the torque constraints (\sref{sec:torque_constraints})
or the dynamics   (\sref{sec:physical_contraints}) keeping them as hard 
constraints, since torque constraints and physical consistency should never be violated. 
%
On the other hand  it is important to allow a certain level
of relaxation for the swing tasks in \sref{sec:swingTask} 
that could be violated if the joint kinematic limits are reached%
\footnote{Using slacks in friction constraints did not result in  significant improvements.}.

To relax the constraints of the swing task, 
first, we augment the decision variables 
$\vc{\tilde{u}} = [\vc{u}^T \ \epsilon^T]^T \in \Rnum^{6+n+k+n_{sw}}$  
with the vector of slack variables  
$\epsilon \in \Rnum^{n_{sw}}$ where we introduce a slack variable  for each direction of the swing tasks. 
Then, we replace the   equality constraint 
$\vc{A}_{sw} u + \vc{b}_{sw}=0$ of the swing tasks,  by two inequality 
constraints: 
%
\begin{align}
\label{eq:slackIneq}\nonumber
-\epsilon \leq \vc{A}_{sw} \tilde{u} &+ \vc{b}_{sw} \leq \epsilon\\
\epsilon & \geq 0.
\end{align}
%
The first inequality in \eref{eq:slackIneq} restricts  the solution to a bounded region around  the original 
constraint while the second one ensures that the slack variables remain non-negative. 
To make sure that there is  a constraint violation only when  the constraints are
conflicting, we minimize the norm of the slack vector adding a 
regularization term $\alpha\Vert \epsilon \Vert^2$ to \eqref{cost}
with a high weight $\alpha$.

To reduce the computational complexity, 
we could have introduced a single slack for each swing task (rather than 
one for each direction).
However, this could create coupling errors in the tracking. 
For instance, since the \gls{hfe} joint (see \fref{fig:blockDiagram}) mainly acts in the $XZ$ plane, 
if it reaches its joint limit, only that plane should be affected leaving the $Y$ direction unaffected.
%
A single slack couples the three directions causing tracking errors also in the $Y$ direction.
%
Conversely, using multiple slacks,  only the directions  the \gls{hfe} acts upon, will be affected.

\section{Results}
\label{sec:expResults}
In this section we validate the capabilities of the
controller under various terrain conditions and locomotion gaits.
The \gls{wbc} and torque control loops run
in real-time threads at \unit[250]{Hz} and \unit[1]{kHz}, respectively.
We set the gains for the swing tasks 
to $\vc{K}_{sw} = \mrm{diag}({2000,2000,2000})$ and $\vc{D}_{sw} =
\mrm{diag}({20,20,20})$,
while for the trunk task we set $\vc{K}_{x} = \mrm{diag}({2000,2000,2000})$
$\vc{D}_{x} = \mrm{diag}({400,400,400})$ and $\vc{K}_{\theta} = \mrm{diag}({1000,1000,1000})$
$\vc{D}_{\theta} = \mrm{diag}({200,200,200})$. These values proved to be working
in both simulation and real experiments. The results are collected in the accompanying 
video\footnote{Link: \href{https://youtu.be/Lg3V_juoE1w}{\texttt{https://youtu.be/Lg3V\_juoE1w}}}.
Additionally, in experimental trials, we also included a low gain joint-space attractor
(PD controller) for the swing task, since imprecise torque tracking of the 
knee joints (due to the low inertia) produce control instabilities in an
operational space implementation (e.g. the one in Section \ref{sec:swingTask}).
\subsection{Constraint Softening through Slack Variables}
\begin{figure}[tb]
\centering
\includegraphics[width=\columnwidth]{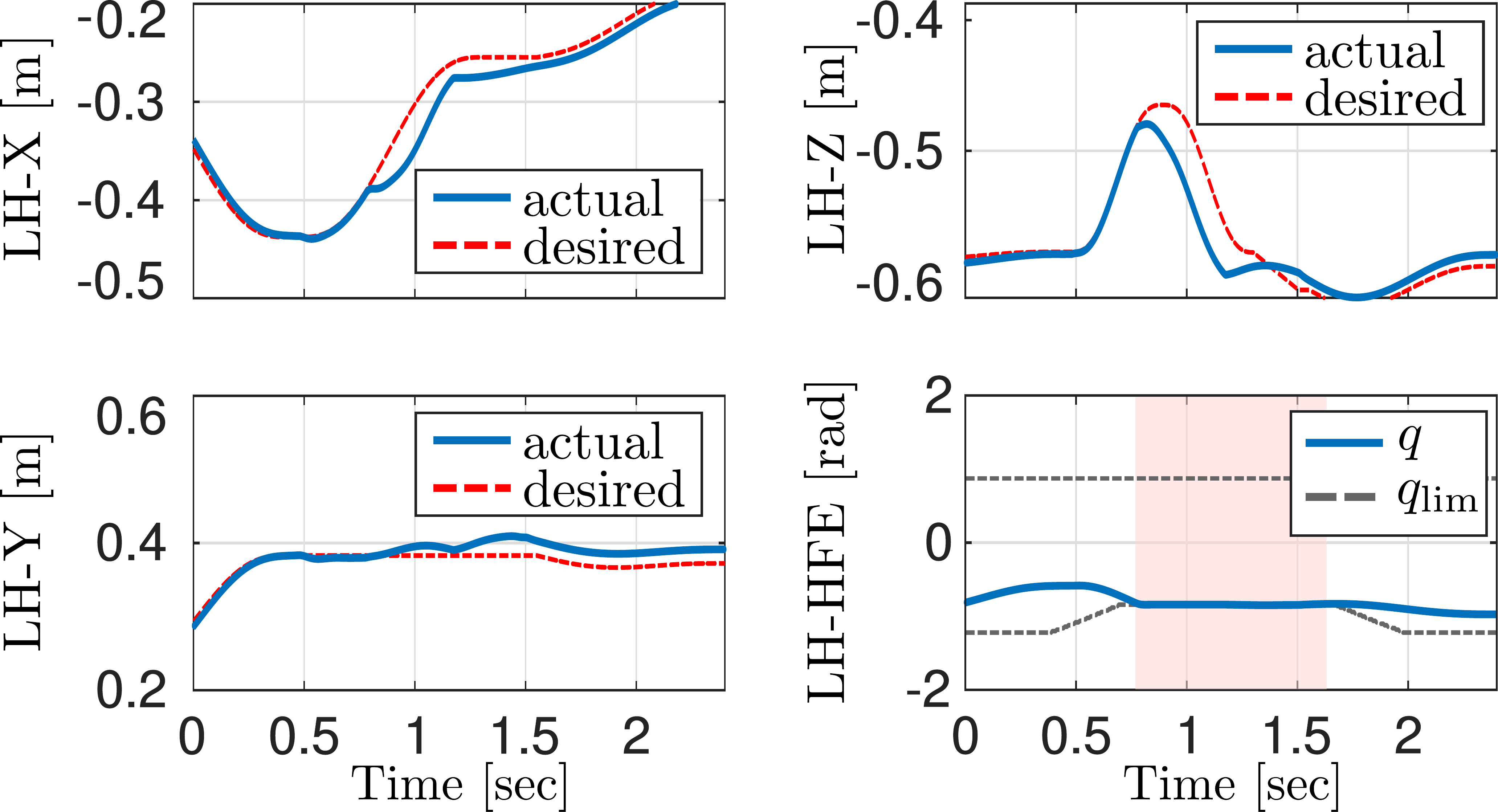}
\caption
[Effect of (kinematic limits) slacks variables on foot tracking.]
{ Simulation. Effect of (kinematic limits) slacks variables on foot tracking. The
upper-left/right and bottom-left plots show the tracking of the desired foot
position (\acrshort{lh} leg) in $X$, $Y$ and $Z$, respectively. Bottom-right
plot depicts the joint limits (black line) and the actual position (blue line) of the \acrshort{hfe} joint.
The red shaded area underlines that when the slack increases, the \acrshort{hfe} joint is properly clamped.}
	\label{fig:effectOfSlacks}
\end{figure} 
In \fref{fig:effectOfSlacks} we artificially incremented the lower limit of the
\acrshort{lh}-\gls{hfe} joint.
When the limit is hit, the bound on the joint acceleration
\eref{eq:accellBounds} produces a desired torque command that stops its motion.
This ``naturally'' clamps the actual joint position to the limit (bottom-left
plot) and   influences the  foot tracking mainly  along the $X$ and $Z$ directions.

\textit{Computational time:} the solution of the \gls{qp}
takes between \unit[90-110]{$\mu s$}  on a Intel  i5 machine without the
slacks variables.
After augmenting the problem with the slack variables and its constraints, 
it increases   \unit[30]{\%} on average (\unit[120-150]{$\mu s$}). However it still
remains suitable for real-time implementation (250 $Hz$).
\subsection{Friction Constraints and Bounded Slippage}
\label{sec:frictionTests}
We evaluated in simulation the controller performance against inaccurate 
friction coefficient estimates $\mu$, which define incorrectly the friction cone
constraints in the \gls{wbc}.
In the accompanying video, we show an example where the robot crawls at
\unit[0.11]{m/s} on a slippery floor ($\mu=0.4$) while we set the friction
coefficient to $\mu=1.0$ in the \gls{wbc}, to emulate an estimation error. 
Simulation results support the fact that foot slippage remains bounded
by the action of the \textit{stance task} (\sref{sec:stanceAccel}).
If we gradually correct $\mu$ the slippage events
completely disappear; allowing an increase of forward velocity up to \unit[0.16]{m/s}.
\subsection{Torque Limits and Load Redistribution}
\label{sec:expTorqueLim}
\begin{figure}[tb]
\centering
\includegraphics[width=\columnwidth]{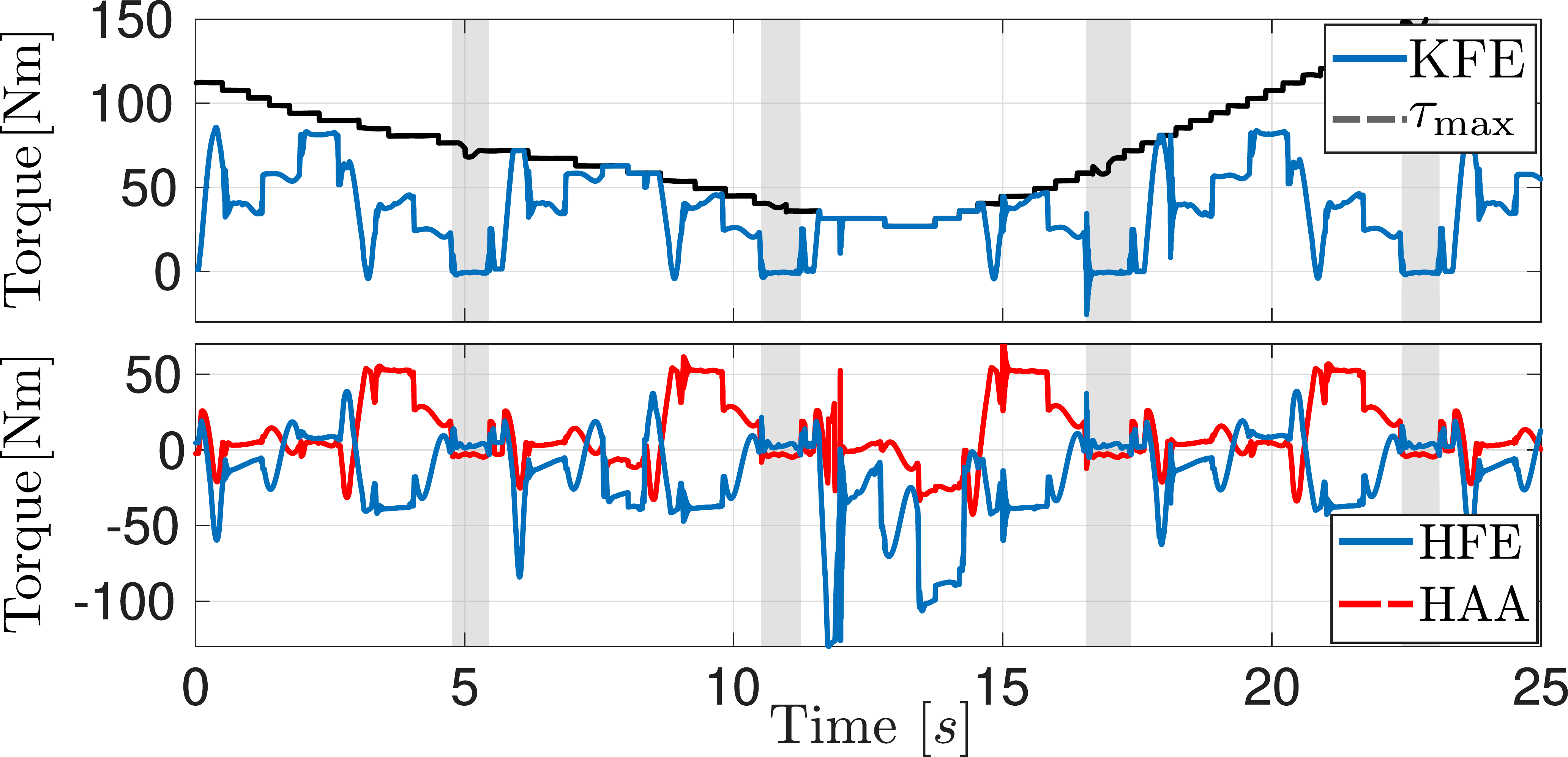}
\caption
[Effect of  introducing an artificial torque limit on the \acrshort{kfe} 
joint of the LF leg during a typical crawl.]
{Simulation. Effect of  introducing an artificial torque limit on the \acrshort{kfe} 
joint of the LF leg during a typical crawl. 
The shaded area represents the swing phase of the leg while the unshaded part is the stance 
phase. 
 We reduced the torque limit down to \unit[26]{Nm}
(black line, upper plot), and as a consequence, the
\acrshort{hfe} torque is increased by the controller (bottom plot).}
\label{fig:torqueLimits}
\end{figure}

\begin{figure}[tb]
	\centering
	\includegraphics[width=\columnwidth]{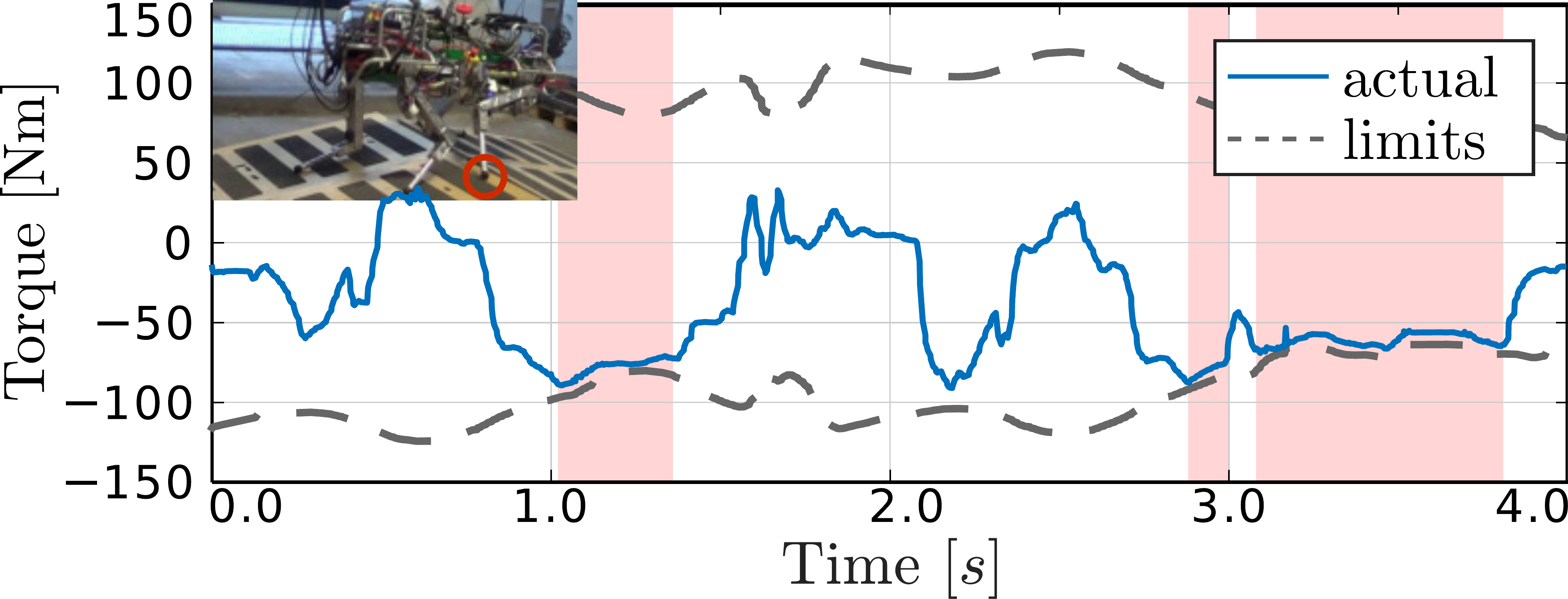}
	\caption[Reaching the torque limits on the RF-HFE  joint
	while climbing up and down two ramps.]{Real experiments. Reaching the torque limits on the RF-HFE  joint
	while climbing up and down two ramps. The \acrshort{hyq} robot reached three times its
	torque limits (red shaded areas). The real torque (blue line) is tracking
	the desired one (not shown) computed from our whole-body controller while satisfying the
	joint limits (black line). The torque limits are time-varying due to the
	joint mechanism.}
	\label{fig:reachingLimits}
\end{figure}

We analyzed in simulation the effect of adding  an artificial torque limit,  in our \gls{wbc}. 
This helps us to derive controllers that are robust against joint damages.
Figure \ref{fig:torqueLimits} shows a reduction of the torque limits down to
\unit[26]{Nm} in the \gls{lf}-\acrshort{kfe} joint and the load redistribution among
the other joints (\acrshort{haa} and \acrshort{hfe}) of the LF leg.
Indeed, while the \acrshort{kfe} joint torque is clamped, the \gls{hfe} is
loaded more (lower plot). This load redistribution did not affect the trunk 
motion and it demonstrates how the controller exploits the torque redundancy by
finding a new load distribution.
%
We carried out also intensive experimental validation in various challenging
terrains. Slopes increase the probability of reaching torque limits because 
of the more demanding kinematic configurations. Indeed, in
 \fref{fig:reachingLimits}, the robot reached three times the torque limits (red
shaded areas). Crossing this terrain would not be possible without enforcing
the torque limits as hard constraints.

\subsection{Different Torque Regularization Schemes}
\label{sec:torqueReg}
By setting different regularizations in  \eref{cost} 
\cite{Focchi2017}, we can either choose to 
maximize the robustness to uncertainties in the friction
parameters (e.g. \gls{grfs} closer to the friction cone normals) or to minimize
the joint torques\footnote{Setting the weighting matrix $R_{kk} = J_{st} S^T
R_{\tau}S J_{st}^T$ where: $R_{kk}$ is the sub-block of $R$   that regularizes for
 \gls{grfs} variables in \eref{eq:qp_problem} and $S$ selects the actuation joints.}.
In the latter case, for instance, we could encourage the controller to use a
particular joint by increasing its corresponding weight.
If we gradually increase the weight of the \gls{kfe} joints (see accompanying
video), the effect of torque regularization becomes visible because the
\gls{grfs} are no longer vertical. Indeed the \gls{grfs} start to point toward
the knee axis in order to reduce its torque command.
\subsection{Comparison with Previous Controller (Quasi-Static)}
\label{sec:staticVSDynamic}
\begin{figure}[!t]
	\centering
	\includegraphics[width=\columnwidth]{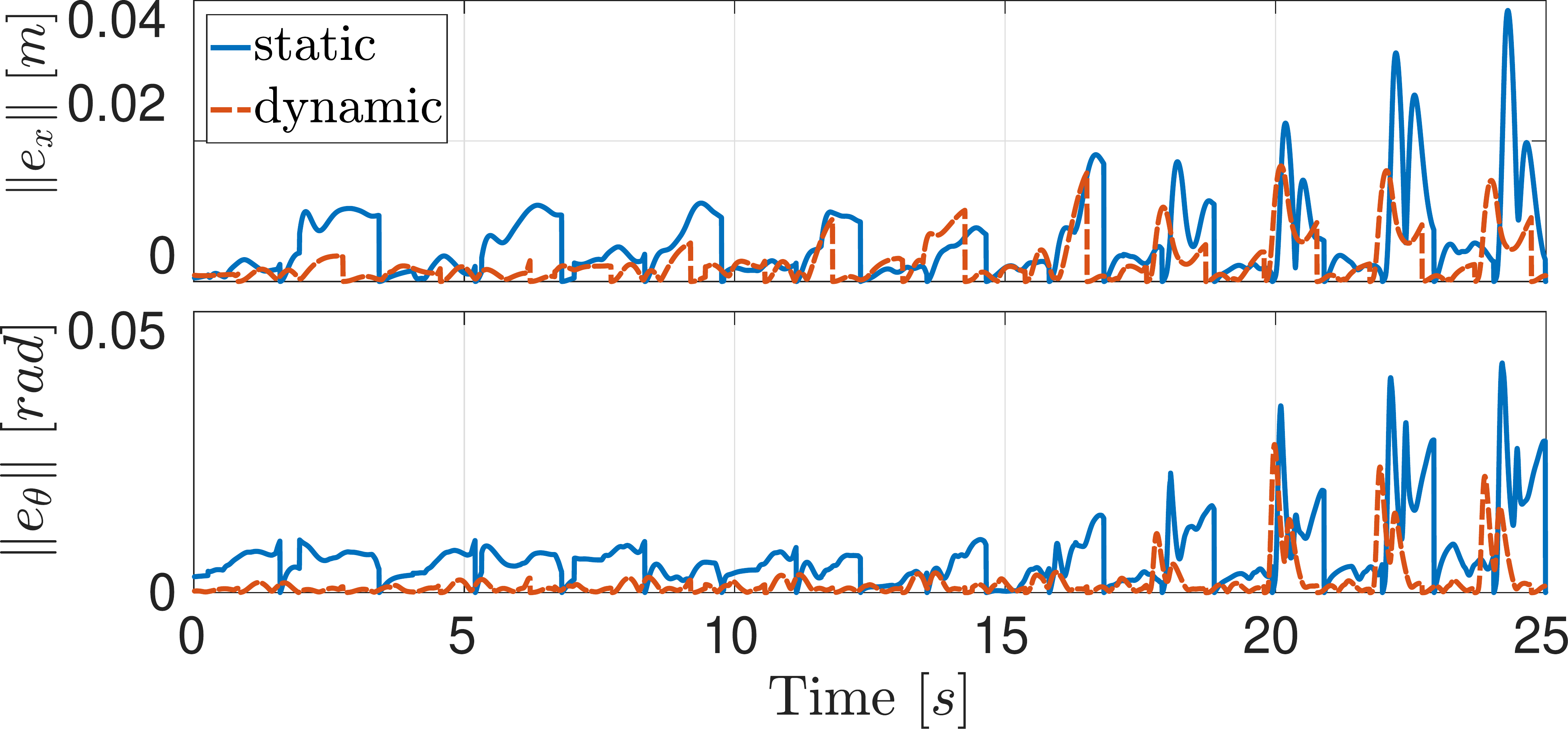}
	\caption[Comparison of tracking errors for the trunk task of a
	quasi-static controller against our whole-body controller
	(dynamic).]
	{Simulation. Comparison of tracking errors for the trunk task of a
	quasi-static controller against our whole-body controller
	(dynamic). $l^2$-norms of linear and angular errors are shown in the top and bottom figures.
	Note that the errors are reset to zero at each step due to re-planning.}
	\label{fig:staticVSdynamic}
\end{figure}

We compare our whole-body controller (dynamic) against a centroidal-based
controller (quasi-static) \cite{Focchi2017}.
As metric we use the $l^2$-norm for the linear $ e_{x} $ and angular
$ e_{\theta}$ tracking errors of the trunk task.
If we increase linearly the forward speed from \unit[0.04]{m/s} to
\unit[0.15]{m/s}, the tracking error is reduced approximately by \unit[50]{\%} in
comparison to the quasi-static controller (\fref{fig:staticVSdynamic}).
This is due to the fact that our \gls{wbc} computes both joint accelerations
and contact forces, which allows a proper mapping of torque commands
(inverse dynamics).
Indeed this results in better accuracy in the execution of more dynamic motions.
\subsection{Disturbance Rejection against Unstable Foothold}
\begin{figure}[tb]
	\centering
	\includegraphics[width=\columnwidth]{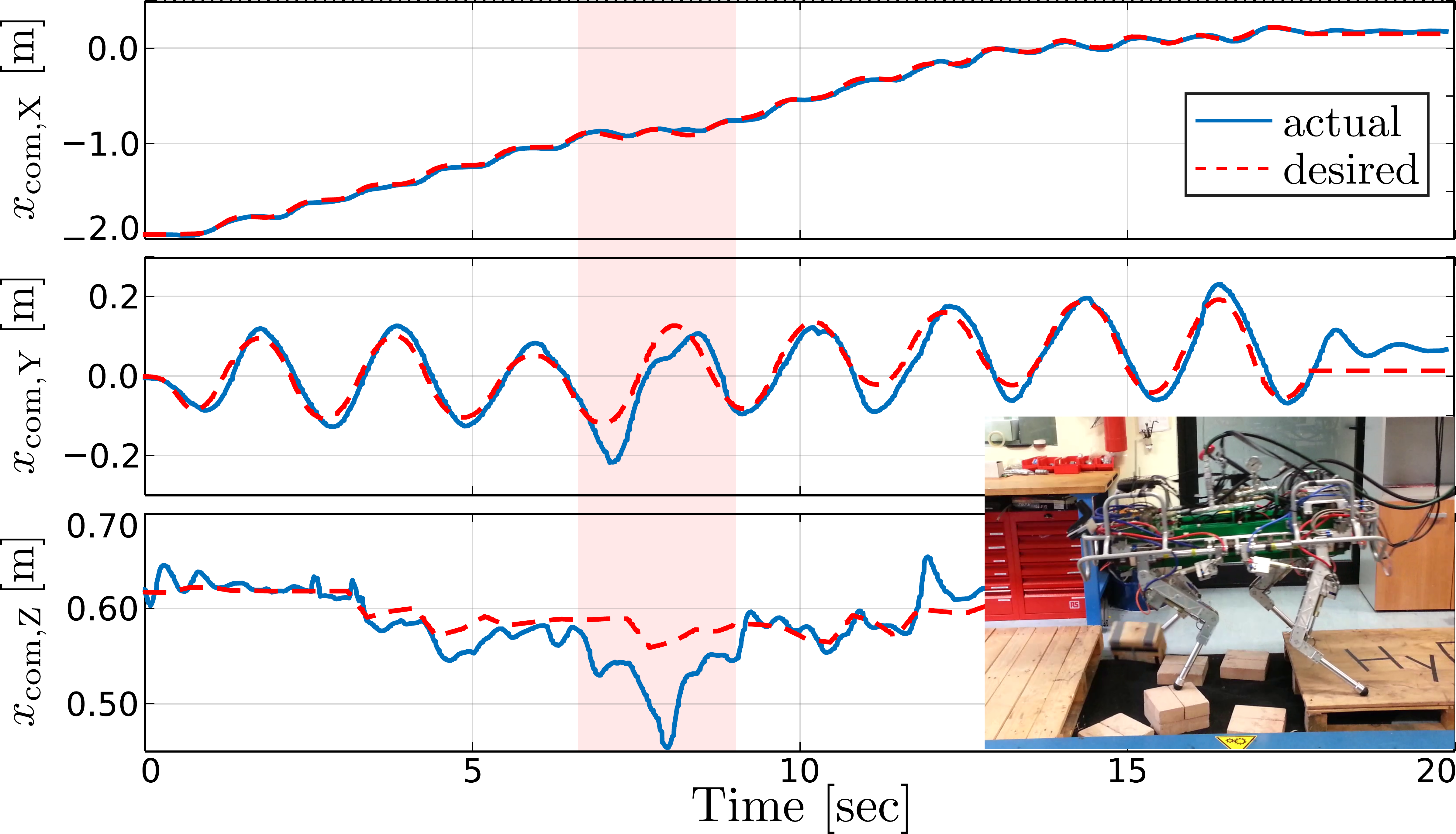}
	\caption[Disturbance rejection against unstable foothold that occurs when a
	stepping-stone rolled under the RF leg.]{Real experiments. Disturbance rejection against unstable foothold that occurs when a
	stepping-stone rolled under the RF leg at \unit[$t=6.5$]{s}. The controller lost
	tracking of the \acrshort{com} height, however, the friction cone constraints keep
	instantaneously the robot's balance. Indeed a good tracking of the horizontal
	motion of the \acrshort{com} is obtained. Note that the red shaded area depicts
	the moments of the disturbance rejection.}
	\label{fig:complianceRejection}
\end{figure}

We encoded compliance tracking of the \gls{com} task through a
\textit{virtual} impedance.
Friction cone constraints help to instantaneously keep the robot's balance
whenever a tracking error happens due to, for instance, an unstable foothold.
Furthermore joint constraints (positions and torques) guarantee feasibility of
the computed torque commands. Figure \ref{fig:complianceRejection} shows how the controller
compliantly tracks the desired \gls{com} trajectory during an unstable footstep (a
rolling stepping-stone) that occurs at \unit[$t=6.5$]{s} (experiments results from \cite{Mastalli2017}). 
This creates  tracking errors on the \gls{com} height, yet, 
good tracking performance is kept for the horizontal \gls{com} motion, due to the
friction cone constraints that maintained the robot's balance along the entire
locomotion.

\subsection{Locomotion over Slopes}
\begin{figure}[tb]
\centering
\includegraphics[width=\textwidth]{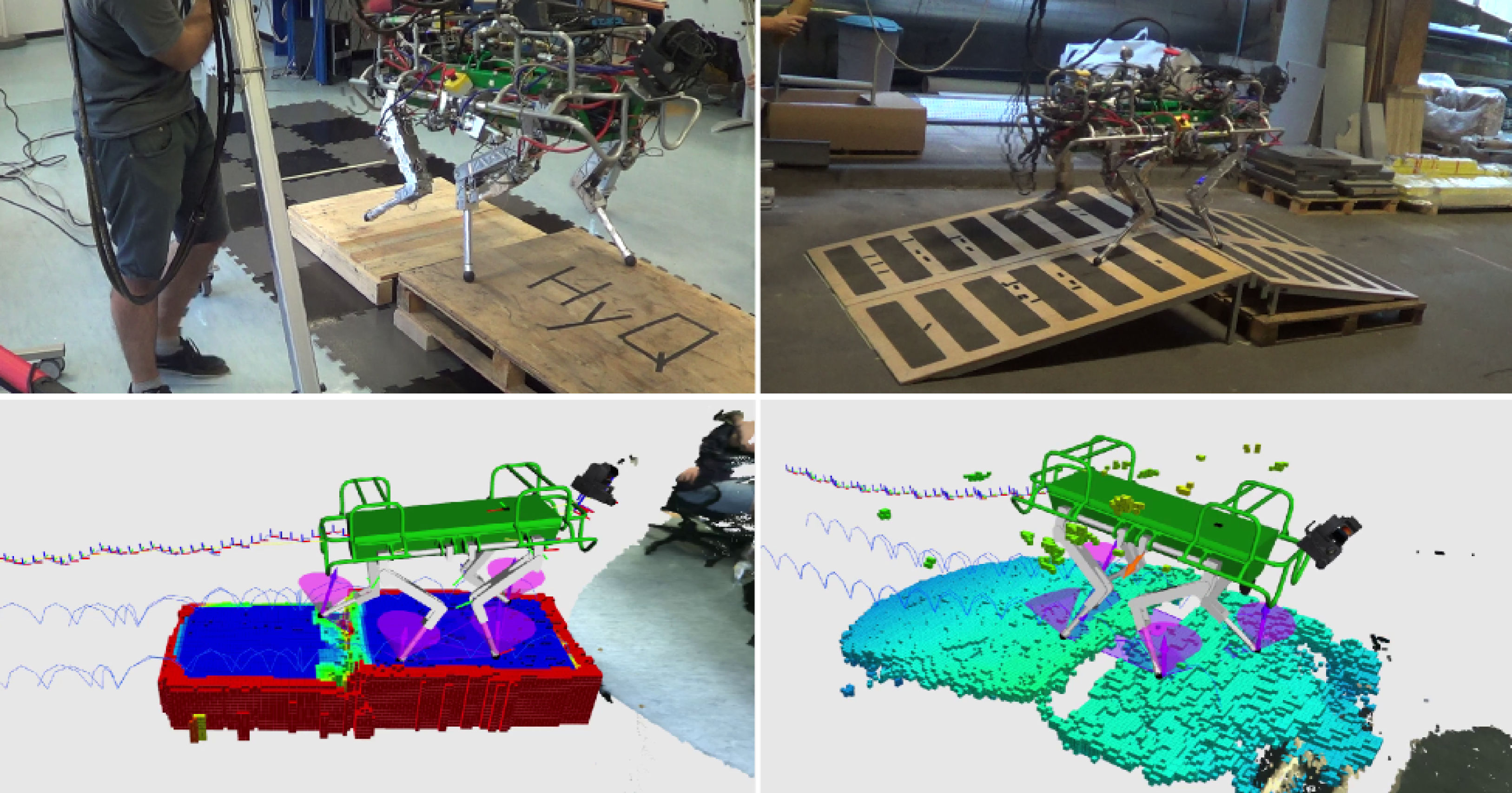}
\caption
[Snapshots of experimental trials to evaluate our \acrshort{wbc} and  the online terrain mapping.]
{Snapshots of experimental trials to evaluate our whole-body control
and  the \textit{online} terrain mapping. Left column: crossing a \unit[22]{cm} gap with a
\unit[7]{cm} step. Right column: traversing two ramps with a \unit[15]{cm}
gap between them.}
\label{fig:controlMapping}
\end{figure}

These experiments have been performed with \textit{online} terrain mapping
\cite{Focchi2018}. Both the terrain mapping and the whole-body controller make
use of a drift-free state estimation algorithm to obtain the body state.
The friction cone constraints of the controller are
described given the real terrain normals provided by an onboard mapping algorithm\footnote{The controller action can be 
greatly improved by setting the real terrain normal (under each foot)
rather than using a default value for all the feet.}.
The friction coefficient has been conservatively set to 0.7 for all the experiments. Figure 
\ref{fig:controlMapping} shows different snapshots of
various challenging terrain used to evaluate our controller.
The centroidal trajectory, gait and footholds are computed simultaneously
as described in \cite{Cabezas2018}.
\subsection{Tracking Performance with Different Gaits}
\begin{figure}[t]
	\centering
	\includegraphics[width=\textwidth]{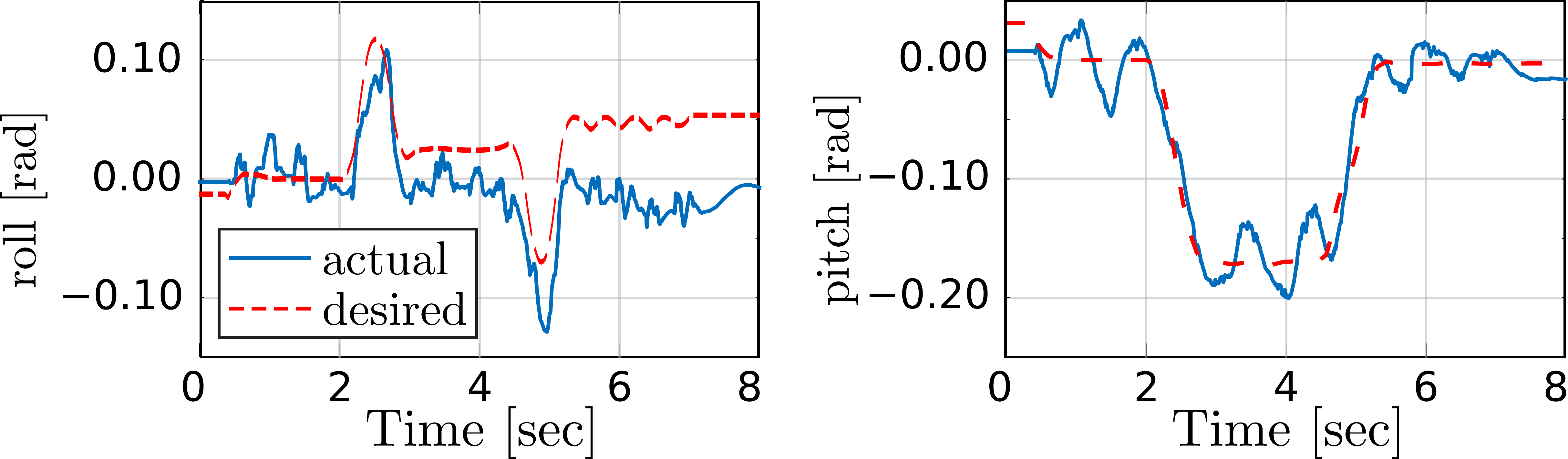}
	\caption[Roll and pitch tracking performance while climbing up a ramp with a
	trotting gait.]{Real experiments. Roll and pitch tracking performance while climbing up a ramp with a
	trotting gait. Although there is under-actuation the controller can still
	track roll and pitch motions.} 
	\label{fig:trottingTracking}
\end{figure}

The quadrupedal trotting gait is difficult to control because the robot uses only two
legs at the time to achieve the tracking of the desired \gls{com} 
motion and of the trunk orientation. Figure \ref{fig:trottingTracking} depicts the roll and pitch tracking for
climbing up a ramp during a trotting gait. Although a trot is an under-actuated gait, 
our controller can still track the desired orientation. Moreover, in
these cases, the orientation error is always below \unit[0.2]{rad}.
\section{Conclusion}\label{sec:conclusion}
This paper presented an experimental validation of our passive \gls{wbc}.
Compared to our previous work \cite{Focchi2017}, the presented \gls{wbc} 
enables higher dynamic motions thanks to the use of the  full dynamics of the robot. 
Although  similar controllers have been proposed in 
the literature (e.g. \cite{Herzog2016,Henze2016,Bellicoso2018}), 
we validated our locomotion controller  on \gls{hyq} over a wide range of challenging terrain (slopes, 
gaps, stairs, 
etc.), using 
different gaits (crawl and trot).
Additionally, we have analyzed the controller capabilities against 
1) inaccurate friction coefficient estimation,
2) unstable footholds, 
3) changes in the regularization scheme and 
4) the load redistribution under restrictive torque limits.
%
%
Extensive experimental results validated the
controller performance together with the \textit{online} terrain mapping and the state estimation.
Moreover, we demonstrated experimentally the superiority of our \gls{wbc} compared
to a quasi-static control scheme \cite{Focchi2017}.
\newpage 
\glsresetall \chapter[STANCE: Locomotion Adaptation over Soft Terrain]{\vspace{-25pt}STANCE:\\ Locomotion Adaptation over Soft Terrain}\label{chap_stance}
\vspace{-20pt}
\stancePaper

\boldSubSec{Abstract}
Whole-Body Control (WBC) has emerged as an important framework in locomotion control for legged robots.
However, most WBC frameworks fail to generalize beyond rigid terrains.  
Legged locomotion over soft terrain is difficult due to the presence of unmodeled contact dynamics 
that WBCs do not account for.
This introduces uncertainty in locomotion and affects the stability and performance of the system. 
In this paper, we propose a novel soft terrain adaptation algorithm called 
STANCE: Soft Terrain Adaptation and Compliance Estimation.
STANCE consists of a WBC that exploits the knowledge of the terrain to generate an optimal solution that is contact consistent
and an online terrain compliance estimator that provides the WBC with terrain knowledge.
We validated STANCE both in simulation and experiment on  the 
Hydraulically actuated Quadruped (HyQ)
robot, and we compared it against the state of the art WBC. 
We demonstrated the capabilities of STANCE with multiple terrains of different compliances, aggressive maneuvers, different forward velocities, 
and external disturbances. STANCE allowed HyQ to adapt online to terrains with different compliances (rigid and soft) without pre-tuning. 
HyQ was able to  successfully deal with the transition between different terrains and showed the ability to  differentiate between compliances~under~each~foot. 

\boldSubSec{Accompanying Video} \href{https://youtu.be/0BI4581DFjY}{\texttt{https://youtu.be/0BI4581DFjY}}
\section{Introduction}
\gls{wbc} frameworks have achieved remarkable results in legged locomotion 
control~{\cite{Farshidian2017a,Bellicoso2017,Fahmi2019}}. 
Their main feature is that they use optimization techniques
to solve the locomotion control problem. 
\gls{wbc} can achieve multiple tasks in an optimal fashion 
by exploiting the robot's full dynamics and reasoning about both the actuation constraints and the contact 
interaction. 
These tasks include balancing,  interacting with the 
environment, and 
performing dynamic locomotion over a wide variety of terrains \cite{Fahmi2019}.
The tasks are executed at the robot's end effectors, 
but can also be utilized for contacts anywhere on the robot's body \cite{Henze2017} or for
a cooperative manipulation task between robots \cite{Bouyarmane2018}.

\begin{figure}[tb]
	\centering
	\includegraphics[width=0.92\textwidth]{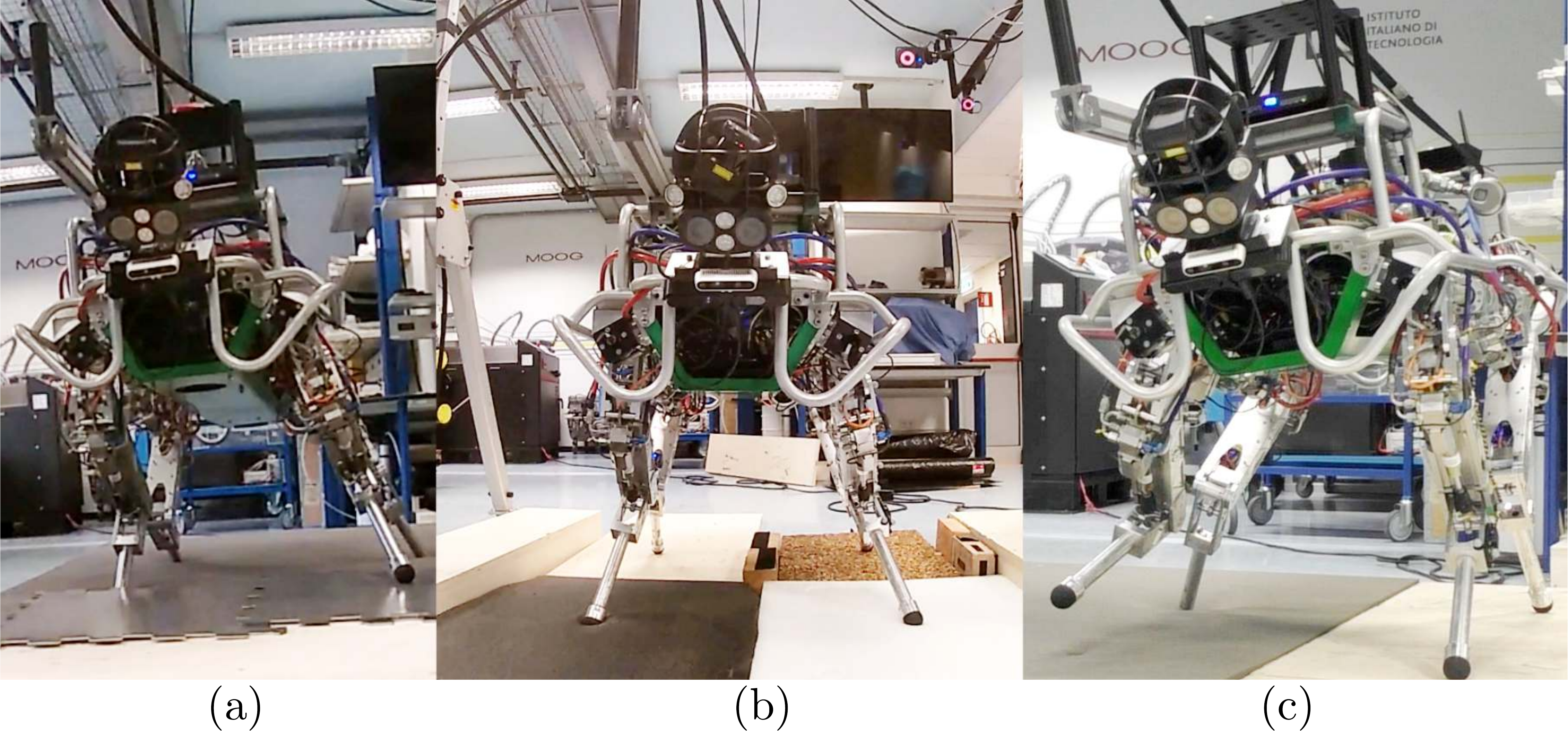}
	\caption{HyQ traversing multiple terrains of different compliances.}
	\label{fig:photos}
\end{figure}

To date, most of the work done on \gls{wbc} assumes that the ground is rigid 
(\ie rigid contact consistent).
However, if the robot traverses soft terrain (as shown in \fref{fig:photos}), 
the mismatch between the rigid assumption and the soft contact interaction can significantly affect the robot's 
performance and locomotion stability. 
This mismatch is due to the unmodeled contact dynamics between the robot and the terrain. 
In fact, under the rigid ground assumption, the controller can generate instantaneous changes to the 
\gls{grfs}. 
This is equivalent to thinking that  the terrain will respond with  an infinite bandwidth.

In order to robustly traverse a wide variety of terrains of different compliances, 
the \gls{wbc} must become \textit{\gls{c3}}. 
Namely, the \gls{wbc} should be terrain-aware. 
That said, a more general \gls{wbc} approach should be developed
that can adapt \textit{online} to the changes in the terrain compliance.

\subsection[Related Work - Soft Terrain Adaptation for Legged Robots]{Related Work: \\ Soft Terrain Adaptation for Legged Robots}
Locomotion over soft terrain can be tackled either from a control  or a planning  perspective.
In the context of locomotion control, 
Henze~\etaL~\cite{Henze2016} presented the first experimental 
attempt using a \gls{wbc} over soft terrain. 
Their~\gls{wbc} is based on the rigid ground 
assumption, but it allows for constraint relaxation.
This allowed the humanoid  robot TORO to adapt to a compliant surface. 
Their approach was further extended in~\cite{Henze2018} by dropping the rigid contact assumption and using an 
energy-tank approach. 
Despite balancing on compliant terrain, 
both approaches were only tested for one type of soft terrain when the robot was standing still. 

Similarly, 
other works explicitly adapt to soft terrain by incorporating terrain knowledge (i.e., contact model) into 
their  balancing controllers. 
For example, Azad~\textit{et~al.}~\cite{Azad2015}  proposed a momentum based controller for balancing on soft terrain by 
relying on a nonlinear soft contact model. 
Vasilopoulos~\etaL~\cite{Vasilopoulos2018} proposed a similar hopping controller that models the 
terrain using a viscoplastic contact model.
However, these approaches were only tested in simulation and for monopods.

In  the context of locomotion planning, 
Grandia~\etaL~\cite{Grandia2019} indirectly adapted to soft terrain by
shaping the frequency of the cost function of their \gls{mpc} formulation.  
By penalizing high frequencies, they generated optimal motion plans that 
respect the bandwidth  limitations due to soft terrain. 
This approach was tested over three types of terrain compliances. 
However, it was not tested during transitions from one terrain to another. 
This approach showed
an improvement in the performance of the quadruped robot in simulation and experiment. 
However, the authors did not offer
the possibility 
to change their
tuning parameters online. 
Thus, they were not able to adapt the locomotion strategy  based on the compliance of the terrain. 

In contrast to the aforementioned work, other approaches relax the rigid ground assumption (hard contact constraint)
but not for soft terrain adaptation purposes. 
For instance, Kim~\etaL~\cite{Kim2019} 
implemented an approach to handle sudden changes in the rigid contact interaction. 
This approach relaxed the 
hard contact assumption in their \gls{wbc} formulation
by
penalizing the contact interaction in the cost function 
rather than incorporating it as a hard constraint. 
For  computational purposes, Neunert~\etaL~\cite{Neunert2018} and Doshi~\etaL~\cite{Doshi2019} 
proposed relaxing the rigid ground assumption.
Neunert~\etaL~used a soft contact model in their nonlinear \gls{mpc} 
formulation
to provide smooth gradients of the contact dynamics
to be more efficiently solved by their gradient based solver. 
The soft contact model did not have a physical meaning and the contact parameters were empirically chosen. 
Doshi~\etaL~proposed a similar approach which incorporates a 
slack variable that expands the feasibility region of the hard constraint.

Despite the improvement in performance of the legged robots over soft terrain in the aforementioned works,
none of them offered the possibility to adapt to the terrain \textit{online}.
Most  of the aforementioned works 
lack a general approach that can
deal with multiple terrain compliances 
or with transitions between them. 
Perhaps, one noticeable work (to date) in online soft terrain adaptation was proposed by 
Chang~\etaL~\cite{Chang2017}.
In that work, an iterative soft terrain adaptation approach was proposed. 
The approach relies on a non-parametric contact model 
that is simultaneously updated alongside an optimization based hopping controller.
The approach was capable of iteratively learning the terrain interaction and 
supplying that knowledge to the optimal controller. 
However, because the learning module was exploiting Gaussian process regression, which is computationally expensive,
the approach did not reach real-time performance and was only tested in simulation, for one leg, under one 
experimental condition (one terrain).
\subsection[Related Work - Contact Compliance Estimation in Robotics]{Related Work:\\Contact Compliance Estimation in Robotics} 
For contact compliance estimation, we need to accurately model the contact dynamics 
and estimate the contact parameters online.
In contact modeling, Alves~\etaL~\cite{Alves2015} presented a detailed overview of the types of parametric soft contact models used in the literature. 
In compliance estimation, Schindeler~\etaL~\cite{Schindeler2018} used a two stage polynomial identification 
approach to estimate the parameters of the~\gls{hc} contact model online.  
Differently, Azad~\etaL~\cite{Azad2016} used
a least square-based estimation algorithm
and compared multiple contact models (including the \gls{kv} and the \gls{hc} models).
Other approaches that are not based on soft contact models
use force observers \cite{Coutinho2014} or neural networks \cite{Coutinho2013}. 
These aforementioned approaches in compliance estimation
were designed for robotic manipulation tasks.   

To date, the only  work on compliance estimation in 
legged locomotion was the one by Bosworth~\etaL~\cite{Bosworth2016}. 
The authors presented two online (in-situ) approaches to estimate 
the ground properties (stiffness and friction).
The results were promising and the approaches were validated on a quadruped robot 
while hopping over rigid and soft terrain. 
However, the estimated stiffness showed a trend, but was not accurate; 
the lab measurements of the terrain stiffness did not match the in-situ ones. 
Although the estimation algorithms could be implemented online, 
the robot had to stop to perform the estimation. 

\begin{figure}[tb]
\centering
\includegraphics[width=0.95\textwidth]{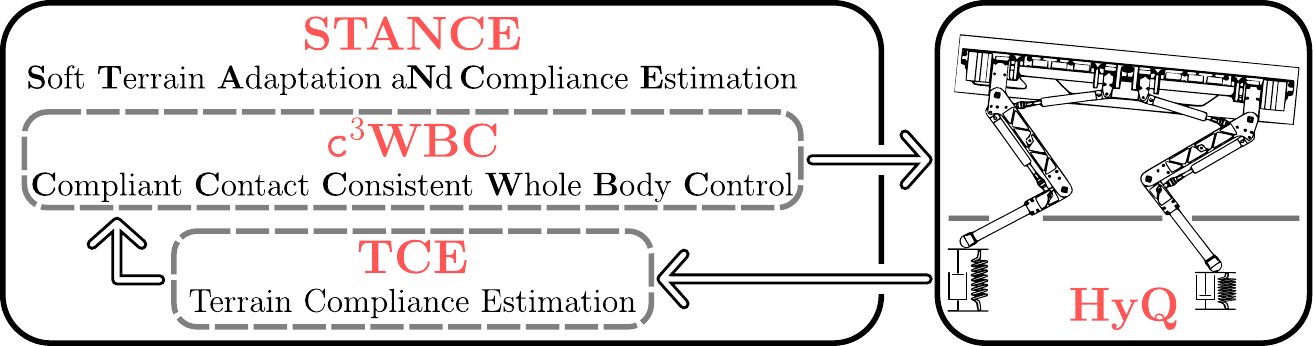}
\caption{An overview of the STANCE algorithm.}
\label{fig:concept}
\end{figure}

\subsection{Proposed Approach and Contribution}
In this work, we propose an online soft terrain adaptation algorithm called: 
\gls{stance}. 
As shown in \fref{fig:concept}, \gls{stance} consists~of
\begin{itemize}[leftmargin=*]
\item A \gls{awbc} that is contact consistent to any type of terrain
\textit{given the terrain compliance}. 
This is done by extending the state-of-the-art \gls{wbc} in \cite{Fahmi2019}, hereafter denoted as the \gls{swbc}. 
In particular, \gls{awbc} incorporates a soft contact model into the \gls{wbc} formulation.
\item A \gls{ste} which is an online learning algorithm that 
provides the  \gls{awbc} with an estimate of the terrain compliance.
It is based on the same contact model that is incorporated in the \gls{awbc}.
\end{itemize}
The main contribution of \gls{stance} is that it can adapt to any type of terrain (stiff or soft)
online without pre-tuning. 
This is done by closing the loop of the \gls{awbc} with the \gls{ste}.
To our knowledge, this is the first implementation of such an approach in legged locomotion.

\gls{stance} is meant to overcome the limitations of the aforementioned approaches in soft terrain 
adaptation for legged robots. 
Compared to previous works on \gls{wbc} that tested 
their approach only during standing
\cite{Henze2016,Henze2018}, 
we test our \gls{stance} approach  during  locomotion. 
Compared to other approaches~\cite{Azad2015,Vasilopoulos2018}
that were tested on monopods in simulation, 
\gls{stance} is implemented and tested 
in experiment on \gls{hyq}.
Compared to previous work on soft terrain adaptation \cite{Grandia2019}, 
\gls{stance} can adapt to soft terrain \textit{online} and was tested on multiple 
terrains with different compliances and with transitions between them. 
Compared to \cite{Chang2017}, our \gls{ste} is computationally inexpensive, which allows \gls{stance} to
run real-time in experiments and simulations. 
Compared to the previous work done on compliance estimation, 
we implemented our \gls{ste} on a legged robot which is, to the  best of our knowledge, 
the first experimental validation of this approach. 
Differently from~\cite{Bosworth2016}, our \gls{ste} approach could be implemented
in parallel with any gait or task. 
We also achieved a more accurate estimation of the terrain
compliance compared to \cite{Bosworth2016}.

As additional contributions, 
we discussed the benefits (and the limitations) of exploiting the knowledge of the terrain in \gls{wbc}
based on the experience gained during extensive experimental trials. 
To our knowledge,  \gls{stance}  is the first work to present legged 
locomotion experiments crossing multiple terrains of different compliances.

\section{Robot model}

Consider a legged robot with $n$ \gls{dofs} and $c$  feet.
The total dimension of the feet operational space $n_{\nmrm{f}}$ can be separated into stance  ($n_{\mrm{st}} = 
3 c_{\mrm{st}}$) and swing feet 
($n_{\mrm{sw}} = 3 c_{\mrm{sw}}$) where $c_{\mrm{st}}$ and $c_{\mrm{sw}}$ are the number of stance and 
swing legs respectively.
Assuming that all external forces are exerted on the stance feet,
the robot dynamics is written as

\begin{eqnarray}
\underbrace{
	\mat{M_{\mrm{com}} & \vc{0}_{3\times 3} & \vc{0}_{3\times n} \\
		\vc{0}_{3\times 3} & \vc{M}_\theta & \vc{M}_{\theta j} \\
		\vc{0}_{n\times 3} & \vc{M}_{\theta j}^T & \vc{M}_j 
	}
}_{\vc{M}(\vc{q})}
\underbrace{
	\mat{\ddot{x}_\mrm{com} \\ \dot{\omega}_{b}\\ \ddot{\vc{q}}_j}
}_{\ddot{\vc{q}}}
+ 
\underbrace{
	\mat{ 
		h_{\mrm{com}} \\ \vc{h}_\theta
		\\ \vc{h}_j}
}_{h(q,\dot{q})} \nonumber \\ 
= \mat{ \vc{0}_{3\times 1} \\ \vc{0}_{3\times 1} \\ \vc{\tau_j}}
+ \underbrace{
	\mat{\vc{J}_\mrm{st,com}^T \\ \vc{J}_{\mrm{st},\theta}^T \\ \vc{J}_{\mrm{st},j}^T}}
_{\vc{J}_\mrm{st}(\vc{q})^T}\vc{\grf}
\label{eq:full_dynamicsCOMSTANCE}
\end{eqnarray}

\noindent
where $\vc{q}~\in~SE(3)~\times~\Rnum^n$ 
denotes the generalized robot states
consisting of 
the \gls{com} position $x_{\mrm{com}}$~$\in~\Rnum^3$,
the base orientation $R_b$~$\in~SO(3)$, and 
the joint positions $q_j$~$\in~\Rnum^n$.
The vector $\vc{\dot{q}}~=~[\dot{x}_\mrm{com}^T~\omega_b^T~\vc{\dot{q}}_j^T]^T~\in~\Rnum^{6+n}$
denotes the generalized velocities consisting of  
the velocity of the \gls{com} $\dot{x}_\mrm{com}$~$\in~\Rnum^3$, 
the angular velocity of the base $\omega_b$~$\in~\Rnum^3$, 
and the joint velocities $\dot{q}_j$~$\in~\Rnum^n$. 
The vector  $\ddot{\vc{q}}~=~[\ddot{x}_{\mrm{com}}^T~\dot{\omega}_{b}^T~\ddot{q}_j^T]^T~\in~\Rnum^{6+n}$ 
denotes the corresponding generalized accelerations.
All Cartesian vectors are expressed in the world frame $\Psi_W$ unless mentioned otherwise. 
%
%
$\vc{M}~\in~\Rnum^{(6+n) \times (6+n)}$ is the  inertia matrix.
$\vc{h}\in \Rnum^{6+n}$ is the force vector that accounts for Coriolis,
centrifugal, and gravitational forces.
$\vc{\tau_j}\in\Rnum^n$ are the actuated joint torques,
$\vc{\grf} \in\Rnum^{n_\mrm{st}}$ is the vector of \gls{grfs} (contact forces).
The Jacobian matrix $J$ $\in\Rnum^{n_\mrm{f} \times(6+n)}$
is separated into swing Jacobian $J_{\mrm{sw}}$~$\in\Rnum^{n_{\mathrm{sw}}~\times~(6+n)}$ and stance 
Jacobian  $J_{\mrm{st}}~\in~\Rnum^{n_{\mathrm{st}}~\times~(6+n)}$ which can be further expanded into 
$J_{\mathrm{st,com}}$  $\in\Rnum^{n_{\mathrm{st}} \times 3}$,
$J_{\mathrm{st,}\theta}$~$\in\Rnum^{n_{\mathrm{st}} \times 3}$,
and $J_{\mathrm{st},j}$  $\in\Rnum^{n_\mathrm{st} \times n}$.
The feet velocities $v~\in\Rnum^{n_\mrm{f}}$ are separated into stance $v_{\mrm{st}}~\in~\Rnum^{n_{\mrm{st}}}$ and swing $v_{\mrm{sw}}~\in~\Rnum^{n_{\mrm{sw}}}$ feet velocities. Similarly, the feet accelerations $\dot{v}~\in\Rnum^{n_\mrm{f}}$ are separated into stance $\dot{v}_{\mrm{st}}~\in~\Rnum^{n_{\mrm{st}}}$ and swing $\dot{v}_{\mrm{sw}}~\in~\Rnum^{n_{\mrm{sw}}}$ feet accelerations.
The feet forces $F~=~[F_{\mrm{st}}^T \ 
F_{\mrm{sw}}^T]^T~\in~\Rnum^{n_\mrm{f}}$ are also separated into stance  $F_{\mrm{st}}~\in~\Rnum^{n_\mrm{st}}$ and swing 
$F_{\mrm{sw}}~\in\Rnum^{n_\mrm{sw}}$ feet forces.
We split the robot dynamics  \eref{eq:full_dynamicsCOMSTANCE} into an unactuated floating base part (the first 6 rows) and an actuated part (the remaining $n$ rows) as
\begin{subequations}
	\begin{eqnarray}
	\vc{M}_u(\vc{q}) \ddot{\vc{q}} + h_u (q,\dot{q})  &=& \vc{J}_{\mrm{st},u}(\vc{q})^T\vc{\grf} \label{eq_unactuated}\\
	\vc{M}_a(\vc{q}) \ddot{\vc{q}} + h_j (q,\dot{q})  &=&  \vc{\tau_j} + \vc{J}_{\mrm{st},j}(\vc{q})^T\vc{\grf}  \label{eq_actuated}
	\end{eqnarray}
\end{subequations}
where $M_u$~$\in~\Rnum^{6\times6+n}$ and  $M_a$~$\in~\Rnum^{n\times6+n}$
are sub matrices of $M$, 
$h_u=[h_{\mrm{com}}^T~h_\theta^T ]^T$~$\in~\Rnum^{6}$ and
$h_j$~$\in~\Rnum^{n}$ are sub vectors of $h$, and
$J_{\mrm{st},u}=[J_{\mrm{st,com}}^T~J_{\mrm{st},\theta}^T]^T$.
Finally, we define the gravito-inertial wrench as
$W_\mrm{com}~=~\vc{M}_u(\vc{q}) \ddot{\vc{q}} + h_u (q,\dot{q})$ $\in \Rnum^6$.

\begin{sidewaysfigure}
	\centering
	\includegraphics[width= \textwidth]{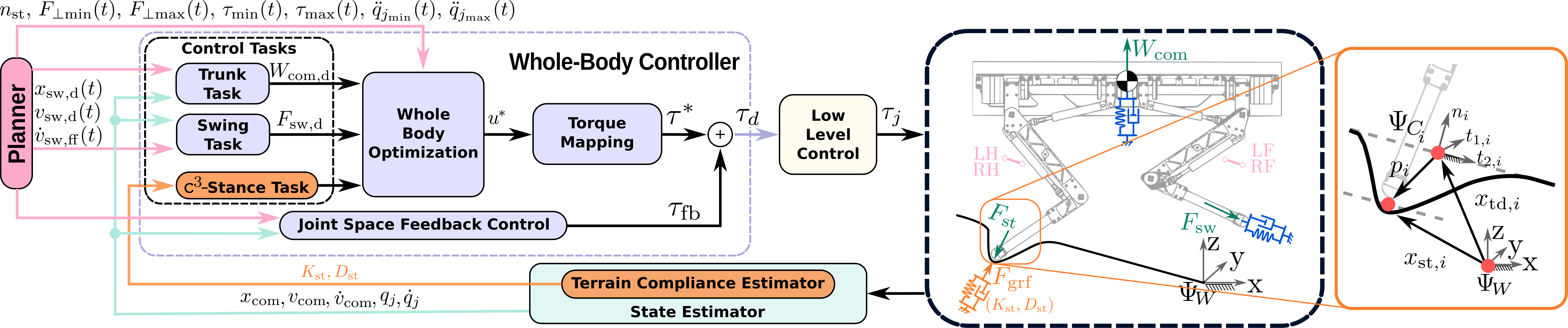}		
	\caption[Overview of the \acrshort{wbc} in	our locomotion framework.]{Overview of the \acrshort{wbc} in	our locomotion framework.  
		The dashed black box presents the definition of the \acrshort{hyq}'s legs (\acrshort{lf}, \acrshort{rf}, \acrshort{lh} and \acrshort{rh}) 
		and the generated wrenches.
		The solid orange box presents the terminologies used for the soft contact model. 
		$\Psi_\mrm{W}$ 	is the world frame and 	$\Psi_{\mrm{C}_i}$ is the local contact frame for the leg $i$ fixed at the touch down position.}
	\label{fig:blockDiagramSTANCE}
\end{sidewaysfigure}

\section{Standard Whole-Body Controller (sWBC)} 
\label{sec:wbc}
This section summarizes the \gls{swbc} as detailed in \cite{Fahmi2019}.
Besides the \gls{wbc}, our locomotion framework includes a locomotion planner, 
state estimator and a low-level
torque controller as shown in \fref{fig:blockDiagramSTANCE}.
Given high-level user inputs, the planner generates the desired trajectories for the \gls{com}, trunk orientation and 
swing legs, and provides them to the \gls{wbc}. 
The state estimation provides the \gls{wbc} with the estimated states of the robot. 

The objective of the \gls{swbc} is to 
ensure the execution of the trajectories 
provided by the planner 
while keeping the robot balanced and reasoning about the robot's dynamics, 
actuation limits and the contact 
constraints \cite{Fahmi2019}. 
We denote the execution of the trajectories provided by the planner as \textit{control tasks}.
	These control tasks alongside the aforementioned constraints define the \gls{wbc} problem.
The control problem is casted as a \gls{wbopt} problem via a \gls{qp} which solves for the optimal generalized 
accelerations 
and contact forces at each iteration of the  control loop~\cite{Herzog2016}.
The optimal solution of the \gls{wbc} is then mapped into joint torques 
that are sent to the low-level torque controller.

\subsection{Control Tasks}
\label{sec:controlTasks}
We categorize the \gls{swbc} \textit{control tasks} into: 
1) a \textit{trunk task} that tracks the desired trajectories 
of the \gls{com} position and trunk orientation, 
and 2) a \textit{swing task} that tracks the swing feet trajectories \cite{Fahmi2019}. 
Similar to a PD+ controller \cite{Ortega2013}, both tasks are achieved by a Cartesian-based impedance controller with a 
feed-forward term.
The feedforward terms are added in order to improve the tracking performance 
	of the tasks when following the trajectories from the planner \cite{Henze2016,Focchi2017}. 
The tracking of the trunk task  is obtained by the desired wrench at the \gls{com}  $W_{\mrm{com,d}} \in 
\Rnum^{6}$. 
This is generated by a Cartesian impedance at the \gls{com}, a gravity compensation 
term, and a feed-forward term.
Similarly, the tracking of the swing task can be obtained by the 
virtual force $F_{\mrm{sw,d}} \in \Rnum^{n_{\mrm{sw}}}$.
This is generated by a Cartesian impedance at the swing foot  
and a feed-forward term. 
As in~\cite{Fahmi2019}, we can also write the swing task at the acceleration level by defining the desired swing feet velocities $\dot{\vc{v}}_{\mrm{sw,d}}$ $\in \Rnum^{n_{\mrm{sw}}}$ as
\begin{eqnarray}
\dot{\vc{v}}_{\mrm{sw,d}} = \dot{\vc{v}}_{\mrm{sw,ff}} +
\mrm{K}_{\mrm{sw}} \Delta x_{\mrm{sw}} +
\mrm{D}_{\mrm{sw}} \Delta v_{\mrm{sw}} 
\label{eq:swingTaskAccSTANCE}
\end{eqnarray}
where $\mrm{K}_{\mrm{sw}}, \mrm{D}_{\mrm{sw}} \in\Rnum^{n_\mrm{sw}\times n_\mrm{sw}}$
are positive definite PD gains,
$\Delta x_{\mrm{sw}} = x_{\mrm{sw,d}} - x_{\mrm{sw}} $ $\in \Rnum^{n_{\mrm{sw}}}$ and $\Delta v_{\mrm{sw}} = v_{\mrm{sw,d}} - 
v_{\mrm{sw}}$ $\in \Rnum^{n_{\mrm{sw}}}$ are tracking 
errors of the swing foot position and velocity, respectively,
and $\dot{\vc{v}}_{\mrm{sw,ff}}$ is a feed-forward term.

\subsection{Whole-Body Optimization} 
\label{sec:wbopt}
To accomplish the \gls{swbc} objective (the control tasks in  \sref{sec:controlTasks} and constraints), 
we formulate the \gls{wbopt} problem presented in \algoref{eqn_wbopt} and detailed in \cite{Fahmi2019}.
\subsubsection{Decision Variables}
As shown in \algoref{eqn_wbopt}, we choose the generalized 
accelerations~$\ddot{\vc{q}}$ and the contact forces~$\grf$ as the decision variables~$\vc{u}~=~[\ddot{\vc{q}}^T~\vc{\grf}^T]^T~\in~\Rnum^{6+n+n_{\mrm{st}}}$. 
Later in this subsection, we will augment the vector of decision 
variables with a slack term  $\eta \in \Rnum^{n_{\mrm{sw}}}$.

\subsubsection{Cost}
The cost function  \eref{wbopt_cost_function} consists of two terms. 
The first term ensures the tracking of the trunk task by minimizing the two-norm of the tracking error between the actual $W_{\mrm{com}}$ 
and desired  $W_{\mrm{com,d}}$ \gls{com} wrenches.
The second term in  \eref{wbopt_cost_function}
regularizes the solution and penalizes the slack variable.

\subsubsection{Physical Consistency}
The equality constraint \eref{wbopt_physical_consistency} enforces the physical consistency between $\grf$ and 
$\ddot{\vc{q}}$ by ensuring that the contact wrenches due to $\grf$ will sum up to $W_\mrm{com}$. 
This is done by imposing the unactuated dynamics \eref{eq_unactuated} as an equality
constraint.

\subsubsection{Stance Task}
To remain  \textit{contact consistent}, we incorporate the \textit{stance task}
that enforces the stance legs to remain in contact with the terrain. 
Since the \gls{swbc} is assuming a rigid terrain, 
the stance feet are forced to remain stationary
in the world frame,  i.e.,~$v_{\mrm{st}} = \dot{v}_{\mrm{st}} = 0$
(see~\cite{Fahmi2019}). 
As a result, we incorporate the rigid contact model
in the \gls{swbc} formulation as an equality constraint at the acceleration level \eref{wbopt_rigid} in
order to have a direct dependency on the decision variables. 
In detail,  since $v_{\mrm{st}} = J_{\mrm{st}} \dot{q} = 0$, differentiating once with respect to time yields
	$\dot{v}_{\mrm{st}} = \vc{J}_{\mrm{st}}\ddot{\vc{q}}+\dot{\vc{J}}_{\mrm{st}}\dot{\vc{q}}=0$. 

\subsubsection{Friction and Normal Contact Force}
The inequality constraint \eref{wbopt_friction} enforces the friction constraints by ensuring that the contact forces lie inside the  friction cones.
This is done by limiting the tangential component of the  \gls{grfs}~$\grfp{\parallel}$.
The inequality constraint \eref{wbopt_contact_force} enforces  constraints 
on the normal component of the \gls{grfs}~$\grfp{\perp}$. 
This includes the unilaterality constraints which encodes that the legs can 
only push on the ground  by setting an  ``almost-zero'' lower bound $F_{\mrm{\min}}$ to 
$\grfp{\perp}$.
They also allow a smooth loading/unloading of the legs, and set
a varying upper bound $F_{\mrm{\max}}$ to   $ \grfp{\perp}$. 
For the detailed implementation of the inequality constraints~\eref{wbopt_friction}~and~\eref{wbopt_contact_force}, refer to \cite{Focchi2017}.

\subsubsection{Swing Task}
We implement the tracking of the swing task  (in \sref{sec:controlTasks})
at the acceleration level  \eref{eq:swingTaskAccSTANCE} rather than the force level
since we can express the swing feet velocities $\dot{v}_{\mrm{sw}}$ as a function of  $\ddot{q}$ which is a decision variable, 
\ie~$\dot{\vc{v}}_{\mrm{sw}}(\vc{q})= \vc{J}_{\mrm{sw}}\ddot{\vc{q}} + 
\dot{\vc{J}}_{\mrm{sw}}\dot{\vc{q}}$. 
This task could be encoded as an equality constraint
$\dot{v}_{\mrm{sw}} =\dot{v}_{\mrm{sw,d}}$. 
Yet, it is important to relax this hard constraint 
when the joint kinematic limits are reached (see~\cite{Fahmi2019}). 
Hence, the swing task is encoded in \eref{wbopt_swing_task} 
by an inequality constraint that bounds the solution around the original hard constraint
and a slack term  $\eta$ that is penalized for its non-zero values in the cost function \eref{wbopt_cost_function}
and is constrained to remain non-negative in \eref{wbopt_swing_task}.

\subsubsection{Torque and Joint Limits}
The torque and joint limits  are enforced in the 
inequality constraints \eref{wbopt_torque_limits}  and  \eref{wbopt_joint_limits}, respectively.

\subsubsection{Torque Mapping}
The \gls{wbopt} \eref{wbopt_cost_function}-\eref{wbopt_rigid},~\eref{wbopt_friction}-\eref{wbopt_joint_limits} generates
optimal joint accelerations $\ddot{\vc{q}}_j^*$  and contact forces 
$\vc{\grf}^*$, that are mapped into optimal joint torques $\tau^*$ 
and sent to the low-level controller
using the actuated dynamics~\eref{eq_actuated}~as
\begin{equation}
\tau^{*} = M_a\ddot{q}^* + h_j - J_{\mrm{st},j}^T \grf^*
\label{eq:torquesSTANCE}
\end{equation}

\begin{algorithm}[t!]
	\centering
	\caption{Whole-Body Optimization: \acrshort{swbc} Vs. \acrshort{awbc}}\label{eqn_wbopt}
	\vspace{-0.9cm}
	\[
	\begin{minipage}[t]{0.15 \linewidth}
	\begin{align*}
	~&\text{\sffamily(Trunk Task)}\\[4pt]
	~&\text{\sffamily(Decision Variables)}\\
	~&~\\
	~&\text{\sffamily(Physical Consistency)}\\[4pt]
	~&\text{\sffamily\cancel{(Stance Task)}}\\
	~&\text{\sffamily\textcolor{RoyalBlue}{\textbf{(\texttt{c}$^3$-Stance Task)}}}\\ ~&\\ ~&\\[2pt] 
	~&\text{\sffamily(Friction)}\\[1pt]
	~&\text{\sffamily(Normal Contact Force)}\\[1pt]
	~&\text{\sffamily(Swing Task)} \\[0pt]
	~&\text{\sffamily(Torque Limits)} \\
	~&\text{\sffamily(Joint Limits)} 
	\end{align*}		
	\end{minipage}
	\begin{minipage}[t]{0.65\linewidth}
	\begin{gather} 
	\underset{\vc{u}} {\text{min~}}  
	\Vert W_{\mrm{com}}-W_{\mrm{com,d}} \Vert^2_Q + \bmcolor{\Vert\vc{u}\Vert^2_{R}}
	\label{wbopt_cost_function} \\
	\vc{u}=[\ddot{\vc{q}}^T \ \vc{\grf}^T  \ \vc{\eta}^T \ \bmcolor{\vc{\epsilon}^T }]^T \nonumber 
	\label{wbopt_dec_variables}\\
	\text{s.t.:}   \hspace{135pt}  \nonumber  \\
	M_u \ddot{\vc{q}} + h_u = J^T_{\mrm{st},u}  \grf  
	\label{wbopt_physical_consistency}\\
	\cancel{\dot{v}_{\mrm{st}}=\vc{J}_{\mrm{st}}\ddot{\vc{q}}+\dot{\vc{J}}_{\mrm{st}}\dot{\vc{q}} = 
		0} \label{wbopt_rigid}\\
	\bmcolor{\grf = K_{\mrm{st}} \epsilon   + D_\mrm{st} \dot{\epsilon}  } 
	\label{wbopt_fgrf_epsilon}\\
	\bmcolor{\dot{v}_{\mrm{st}} =  
		\vc{J}_{\mrm{st}}\ddot{\vc{q}}+\dot{\vc{J}}_{\mrm{st}}\dot{\vc{q}} = 
		- \ddot{\epsilon} } 
	\label{wbopt_ddot_epsilon}\\
	\bmcolor{ \epsilon \geq 0  } \label{wbopt_penetration_slack} \\
	\vert \grfp{\parallel} \vert \leq \mu |\grfp{\perp} | \label{wbopt_friction} \\
	F_{\mrm{\min}} \leq \grfp{\perp} \leq F_{\mrm{\max}} \label{wbopt_contact_force}\\
	-\eta\leq \dot{v}_{\mrm{sw}} - \dot{\vc{v}}_{{\mrm{sw,d}}}  \leq\eta ,~ \eta  \geq 0  
	\label{wbopt_swing_task}\\
	\vc{\tau}_{\min} \leq \vc{\tau}_j \leq \vc{\tau}_{\max}  \label{wbopt_torque_limits} \\
	\ddot{\vc{q}}_{j_{\min}} \leq \ddot{\vc{q}}_j \leq \ddot{\vc{q}}_{j_{\max}}  
	\label{wbopt_joint_limits}
	\end{gather}
	\end{minipage}
	\nonumber
	\]
	\vspace{-0.2cm}
\end{algorithm}

\subsection{Feedback Control}
The computation of the optimal torques  $\tau^*$ relies on the inverse dynamics 
in \eref{eq:torquesSTANCE} which might be prone to model inaccuracies \cite{Righetti2013}. 
In order to tackle this issue, the desired torques $\tau_d$ sent to the lower level 
control could combine the optimal torques $\tau^*$ in  \eref{eq:torquesSTANCE}
with a feedback controller $\tau_{\mrm{fb}}$ as shown in \fref{fig:blockDiagramSTANCE}. 
The feedback controller improves the tracking performance if the dynamic model of the robot becomes less accurate \cite{Righetti2013}. 
The feedback controller is a proportional-derivative (PD) joint space impedance controller \cite{Focchi2017}. 
\begin{remark}
Throughout this work and similar to \cite{Fahmi2019}~and~\cite{Herzog2016}, we found it sufficient to use only the inverse-dynamics term  
(the optimal torques $\tau^*$) and not the joint feedback part. 
This is due to the fact that we can identify the parameters 
of our dynamic model with sufficient accuracy as detailed in~\cite{Tournois2017}. 
That said, we carried out the simulation and experiment without any need of the feedback loop. 
\end{remark}

\section{\texorpdfstring{C$^3$}{C3} Whole-Body Controller} \label{wbc:soft}
Over soft terrain, the feet positions are non-stationary  and are allowed to deform the terrain. 
Thus, the rigid contact assumption of the stance task \eref{wbopt_rigid}
in the \gls{swbc} does not hold anymore and should be dropped.
To be \gls{c3}, the interaction between the stance feet 
and the soft terrain must be governed not just by  the robot dynamics but also by 
the soft contact dynamics. 
That said, the \gls{awbc} extends the \gls{swbc} by: 1)~modeling the soft contact dynamics  
and incorporating it as a \textit{stance task} similar to the control tasks in \sref{sec:controlTasks},
and 2)~encoding the stance task in the  \gls{wbopt} as a function of the decision
variables. 
The differences between the \gls{swbc} and the \gls{awbc} are highlighted in 
\textbf{\textcolor{RoyalBlue}{boldface}} in 
\algoref{eqn_wbopt}.
\begin{remark}
The term \textit{contact consistent} is a well-established term in the literature that 
was initially introduced in \cite{JaeheungPark2006}. 
It implies formulating the control structure to account for the contact with the environment.
The term \gls{c3} is an extension of the term contact consistent. 
Hence, \gls{c3} implies formulating the control structure to account for the \textit{compliant} contact with the 
environment.
\end{remark}

\subsection{\texorpdfstring{\texttt{c}$^3$}{c3}-Stance Task}
\label{sec:contactmodelling}
We model the soft contact dynamics with a simple explicit model (the \gls{kv} model). 
This consists of 3D linear springs and dampers 
normal and tangential to the contact point \cite{Neunert2018}. 
The normal direction of this impedance 
emulates the normal terrain deformation while 
the tangential ones emulate the shear deformation.  
Although several models that accurately emulate contact dynamics are available
\cite{Azad2010,Ding2013,Alves2015}, we implemented the \gls{kv} model for several reasons. 
First, since the model is linear in the parameters, it fits our \gls{qp} formulation. 
Second, estimating the parameters of the \gls{kv} model is computationally efficient. 
As a result, using this model,  we can run a learning algorithm online which would be challenging
if a model similar to \cite{Chang2017} is used. 
For a legged robot with point-like feet,  
for each stance leg $i$, we formulate the contact model in the world frame as
\begin{eqnarray}
\grfp{i} =    k_{\mrm{st},i}  p_i  + d_{\mrm{st},i} \dot{p}_i
\end{eqnarray}
where $k_{\mrm{st},i} \in \Rnum^{3\times3}$,
$d_{\mrm{st},i} \in \Rnum^{3\times 3}$,
$\grfp{i} \in \Rnum^3$, $ p_i \in \Rnum^3$, and $ \dot{p}_i  \in \Rnum^3$ 
are the terrain stiffness, the terrain damping,
the \gls{grfs}, the penetration 
and the penetration rate of the $i$-th stance leg, all expressed in the world frame, respectively (see \fref{fig:blockDiagramSTANCE}).
We define $ p_i$ and $ \dot{p}_i$ as
\begin{eqnarray}
p_i =  x_{\mrm{td},i} - x_{\mrm{st},i} \text{,} \hspace{5em } \dot{p}_i = 0 - v_{\mrm{st},i}
\label{eq:penetrationDef}
\end{eqnarray}
where  $x_{\mrm{td},i} \in \Rnum^3$ denotes the position of 
the contact point of foot $i$ at the touchdown in the world frame. 
By appending all of the stance feet, the contact model can be re-written in a compact form as
\begin{equation}
\grf = K_{\mrm{st}} p + D_{\mrm{st}} \dot{p} =  K_{\mrm{st}} (x_{\mrm{td}} - x_{\mrm{st}}) -  D_{\mrm{st}} v_{st}
\label{contactmodel1}
\end{equation}
where $K_{\mrm{st}}$ $\in R^{n_{\mrm{st}} \times n_{\mrm{st}}}$ and $D_{\mrm{st}}$ $\in R^{n_{\mrm{st}} \times 
n_{\mrm{st}}}$ are the block-diagonal  stiffness and 
damping matrices of the terrain of all the stance feet, respectively,
and $x_{\mrm{td}}$ $\in \Rnum^{n_{st}}$ are the touchdown positions of all the stance feet.

Similar to \sref{sec:controlTasks}, we deal with the contact model~\eref{contactmodel1} as 
another \gls{wbc} task (alongside the trunk and swing \eref{eq:swingTaskAccSTANCE} tasks). 
We can think of  \eref{contactmodel1} 
as a desired stance task that keeps the \gls{wbc} \gls{c3}. 
This stance task  is achieved by a Cartesian impedance at 
the stance foot which is represented by the 
impedance of the terrain ($K_{\mrm{st}}$  and $D_{\mrm{st}}$). 
This similarity makes us encode the contact model in the \gls{wbopt} as 
a stance constraint similar to what we 
did for the swing task in \sref{sec:wbopt}.
Hereafter, we refer to this stance task as the~\gls{c3}-stance task (see \fref{fig:blockDiagramSTANCE}).

\subsection{Whole-Body Optimization Revisited}
The \gls{c3}-stance task is included in the \gls{wbopt}  by writing the soft contact model 
\eref{contactmodel1} as a function of the decision variables. 
Ideally, we can directly reformulate  \eref{contactmodel1} 
as a function of $\grf$ and $\dot{v}_{\mrm{st}}$.
Indeed, $\dot{v}_{\mrm{st}}$ can be expressed as a function of the 
joint accelerations $\ddot{q}$ which is a decision variable (as explained in~\cite{Fahmi2019}).  
By numerically integrating $\dot{v}_{\mrm{st}}$ 
(once to obtain $v_{\mrm{st}}$ and twice to obtain $x_{\mrm{st}}$),
we can associate $\grf$ with $\dot{v}_{\mrm{st}}$.
This approach requires the knowledge of  $x_{\mrm{td}}$  to compute $p$ 
which might be prone to estimation errors and it requires  a reset  of the integrator at every touchdown.

We choose a more convenient approach which  
is to add the \textit{desired} foot penetration $\epsilon$
as an extra decision variable  in the \gls{wbopt} formulation. 
The difference between  $p$ and $\epsilon$ is that~$p$ is
the \textit{actual} 	penetration due to the interaction with the soft contact
while $\epsilon$ is the \textit{desired} penetration in the world frame generated from the optimization problem. 
Both variables imply the same physical phenomenon (the soft contact deformation). 
That said, 
we can rewrite $\grf$ in \eref{contactmodel1} 
as a function of $\epsilon$ and $\dot{\epsilon}$ (by numerically differentiating $\epsilon$)
without the previous knowledge of $x_{\mrm{td}}$, which is advantageous. 

To do so, 
$\epsilon$ is appended to the vector of decision variables~$u$ and regularized in \eref{wbopt_cost_function}.
Then, we incorporate \eref{contactmodel1} directly as a function of  $\grf$ and $\epsilon$ 
as in the equality constraint~\eref{wbopt_fgrf_epsilon}.
We numerically differentiate $\epsilon$ to obtain 
$\dot{\epsilon}_k = \frac{\epsilon_k - \epsilon_{k-1}}{\Delta t}$. 
To maintain physical consistency, we need to enforce an additional
constraint 
between the desired penetration $\epsilon$ and
the contact acceleration
($\ddot{\epsilon} = -\dot{v}_{\mrm{st}}$). 
This is encoded as an equality constraint as shown in \eref{wbopt_ddot_epsilon}. 
To do so, we numerically differentiate 
$\epsilon$ twice to obtain $\ddot{\epsilon}_k = \frac{\epsilon_k - 2\epsilon_{k-1} + \epsilon_{k-2}}{\Delta t^2}$. 
We also ensure the consistency of the physical contact model throughout the optimization
problem by ensuring that the penetration is always positive in 
\eref{wbopt_penetration_slack}. 

We also consider
the loading and unloading phase, explained in \cite{Fahmi2019}
 and \cite{Focchi2017}, to be terrain-aware. 
We tune the loading and unloading phase period  $T_{l/u}$ for each leg
to follow the settling time of a second order system response
that is a function of the terrain compliance
and the robot's mass \cite{Franklin2014}. 
Hence, $T_{l/u}$ is
\begin{equation}
	T_{l/u} = 4.6 / \sqrt{\frac{k_{\mrm{st,i}}}{m_e}}
\end{equation}
where $m_e$ is the equivalent mass felt at the robot's 
feet (\ie the weight of the robot $m_{R}$ spread 
across its stance feet~$m_e~=~m_{R} / n_{\mrm{st}}$)
and the constant in the numerator represents a $1\%$ steady-state error.

Finally,
the \gls{wbopt} 
\eref{wbopt_cost_function}, \eref{wbopt_physical_consistency}, \eref{wbopt_fgrf_epsilon}-\eref{wbopt_joint_limits} 
generates
optimal joint accelerations $\ddot{\vc{q}}_j^*$  and contact forces 
$\vc{\grf}^*$, that are  mapped into optimal joint torques $\tau^*$ 
and sent to the low-level controller
using the actuated part of the 
robot's dynamics as shown in~\eref{eq:torquesSTANCE}. 
Note that similar to the \gls{swbc}, we  found it sufficient to use only the inverse-dynamics term  
	(the optimal torques $\tau^*$) and not the joint feedback part.

As explained above, adding $\epsilon$ as a decision variable
involved adding  two constraints in the optimization
which increases the problem size and the computation time. 
Yet, we are still able to run the \gls{awbc} in real-time.  
The advantage of our approach is that the knowledge of the 
touchdown position  $x_{\mrm{td}}$ is not required.
We only need the previous two time instances of the penetration $\epsilon_{k-1}$
and $\epsilon_{k-2}$ that we already computed in the previous control loops.

\section{Terrain Compliance Estimation} 
\label{sec:softterrainestimation}
\begin{sidewaysfigure}
	\centering
	\includegraphics[width=0.98\textwidth]{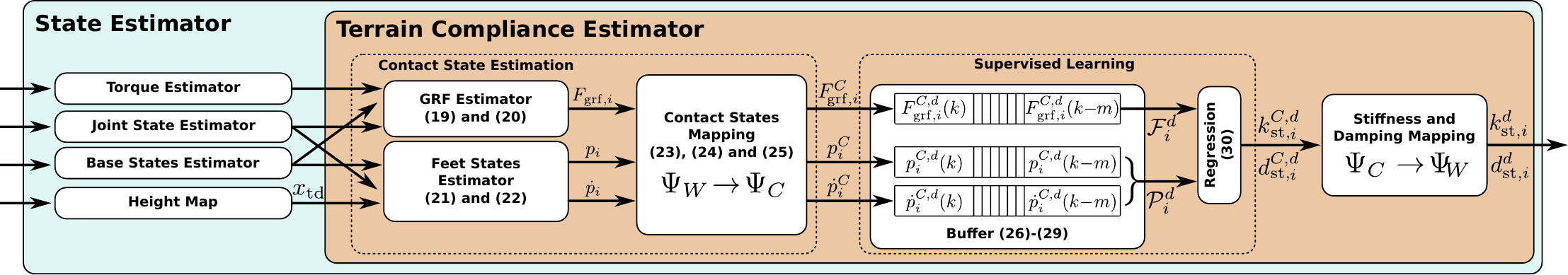}
	\caption{Overview of the \acrshort{ste}'s architecture inside the state estimator.}
	\label{tce_arch}
\end{sidewaysfigure}
The purpose of the \gls{ste} is to estimate \textit{online} the terrain parameters 
(namely $K_{\mrm{st}}$ and $D_{\mrm{st}}$) based on the states of the robot. 
It is a stand-alone algorithm that is decoupled from the \gls{awbc}.
The \gls{ste} uses the contact model \eref{contactmodel1}.
	Based on that, the current measurement of the
	\textit{contact states} (contact status $\alpha$, \gls{grfs} $\grf$, the penetration $p$, and the penetration rate $\dot{p}$) of each leg $i$ at every time step are required.
Given the contact states, we use supervised learning to learn the terrain parameters.
As shown in \fref{tce_arch}, the \gls{ste} consists of two main modules: 
	contact state estimation (\sref{sec_contact_state})
	and supervised learning (\sref{sec_supervised_learning}). 
	The contact state estimation module estimates the contact states and provides it to the supervised learning
	module that collects these data and computes the estimates of the terrain parameters.

\subsection{Contact State Estimation}
\label{sec_contact_state}
The contact states are estimated solely from the current states of the robot
	by the state estimator.
	The \grfs are estimated from the torques and the joint states, and
	the penetration and its rate are estimated from the floating base (trunk) states and the joint states. 
\subsubsection{GRFs Estimation}
To estimate the GRF, we use actuated part of the dynamics in \eref{eq_actuated} as

\begin{equation}
F_{\mrm{grf},i} = \alpha_i \vc{J}_{j,i}^{-T} (M_{a,i}\ddot{\vc{q}}_{i} + \vc{h}_{j,i} - \vc{\tau}_{j,i})
\label{grf_est}
\end{equation}
where $F_{\mrm{grf},i}$, $\vc{J}_{j,i}$, $M_{a,i}$, $\ddot{q}_{i}$,  $\vc{h}_{j,i}$, and $\vc{\tau}_{j,i}$
correspond to 
$F_{\mrm{grf}}$, $\vc{J}_{\mrm{st},j}$, $M_a$, $\ddot{q}$, $\vc{h}_j$, and $\vc{\tau}_j$ 
for the $i$-th leg, respectively. Additionally,  $\alpha_i$ is the contact status
variable that detects if there is a contact in the $i$-th leg or not. 
The contact is detected when the \gls{grfs} exceed a certain threshold $F_{\min}$.
Hence, $\alpha_i$ computed along the normal direction of the $i$-th leg $n_i$ as:
\begin{equation}
\alpha_i =\begin{cases}
1, & \text{if $n_i^T (\vc{J}_{j,i}^{-T} (M_{a,i}\ddot{\vc{q}}_{i} + \vc{h}_{j,i} - \vc{\tau}_{j,i}) \geq F_{\min}$}\\
0, & \text{otherwise}
\end{cases}
\label{contact_status}
\end{equation}

\subsubsection{Penetration Estimation}

	As shown in \eref{eq:penetrationDef}, we estimate the penetration and its rate
	using the stance feet positions $x_{\mrm{st},i}$ and velocities $v_{\mrm{st},i}$,
	and the touchdown position $x_{\mrm{td},i}$ all in the \textit{world} frame.
	To estimate the feet states in the world frame, we use the forward kinematics and the base state in the world frame.
	Thus, the penetration and its rate are written as
\begin{eqnarray}
	p_i &=& x_{\mathrm{td},i} - x_{\mrm{st},i}
	=x_{\mathrm{td},i} - x_{b} -  R_B^W x_{\mrm{st},i}^B
	\label{p_est}\\
	\dot{p}_i &=& - v_{\mrm{st},i}
	= - v_b - R_B^W v_{\mrm{st},i}^B - (\omega_b \times R_B^W) x_{\mrm{st},i}^B
	\label{dotp_est}
	\end{eqnarray}
	where
	$x_{b}$ $\in \Rnum^3$ and
	$v_b$ $\in \Rnum^3$
	are the base position and velocity in the world frame, respectively. 
	The terms
	$x_{\mrm{st},i}^B$ and
	$v_{\mrm{st},i}^B$
	are the stance feet position and velocity of the $i$-th leg in the base frame, respectively. 
	The terms
	$R_B^W$ $\in SO(3)$ and 
	$\omega_b$ are the rotation matrix mapping vectors from the base frame to 
	the world frame and the base angular velocity, respectively.
	$
	$
	The  touch down positions are  obtained using a height map.

\subsubsection{Contact States Mapping}
	Since, the \gls{kv} model consists of 3D linear springs and dampers,  
		normal and tangential to the contact point, this makes the stiffness and damping 
		matrices diagonal with respect to the contact frame. 
		However,  if expressed in the world frame, the 
		stiffness and damping matrices become dense.
		Thus, if we formulate the \gls{kv} model  in the contact frame rather than the world frame,
		we estimate less number of elements per matrix per leg: three elements instead of nine.
		Henceforth, the \gls{kv} model in the \gls{ste} should be formulated with respect to the contact frame
rather than the world frame to reduce the computational complexity.
To do so, the \grfs \eref{grf_est}, the penetration \eref{p_est}  and its rate \eref{dotp_est} 
	of the $i$-th leg 
	are transformed from the world frame $\Psi_W$ to the contact frame $\Psi_{C_i}$ as (see \fref{fig:blockDiagramSTANCE})
\begin{eqnarray}
F_{\mathrm{grf},i}^C &=& R_W^{C_i} \grfp{i} \label{grf_trans}\\
p_{i}^C &=&  R_W^{C_i} p_i \label{p_trans}\\
\dot{p}_{i}^C &=& R_W^{C_i} \dot{p}_i \label{pdot_trans}
\end{eqnarray}
where the superscript $\bullet^C$ refers to the contact frame and~$R_W^{C_i}$ is the rotation matrix mapping from the world $\Psi_W$ to the contact~$\Psi_{C_i}$ frames for the $i$-th leg. 
Note that the transformation~\eref{pdot_trans} 
is linear
since the contact frame is \textit{fixed} with respect to the world frame at the touch down position (\ie $\dot{R}_W^{C_i} = 0$).

\subsection{Supervised Learning}
\label{sec_supervised_learning}
Considering the contact model in the contact frame
and using the estimated contact states  \eref{grf_trans}-\eref{pdot_trans},
we learn the terrain parameters online via supervised learning.
In particular, we use weighted linear least squared regression.
The algorithm is treated as a batch algorithm with $m$-examples such that, 
at every time instant $k$, we gather samples from the 
previous $m$ time instances and 
compute the terrain parameters
\cite{Stulp2015}.

For the $k$-th time instant,
of the $i$-th leg in the $d\text{-th}$ direction ($d \in \{n_i,t_{1,i},t_{2,i}\}$, 
see \fref{fig:blockDiagramSTANCE}),
	the terms $F_{\mathrm{grf},i}^{C,d}(k)$, $p_{i}^{C,d}(k)$, and $\dot{p}_{i}^{C,d}(k)$
	are estimated as shown in \sref{sec_contact_state} where~$\bullet_i^d(k)$ refers to the $k$-th time instance of the $i$-th leg in the $d$-th direction.
That said, we construct the following objects (buffers) with size $m$
\begin{eqnarray}
\mathcal{F}_i^d &=& \mat{ F_{\mathrm{grf},i}^{C,d}(k) & \cdots &F_{\mathrm{grf},i}^{C,d}(k-m) }^T \label{tce1}\\
\mathrm{P}_i^d &=& \mat{p_{i}^{C,d}(k) & \cdots & p_{i}^{C,d}(k-m) }^T \label{tce2}\\
\dot{\mathrm{P}}_i^d &=& \mat{ \dot{p}_{i}^{C,d}(k) & \cdots & \dot{p}_{i}^{C,d}(k-m) }^T \label{tce3}\\
\mathcal{P}_i^d &=& \mat{\mathrm{P}_i^{d}&  \dot{\mathrm{P}}_i^{d}} \label{tce4}
\end{eqnarray}
where $\mathcal{F}_i^d \in \Rnum^{m}$
is the \gls{grfs} buffer and $\mathcal{P}_i^d \in \Rnum^{m \times 2}$
is the penetration, and penetration rate buffer.
Given $\mathcal{F}_i^d$ and  $\mathcal{P}_i^d$ as inputs and outputs of the learning algorithm respectively, 
we estimate the terrain impedance parameters as
$I_i^d   =\mat{k_{\mathrm{st},i}^{C,d} ~ d_{\mathrm{st},i}^{C,d}}^T \in \Rnum^{2}$
using the analytical solution
\begin{equation}
	I_i^{d}   =  (\mathcal{P}_i^{dT} W \mathcal{P}_i^d  )^{-1} \mathcal{P}_i^{dT} W  \mathcal{F}_i^d
	\label{tce5}
\end{equation}
where $k_{\mathrm{st},i}^{C,d}\in \Rnum$ and $d_{\mathrm{st},i}^{C,d} \in \Rnum$ are the terrain stiffness and damping parameters expressed in the contact frame.
The matrix $W \in \Rnum^{m \times m}$ is a weighting matrix used to 
penalize the error on most recent sample compared to the less 
recent ones and thus, giving more importance to  the most recent samples. 

All of the legs in the learning algorithm are decoupled. 
We found it advantageous to treat each leg separately because the robot can 
be  standing on a different terrain at each foot.

\subsection{Implementation Details}
\algoref{tce_alg} sketches the entire \gls{ste} process.
To initialize the buffers, we acquire samples when the robot is at full stance
	and return the first estimate of $I_i^d$ once the buffers are full.
	After initialization, 
	we acquire samples and update the buffers only when the leg is at stance. 

The buffers are continuously updated in a sliding window fashion. 
	When a leg finishes the swing phase and is at a new touch down,
	it continues to use the previous samples from the previous stance phase.  This is advantageous since it gives a smooth transition between terrains, but it adds a delay. 
\begin{remark}
	Since the \gls{awbc} formulation is based in the world frame, it is essential 
		to map the estimated stiffness and damping matrices
		back to the world frame before providing them to
		the \gls{awbc} (see \fref{tce_arch}). 
	
\end{remark}

\begin{remark}
	The \gls{ste} can be used with any arbitrary terrain geometry given the terrain normal and thus $R_W^{C_i}$. The terrain normal $n_i$ at the contact point $i$ can be
	provided by a height map that is generated via an RGBD sensor.
	\end{remark}

\begin{algorithm}[t!]
	\caption{Terrain Compliance Estimation}\label{tce_alg}
	\begin{algorithmic}[1]
		\State initialize the buffers ($\mathcal{F}_i^d$ and $\mathcal{P}_i^d$) and $I_i^d$ 
		\For{each iteration $k$} 
		\For{each leg $i$} 
		\If{leg is in contact ($\alpha_i == 1$)}\hfill  \eref{contact_status}
		\For{each direction $d$} 
		\State estimate  $\grfp{i}^{d}(k)$ \hfill \eref{grf_est}
		\State estimate $ p_{i}^d(k)$  \hfill \eref{p_est}
		\State estimate $\dot{p}_{i}^d(k)$ \hfill \eref{dotp_est}
		\State transform $\grfp{i}^{d}(k)$ into 	$F_{\mathrm{grf},i}^{C,d}(k)$       \hfill \eref{grf_trans}
		\State transform $p_{i}^{d}(k)$ into 	$p_{i}^{C,d}(k)$           \hfill \eref{p_trans}
		\State transform $\dot{p}_{i}^{d}(k)$ into 	$\dot{p}_{i}^{C,d}(k)$ \hfill \eref{pdot_trans}
		\State update buffers $\mathcal{F}_i^d$ and  $\mathcal{P}_i^d$ \hfill \eref{tce1}-\eref{tce4}
		\State solve for $I_i^d$ \hfill \eref{tce5}
		\EndFor
		\State map the estimated parameters to $\Psi_W$
		\EndIf
		\EndFor 
		\EndFor 
	\end{algorithmic}
\end{algorithm}

\section{Experimental Setup}
\label{sec:exp_setup}
\subsection{State Estimation}
\label{sec_exp_setup_state}
We implemented our approach on \acrshort{hyq} 
\cite{Semini2011} which is equipped with a variety of sensors. 
Each leg contains two load-cells, one torque sensor, 
and three high-resolution optical encoders.
A tactical-grade \gls{imu} (KVH 1775) is mounted on its trunk.
Of particular importance to this experiment is the Vicon \gls{mcs}.
It is a multi-camera infrared system capable of measuring the pose of an object with high accuracy.
During experiments, an accurate and non-drifting 
estimate of the position of the feet in the world frame 
is required to calculate the real penetration for the \gls{ste}.
Typically, \acrshort{hyq} works independently of external sensors (\eg \gls{mcs} or GPS), 
however, soft terrain presents problems for state estimators \cite{Henze2016}.
This was re-affirmed in experiment. 

The current state estimator~\cite{Nobili2017}
relies upon fusion of \gls{imu} and leg odometry data at a high frequency and 
uses lower frequency feedback from cameras or lidars to correct the drift.  
The leg odometry makes the assumption that the ground is rigid. 
On soft terrain, the estimator has difficulties in determining when a foot is in 
contact with the ground (\ie is the foot in the air, or compressing the surface?). 
These errors cause the leg odometry signal to drift jeopardizing the 
estimation. Although, incorporating vision information could be a possibility 
to correct for the drift in the estimation,
improving state estimation on soft 
terrain is an ongoing  area of research and is  out 
of the scope of this paper. 

Despite the drifting problem, we used the current state estimator \cite{Nobili2017} in our \gls{wbc}
because the planner in \fref{fig:blockDiagramSTANCE} 
has a re-planning feature that
makes our \gls{wbc} robust against a drifting state estimator \cite{Focchi2018}.
However, the \gls{ste} still requires an accurate and non drifting 
estimate of the feet position in the world frame. 
Therefore, to validate the \gls{ste}, we 
used an external \gls{mcs} that completely eliminates the drift problem.
The \gls{mcs} measures the pose of a special marker array placed 
on the head of the robot. Then  the position 
of the feet in the world frame was calculated online 
by using the \gls{mcs} measurement and 
the forward kinematics of the robot. 

\subsection{Terrain Compliance Estimator~(TCE) Settings}
In this work we used a sliding window of
$m = 250$~samples (or 1 \unit{s} for a control loop 
running at 250 \unit{Hz}).
Despite the general formulation, in this paper we estimate the terrain parameters
only for the direction normal to the terrain, 
and assume that the 
tangential directions are the same. 
We carried out the simulation and experiment on a horizontal plane. Thus, the rotation matrix $R_W^{C_i}$ is identity. 
Furthermore,
we  did not estimate the damping parameter due to the inherent noise in the feet velocity signals
that would jeopardize the estimation. 
The damping term $D_{\mrm{st}} v_{st}$ in \eref{contactmodel1}
is less dominant in computing the \grfs  
compared to the stiffness term. 
This is because
the feet velocities in the world frame $v_{st}$ are usually orders of magnitude smaller than the penetration
during stance, and
the damping parameter $D_{\mrm{st}}$  is usually orders of magnitude smaller
than the stiffness parameter as shown in \cite{Ding2013}.

\subsection{Tuning of the Low Level Control}
During experiments, we found that the low level torque loop
creates system instabilities when interacting with soft environments.

In particular, when we used the same set of (high) torque 
gains in the low level control loop tuned for rigid terrain, we noticed 
joint instabilities when walking over soft terrain.
This is because interacting with soft terrain reduces the stability margins of the system. 
Thus, keeping a high bandwidth in the inner torque loop given the 
reduced stability margin will cause system instability. 
In our previous work~\cite{Focchi2016}, we experimentally validated that increasing 
the torque gain of the inner loop can 
indeed cause system instabilities.
In fact, this is a well know issue in haptics \cite{Hulin2014}. 
As a result, reducing the bandwidth by decreasing the torque gains in the inner torque loop was necessary to address these instabilities.

Our control design is a nested architecture consisting of the \gls{wbc} 
and the low level torque control in which,
both control loops contribute to the system stability \cite{Focchi2016,Mosadeghzad2013}.
Over soft terrain, 
the dynamics of the environment also plays a role 
and must be considered in analyzing the stability of the system. 
That said, there is a nontrivial relationship between soft terrain and 
the stability of a nested control loop architecture, 
and a formal and thorough analysis is an ongoing work.

\section{Results}
\label{sec:results}
In this section, we evaluate the proposed approach on \gls{hyq} in simulation and experiment. 
We compare \textit{three} approaches:
the \gls{swbc} which is the baseline,
the \gls{awbc} which is our proposed \gls{wbc} without the \gls{ste},
and \gls{stance} which incorporates both the \gls{awbc} and \gls{ste}.
We show the extent of improvement given by the \gls{awbc} controller with respect to the \gls{swbc} as well as the 
importance of the \gls{ste} 
during locomotion over multiple terrains with different compliances. 
We set the same parameters and gains throughout the entire simulations and experiments, 
unless mentioned otherwise. 
The results are shown in the accompanying 
video\footnote{Link: \href{https://youtu.be/0BI4581DFjY}{\texttt{https://youtu.be/0BI4581DFjY}}}.

\subsection{Simulations}
\label{sec:sim}
To render soft terrain in simulation, we used the \gls{ode} physics engine \cite{Smith2005}. 
We used \gls{ode} because it is easily integrable with our framework, and
it is numerically fast and stable for stiff and soft contacts \cite{Catto2011}. 
Moreover, \gls{ode} can render soft contacts that emulates physical
parameters (using the SI units  \unit{N/m} and  \unit{Ns/m} for springs and dampers, respectively)
unlike other engines that uses non-physical ones \cite{Erez2015}.
\gls{ode}'s implicit solver uses linear springs and dampers for their soft constraints
which fits perfectly with our contact model \eref{contactmodel1}.
In this way, we have a controlled simulation environment   
where we can emulate any terrain compliance 
by manipulating its stiffness $K_t$ and damping~$D_t$ parameters
similar to our contact model. 
Throughout the simulation, 
we use four types of terrains with the following parameters: 
soft $T_1$ ($K_t=3500$~\unit{N/m}), 
moderate $T_2$ ($K_t = 8000$~\unit{N/m}), 
stiff $T_3$ ($K_t = 10000$~\unit{N/m}), 
and rigid $T_4$ ($K_t =  2\times 10^6$~\unit{N/m})
 all with the  same damping ($D_t = 400$~\unit{Ns/m}).

\subsubsection{Locomotion over Multiple Terrains} 
\label{sec:sim3terrain}
%

\begin{sidewaysfigure}
	\centering
	\includegraphics[width=\textwidth]{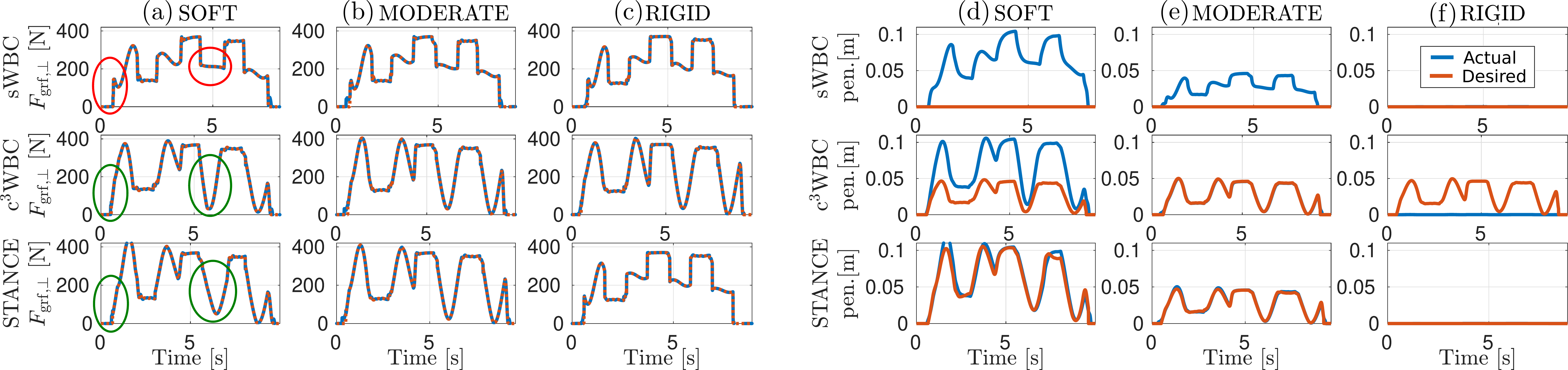}
	\caption[Comparison of \acrshort{swbc}, \acrshort{awbc}, and \acrshort{stance} over three type of terrains]{Simulation. Comparison of \acrshort{swbc}, \acrshort{awbc}, and \acrshort{stance} over three type of 
		terrains: 
		soft $T_1$ ($K_t=3500$ \unit{N/m}), 
		moderate $T_2$ ($K_t = 8000$ \unit{N/m}), 
		and rigid $T_4$ ($K_t =  2\times 10^6$ \unit{N/m})
		all with the  same damping ($D_t = 400$ \unit{Ns/m}).
		(a)-(c): The actual and desired  contact forces in the normal direction of the \acrshort{lf} leg for one gait cycle.
		(d)-(f): The actual $p$ and desired $\epsilon$ penetration in the normal direction of the \acrshort{lf} leg for one gait 
		cycle.
		The red and green ellipses highlight the performance of the three approaches in adapting to soft terrain. 	  
	}
	\label{fig:softTerrainSimComparison}
\end{sidewaysfigure}

\begin{table}[!t]
	\renewcommand{\arraystretch}{2}
	 \captionsetup{justification=centering}
	\caption{Mean Absolute Tracking Error (MAE) [N] of the \acrshort{grfs}
		 in Simulation using \acrshort{swbc}, \acrshort{awbc} and \acrshort{stance}
		 over Multiple Terrains.
		 }
	\label{tab_SIMmeanAbsErrorGRFs}
		\centering
	\begin{tabular}{cccc}
		\hline \hline	
		Terrain & \textbf{\acrshort{swbc}} & \textbf{\acrshort{awbc}} & \textbf{\acrshort{stance}}\\
		\hline
		Soft         	&7.7261  &   7.4419  &  \textbf{6.3547} \\
		Moderate    	& 8.0594 &   \textbf{7.4585}  &  7.9889 \\
		Rigid  	        &\textbf{4.889}   & 6.6523     & 5.128 \\		
		\hline \hline
	\end{tabular}
\end{table}
We evaluate the three approaches with the 
robot walking at $0.05$ \unit{m/s} over the terrains:
 $T_1$ (soft), $T_2$ (moderate), and $T_4$ (rigid).
We provided the \gls{awbc} with the terrain parameters of the moderate terrain~$T_2$
for all the three simulations. 
We do that in order to test the performance of \gls{awbc} if given
the real terrain parameters (in case of $T_2$)
or inaccurate parameters (in case of  $T_1$ and $T_4$).
In this simulation, 
we compare the actual values of $ \grfp{\perp}$ against the optimal values 
$\grfp{\perp}^*$(solution of the \gls{wbopt}) as well as the actual penetration $p$ against the 
desired penetration~$\epsilon$ of the \gls{lf} leg.
We have omitted the other three feet for space as all four legs have the same performance.
The results are shown in \fref{fig:softTerrainSimComparison}.
The \gls{mae} of the \grfs in these simulations are presented in 
\tref{tab_SIMmeanAbsErrorGRFs}.
The \gls{mae} of the \grfs is defined as: $\text{MAE} = \frac{1}{T} \int_{0}^{T} \vert
	F_{\mrm{grf}} - F_{\mrm{grf}}^* \vert dt $.

\fref{fig:softTerrainSimComparison}a
captures the effect of the three approaches
on the \grfs over soft terrain. 
We can see that a \gls{wbc} based on a rigid contact assumption (\gls{swbc}) assumes that it can achieve
an infinite bandwidth from the terrain and thus supplying an instantaneous change in the \gls{grfs} as highlighted by 
the red ellipses 
in \fref{fig:softTerrainSimComparison}a. 
On the other hand, \gls{stance} and \gls{awbc} were both 
capable of attenuating this effect as highlighted by the green ellipses.
For the reasons explained earlier in this paper and in \cite{Grandia2019},
instantaneous changes in the \gls{grfs} are undesirable over soft terrain. 
This resulted in an improvement in the tracking of the \grfs  
in \gls{stance} and \gls{awbc} compared to \gls{swbc} as shown
 in \tref{tab_SIMmeanAbsErrorGRFs}. 
Moreover, by comparing \gls{awbc} and \gls{stance} over soft terrain $T_1$, 
we can see that the shape of the \grfs did not differ. 
However, the tracking of the \grfs in \gls{stance} is better than the 
\gls{awbc}.
This shows that suppling the \gls{awbc} with the incorrect values of the terrain 
parameters 
deteriorates the \grfs tracking performance.
\fref{fig:softTerrainSimComparison}b shows the \grfs on a moderate terrain.
Since the \gls{awbc} is provided with the exact terrain parameters of $T_2$, 
we can perceive the \gls{awbc} as \gls{stance} with a perfect \gls{ste} on moderate terrain. 
As a result, \tref{tab_SIMmeanAbsErrorGRFs} 
shows that in this set of simulations, 
\gls{awbc} outperformed \gls{stance}
in
the \grfs tracking.
 This shows  that a  more accurate \gls{ste}
 can result in a better \grfs tracking.
Additionally, 
\fref{fig:softTerrainSimComparison}c shows the \grfs on rigid terrain.
We can see that  the \gls{swbc} resulted in a typical (desired) shape of the \grfs 
for a crawl motion in rigid terrain \cite{Focchi2018}. 
\gls{stance} showed a shape of the \grfs similar to the \gls{swbc} which
is expected since the \gls{ste} provided \gls{stance} with  parameters similar to the rigid terrain.
However, for \gls{awbc}, the \grfs shape did not change compared to the other three terrains. 
As shown in \tref{tab_SIMmeanAbsErrorGRFs}, the best tracking to the \grfs was by the 
\gls{swbc}, which was expected since the \gls{swbc} was designed for rigid terrain. 
However, \gls{swbc} was only slightly better than \gls{stance} due to small estimation errors from the \gls{ste}.

\fref{fig:softTerrainSimComparison}a-c show the superiority of \gls{stance} compared
to \gls{swbc} and \gls{awbc}. \gls{stance} adapted to the three terrains
by estimating their parameters and supplying them to the \gls{wbc}.
This resulted in changing the shape of the \grfs accordingly 
that improved the tracking of the  \gls{grfs}.  
Unlike \gls{stance}, the \gls{swbc} and the \gls{awbc} both are contact consistent for only \textit{one} type of terrain
which resulted in a deterioration of the \grfs tracking over the other types of terrains.
The advantages of \gls{stance} compared to \gls{swbc} and \gls{awbc} are also shown in 
\fref{fig:softTerrainSimComparison}d-f. 
Since the \gls{swbc} is always assuming a rigid contact, 
the penetration $\epsilon$
was always zero throughout the three terrains. 
Similarly, since the \gls{awbc} alone is aware only of one type of terrain, 
it is always assuming the same contact model, 
in which the desired penetration $\epsilon$ was similar throughout the three terrains.  
\gls{stance}, however, was capable of predicting the 
penetration correctly for all the three terrains. 

In general,
even if the contact model is for soft contacts, \gls{stance} was capable of correctly
predicting the penetration of the robot even in rigid terrain (zero penetration). 
This resulted in \gls{stance} adapting to rigid, soft and moderate terrains by means
of adapting the \grfs and correctly predicting the penetration. 

\subsubsection{Longitudinal Transition Between Multiple Terrains}
\label{sec:simCrossingTerrain}
\begin{figure}[!t]
	\centering
	\includegraphics[width=\columnwidth]{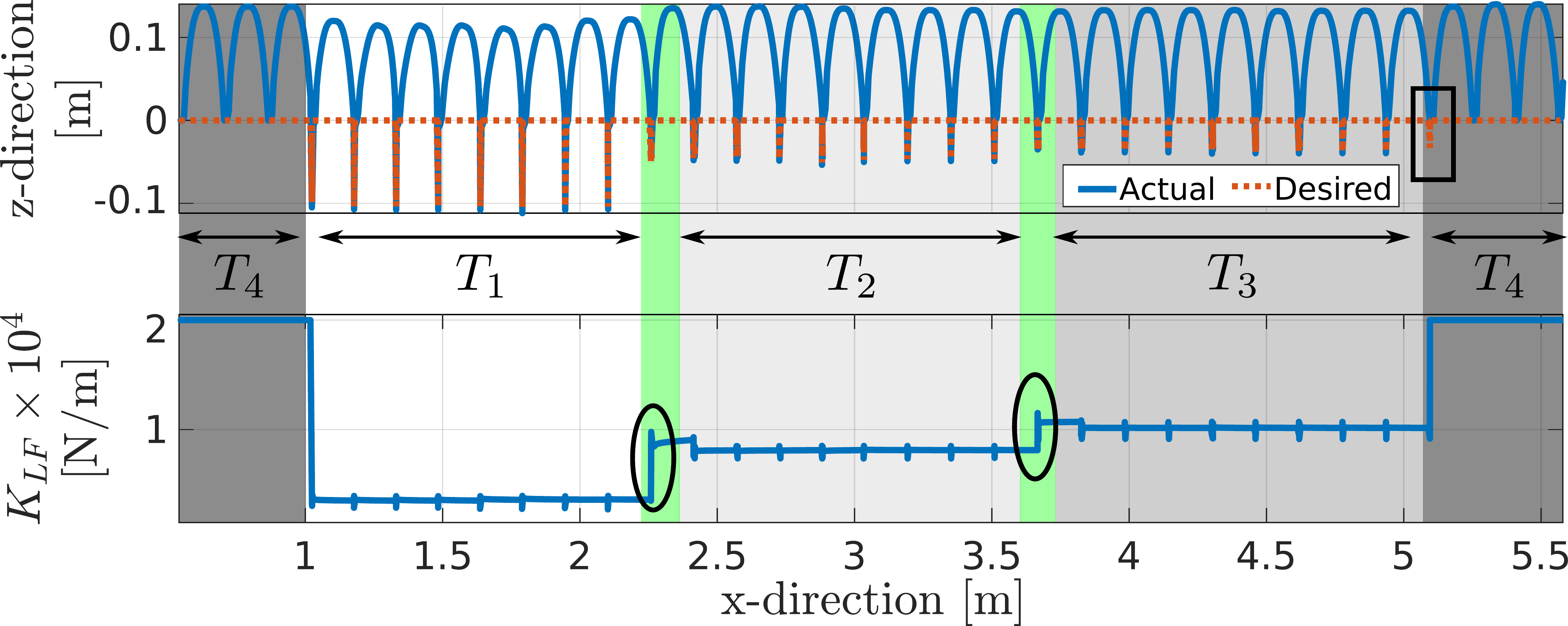}
	\caption[Traversing multiple terrains of different compliances.]{Simulation. Traversing multiple terrains of different compliances ($T_4$, $T_1$, $T_2$, $T_3$, 
	$T_4$). 
		Top: Tracking of the desired terrain penetration of the \acrshort{lf} leg  in the xz-plane. 
		Bottom: Estimated terrain stiffness of the \acrshort{lf} leg. For readability purposes we only plot estimated values less than $2\times10^4$.
		The green shaded areas highlight the overlap 
		between terrains that results in higher estimated stiffness (black ellipses).}
	\label{fig:simCrossingTerrain}
\end{figure}
We show the adaptation of \gls{stance} when walking and transitioning between 
multiple terrains. 
We test the accuracy of the \gls{ste} and 
the effect of closing the loop of the \gls{awbc} with the \gls{ste} 
on the feet trajectories and terrain penetration. 
In this simulation, \gls{hyq} is traversing five different terrains, starting and
ending with a rigid terrain: $T_4$, $T_1$, $T_2$, $T_2$, $T_4$.
The results are presented in \fref{fig:simCrossingTerrain}. 
The top plot presents the actual foot position against the desired penetration
$\epsilon$
of the \gls{lf} leg in the xz-plane of the world frame. 
The origin of the z-direction (normal direction) 
is the uncompressed terrain height. 
Thus, trajectories below
zero represent the penetration of the \gls{lf} leg. 
The bottom plot shows the history of the
estimated terrain stiffness of the  \gls{ste} of the
\gls{lf} leg.  
Table \ref{tab_stesim} reports the mean, standard deviation, 
and percentage error\footnote{ The percentage error is defined as:
	\% Error $= \vert \frac{\text{Estimate}-\text{Actual}}{\text{Actual}} \vert \times100$
	}
of the estimated terrain stiffness of the \gls{lf} leg 
against the ground truth value set in \gls{ode}.
The table shows that the \gls{ste} had an estimation accuracy below $2$\% for the soft terrains $T_1$, $T_2$ and $T_3$. 
However, the estimation accuracy of the rigid terrain was lower than that of the soft terrains. 
This is expected since on a rigid terrain, the penetrations are (almost) zero. 
Thus, a small inaccurate penetration estimation due to any model errors could 
result in a lower estimation accuracy. 
Apart from the rigid case, the standard deviation 
is always below~6$\%$ of the ground truth value. 
\fref{fig:simCrossingTerrain} shows that \gls{stance}
is always \gls{c3},  
the actual foot position is always consistent with the 
desired penetration
during stance. 
We can see that, when \gls{hyq} is standing over rigid terrain, 
both the actual foot position and desired penetration are zero. 
As \gls{hyq} walks, over the soft terrains, the penetration is highest in the softest terrain
and smallest in the stiffest terrain. 

In the simulation environment, we overlapped the 
terrains to prevent the feet from getting stuck between them. 
This overlap created a transition (highlighted in green in the figure) which resulted in a stiffer 
terrain. 
The overlap was captured by the \gls{ste} and resulted in a slight increase in the estimated 
parameters as highlighted by the two ellipses in the lower plot.
We also noticed a lag in estimation, 
due to a filtering effect,
since the \gls{ste} is using the most recent $m$-samples.
As highlighted by the black box in \fref{fig:simCrossingTerrain}, 
\gls{hyq} was on rigid terrain (actual penetration is zero) while \gls{stance}  still perceived
it as being on $T_3$ (desired penetration is non-zero).

\begin{table}
	\centering
	\caption{Mean $\mu$ [N/m], Standard Deviation $\sigma$ [N/m], and Percentage Error  of the 
		Estimated Terrain Stiffness 
		of the \acrshort{lf} Leg in Simulation.}
	\label{tab_stesim}
	\renewcommand{\arraystretch}{2}
	\begin{tabular}{c ccc}
		\hline \hline
		Terrain & Actual Stiffness & Mean $\mu$ $\pm$ STD $\sigma$ & \% Error\\
		\hline
		$T_1$ & 3500 & 3530 $\pm$ 200 & 0.9\% \\ 
		$T_2$ & 8000 & 8110 $\pm$ 400 & 1.4\% \\
		$T_3$ & 10000 &  10110 $\pm$ 400~& 1.1\%  \\
        $T_4$ & 2000000 & 2240000 $\pm$ 740000 & 12\%  \\
		\hline \hline
	\end{tabular}
\end{table}

\subsubsection{Aggressive trunk maneuvers} 
\label{sec:simAggressiveTrunk}
\begin{figure}[!t]
	\centering
	\includegraphics[width=0.88\columnwidth]{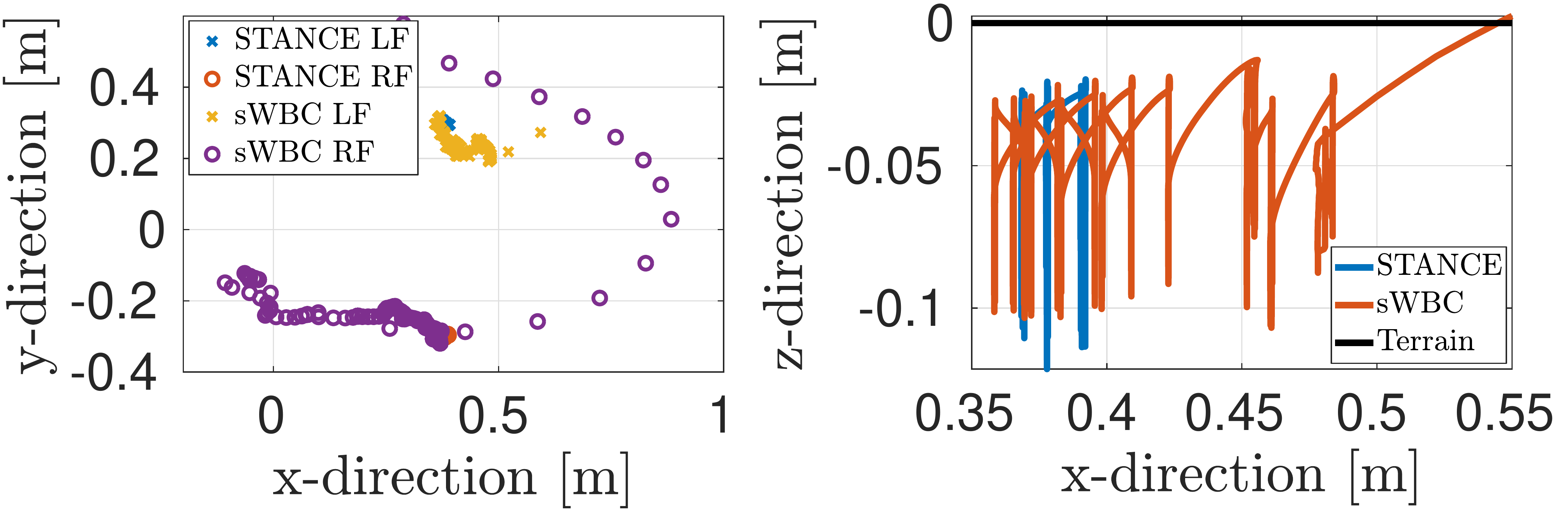}
	\caption[Comparing \acrshort{swbc} and \acrshort{stance}
	under aggressive trunk maneuvers.]{Simulation. 
		Comparing \acrshort{swbc} and \acrshort{stance}
		under aggressive trunk maneuvers. 
		Left: Top view of the front feet (\acrshort{rf} and \acrshort{lf}) positions. 
		Right: Side view of the  \acrshort{lf}  position.
			}
	\label{fig:simAggressiveTrunkTraj}
\end{figure}
We tested \gls{swbc} and \gls{stance}  under aggressive trunk maneuvers by
commanding  desired sinusoidal trajectories at the robot's height ($0.05$ \unit{m} amplitude and 
$1.8$ \unit{Hz} frequency) and at roll orientation ($0.5$ \unit{rad} amplitude and $1.5$ \unit{Hz} frequency) 
over the soft terrain $T_1$.
The results are shown in \fref{fig:simAggressiveTrunkTraj}. 
The left plot shows a top view of the actual front feet (\gls{lf} and \gls{rf}) positions in the world frame.
The right plot shows a side view of the actual  \gls{lf} foot position in the world frame. 
We notice that 
the feet of \gls{hyq} are always in contact with the terrain
in \gls{stance}
which is expected since \gls{stance} is \gls{c3}.
Unlike \gls{stance},
the feet did not remain in contact with the terrain
in the \gls{swbc}. 
This is clearly seen in \fref{fig:simAggressiveTrunkTraj} where
\gls{hyq} lost  contact multiple times.
 This resulted
in the robot falling over in the \gls{swbc} case as shown in the video.

\subsubsection{Speed Test}
\label{sec:simSpeedTest}
\begin{figure}[!t]
	\centering
	\includegraphics[width=0.9\columnwidth]{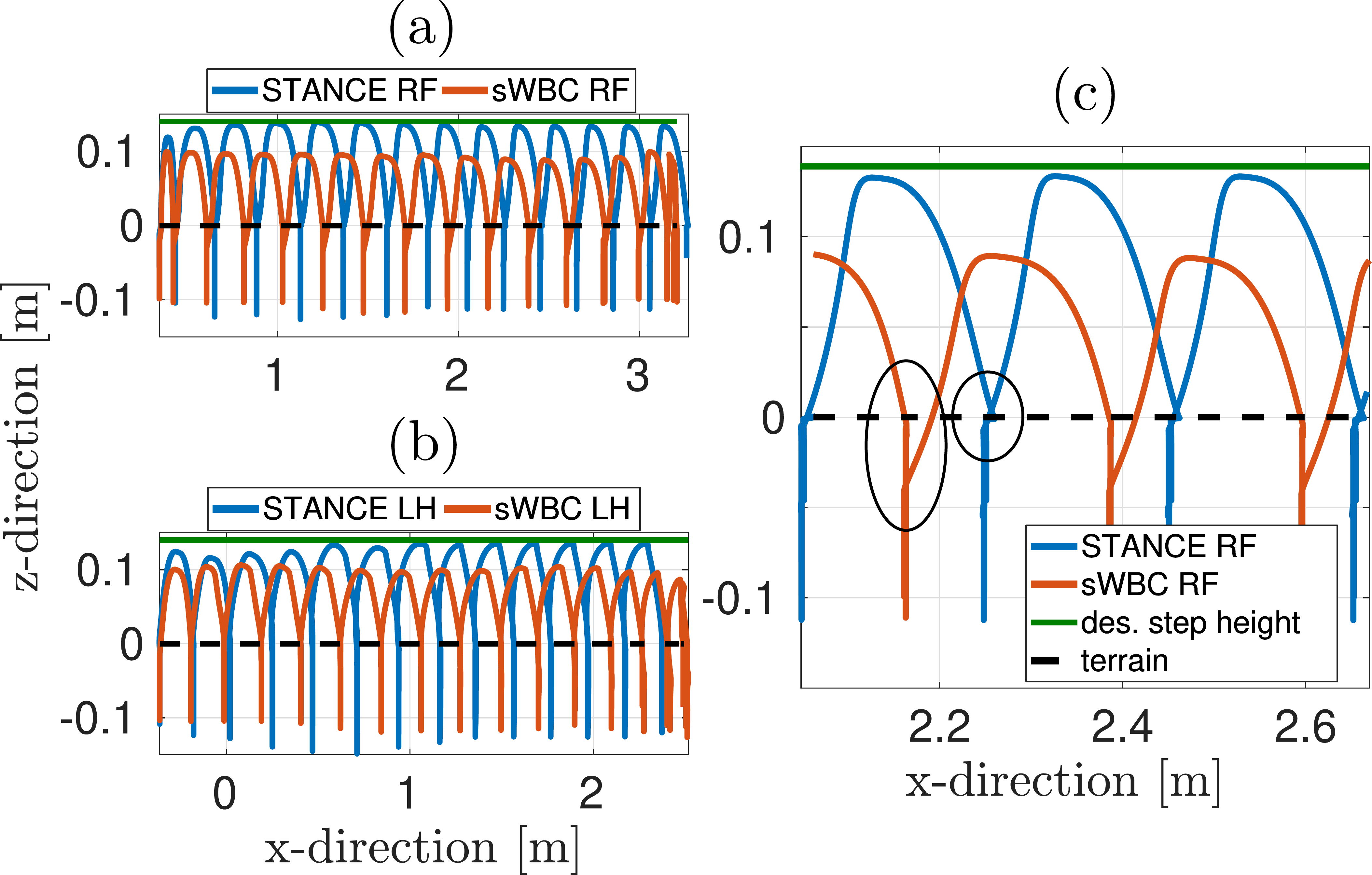}
	\caption[Speed test. 
	Increasing the desired forward velocity from $0.05$ to $0.3$ \unit{m/s}.]
	{Simulation. Speed test. Increasing the desired forward velocity 
		from $0.05$ to $0.3$ \unit{m/s}.
	Left: Side view of the \acrshort{rf} (a) and \acrshort{lh} (b) positions. 
	Right:~(c)~Closeup section of the top left plot. 
	The green lines are the desired step height. 
	The black dashed lines are the terrain height.
}
	\label{fig:speedTest}
\end{figure}
We carried out a simulation 
where \gls{hyq} walks over soft terrain $T_1$,
starting with a forward 
velocity of $0.05$  \unit{m/s} until it reaches $0.3$  \unit{m/s} 
with an acceleration of~$0.005$~\unit{m/s$^2$}.
In this simulation, 
we compare \gls{stance} against the \gls{swbc}. 
\fref{fig:speedTest}a and \fref{fig:speedTest}b
show the actual
trajectories
of the  \gls{rf} and \gls{lh} legs
in the world frame, respectively. 
\fref{fig:speedTest}c shows a closeup  section of
the
\gls{rf} leg's trajectory. 
The simulation shows that \gls{stance} was \gls{c3} over the entire simulation
while the \gls{swbc} was not.

In particular, \gls{stance} was able to remain in contact with the terrain that allowed \gls{hyq} to  start the swing phase directly from the terrain height. 
Unlike \gls{stance},   the \gls{swbc} is not terrain aware
and did not remain \gls{c3} which resulted in
	starting the swing trajectory while still being inside the deformed terrain. 
This is highlighted by the two ellipses in the right plot.
Additionally, the compliance contact consistency property of \gls{stance} enabled the robot to 
maintain the desired step clearance
(i.e., achieving the desired step height of $0.14$ \unit{cm})
compared to \gls{swbc}. 
Most importantly, 
as shown in the accompanying video, 
the \gls{swbc} failed to complete the simulation and could not achieve the final desired forward velocity;  It fell at a speed of $0.21$~\unit{m/s}.
Note that  both approaches could reach higher velocities
with a more dynamic gait (trot). However, 
this simulation is not focusing on analyzing the
maximum speed that the two approaches can reach
but rather the differences between these approaches
at a higher crawl speeds. 

\subsubsection{Power Test}
\label{sec:powertest}
\begin{figure}[t!]
	\centering
	\includegraphics[width=\columnwidth]{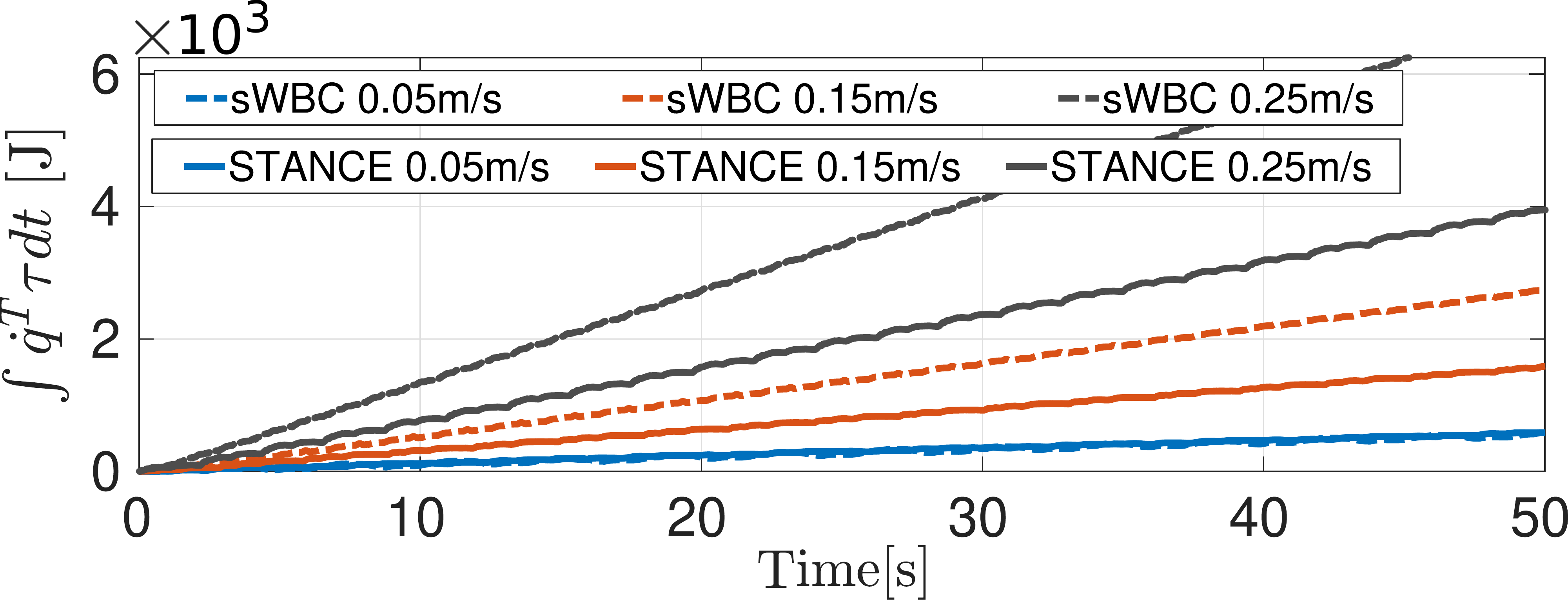}
	\caption{Simulation. Power consumption comparison between
		\acrshort{swbc} and \acrshort{stance} with different forward velocities (0.05~\unit{m/s}, 0.15~\unit{m/s} and 	 0.25~\unit{m/s}). 
	}
	\label{fig:powerConsumption}
\end{figure}
In this test, 
we compare the power consumption using
\gls{stance} and \gls{swbc}
on
\gls{hyq} during walking over the  soft terrain $T_1$
at different forward velocities ($0.05$~\unit{m/s}, $0.15$ \unit{m/s} and $0.25$ \unit{m/s}).
\fref{fig:powerConsumption} presents the energy plots of \gls{stance} and \gls{swbc}. 
The plot shows that \gls{stance}  requires less power than the \gls{swbc} 
because it knows how the terrain will deform. 
\gls{stance} exploits the terrain interaction
to achieve the motion. 
The difference in consumed energy is 
negligible at $0.05$ \unit{m/s} but becomes significant
at higher speeds.

\subsection{Experiment}
\label{sec:exp}
We validated the simulation presented in \sref{sec:sim} on the real platform. 
We analyzed \gls{swbc},  \gls{awbc} (with fixed terrain parameters) and \gls{stance}
as well as the performance of the \gls{ste} module itself. 

A foam block of 
$160$~\unit{cm}~$\times~120$~\unit{cm}~$\times~20$~\unit{cm} 
was selected as a soft terrain
for these experiments.
To obtain a ground truth of the foam stiffness, 
we carried out indentation tests on a $50$ \unit{cm}$^3$ sample  of 
the foam with a stress-strain machine that covers the 
range of penetration of interest for our robot (below $0.15$ \unit{cm}). 
The indentation test  showed a softening behavior of the foam with an average stiffness 
of $2400$ \unit{N/m}.
The \gls{mae} of the \grfs of the upcoming experiments are shown in
 \tref{tab_EXPmeanAbsErrorGRFs}.

\subsubsection{Locomotion over Soft Terrain}
\label{sec:expsSoftTerrain}
\begin{figure}[!t]
	\centering
	\includegraphics[width=\columnwidth]{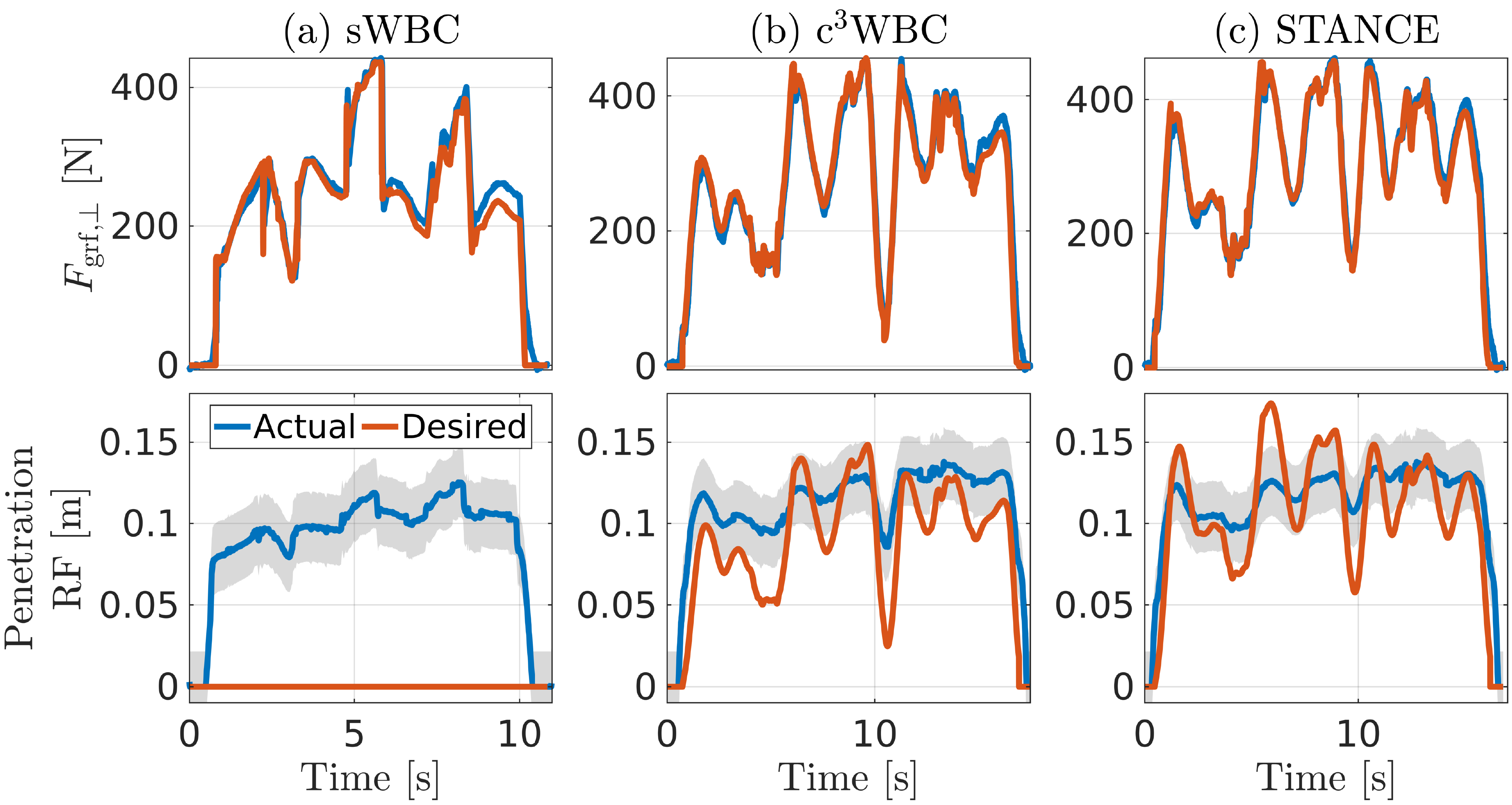}
	\caption
	[Comparing \acrshort{swbc}, \acrshort{awbc}   
	and \acrshort{stance} over a soft foam  
	block.]
	{Experiment. 
		Comparing \acrshort{swbc}, \acrshort{awbc}   
		and \acrshort{stance} over a soft foam  
		 block ($K_t = 2400$ \unit{N/m}).  
Top: Tracking of the \grfs of the \acrshort{rf} leg. 
		Bottom: Tracking of the foot penetration. 
		The gray shaded areas represent the uncertainty of the measurements.
	}
	\label{fig:expsSoftTerrain}
\end{figure}
\begin{table}[t]
	\centering
	\caption{Mean Absolute Tracking Error (MAE) [N] of the \acrshort{grfs}
		using \acrshort{swbc}, \acrshort{awbc} and \acrshort{stance}
		under Different Sets of Experiments.}
	\label{tab_EXPmeanAbsErrorGRFs}
	\renewcommand{\arraystretch}{2}
	\begin{tabular}{lccc}
		\hline \hline	
		Description & \textbf{\acrshort{swbc}} & \textbf{\acrshort{awbc}} & \textbf{\acrshort{stance}}\\
		\hline
		Soft Terrain (Sec. \ref{sec:expsSoftTerrain})    &	73.9042		& 68.5581 	 & 
		\textbf{61.8207	}	\\
		Longitudinal Trans. (Sec. \ref{sec:expCrossingTerrain})     &	 70.5276	&  64.2636 	 & 	
		\textbf{60.6285} 		\\
		Lateral Trans. (Sec. \ref{exp_lat})  		&	73.0766		& 	- 	 & 	\textbf{53.0107}		
		\\
		\hline \hline
	\end{tabular}
\end{table}
\begin{table}
	\centering
	\caption{Mean $\mu$ [N/m], Standard Deviation $\sigma$ [N/m], and Percentage Error of the 
		Estimated Terrain Stiffness of the Four Legs in Experiment over Soft Terrain ($2400$~\unit{N/m}).}
	\label{tab_ste_exp_soft}
	\renewcommand{\arraystretch}{2}
	\begin{tabular}{c ccc}
		\hline \hline
		Leg & Mean $\mu$ $\pm$ STD $\sigma$ & \% Error\\
		\hline
		LF & 2186 $\pm$ 166 & 9\% \\ 
		RF & 2731 $\pm$ 173 &14\% \\
		LH & 2368 $\pm$ 317~& 1\%  \\
		RF & 2078 $\pm$ 331 & 13\%  \\
		\hline \hline
	\end{tabular}
\end{table}

In this experiment, \gls{hyq} is walking over the foam with a forward velocity of $0.07$~\unit{m/s}
using the three approaches. 
The results  are presented in \fref{fig:expsSoftTerrain}
that shows the actual and desired $\grfp{\perp}$ and penetration of the \gls{rf} leg.
The shaded gray area in the lower plots of \fref{fig:expsSoftTerrain} represents the uncertainty in the estimation 
of the foot position 
(see \sref{sec_exp_setup_state}).
In these experiments, all three approaches performed well; none of them failed. 
However, the shape of \grfs 
were different within the three approaches. 
As in \sref{sec:sim3terrain}, 
since \gls{swbc} is rigid contact consistent, the desired \grfs were designed for rigid contacts. 
Unlike \gls{swbc}, \gls{stance} is \gls{c3}, which was capable of changing the shape of the \gls{grfs}. 
This is highlighted in \tref{tab_EXPmeanAbsErrorGRFs} in which \gls{stance} 
outperformed \gls{swbc} in the tracking of the \gls{grfs}. 

In simulation,
 when we provided the \gls{awbc} with the true value
of the  stiffness, the
\gls{mae} of the \gls{grfs} was better. 
However, in this experiment, providing the value obtained
from the indentation tests to the \gls{awbc} resulted in a worse \grfs \gls{mae}.
This outperformance of \gls{stance} compared to the \gls{awbc}
in this experiment could be because of the \gls{ste}. 
To clarify, 
the actual terrain compliances are not constant, 
but since the \gls{ste} is online, it is able to capture these changes in the 
terrain compliances as well as model errors. 
As shown in the accompanying video, 
\gls{stance} had a smoother transition during crawling compared to \gls{swbc}. 
We found the robot transitioning from swing to stance more aggressively in \gls{swbc} 
than \gls{stance}. Such smooth behavior was also noticed in \cite{Grandia2019}.

\tref{tab_ste_exp_soft} shows the mean, standard deviation, 
and percentage error 
of the estimated terrain stiffness of all the four legs 
against the ground truth value ($2400$~\unit{N/m}) obtained from the indentation tests.
The table shows that the accuracy of the \gls{ste} in simulation is better than in experiments. 
This is expected since in simulation, the \gls{ste} has a perfect knowledge of the feet penetration. 
However, the accuracy of our \gls{ste} is better compared to \cite{Bosworth2016} in which the percentage error
exceeded 50\% (the actual stiffness was more than double that of the estimated one in \cite{Bosworth2016}).

\subsubsection{Longitudinal Transition Between Multiple Terrains}
\label{sec:expCrossingTerrain}
\begin{figure}[!t]
	\centering
	\includegraphics[width=1.0\columnwidth]{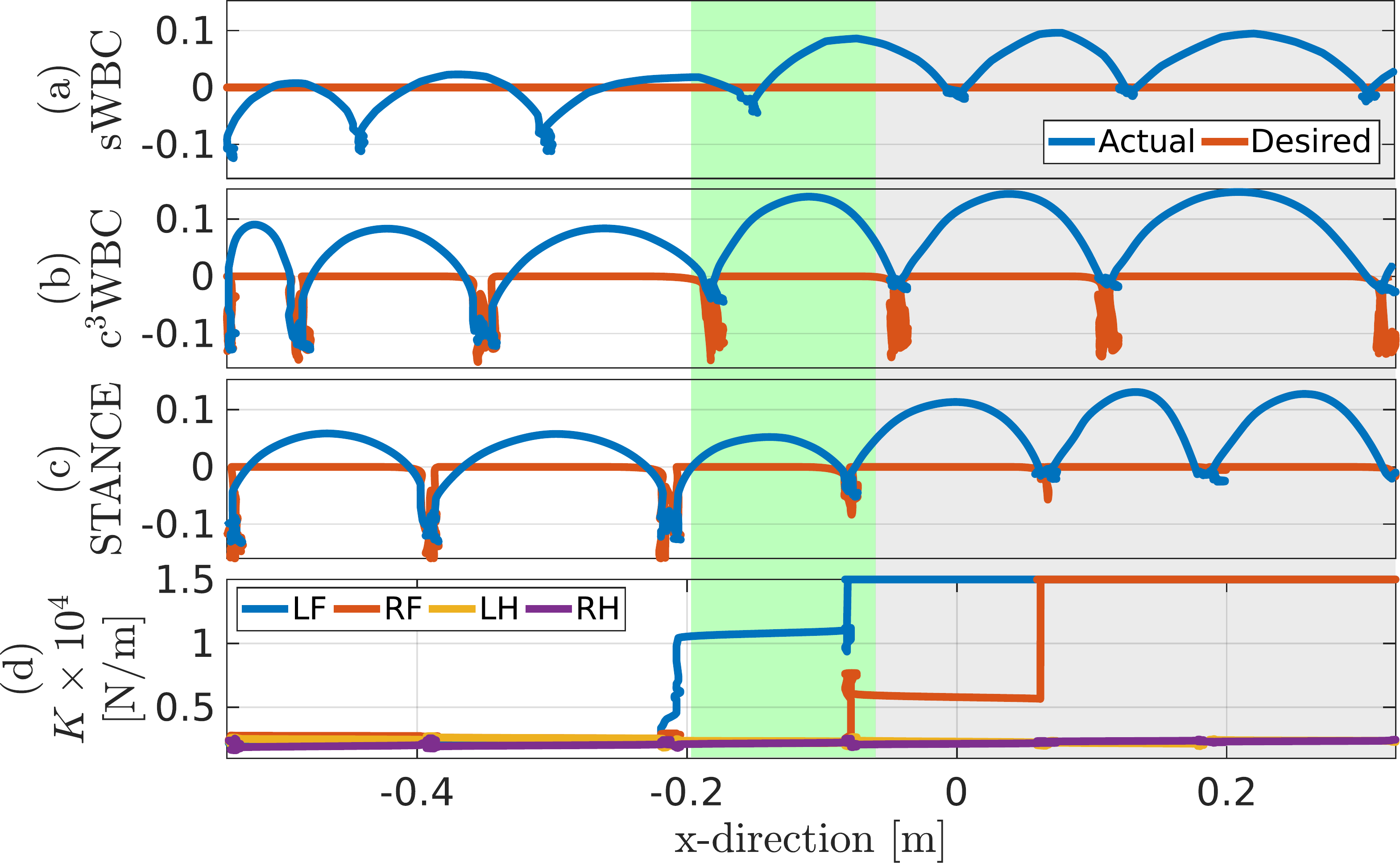}
	\caption[Longitudinal transition from soft to rigid terrain.]
	{Experiment.
		Longitudinal transition from soft to rigid terrain. 
		The first three plots show the tracking of the 
		desired foot penetration of \acrshort{rf} leg using the three approaches
		(the \acrshort{swbc}, \acrshort{awbc} with fixed terrain stiffness  and \acrshort{stance}.
		The fourth plot shows the stiffness estimated by the \acrshort{ste} for the four legs.}
	\label{fig:longitudinalArrangement}
\end{figure}
Similar to \sref{sec:simCrossingTerrain},  we compare the three approaches while
transitioning between the foam block and a rigid pallet. 
We added a pad between between the two terrains to avoid the 
foot getting stuck (see \fref{fig:photos}a).
\fref{fig:longitudinalArrangement}a-c
show the actual position and the desired penetration of the \gls{rf} leg 
in the  xz-plane for the three approaches. 
\fref{fig:longitudinalArrangement}d shows the estimated terrain stiffness of the \gls{ste} for all four  feet.

From \fref{fig:longitudinalArrangement}a-b we see that both \gls{swbc} and \gls{awbc} 
did not adapt to terrain changes. 
Since both controllers are designed for a specific constant terrain,
the desired penetration did not change from soft to rigid. 
In the \gls{swbc}, there is no tracking of the penetration,
and in the \gls{awbc}, the tracking of the penetration is good only when the leg is on the foam 
where the stiffness is consistent to the one used in the controller.
On the other hand, as shown in \fref{fig:longitudinalArrangement}c-d, \gls{stance} changes its parameters when 
facing a 
different terrain; it was capable of adapting its desired penetration to the type of terrain. 
In fact, the desired penetration was non-zero on soft terrain and was almost zero on rigid terrain. 
This again resulted in \gls{stance} achieving the 
best \grfs tracking as shown in  \tref{tab_EXPmeanAbsErrorGRFs}.

\fref{fig:longitudinalArrangement}d shows the importance of having a \gls{ste} for each leg. 
The estimated terrain parameters are different between the legs where 
the hind legs are on the foam while the rigid ones transitioning from foam to rigid. 
The figure also shows that the \gls{lf} leg walked over the rigid terrain before the \gls{rf} 
and that the \gls{ste} captures the 
intermediate stiffness estimation due to the rubber pad (see video).

\subsubsection{Lateral Transition Between Multiple Terrains}
\label{exp_lat}

Unlike the previous experiment, we set the foam and the pallet laterally as shown in \fref{fig:photos}c 
and in the accompanying video.
This is a more  challenging scenario for stability reasons. 
In particular, the robot must extend its leg further in the soft terrain 
maintain the trunk's balance. 
Consequently, since the width of \gls{hyq}'s torso is smaller than its length, 
the \gls{zmp} is more likely to get out of the support polygon.
The \gls{grfs} \gls{mae} in \tref{tab_EXPmeanAbsErrorGRFs} show that \gls{stance} can outperform \gls{swbc} during 
both
longitudinal and lateral transitions.

\subsubsection{External Disturbances over Soft Terrain}
\begin{figure}[!t]
\centering
\includegraphics[width=0.95\columnwidth]{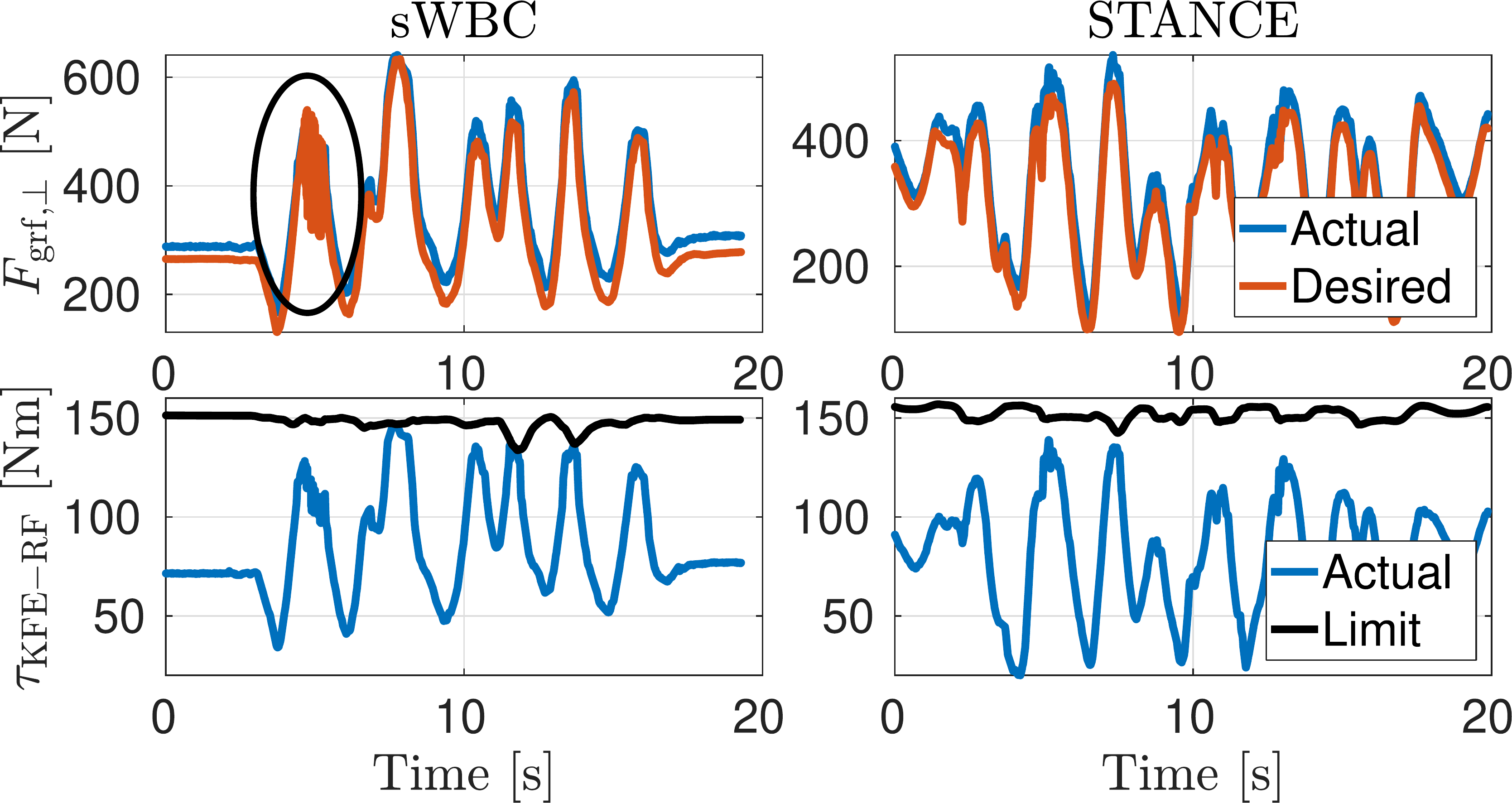}
\caption[The \acrshort{swbc} and \acrshort{stance}
under disturbances over soft terrain.]
{Experiment. The \acrshort{swbc} and \acrshort{stance}
under disturbances over soft terrain. 
Top: 
The actual and desired $\grfp{\perp}$ in \acrshort{swbc} and \acrshort{stance}, respectively.
Bottom: 
The actual torque and torque limits of 
the Knee Flexion-Extension (KFE) joint of the \acrshort{rf} 
leg
in \acrshort{swbc} and \acrshort{stance}, respectively.
}
\label{fig_exp_dist}
\end{figure}
In this experiment, we test the \gls{swbc} and \gls{stance} 
when the user applies a disturbance on \gls{hyq}.
The results are shown in \fref{fig_exp_dist}. 
The top plots show the actual and desired $\grfp{\perp}$ in \gls{swbc} and \gls{stance}, respectively.
The bottom plots show the actual torque and torque limits of the \gls{kfe} joint of the \gls{rf} leg
in \gls{swbc} and \gls{stance}, respectively.
In the accompanying video, we can qualitatively see that with \gls{stance}, the feet of \gls{hyq} keep moving 
to remain \gls{c3} with the terrain. 
On the other hand, the \gls{swbc} kept its feet stationary. 
This behavior was also reported by \cite{Henze2016}. 

Most importantly,   we noticed that \gls{hyq}
reaches the torque limits in the \gls{swbc}
as shown  in \fref{fig_exp_dist}.
 However, in \gls{stance}, since the robot was
constantly moving its feet, hence redistributing its forces, 
the torque limits were not reached. 
This behavior was also reflected on the \grfs 
in which, the \grfs were resonating in the \gls{swbc} as highlighted by 
the ellipse in \fref{fig_exp_dist}.

\subsubsection{\gls{ste}'s Performance over Multiple Terrains}

\begin{table}[t!]
\centering
\caption{Mean $\mu$  \unit{[N/m]} and Standard Deviation $\sigma$ [\unit{N/m}]
of the Estimated Terrain Stiffness of the Four Legs in Experiments (see \fref{fig:photos}b).}
\label{tab_ste_exp}
\renewcommand{\arraystretch}{2}
\begin{tabular}{ccc}		
\hline \hline	
Leg & Mean $\mu$ $\pm$ STD $\sigma$ \\
\hline
LF &	448400 $\pm$ 165100    \\     
RF 		&55200 $\pm$ 48400\\
LH 	&2645000 $\pm$ 336000\\
RH  &  1393000 $\pm$ 442000 \\
\hline \hline
\end{tabular}
\end{table}

We analyze the performance of the \gls{ste} on \gls{hyq} over multiple terrains with various softnesses. 
The softness of the four used terrains are shown in \fref{fig:photos}b. 
The estimated stiffness (mean and standard deviation) under each leg is shown in \tref{tab_ste_exp}.
As shown in the table, the robot can  differentiate between the types of 
terrain.
Although we did not measure the true stiffness value of these terrains,
we can observe their softness 
in the video and \fref{fig:photos}b
and compare it to the values in  \tref{tab_ste_exp}.

\subsection{Computational Analysis}

\gls{stance} is running online which means that we can
estimate the terrain compliance (using the TCE) continuously while walking, 
and run the entire framework without breaking real-time requirements. 
We validated the first argument by showing that indeed the \gls{ste} 
can continuously estimate the terrain compliance. 
Hereafter, we validate the second argument by analyzing the computational complexity of 
\gls{stance} and compare it against the \gls{swbc}. 
Since our \gls{wbc} framework is running at $250~\unit{Hz}$, 
it is essential that the computation does not exceed the $4~\unit{ms}$ time frame.  
Hence, 
we conducted a simulation in which we calculated the time 
taken to process the entire framework 
without the lower level control 
(ie., the state estimator, the planner and the \gls{wbc})
that is running on a different real-time thread at $1~\unit{kHz}$. 
We compared the computation time 
on an Intel Core i$7$ quad core CPU
in the case of \gls{stance} 
(the \gls{awbc} and the \gls{ste}) and the \gls{swbc}.
We used the same parameters and gains as in 
\sref{sec:sim3terrain}.
The results show that the average processing time taken was 
$0.68~\unit{ms}$ and $0.74~\unit{ms}$
for the \gls{swbc} and \gls{stance} respectively. 
In both cases, the maximum computation time was always below
$2~\unit{ms}$.

\section{Conclusions}
We presented a soft terrain adaptation algorithm called \gls{stance}:
\textbf{S}oft \textbf{T}errain \textbf{A}daptation a\textbf{N}d \textbf{C}ompliance \textbf{E}stimation.
\gls{stance} can adapt online to any type of terrain compliance (stiff or rigid). 
\gls{stance} consists of two main modules:
a compliant contact consistent whole-body controller~(\gls{awbc})
and a terrain compliance estimator~(\gls{ste}).
The \gls{awbc} 
extends our previously implemented \gls{wbc} (\gls{swbc}) \cite{Fahmi2019}, 
such that it is contact consistent
to any type of compliant terrain given the terrain parameters.
The \gls{ste} estimates online the terrain compliance and closes the loop with the \gls{awbc}. 
Unlike previous works on \gls{wbc}, \gls{stance} does not assume that the ground is rigid. 
Stance is computationally lightweight and it overcomes the limitations of the  
previous  state of the art approaches. 
As a result, \gls{stance} can efficiently traverse multiple terrains with different compliances. 
We validated \gls{stance} on our quadruped robot \gls{hyq} 
over multiple terrains of different stiffness in simulation and experiment. 
This, to the best of the authors' knowledge, is the first experimental validation on a legged robot
of closing  the loop  with a terrain estimator.

Incorporating the  terrain knowledge makes \gls{stance}  \gls{c3}.
This means that 
\gls{stance} is able to
generate smooth \grfs that are physically consistent with the terrain,
and continuously adapt the robot's feet to remain
in contact with the terrain.
As a result, 
the tracking error of the \grfs and the power consumption were reduced, 
and the impact during contact interaction was attenuated. 
Furthermore, 
\gls{stance} is more robust in challenging scenarios.
As demonstrated, \gls{stance} made it possible to perform
aggressive maneuvers and walk at high walking speeds over soft terrain
compared to the state of the art \gls{swbc}. 
In the standard case, the contact is lost because 
the motion of the terrain is not taken into account. 
On the other hand, there are minor differences
in performance between \gls{stance} and the \gls{swbc}
for less dynamic motions.

\gls{stance} can efficiently transition between multiple terrains 
with different compliances,
and each leg was able to independently 
sense and adapt to the change in terrain compliance. 
We also tested the capability of the \gls{ste} in discriminating between different terrains. 
The insights gained in simulation have been confirmed in experiment. 
 
In future works, 
we plan to implement an algorithm to improve the \gls{ste}. 
In particular, we plan on using 
onboard sensors, such as a camera,
 instead of relying on the external measurements from an MCS.
We also plan to explore other non-linear contact models in 
the \gls{ste} and the \gls{awbc}. 

%

\newpage~\thispagestyle{plain} \newpage
\glsresetall \chapter[On State Estimation for Legged Locomotion over Soft~Terrain]{On State Estimation\\ for Legged Locomotion over Soft~Terrain}\label{chap_lsens}

\lsensPaper
\boldSubSec{Abstract}
Locomotion over soft terrain remains a challenging problem for legged robots. 
Most of the work done on state estimation for legged robots 
is designed for rigid contacts, 
and does not take into account the physical parameters of the terrain.
That said, this letter answers the following questions: 
how and why does soft terrain affect state estimation for legged robots? 
To do so, 
we utilized a state estimator that fuses IMU measurements with leg odometry
that is designed with rigid contact assumptions.
We experimentally validated the state estimator with the HyQ robot trotting
over both soft and rigid terrain. 
We demonstrate that soft terrain negatively affects state estimation for legged robots, 
and that the state estimates have a noticeable drift over soft terrain compared to rigid terrain.
\newpage
\section{Introduction}\label{sec_introduction}
Quadruped robots are advancing towards being fully autonomous
as can be seen by their recent development 
in research and industry, and their remarkable agile capabilities
\cite{Semini2019,Bledt2018,Raibert2008}. 
This demands quadruped robots to be robust while 
traversing a wide variety of unexplored complex non-flat terrain.
The terrain may not just vary in geometry, but also in its physical properties 
such as terrain impedance or friction.
Reliable state estimation is a major aspect
for the success of the deployment of quadruped robots
because
most locomotion planners and control strategies
rely on an accurate estimate of the pose and velocity
of the robot. 
Furthermore, reliable state estimation is essential, 
not only for locomotion (low-level state estimation),
but also for autonomous navigation and inspection tasks that are 
emerging applications for quadruped robots (task-level state estimation).

To date, 
most of the work done on state estimation for legged robots 
are based on filters that fuse multiple sensor modalities.
These sensor modalities mainly include 
high frequency inertial measurements and kinematic measurements (\eg leg odometry),
as well as other low frequency modalities (\eg cameras and lidars) to correct the drift.

For instance, 
an \gls{ekf}-based sensor fusion algorithm has been proposed by~\cite{Nobili2017} 
that fuses \gls{imu} measurements, leg odometry, stereo vision, and lidar.
In \cite{Ma2016}, a similar algorithm has been proposed that fuses 
\gls{imu} measurements, leg odometry, stereo vision, and GPS.
In \cite{Bledt2018}, a nonlinear observer 
that fuses \gls{imu} measurements and leg odometry has been proposed.
In \cite{Fink2020}, a state estimator fuses a \gls{ges} nonlinear attitude
observer based on \gls{imu} measurements with leg odometry to provide bounded velocity estimates.
The global stability is important for cases when the robot may have fallen over whereas typical 
\gls{ekf}-based works may diverge. 
The bounded velocity estimates help to decrease drift in the unobservable position estimates. 
Finally, an approach similar to \cite{Fink2020} has been proposed in \cite{Hartley2020}.
This approach proposed an invariant \gls{ekf}-based sensor fusion algorithm that includes IMU measurements, contact sensor dynamics, and leg odometry. 

\begin{figure}[tb]
\centering
\includegraphics[width=0.92\textwidth]{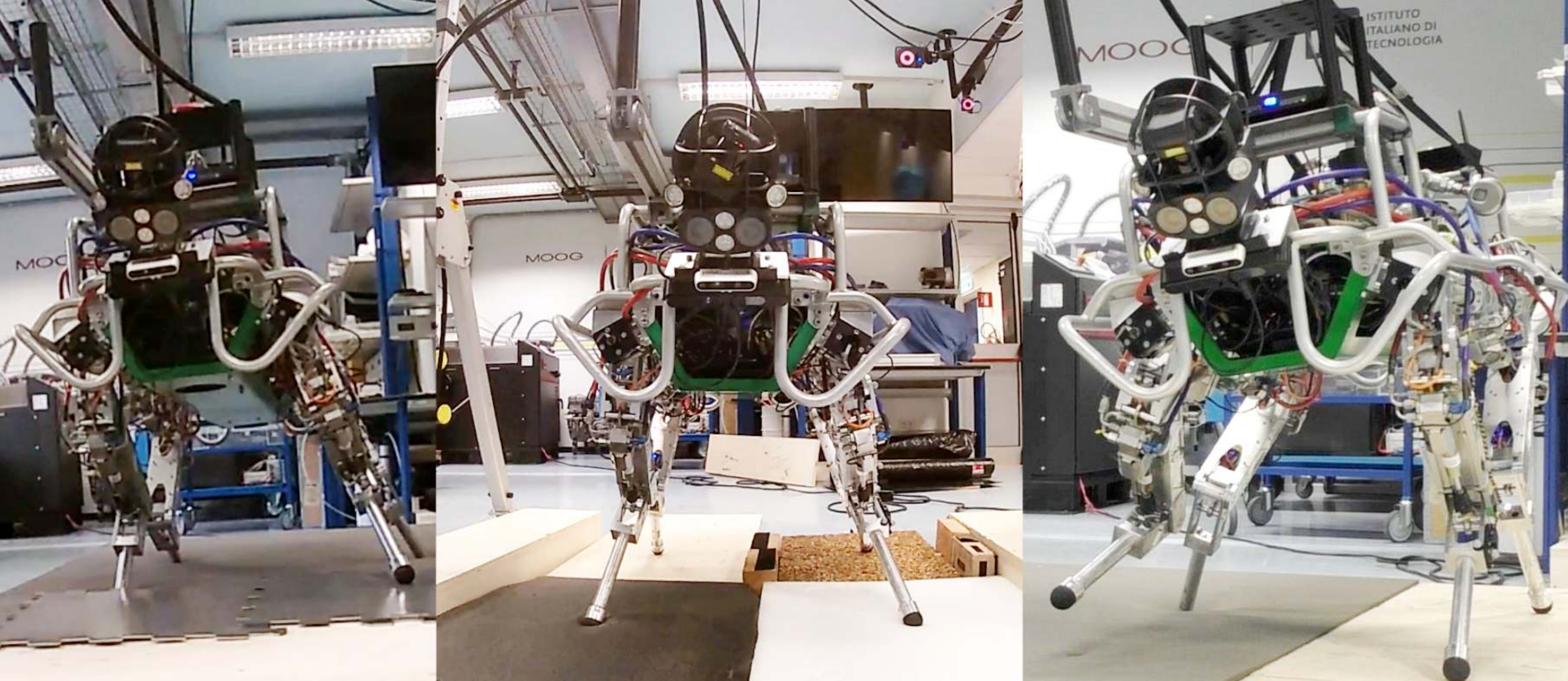}
\caption{HyQ traversing multiple terrains of different compliances.\label{fig_hyq_soft_terrain}}
\end{figure}

The aforementioned state estimators 
are shown to be reliable on stiff terrain. 
Yet, 
over soft terrain (as shown in~\fref{fig_hyq_soft_terrain}), 
the performance of these state estimators starts to decline. 
Over soft terrain,
the state estimator has difficulties 
determining when a foot is in contact with the ground. 
For instance, the state estimator has difficulties determining 
if the foot is in the air, 
if the foot is applying more force than the terrain (terrain compression),
if the terrain itself is applying more force than the foot (terrain expansion), 
or if the foot and the terrain are applying the same force (rigid terrain). 
This results in a large position estimate drift, and it was reported in our previous work~\cite{Fahmi2020} where  
we noted that we encountered difficulties because of state estimation over soft terrain.
Apart from our previous work, 
other works also mention that state estimation over soft terrain is a challenging task, 
\eg~\cite{Wish2020, Henze2018}. 
Yet, 
to the authors' knowledge,
literature has not yet discussed the question on how soft terrain affects the state estimation.

The contributions of this work are
the experimental analysis and formal study on:
the effects of soft terrain on state estimation, the reasons behind these effects, and simple ways to improve state estimation. 
This letter is building upon our previous work on soft terrain adaptation~\cite{Fahmi2020} 
and on state estimation~\cite{Fink2020}.

The rest of this letter is organized as follows: 
\sref{sec_modeling} describes
the robot model,
the onboard sensors,
and how to estimate the \gls{grfs} acting on the robot.
\sref{sec_state_est} explains the state estimator used in this letter, 
and how to estimate the base velocity of the robot using leg odometry. 
\sref{sec_results} details the results of our experiment and demonstrates how soft terrain affects state estimation.
Finally, \sref{sec_conclusion} presents our conclusions.

\section{Modeling, Sensing, and Estimating}
\label{sec_modeling}

In this letter,
we consider the quadruped robot HyQ~\cite{Semini2011} 
shown in~\fref{fig_hyq_soft_terrain}.
Each leg has three actuated joints.
Despite experimenting on a specific platform, the problem is generic in nature and it applies equally to any legged robot.
Furthermore, by using the $90$~kg HyQ robot, a~heavy and strong platform, we are exciting more dynamics. 

\subsection{Notations}
We introduce the following reference frames:
the body frame $\BF$ which is located at the geometric center of the trunk (robot torso), and
the navigation frame $\NF$ which is assumed inertial (world frame).
The basis of the body frame are orientated forward, left, and up.
To simplify notation, the \gls{imu} is located such that the accelerometer measurements are directly measured in $\BF$.

\subsection{Kinematics and Dynamics}
Assuming that all of the external forces are exerted on the feet, the dynamics of the robot is
\begin{equation}\label{eq:fbdynamics}
M(\bar{x})\ddot{\bar{x}}+ h(\bar{x},\dot{\bar{x}})=\bar\tau
\end{equation}
where
$\bar{x}=\begin{bmatrix}x^T&\eta^T&q^T\end{bmatrix}^T\in\R^{18}$ is the generalized robot states,
$\dot{\bar{x}}\in\R^{18}$ is the corresponding generalized velocities,
$\ddot{\bar{x}}\in\R^{18}$ is the corresponding generalized accelerations,
$x\in\R^3$ is the position of the base,
$\eta\in\R^3$ is the attitude of the base,
$q\in\R^{12}$ is the vector of joint angles of the robot,
$M\in\R^{18\times18}$ is the joint-space inertia matrix,
$h$ is the vector of Coriolis, centrifugal and gravity forces,
$\bar\tau=(\begin{bmatrix}0&\tau^T\end{bmatrix}^T-JF)\in\R^{18}$,
$\tau\in\R^{12}$ is the vector of actuated joint torques,
$J\in\R^{18\times 12}$ is the floating base Jacobian, and
$F\in\R^{12}$ is the vector of external forces (\ie \gls{grfs}).

We solve for the \gls{grfs} $F_\ell$ of each leg $\ell$ using the actuated part of the dynamics in~\eqref{eq:fbdynamics}.
\begin{equation}\label{eq:grf}
F_\ell = -\alpha_\ell(J_\ell^T(q_\ell))^{-1}(\tau_\ell-h_\ell(\bar{x}_\ell,\dot{\bar{x}}_\ell))
\end{equation}
\mbox{$F_\ell\in\R^3\subset F$} is the \gls{grfs} for $\ell$ in $\BF$,
\mbox{$J_\ell\in\R^{3\times3}\subset J$} is the foot Jacobian of $\ell$,
\mbox{$\tau_\ell\in\R^3\subset \tau$} is the vector of joint torques of $\ell$,
\mbox{$h_\ell\in\R^3\subset h$} is the vector of centrifugal, Coriolis, gravity torques of $\ell$ in $\BF$,
and 
\mbox{$\alpha_\ell\in\{0,1\}$} selects if the foot is on the ground or not.
A threshold of $F_\ell$ is typically used to calculate $\alpha_\ell$.
\begin{equation}\label{eq:grfalpha}
\alpha_\ell = \begin{cases}
1 & ||(J_\ell^T)^{-1}(\tau_\ell-h_\ell)||>\epsilon \\
0 & \text{otherwise}
\end{cases}
\end{equation}
where $\epsilon>0\in\R$ is the threshold.

\begin{assumption}
	There exists a force threshold $\epsilon$ that determines if the foot is in contact with the environment.
	\label{lsens_ass1}
\end{assumption}

The translational and rotational kinematics, and the translational dynamics 
of the robot as a single rigid body
in $\NF$ are
\begin{align}\dot{x}^n=v^n&&\dot{v}^n=a^n+g^n&&\dot{R}^n_b=R^n_bS(\omega^b)\label{eq:pvdyn}\end{align}
where $x^n\in\R^3$, $v^n\in\R^3$, $a^n\in\R^3$ are the position, velocity, and acceleration of the base in $\NF$, respectively, $R_b^n\in \SO(3)$ is the rotation matrix from $\BF$ to $\NF$, and $\omega^b$ is the angular velocity of the base in $\BF$.
The skew symmetric matrix function is $S()$.

\subsection{Sensors}
The modeling assumes that the quadruped robot is equipped with 
a six-axis \gls{imu}  on the trunk
(3~\gls{dofs} gyroscope and 3~\gls{dofs} accelerometer), and that every joint contains an encoder and a torque sensor.
The accelerometer measures specific force $f_s^b\in\R^3$ 
\begin{equation}\label{eq:fs}
f_s^b = a^b + g^b
\end{equation}
where $a^b\in\R^3$ is the acceleration of the body in $\BF$ and $g^b\in\R^3$ is the acceleration due to gravity in $\BF$.
The gyroscope directly measures angular velocity $\omega^b\in\R^3$ in $\BF$.
The encoders are used to measure the joint position $q_i\in\R$ and joint speed $\dot{q}_i\in\R$.
The pose of each joint (\ie the forward kinematics) is assumed to be exactly known.
The torque sensors in the joints directly measure torque $\tau_i\in\R$.

The measured values of all of the sensors differ from the theoretical values in that they contain a bias and noise: $\tilde{x} = x + b_x + n_x$
where $\tilde{x}$, $b_x$, and $n_x$ are the measured value, bias, and noise of $x$, respectively.
All of the biases are assumed to be constant or slowly time-varying, and all of the noise variables have zero mean and a Gaussian distribution.

\section{State Estimator} \label{sec_state_est}
To compare the effect of different terrains, we use the state-of-the-art low-level state estimator from~\cite{Fink2020}.
It includes input from three proprioceptive sensors: an \gls{imu}, encoders, and torque sensors.
For reliability and speed no exteroceptive sensors are used.
The state estimator consists of three major components: an attitude observer, leg odometry, and a sensor fusion algorithm.

\subsection{Non-linear Attitude Observer}
Typically in the quadruped robot literature an \gls{ekf} is used for attitude estimation, \eg\cite{Nobili2017,Ma2016,Bloesch2013}.
However, 
our \textit{attitude observer}~\cite{Fink2020} is \gls{ges}, and it consists of a \textit{\gls{nlo}}~\cite{Grip2015} and an \textit{\gls{xkf}}~\cite{Johansen2017}. 
The \gls{nlo} is
\begin{align}\label{eq:NLO}
	\begin{split}
	\dot{\hat{R}}^n_b &= \hat{R}^n_bS(\omega^b-\hat b^b)+\sigma  K_p J_s(\hat{R}^n_b)\\ 
	\dot{\hat{b}}^b &= \Proj\left(\hat{b}^b,-k\vex\left(\mathbb{P}\left({\hat{R}^{nT}_{bs}} K_pJ_s(\hat{R}^n_{b})\right)\right)\right)\\ 
	J_s(\hat{R}^n_b) &= \sum_{j=1}^k(y_j^n-\hat{R}^n_by_j^b){y_j^b}^T
	\end{split}
    \end{align}
where $K_p\in\R^{3\times 3}$ is a symmetric positive-definite gain matrix, $k>0\in\R$ is a scalar gain, $\sigma\ge 1\in\R$ is a scaling factor, $\hat{R}^n_{bs}=\sat(\hat{R}^n_b)$, the function $\sat(X)$ saturates every element of $X$ to $\pm1$,
$\Proj$ is a parameter projection that ensures that $||\hat{b}||<M_b$, $M_b>0\in\R$ is a constant known upper bound on the gyro bias, $\mathbb{P}(X)=\frac{1}{2}(X+X^T)$ for any square matrix $X$, and $J_s$ is the stabilizing injection term.
The observer is \gls{ges} for all initial conditions assuming there exists $k>1$ non-collinear vector measurements, \ie $\left| y_i^n\times y_j^n\right|> 0$
where $i,j\in\{1,\cdots,k\}$.
Furthermore, if there is only one measurement the observer is still \gls{ges} if the following \gls{pe} condition holds:
if there exist constants $T>0\in\R$ and $\gamma>0\in\R$ such that, for all $t\ge 0$,
$\int_t^{t+T}y_1^n(\tau)y_1^n(\tau)^T\dtau\ge \gamma I$
holds then $y_1^n$ is \gls{pe}.  See~\cite{Grip2015} for proof.

The \gls{xkf}~\cite{Johansen2017} is similar to an \gls{ekf} in that it linearizes a nonlinear model about an estimate of the state and then applies the typical \gls{ltv} 
Kalman~filter
to the linearized model.
If the estimate is close to the true state then the filter is near-optimal.
However, if the estimate is not close to the true state, the filter can quickly diverge.
To overcome this problem, the \gls{xkf} linearizes about a globally stable exogenous signal from a \gls{nlo}.
The cascaded structure maintains the global stability properties from the \gls{nlo} and the near-optimal properties from the Kalman~filter. 
The observer is
\begin{align}\label{eq:XKF}
	\begin{split}
	\dot{\hat{x}} &= f_x+C(\hat{x}-\breve{x})+K\left(z-h_x-H(\hat{x}-\breve{x})\right)\\
	\dot{P} &=CP+PC^T-KHP+Q\\
	K&=PH^TR^{-1}
	\end{split}
\end{align}
where $C=\left.\partial f_x/\partial{x}\right|_{\breve{x},u}$, $H=\left.\partial h_x/\partial{x}\right|_{\breve{x},u}$, $\breve{x}\in\R^n$ is the bounded estimate of $x$ from the globally stable NLO.
See~\cite{Johansen2017} for the stability proof.

\subsection{Leg Odometry}
Leg odometry computes the overall base velocity~$\dot x^b$ of the robot by combining the contribution of each foot velocity~$\dot x_{\ell}^b$.
Each leg $\ell$ only contributes to the leg odometry when it is in contact~$\alpha_\ell$.
Thus, we calculate the overall base velocity $\dot x^b$ as
\begin{equation}
\dot x_{\ell}^b = -\alpha_\ell\left(J_\ell(q_\ell)\dot q-\omega^b \times x_{\ell}^b\right)
\hspace{20pt}
\dot x^b = \dfrac{1}{n_s}\sum_{\ell} \dot x_{\ell}^b
\label{eq:lo}
\end{equation}
where $n_s=\sum\limits_\ell\alpha_\ell$ is the number of stance legs.

\begin{assumption}
The leg odometry assumes
	that the robot is always in rigid contact with the terrain. 
	This implies 
	that the stance feet do not move in $\NF$,
	there is no slippage, 
	the terrain does not expand or compress, 
	and the robot does not jump or fly.
	\label{ass2_lsens}
\end{assumption}

\subsection{Sensor Fusion}
Lastly, the inertial measurements~\eqref{eq:fs} are fused with the leg odometry~\eqref{eq:lo}. The main advantage of decoupling the attitude from the position and linear velocity is that the resulting dynamics is \gls{ltv}, and thus has guaranteed stability properties.  \ie the filter will not diverge in finite time.

We use a \gls{ltv} Kalman~filter 
with the dynamics~\eqref{eq:pvdyn}, the accelerometer~\eqref{eq:fs}, and leg odometry~\eqref{eq:lo}.
\begin{align}\label{eq:kf}\begin{split}
\dot{\hat{\underline{x}}}&=f_{\underline{x}}+\underline{K}(\underline{z}-h_{\underline{x}})\\
\dot{\underline{P}} &= \underline{CP} + \underline{PC^T} -\underline{K}\underline{H}\underline{P}+\underline{Q}\\
\underline{K} &= \underline{P}\underline{H}^T\underline{R}^{-1}
\end{split}\end{align}
where the state $\underline{x}=\begin{bmatrix}{x^n}^T&{v^n}^T\end{bmatrix}^T\in\R^6$ is position and velocity of the base, the input $u=(R_b^nf_s^b-g^n)\in\R^3$ is the acceleration of the base, the measurement $z=R_b^nx_\ell^b\in\R^3$ is the leg odometry, $\underline{K}\in\R^{6\times 3}$ is the Kalman gain, $P\in\R^{6\times 6}$ is the covariance matrix, $Q\in\R^{6\times 6}$ is the process noise and $R\in\R^{3\times 3}$ is the measurement noise covariance, and
\begin{eqnarray*}
f_{\underline{x}}=\begin{bmatrix}v^n\\u\end{bmatrix}\quad
\underline{C}=\begin{bmatrix}0_3&I_3\\0_3&0_3\end{bmatrix}\quad
\underline{H}=\begin{bmatrix}0_3&I_3\end{bmatrix}
\end{eqnarray*}
and $I_3$ and $0_3$ are the $3\times 3$ identity matrix and matrix of all zeros, respectively.

\section{Experimental Results} \label{sec_results}
To analyze the differences in state estimation between rigid and soft terrain, 
we used HyQ and our state estimator.
HyQ has twelve torque-controlled joints powered by hydraulic actuators.
HyQ has three types of on board proprioceptive sensors:
joint encoders, force/torque sensors, and \glspl*{imu}. 
Every joint has 
an absolute and a relative encoder to measure the joint angle and speed. 
The absolute encoder (AMS Programmable Magnetic Rotary Encoder~-~AS5045) 
measures the joint angle when the robot is first turned on, 
while the
relative encoder (Avago Ultra Miniature, High Resolution Incremental Encoder~-~AEDA-3300-TE1)
measures how far the joint has moved at every epoch. 
Every joint contains a force or torque sensor. Two joints have a load
cell (Burster Subminiature Load Cell~-~8417-6005) and one joint has a custom designed
torque sensor based on strain-gauges. 
In the trunk of the robot there is a fibre optic-based, military grade 
KVH 1775 IMU. 

We used the state estimator \eref{eq:NLO}-\eref{eq:kf} on the
\emph{Soft Trot in Place} and the \emph{Rigid Trot in Place} dataset from the dataset published in~\cite{Fin19}. 
HyQ was manually controlled to trot on a foam block of $160\times120\times20$~cm, and on a rigid ground. 
An indentation test of the foam shows the foam has an average stiffness of 2400~N/m.
All of the sensors were recorded at~1000~Hz.
A \gls{mcs} recorded the ground truth data with millimetre accuracy at~250~Hz.

\begin{figure}[t!]\label{sec_conc}
	\centering
	\includegraphics[width= 0.95 \textwidth]{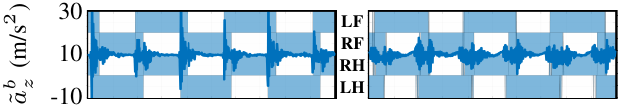}
	\includegraphics[width= 0.95 \textwidth]{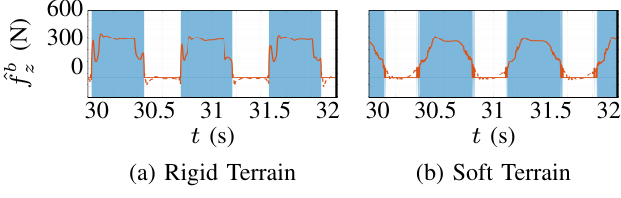}
	\caption
	[The measured specific force $\tilde{a}_z^b$, and 
	the estimated ground reaction forces $\hat{f}_z^b$,
	of \acrshort{hyq} during a trotting experiment.]
	{The $z$ component of the measured specific force $\tilde{a}_z^b$ (top),
		and the estimated ground reaction forces $\hat{f}_z^b$ (bottom),
		in the body frame $\BF$ of \acrshort{hyq} during a trotting experiment.
		The highlighted regions show when the given foot is in stance, and the feet are denoted as left-front~(LF), right-front~(RF), left-hind~(LH), and right-hind~(RH).\label{fig:grfz}\label{fig:accz}}
	\label{fig_heightmap_evaluation}
\end{figure}

The experiments confirmed our original hypothesis that soft terrain negatively impacts state estimation and also allowed us to investigate why. 
It is important to note that rigid versus soft terrain had no impact on the attitude estimation. 
For space reasons, all attitude plots have been omitted.

The first distinct difference between soft and rigid terrain is the specific force measurement of the body as seen in \fref{fig:accz}.  On the rigid terrain there are large impacts and then vibrations every time a foot touches down.  Whereas the soft terrain damped out these vibrations.  Next, on the soft terrain more prolonged periods of positive and negative acceleration can be seen.  This acceleration can also be seen in the plots of the \gls{grfs} in \fref{fig:grfz} where the \gls{grfs} on the soft terrain are more continuous when compared to the rigid terrain. In other words, there are longer loading and unloading phases. 

\begin{sidewaysfigure}
\centering
\includegraphics[width= 0.49 \textheight]{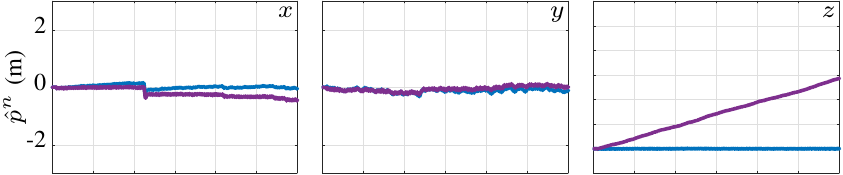}
\includegraphics[width= 0.49 \textheight]{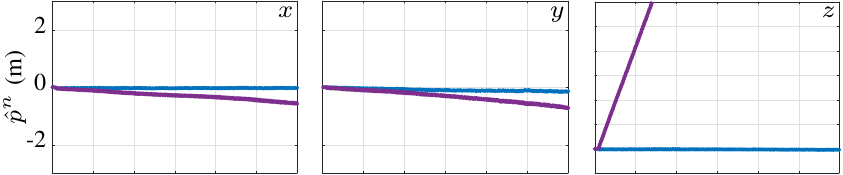}\\
\includegraphics[width= 0.49 \textheight]{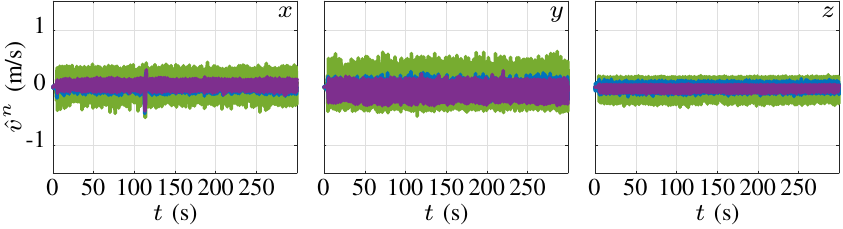}
\includegraphics[width= 0.49 \textheight]{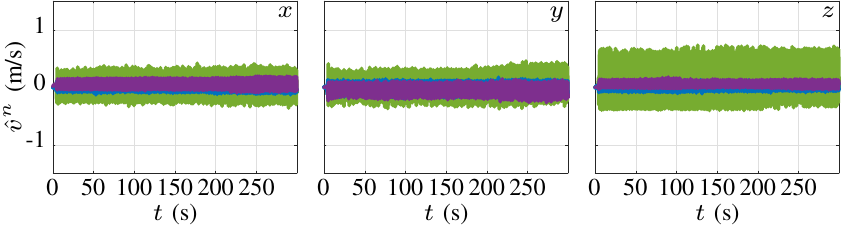}\\
\includegraphics[width= 0.49 \textheight]{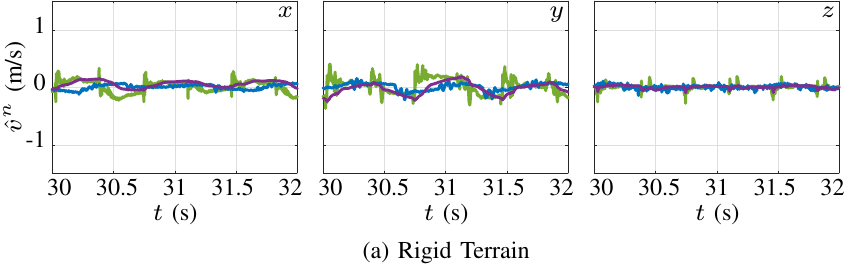}
\includegraphics[width= 0.49 \textheight]{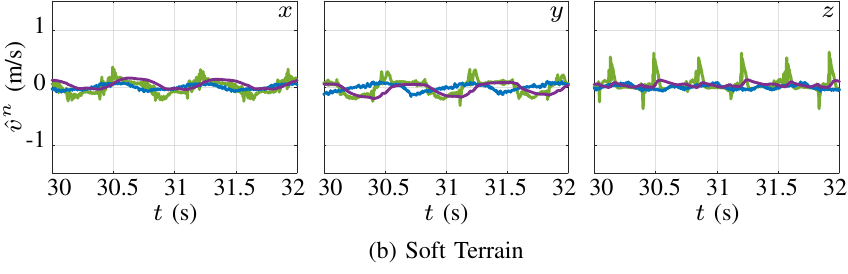}
\caption
[The estimated trunk position $\hat{x}^n$,
and the estimated trunk velocity $\hat{v}^n$,
of \acrshort{hyq} during a trotting experiment.]
{
The estimated trunk position $\hat{x}^n$ (top),
and the estimated trunk velocity $\hat{v}^n$ (middle and bottom),
in the navigation frame $\NF$ of \acrshort{hyq} during a trotting experiment using sensor fusion (purple) versus the raw leg odometry (green), and the motion capture system (blue).
The first two rows show the full experiment ($0\le t \le300$)~s and the bottom row is zoomed in ($30\le t \le32$)~s.\label{fig:posvel}}
\end{sidewaysfigure}

The most important differences between soft and rigid terrain are seen in the velocity and position estimates as shown in \fref{fig:posvel}. We can see that the leg odometry has large erroneous peaks in $z$ velocity at both touch-down and lift-off.  These peaks in velocities can then be seen in the position estimates as a drift.  
On the other hand,
the~$x$~and~$y$ position estimates are quite accurate and only have a slow drift.

In the figures, we can also see multiple of the state estimators assumptions being broken.  
First, there does not exist a constant~$\epsilon$ that can describe when the foot is in contact with the ground, which is contradicting \assref{lsens_ass1}.  
The contact~$\epsilon$ is no longer binary 
(\ie supporting/not-supporting the weight of the robot),
but the contact is now a continuous value with varying amounts of the robot's weight being supported and sometimes even pushed.  When trying to use the previous simple model, the contact ignores a large portion of the loading and unloading phase.  Furthermore, it often chatters rapidly between contact/non-contact when the force is close to~$\epsilon$.  %
Second,
the foot is moving for almost the entire contact (\ie non-zero acceleration) on soft terrain
as shown in \fref{fig:accz}.
This contradicts \assref{ass2_lsens} that the foot velocity is zero when in contact.  
Third, \eqref{eq:lo} is broken. It assumes that all of the velocity (and all of the acceleration) is a result of the \gls{grfs}, but not all of the acceleration due to gravity is being accounted for.  Hence, the robot appears to drift up and away from the ground.

There are a few simple ways to try to improve the estimates of this or other similar state estimators. The first is to tune~$\epsilon$ in~\eqref{eq:grfalpha}.  By increasing~$\epsilon$ there would be less erroneous velocity, but in doing so it would also ignore part of the leg odometry.
In general, 
on a planar surface,
a reduced drift in the~$z$ direction comes at the cost of an increased error in the~$x$ and~$y$ directions.
A second method could be to have an adaptive velocity bias for the leg odometry. However, the bias is not constant and it depends on both the gait and the terrain. Thus, the problem of estimating the body velocity of the robot using leg odometry remains open.

\section{Conclusions}\label{sec_conclusion}
In this letter,
we present an experimental validation and a formal study on  
the influence of soft terrain on state estimation for legged robots.
We utilized a state-of-the-art state estimator that fuses \gls{imu} measurements with leg odometry. 
We experimentally analyzed the differences between soft and rigid terrain using our state estimator and a dataset of the HyQ robot.
That said, we report three main outcomes.
First, we showed that soft terrain results in a larger drift in the position estimates,
and larger errors in the velocity estimates compared to rigid terrain. 
These problems are caused by the broken legged odometry contact assumptions on soft terrain.
Second, we also showed that over soft terrain, the contact with the terrain is no longer binary
and it often chatters rapidly between contact and non-contact.
Third, we showed that soft terrain 
	affects many states besides the robot pose. 
	This includes the contact state and the \gls{grfs} which are essential for the control of legged robots.
Future works include extending the state estimator to incorporate the terrain impedance in the leg odometry model. 
Additionally, further datasets will be recorded to investigate the long-term drift in 
the forward and lateral directions.


\part{Exteroceptive Terrain-Aware Locomotion}
\glsresetall \chapter[ViTAL: Vision-Based Terrain-Aware Locomotion]
{ViTAL:\\Vision-Based Terrain-Aware Locomotion\\for Legged Robots}\label{chap_vital}
\vspace{-20pt}
\vitalPaper
\boldSubSec{Abstract}
This work is on vision-based planning strategies for legged robots that separate locomotion planning into foothold selection and pose adaptation. 
Current pose adaptation strategies optimize the robot's body pose relative to \textit{given} footholds. 
If these footholds are not reached, the robot may end up in a state with no reachable safe footholds. 
Therefore, we present a~\gls{vital} strategy that consists of novel pose adaptation and foothold selection algorithms. 
\gls{vital} introduces a different paradigm in pose adaptation that does not optimize the body pose relative to given footholds, but the body pose that maximizes the chances of the legs in reaching safe footholds. 
\gls{vital}~plans footholds and poses based on skills that characterize the robot's capabilities and its terrain-awareness.
We use the \unit[90]{kg}~\acrshort{hyq} and \unit[140]{kg}~\acrshort{hyqreal} quadruped robots to validate \gls{vital}, 
and show that they are able to climb various obstacles including 
stairs, gaps, and rough terrains at different speeds and gaits. 
We compare \gls{vital} with a baseline strategy that selects the robot pose based on given selected footholds, and show that \gls{vital} outperforms the baseline. 

\boldSubSec{Accompanying Video}  \url{https://youtu.be/b5Ea7Jf6hbo}

\glsresetall

\newpage

\section{Introduction}
\begin{figure}\centering
\includegraphics[width=0.88\columnwidth]{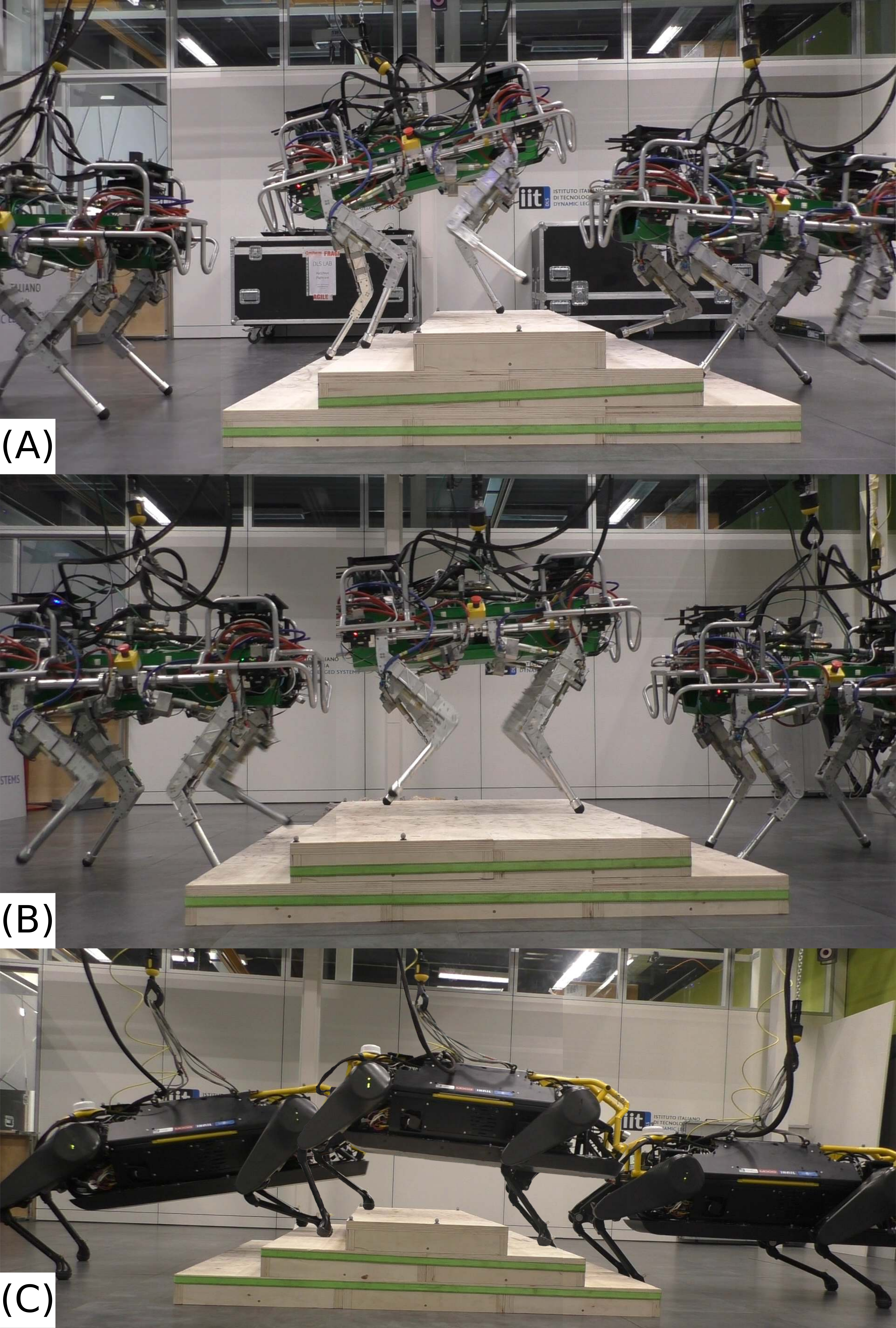}		
\caption{The \acrshort{hyq}~and~\acrshort{hyqreal} quadruped robots climbing stairs using~\gls{vital}.}
\label{fig_1}
\end{figure}

Legged robots have shown remarkable agile capabilities
in academia~\cite{Lee2021,Yang2020,Semini2019,Katz2019,Bledt2018,Hutter2016}
and industry~\cite{BostonDynamics2021,AgilityRobotics2021,UniTree2021}.
Yet, to accomplish breakthroughs in dynamic whole-body locomotion, 
and to robustly traverse unexplored environments, legged robots have to be \textit{terrain aware}. 
\gls{tal} implies that the robot is capable of taking decisions based on the terrain~\cite{Fahmi2021T}. 
The decisions can be in planning, control, or in state estimation,  
and the terrain may vary in its geometry and physical 
properties~\cite{Buchanan2021,Fahmi2021,Wang2020,Ahmadi2020,Lin2020,Fahmi2020,Paigwar2020,Wellhausen2019,Bosworth2016}.  
\gls{tal}~allows the robot to use its on-board sensors to perceive its surroundings and act accordingly. 
This work is on \textit{vision-based \gls{tal} planning strategies}
that plan the robot's motion (body and feet) based on the terrain information that is acquired using vision (see~\fref{fig_1}).

\subsection{Related Work - Vision-Based Locomotion Planning}\label{sec_related_work_vpa}
Vision-based locomotion planning can either be \textit{coupled} or~\textit{decoupled}.
The coupled approach jointly plans the body pose and footholds in a single algorithm. 
The decoupled approach independently plans the body pose and footholds in separate algorithms.
The challenge in the coupled approach is that it is computationally expensive to solve in real-time. 
Because of this, the decoupled approach tends to be more practical 
since the high-dimensional planning problem is split into multiple low-dimensional problems. 
This also makes the locomotion planning problem more tractable. 
However, this raises an issue with the decoupled approach because the 
plans may conflict with each other since they are planned separately.
Note that both approaches could be solved using optimization, learning, or heuristic methods.

\gls{to} is one way to deal with coupled vision-based locomotion planning.
By casting locomotion planning as an optimal control problem,
\gls{to}~methods can optimize the robot's motion
while taking into account the terrain information~\cite{Fan2021, Melon2021, Ponton2021, Mastalli2020b, Winkler2018b}.
The locomotion planner can generate trajectories
that prevent the robot from slipping or colliding with the terrain
by encoding the terrain's shape and friction properties in the optimization problem~\cite{Winkler2018b}. 
\gls{to}~methods can also include a model of the terrain as a cost map in the optimization problem, 
and generate the robot's trajectories based on that~\cite{Mastalli2020}. 
\gls{to} methods provide guarantees on the optimality and feasibility of the devised motions, 
albeit being computationally expensive; performing these optimizations in real-time is still a challenge.
To overcome this issue,~\gls{to} approaches often implement hierarchical (decoupled) approaches. 
Instead of decoupling the plan into body pose and footholds, the hierarchical approaches 
decouple the plan into short and long-horizon plans~\cite{Brunner2013, Li2020}.
Additionally, 
other work relies on varying the model complexity to overcome the computational issue with~\gls{to}\cite{Li2021}.

\gls{drl} methods mitigate the computational burden of \gls{to} methods by 
training function approximators that learn the locomotion 
plan~\cite{Rudin2022, Yu2022, Miki2021, Gangapurwala2021, Kumar2021, Siekmann2021b, Tsounis2020}.  
Once trained, 
an~\gls{drl} policy can generate body pose and foothold sequences based on proprioceptive and/or visual information. 
Yet, \gls{drl}~methods may require tedious learning (large amounts of data and training time) 
given its high-dimensional state representations.

As explained earlier, decoupled locomotion planning can mitigate
the problems of \gls{to} and \gls{drl} 
by separating the locomotion plan into feet planning 
and body planning~\cite{Kalakrishnan2011,Fankhauser2018,Villarreal2019,Buchanan2020b, Fernbach2018, Tonneau2018}.
Thus, one can develop a more refined and tractable algorithm for every module separately.
In this work, planning the feet motion (foothold locations) is called \textit{foothold selection}, 
and planning the body motion is called \textit{pose adaptation}.

\subsection{Related Work - Foothold Selection and Pose Adaptation}
Foothold selection strategies choose the best footholds
based on the terrain information and the robot's capabilities. 
Early work on foothold selection was presented 
by Kolter~\textit{et~al.}~\cite{Kolter2008} and Kalakrishnan~\textit{et~al.}~\cite{Kalakrishnan2009}
where both approaches relied on motion capture systems 
and an expert user to select (label) the footholds. 
These works were then extended in~\cite{Belter2011} 
using unsupervised learning, on-board sensors,
and considered the terrain information such as 
the terrain roughness (to avoid edges and corners) and friction (to avoid slippage). 
Then, Barasuol~\textit{et~al.}~\cite{Barasuol2015} extended the aforementioned work by selecting footholds
that not only considers the terrain morphology, but also considering leg collisions with the terrain.
Further improvements in foothold selection strategies added other evaluation criteria
such as the robot's kinematic limits.
These strategies use
optimization~\cite{Fankhauser2018,Jenelten2020,Song2021}, 
supervised learning~\cite{Villarreal2019, Belter2019, Esteban2020},
\gls{drl}~\cite{Lee2021,Paigwar2020}, 
or heuristic~\cite{Kim2020} methods.

Similar to foothold selection, pose adaptation strategies 
optimize the robot's body pose based on the terrain information and the robot's capabilities. 
An early work on vision-based pose adaptation was presented in~\cite{Kalakrishnan2011}.
The goal was to find the optimal pose that 
maximizes the reachability of \textit{given} selected footholds, 
avoid collisions with the terrain, 
and maintain static stability. 
The given footholds are based on a foothold selection algorithm 
that considers the terrain geometry.
Another approach was presented in~\cite{Belter2012} that
finds the optimal body elevation and inclination \textit{given}
the selected footholds, and the robot location in the map. 
The pose optimizer 
maximizes different margins that 
increase the kinematic reachability of the legs and static stability,
and avoids terrain collisions. 
This approach was then extended in~\cite{Belter2019b}
with an improved version of the kinematic margins.
A similar approach was presented in~\cite{Fankhauser2018} where
the goal was to find an optimal pose that can maximize 
the reachability of \textit{given} selected footholds. 
The reachability term is accounted for in the cost function 
of the optimizer by penalizing the difference between
the default foothold position and the selected one. 
The work in~\cite{Buchanan2020b} builds on top of 
the pose optimizer of~\cite{Fankhauser2018}
to adapt the pose of the robot in confined spaces using 3D terrain maps. 
This is done using a hierarchical approach 
that first samples body poses that allows the robot to navigate through 
confined spaces, then smooths these poses using a gradient descent method 
that is then augmented with the pose optimizer of~\cite{Fankhauser2018}.
The work presented in~\cite{Villarreal2020} generates vision-based pose references that
also rely on \textit{given} selected footholds
to estimate the orientation of the terrain and send it as a pose reference.
Alongside the orientation reference, 
the body height reference is set at a constant vertical distance (parallel to gravity) 
from the center of the approximated plane that fits through the selected footholds.

The aforementioned pose adaptation strategies 
focus on finding \textit{one} optimal solution 
based on \textit{given} footholds;
footholds have to be first selected and \textit{given} to the optimizer. 
Despite selecting footholds that are safe, 
there are no guarantees on what would happen during execution 
if the footholds are not reached or if the robot deviates from its planned motion.
If any of these cases happen, the robot might end up in a pose where no safe footholds can be reached.
This would in~turn compromise the safety and performance of the robot.
Even if the strategy can re-plan, 
reaching a safe pose might not be possible if the robot is already in an unsafe state.

\subsection{Proposed Approach}
We propose \acrfull{vital}, an online whole-body locomotion planning strategy
that consists of a foothold selection and a pose adaptation algorithm. 
The foothold selection algorithm used in this work is an extension of the \gls{vfa} algorithm 
of the previous work done by Villarreal \textit{et~al.}~\cite{Villarreal2019}
and Esteban \textit{et~al.}~\cite{Esteban2020}. 
Most importantly, 
we propose a novel \gls{vpa} algorithm that
introduces a different paradigm to overcome 
the drawbacks of the state-of-the-art pose adaptation methods. 
\textit{Instead of finding body poses that are optimal for given footholds, 
we propose finding body poses that maximize the chances of reaching safe footholds
in the first place.}
Hence, we are interested in putting the robot in a state 
in which if it deviates from its planned motion, 
the robot remains around a set of footholds that are still 
reachable and safe. 
The notion of safety emerges from \textit{skills} that characterize the robot's capabilities.

\gls{vital} plans footholds and body poses 
by sharing the same robot skills (both for the \gls{vpa} and the \gls{vfa}). 
These skills characterize what the robot is capable of doing.
The skills include, but are not limited to:
the robot's ability to avoid edges, corners, or gaps (\textit{terrain roughness}), 
the robot's ability to remain within the workspace of the legs 
during the swing and stance phases (\textit{kinematic limits}), and
the robot's ability to avoid colliding with the terrain (\textit{leg collision}).
These skills are denoted by \textit{\criteria}.
Evaluating the \criteria is usually computationally expensive. 
Thus, to incorporate the \criteria in \gls{vital}, 
we rely on approximating them with~\glspl{cnn} that are trained via supervised learning.
This allows us to continuously adapt both the footholds and the body pose.
The~\gls{vfa} and the~\gls{vpa} are decoupled and can run at a different update rate. 
However, they are non-hierarchical, they run in parallel, 
and they share the same knowledge of the robot skills (the \criteria). 
By that, we overcome the limitations that result from hierarchical planners as mentioned in~\cite{Buchanan2020b},
where high-level plans may conflict with the low-level ones causing a different robot behavior.

The \gls{vpa} utilizes the \criteria to approximate a function that provides the number of safe footholds for the legs. 
Using this function, we cast a \textit{pose optimizer} which solves a non-linear optimization problem
that maximizes the number of safe footholds for all the legs subject to constraints added to the robot pose. 
The pose optimizer is a key element in the \gls{vpa} since it 
adds safety layers and constraints to the learning part of our approach. 
This makes our approach more tractable 
which mitigates the issues that might arise from end-to-end policies in \gls{drl} methods.

\subsection{Contributions}\label{contribs}
\gls{vital} mitigates the 
above-mentioned conflicts that exist in other decoupled
planners~\cite{Buchanan2020b, Jenelten2020, Tonneau2018, Fankhauser2018}.
This is because both the \gls{vpa} and the \gls{vfa} share the same skills encoded in the \criteria. 
In other words, the \gls{vpa} and the \gls{vfa}
will not plan body poses and footholds that may conflict with each other
because both planners share the same logic. 
In this work, the formulation of the \gls{vpa} allows \gls{vital} to reason about the leg's capabilities and the terrain information.
However, the formulation of the \gls{vpa} could be further augmented by other body-specific skills. 
For instance, the \gls{vpa} could be reformulated to reason about the body collisions with the environment 
similar to the work in~\cite{Tonneau2018, Buchanan2020}. 
The paradigm of the~\criteria can also be further augmented to consider other skills. 
We envision that some skills are best encoded via heuristics while others are well suited 
through optimization. 
For this reason, 
the~\criteria can also handle optimization-based foothold objectives such as the ones in~\cite{Jenelten2020}.

Following the recent impressive results in~\gls{drl}-based locomotion controllers, 
we envision \gls{vital} to be inserted as a module into such control frameworks. 
To elaborate, current \gls{drl}-based locomotion controllers~\cite{Rudin2022, Miki2021, Lee2021, Kumar2021} are of a single network;
the~\gls{drl} framework is a single policy that maps the observations (proprioceptive and exteroceptive) to the actions. 
This may be challenging since it requires careful reward shaping, 
and generalizing to new tasks or different sensors (observations) makes the problem harder~\cite{Green2021}.
For this reason, and similar to Green~\textit{et~al.}~\cite{Green2021}, 
we envision that~\gls{vital} can be utilized as a planner for~\gls{drl} controllers where the~\gls{drl} controller
will act as a reactive controller that then receives guided (planned) commands in a form of optimal poses and footholds from \gls{vital}.

\gls{vital} differs from \gls{to} and optimization-based methods in several aspects. 
The \criteria is designed to independently evaluate every skill (criterion). 
Thus, one criterion can be optimization-based while other could be using logic or heuristics. 
Because of this, \gls{vital} is not restricted by solving an optimization problem that handles all the skills at once. 
Another difference between \gls{vital} and \gls{to} is in the way the body poses are optimized. 
In \gls{to}, the optimization problem optimizes a single pose to follow a certain trajectory. 
The \gls{vpa} in \gls{vital} optimizes for the body poses that maximizes the chances of the legs in reaching safe footholds. 
In other words, the \gls{vpa} finds a body pose that would put the robot in a configuration where the legs have the maximum possible number of safe footholds. 
In fact, this paradigm that the \gls{vpa} of \gls{vital} introduces may be also encoded in \gls{to}.
Additionally, \gls{to} often finds body poses that considers the leg's workspace, 
but to the best of the authors' knowledge, there is no \gls{to} method 
that finds body poses that consider the legs' collision with the terrain, and the feasibility of the swinging legs' trajectory.

\noindent To that end, the \textit{contributions} of this article are
\begin{itemize}
\item \gls{vital}, an online vision-based locomotion planning strategy
that simultaneously plans body poses and footholds based on shared knowledge of robot skills (the~\criteria).
\item 
An extension of our previous work on the \gls{vfa} algorithm for foothold selection
that considers the robot's body twist and the gait parameters.
\item 
A novel pose adaptation algorithm called the \gls{vpa} 
that finds the body pose that maximizes the number of safe footholds for the robot's legs.
\end{itemize}

\section{Foothold Evaluation Criteria (FEC)}\label{sec_heuristics}
The \criteria is main the building block for the \gls{vfa} and the \gls{vpa}. 
The \criteria 
are sets of skills that 
evaluate footholds within heightmaps. 
The skills include 
the robot's ability to assess the terrain's geometry, avoid leg collisions, and avoid reaching kinematic limits.
The \criteria can be 
model-based as in this work and~\cite{Kalakrishnan2011, Villarreal2019}, 
or using optimization techniques as in~\cite{Fankhauser2018, Jenelten2020}. 
The \criteria of this work extends the criteria used in our previous 
work~\cite{Barasuol2015,Villarreal2019,Esteban2020}.

The \criteria takes a tuple~$\fectuple$ as an input,
evaluates it based on multiple criteria, 
and outputs a boolean matrix~$\fecout$.
The input tuple~$\fectuple$ is defined as
\begin{equation}
\fectuple = (\hmap, \hheight, \bodyvel, \gaitparams)
\label{fec_tuble}
\end{equation}
where 
$\hmap\in\Rnum^{h_x\times{h_y}}$ is the heightmap of dimensions $h_x$~and~$h_y$,
$\hheight\in\Rnum$ is the \textit{\hipheight} of the leg (in the world frame), 
${\bodyvel\in\Rnum^6}$ is the base twist,
and $\gaitparams$ are the gait parameters 
(step length, step frequency, duty factor, and time remaining till \td). 
The heightmap $\hmap$ 
is extracted from the terrain elevation map, 
and is oriented with respect to the horizontal frame of the robot~\cite{Barasuol2013}. 
The horizontal frame coincides with the base frame of the robot,
and its $xy$-plane is perpendicular to the gravity vector. 
Each cell (pixel) of~$\hmap$ denotes the terrain height 
that corresponds to the location of this cell in the terrain map. 
Each cell of~$\hmap$ also corresponds to 
a \textit{candidate foothold}~$\candidate\in\Rnum^3$ for the robot.

In this work, we only consider the following \criteria:
\textit{\gls{tr}}, \textit{\gls{lc}}, \textit{\gls{kfis}}, and \textit{\gls{fc}}. 
Each criterion~$C$ 
outputs a boolean matrix~$\mu_C$.
Once all of the criteria are evaluated, the final output~$\fecout$ 
is the element-wise logical AND ($\wedge$)
of all the criteria. 
The output matrix~$\fecout\in\Rnum^{h_x\times{h_y}}$ 
is a boolean matrix with the same size as the input heightmap~$\hmap$.
$\fecout$ indicates the candidate footholds 
(elements in the heightmap $\hmap$) that are \textit{safe}.
An element in the matrix~$\fecout$ that is true, 
corresponds to a candidate foothold $\candidate$
in the heightmap~$\hmap$ that is safe. 
The output of the \criteria is
\begin{equation}
\fecout = 
\mu_\mathrm{TR} \wedge \mu_\mathrm{LC} \wedge \mu_\mathrm{KF} \wedge \mu_\mathrm{FC} .
\label{fec_evaluation}
\end{equation}

An overview of the criteria used in this work is shown in \fref{fig_2}{(A)}
and is detailed below.

\boldSubSec{Terrain Roughness (TR)}
This criterion checks edges or corners in the heightmap that are unsafe for the robot to step on. 
For each candidate foothold $\candidate$ in $\hmap$, we evaluate
the mean and standard deviation of the slope relative to its neighboring footholds, 
and put a threshold that defines whether a $\candidate$ is safe or not. 
Footholds above this threshold are discarded.

\boldSubSec{Leg Collision (LC)}
This criterion selects footholds that do not result in leg collision with the terrain 
during the entire gait cycle (from \lo, during swinging, \td and till the \nlo). 
To do so, we create a bounding region around the leg configuration that corresponds to 
the candidate foothold~$\candidate$ and the current hip location. 
Then, we check if the bounding region collides with the terrain (the heightmap) 
by measuring the closest distance between them.
If this distance is lower than a certain value, then the candidate foothold is discarded.

\boldSubSec{Kinematic Feasibility (KF)}
This criterion selects footholds that are kinematically feasible. 
It checks whether a candidate foothold~$\candidate$ 
will result in a trajectory that remains within 
the workspace of the leg during the entire gait cycle.
To do so, we check if the candidate foothold~$\candidate$
is within the workspace of the leg during \td and \nlo. 
Also, we check if the trajectory of the foot from the \lo position $p_{lo}$
till the \td position at the candidate foothold $\candidate$
is within the workspace of the leg. 
In the initial implementation in \cite{Villarreal2019}, 
this criterion was only evaluated during \td. 
In this work, we consider this criterion during the entire leg step-cycle.

\boldSubSec{Foot Trajectory Collision (FC)}
This criterion 
selects footholds that 
do not result in foot trajectory collision with the terrain.
It checks whether the foot swing trajectory 
corresponding to a candidate foothold $\candidate$
is going to collide with the terrain or not. 
If the swing trajectory collides with the terrain, 
the candidate foothold $\candidate$ is discarded.

\begin{remark}
There are three main sources of uncertainty that can affect
the foothold placement~\cite{Barasuol2015}. 
These sources of uncertainty are due to 
trajectory tracking errors, foothold prediction errors, and drifts in the map. 
To allow for a degree of uncertainty, after computing~$\fecout$, 
candidate footholds that are within a radius of unsafe footholds are also discarded.
This is similar to the erosion operation in image processing.
\label{remark_uncertainty_margin}
\end{remark}

\begin{remark}
The initial implementation of the \criteria in~\cite{Villarreal2019}
only considered the heightmap~$\hmap$ as an input;
the other inputs of the tuple~$\fectuple$ in~\eref{fec_tuble} were kept constant.
This had a few disadvantages that we reported in~\cite{Esteban2020}
where we extended the work of \cite{Villarreal2019} by  
considering the linear body heading velocity. 
In this work, we build upon that by considering
the full body twist~$\bodyvel$ and the gait parameters~$\gaitparams$ as expressed by $\fectuple$ in~\eref{fec_tuble}. 
\label{remark_vfa_differences}
\end{remark}

\begin{sidewaysfigure}
\centering
\includegraphics[width=0.8\columnwidth]{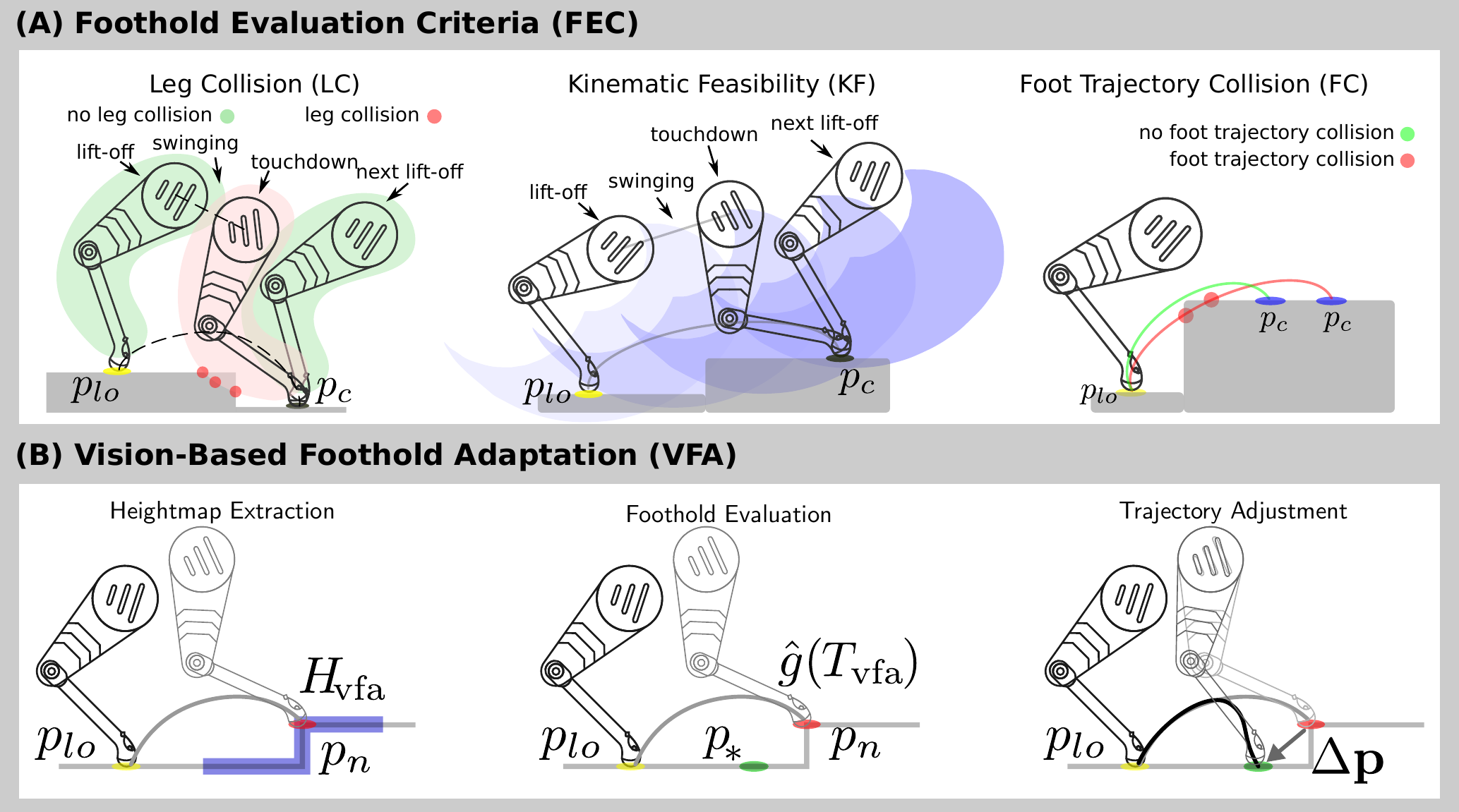}		
\caption
[Overview of \acrshort{vital}.]
{
Overview of \acrshort{vital}. Illustrations are not to scale. 
(A)~The Foothold Evaluation Criteria (\acrshort{fec}):
Leg Collision~(LC), Kinematic Feasibility~(KF), and Foot Trajectory Collision~(FC). 
(B)~The Vision-based Foothold Adaptation (VFA) pipeline.
First, we extract the heightmap $\hmapvfa$ around the nominal foothold $\nominal$.
Then, we evaluate the heightmap either using the exact evaluation $\g$ or
using the \acrshort{cnn} as an approximation $\ghat$. 
Once the optimal foothold $\optimal$ is selected, the swing trajectory is adjusted.
}
\end{sidewaysfigure}

\begin{sidewaysfigure}\ContinuedFloat
\centering
\includegraphics[width=0.8\columnwidth]{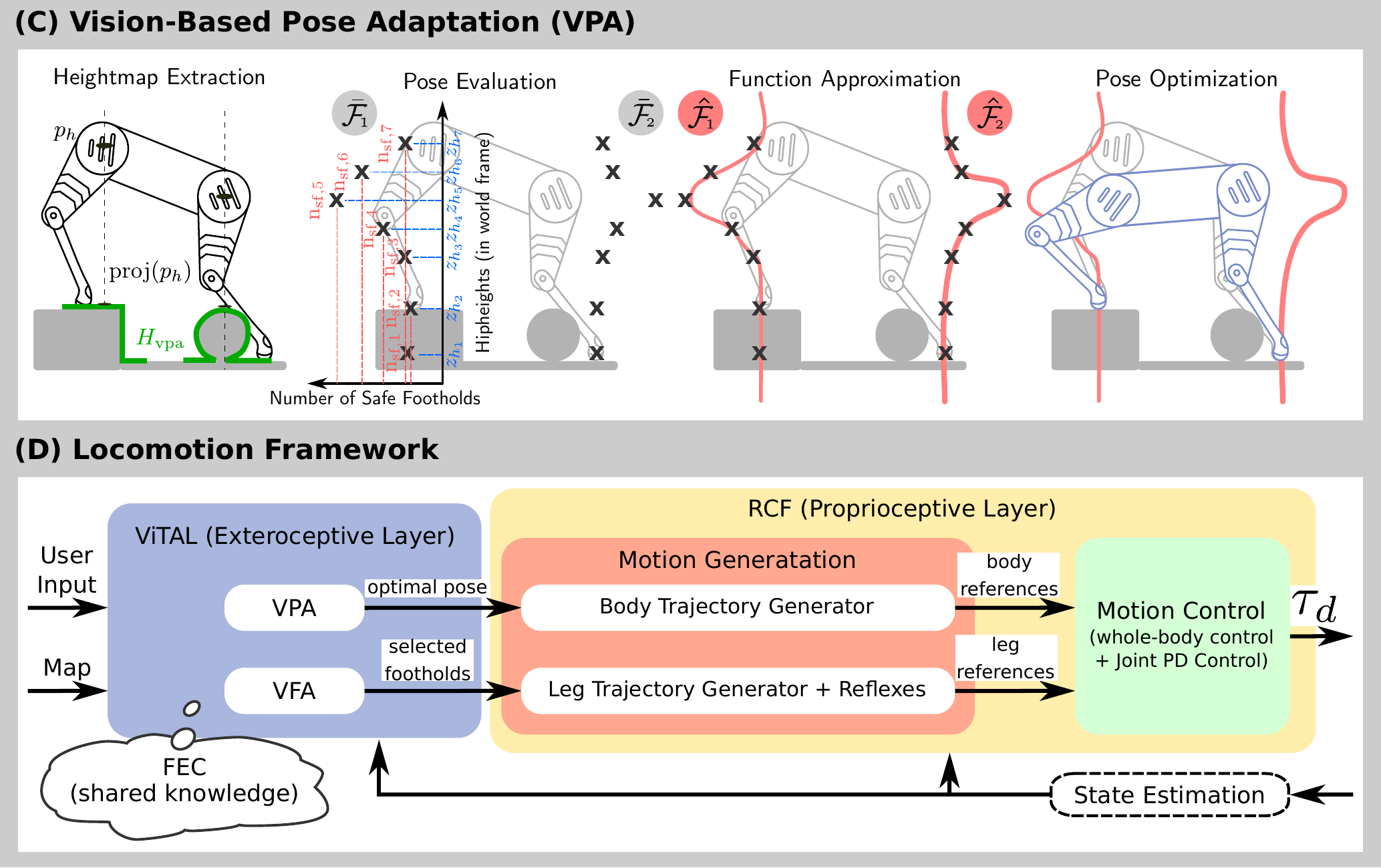}		
\caption
[Overview of \acrshort{vital} (continued).]
{
Overview of \acrshort{vital} (continued). Illustrations are not to scale. 
(C)~The Vision-Based Pose Adaptation (VPA) pipeline.
First, we extract the heightmap $\hmapvpa$ for all the legs. The heightmaps are centered around the projection of the leg hip locations. 
Then, we evaluate the \acrshort{fec} to compute $\finitesetfeasibles$ 
for all the hip heights of all the legs (pose evaluation).
Then, we approximate a continuous function $\rbfn$ from  $\finitesetfeasibles$ (function approximation). 
The pose optimizer finds the pose that maximizes $\rbfn$ for all of the legs~(pose optimization).
(D)~Our locomotion framework.
\acrshort{vital} consists of the \acrshort{vpa} and the \acrshort{vfa} algorithms.
Both algorithms rely on the robot skills which we denote by \acrshort{fec}.
$\tau_d$ are the desired joint torques that are sent to the robot.
}
\label{fig_2}
\end{sidewaysfigure}

\section{Vision-Based Foothold Adaptation (VFA)}\label{vfaa}
The \gls{vfa} evaluates the \criteria to select the optimal foothold
for each leg~\cite{Barasuol2015, Villarreal2019, Esteban2020}.
The \gls{vfa} has three main stages as shown in \fref{fig_2}{(B)}:
\textit{heightmap extraction},  \textit{foothold evaluation}, and \textit{trajectory adjustment}.

\boldSubSec{Heightmap Extraction}
Using the current robot states and gait parameters, 
we estimate the \td position of the swinging foot in the world frame as detailed in \cite{Villarreal2019}.
This is denoted as the \textit{nominal foothold} $\nominal\in\Rnum^3$.
Then, we extract a heightmap $\hmapvfa$ that is centered around $\nominal$.

\boldSubSec{Foothold Evaluation}
After extracting the heightmap, 
we compute the \textit{optimal foothold} $\optimal\in\Rnum^3$ for each leg.
We denote this by foothold evaluation which is the mapping 
\begin{equation}
\g: \heurtuple \rightarrow \optimal
\end{equation}
that takes an input tuple~$\heurtuple$ that is defined as
\begin{equation}
\heurtuple =
(\hmapvfa, \hheight, \bodyvel, \gaitparams, \nominal) .
\label{vfa_tuble}
\end{equation}

\begin{sidewaysfigure}
\centering
\includegraphics[width=0.8\columnwidth]{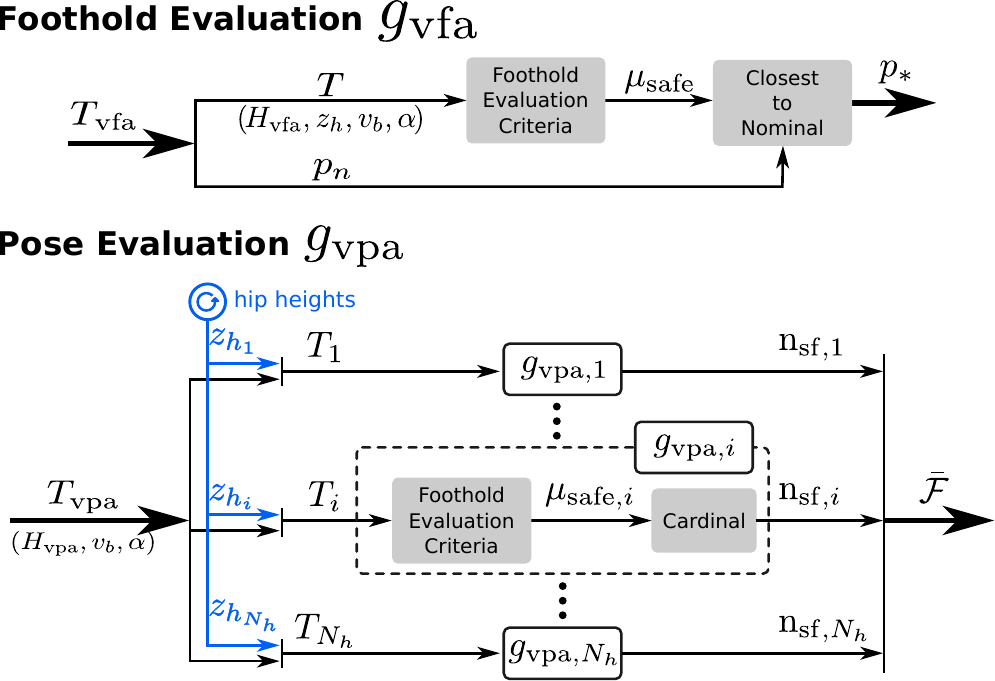}		
\caption{Overview of the foothold evaluation stage in the VFA algorithm, and the pose evaluation stage in the VPA algorithm.}
\label{vfa_footholdeval_vpa_pose_eval}
\end{sidewaysfigure}

Once we evaluate the \criteria in~\eref{fec_evaluation}, 
from all of the safe candidate footholds in $\fecout$, 
we select the optimal foothold~$\optimal$ as the one 
that is \textit{closest to the nominal} foothold $\nominal$.
The aim is to minimize the deviation from the original trajectory 
and thus results in a less disturbed or aggressive motion. 
An overview of the foothold evaluation stage is shown in~\fref{vfa_footholdeval_vpa_pose_eval}
where the tuple~$\fectuple$ of the \criteria in~\eref{fec_tuble} 
is extracted from the \gls{vfa} tuple $\heurtuple$ in~\eref{vfa_tuble}
to compute $\fecout$.
Then, using $\nominal$ and $\fecout$, we extract $\optimal$ as the safe foothold that is closest to $\nominal$.

\boldSubSec{Trajectory Adjustment}
The leg's swinging trajectory is adjusted once~$\optimal$ is computed.

\begin{remark}To compute the foothold evaluation, one can directly apply the exact mapping~$\g$. 
Yet, computing the foothold evaluation leads to 
evaluating the \criteria which is generally computationally expensive. 
Thus, to speed up the computation and to continuously run the \gls{vfa} online, 
we train a~\gls{cnn} to approximate the foothold evaluation~$\ghat$ using supervised learning.
Once trained, the \gls{vfa} can then be executed online using the \gls{cnn}.
The \gls{cnn} architecture of the foothold evaluation is explained in~\appref{cnn_approx}.
\label{remark_cnn_exact}
\end{remark}

\section{Vision-Based Pose Adaptation (VPA)}\label{sec_vpa}
The \gls{vpa} generates pose references that maximize the chances of the legs to reach safe footholds. 
This means that the robot pose has to be aware of what the legs are capable of and adapt accordingly. 
Therefore, the goal of the \gls{vpa} is to adapt the robot pose based on 
the same set of skills in the \criteria used by the~\gls{vfa}.

\subsection{Definitions and Notations}
\boldSubSec{Number of Safe Footholds}
As explained earlier, the \criteria takes a tuple~$\fectuple$ as an input 
and outputs the matrix $\fecout$.
Based on that, let us define the \textit{\gls{nsf}} 
\begin{equation}
\nsf := \mathrm{cardinal}( \{e \in \fecout : e=1   \} )
\label{cardinal}
\end{equation}
as the number of \textit{true} elements in 
the boolean matrix~$\fecout$.

\boldSubSec{Set of Safe Footholds}
Consider a set of tuples~$\settuple$
where each element $\elementtuple \in \settuple$ is a tuple defined as
\begin{equation}
T_i = (H, \elementheights,\bodyvel, \gaitparams)
\label{vpa_eq_2}
\end{equation}
and  $\elementheights \in \setheights$ is a hip height element in 
the set of hip heights~$\setheights$ (in the world frame).
All the tuple elements~$\elementtuple \in \settuple$
share the same heightmap~$\hmap$, body twist~$\bodyvel$ and gait parameters~$\gaitparams$.

Evaluating the \criteria~in~\eref{fec_evaluation} 
for every~$\elementtuple \in \settuple$ that corresponds to~$\elementheights \in \setheights$,
and computing the cardinal~in~\eref{cardinal} yields~$\elementfeasibles$ for every~$\elementtuple$.
This yields the~\textit{\gls{feasibles}}
which is a set containing 
the number of safe footholds $\nsf$ 
that are evaluated based on the \criteria 
given the set of tuples $\settuple$
that corresponds to the set of hip heights $\setheights$
but shares the same heightmap~$\hmap$, body twist~$\bodyvel$ and gait parameters~$\gaitparams$.

\subsection{From the Set of Safe Footholds to Pose Evaluation}\label{pose_evaluation}
The set of safe footholds \gls{feasibles} is one of the building blocks of the \gls{vpa}. 
To compute \gls{feasibles}, we compute the input tuple~$\heurtuplevpa$ 
\begin{equation}
\heurtuplevpa = (\hmapvpa,\bodyvel, \gaitparams)
\label{vpa_tuble}
\end{equation}
that we then augment with the hip heights $\elementheights$ 
in the hip heights set $\setheights$
yielding the set of tuples~$\settuple$. 
Then, we evaluate the \criteria~in~\eref{fec_evaluation} 
for every~$\elementtuple \in \settuple$,
and computing the cardinal~in~\eref{cardinal}.
This can be expressed by the mapping
\begin{equation}
{\gvpa~:~\heurtuplevpa\rightarrow\mathcal{F}}
\label{vpa_eq_3}
\end{equation}
which is referred to as \textit{pose evaluation}.
We can express $\setfeasibles$ as
\begin{equation}
\setfeasibles = 
\{ \elementfeasibles = g_{\mathrm{vpa},i}(\elementtuple)~\forall \elementtuple \in \settuple \}.
\label{vpa_4}
\end{equation}

\begin{remark}
Since $\setheights$ is an infinite continuous set,
so is~$\setfeasibles$
which is not numerically feasible to compute.
Hence, we sample a finite set~$\finitesetheights$ of $\hipheightsamples$ samples of hip heights
that results in a finite set of safe footholds~$\finitesetfeasibles$.
To use the set of safe footholds in an optimization problem, we need a continuous function. 
Thus, after we compute~$\finitesetfeasibles$,  
we estimate a continuous function~$\rbfn$
as explained next.
\label{rem3}
\end{remark}

An overview of the pose evaluation is shown in~\fref{vfa_footholdeval_vpa_pose_eval}
where the tuple~$\heurtuplevpa$ in~\eref{vpa_tuble} is augmented with the 
hip heights $\elementheights$
from~$\finitesetheights$ to construct the \criteria tuples~$\elementtuple$ in~\eref{vpa_eq_2}. 
For every tuple~$\elementtuple$, we evaluate the \criteria using~\eref{fec_evaluation} and 
compute~$\elementfeasibles$ in~\eref{cardinal} using the mapping in~\eref{vpa_eq_3}.
Finally, the set $\finitesetfeasibles$ includes all the elements~$\elementfeasibles$
as in~\eref{vpa_4}.

\subsection{Vision-based Pose Adaptation (VPA) Formulation}\label{sec_vpa_pipeline}
The \gls{vpa} has four main stages as shown in~\fref{fig_2}{(C)}.
First, \textit{heightmap extraction} that is similar to the \gls{vfa}.
Second, \textit{pose evaluation} where we compute $\finitesetfeasibles$. 
Third, \textit{function approximation} where we estimate~$\rbfn$ from~$\finitesetfeasibles$.
Fourth, \textit{pose optimization} where the optimal body pose is computed.

\boldSubSec{Heightmap Extraction}
We extract one heightmap~$\hmapvpa$ per leg
that is centered around the projection of the leg's hip location in the terrain map
($\mathrm{proj.}(p_h)$ instead of~$\nominal$).

\boldSubSec{Pose Evaluation}
After extracting the heightmaps, we compute~$\finitesetfeasibles$
from the mapping in~\eref{vpa_eq_3}
of the pose evaluation.
In the pose evaluation, the \criteria are evaluated 
for all hip heights in~$\finitesetheights$
given the input tuple~$\heurtuplevpa$ as shown in~\fref{vfa_footholdeval_vpa_pose_eval}.

\boldSubSec{Function Approximation}
In this stage, 
we estimate the continuous function~$\rbfn$ from~$\finitesetfeasibles$,
as explained in Remark~\ref{rem3}.
This is done by training a parameterized model
of the inputs $\elementtuple \in \finitesettuple$
and the outputs $\elementfeasibles \in \finitesetfeasibles$.
The result is the function (model)~$\rbfn$
that is parameterized by the model parameters~$w$.
The function approximation is detailed later in this section.

\boldSubSec{Pose Optimization}
Evaluating the~\criteria and approximating it with the function~$\rbfn$,
introduces a metric that represents the possible number of safe footholds for every leg. 
Based on this, the goal of the pose optimizer is to find the optimal pose that
will maximize the number of safe footholds for every leg (maximize~$\rbfn$)
while ensuring robustness.
The pose optimizer is detailed later in this section.

\begin{remark}
Similar to \remref{remark_cnn_exact},
one can directly apply the exact evaluation $\gvpa$ for a given~$\heurtuplevpa$. 
Yet, since this is computationally expensive, we rely 
on approximating the evaluation $\ghatvpa$ using a \gls{cnn}.
In fact,  the learning part is applied to both the pose evaluation 
and the function approximation. 
This means that the pose optimization is running online, 
outside the \gls{cnn}. 
The \gls{cnn} architecture of the pose evaluation is explained in~\appref{cnn_approx}.
\label{remark_cnn_exact_vpa}
\end{remark}

\subsection{Function Approximation}\label{sec_func_approx}
The goal of the function approximation is to approximate 
the set of safe footholds $\setfeasibles$
from the discrete set $\finitesetfeasibles$ computed in the pose evaluation stage.
This is done to provide the pose optimizer with a continuous function.
Given a dataset~$(\finitesetheights, \finitesetfeasibles)$ 
of hip heights~$z_{h_i}\in\finitesetheights$
and number of safe footholds~${\mathrm{n}_{\mathrm{sf},i}\in\finitesetheights}$,
the function approximation estimates a function $\rbfn(z_{h_i}, w)$
that is parameterized by the weights $w$.
Once the weights~$w$ are computed, 
the function estimate~$\rbfn(z_{h_i}, w)$ is then reconstructed and sent to the pose optimizer.

It is important to choose a function~$\rbfn$ that 
can accurately represent the nature of the number of safe footholds. 
The number of safe footholds approaches zero when the hip heights approach~$0$~or~$\infty$.
Thus, we want a function that fades to zero at the extremes
(Gaussian-like functions), and captures any asymmetry or flatness in the distribution. 
Hence, we use radial basis functions of Gaussians. 
With that in mind, we are looking for the weights~$w$ 
\begin{equation}
w = \text{arg}~\text{min}~S(w)
\label{eq_func_aprox_1}
\end{equation}
that minimize the cost~$S(w)$
\begin{equation}
S(w) = \sum_{i=1}^{N_h} (\mathrm{n}_{\mathrm{sf}, i} - \rbfn(z_{h_i}, w))^2
\label{eq_func_aprox_2}
\end{equation}
which is the sum of the squared residuals of 
$\mathrm{n}_{\mathrm{sf}, i}$ and $\rbfn(z_{h_i}, w)$.
$N_h$ is the number of samples (the number of the finite set of hip heights).
The function~$\rbfn(z_{h_i}, w)$ is the regression model 
(the approximation of~$\mathcal{F}$)
that is parameterized by~$w$.
The function~$\rbfn(z_{h_i}, w)$ is the weighted sum of the basis functions
\begin{equation}
\rbfn(z_{h_i}, w) = \sum_{e=1}^{E} w_e \cdot g(z_{h_i}, \Sigma_e, c_e)
\label{eq_func_aprox_3}
\end{equation}
where $w\in\Rnum^E$, and $E$ is the number of basis functions. The basis function is a radial basis function of Gaussian functions%
\begin{equation}
g(z_{h_i}, \Sigma_e, c_e) = \text{exp}(-0.5 (z_{h_i}-c_e)^T \Sigma_e^{-1} (z_{h_i} -c_e))
\label{eq_func_aprox_4}
\end{equation}
where $\Sigma_e$ and $c_e$ are the parameters of the Gaussian function.
Since the function model in~\eref{eq_func_aprox_3} is linear in the parameters, 
the weights of the function approximation can be solved analytically 
using least squares. 
In this work, we
keep the parameters of the Gaussians ($\Sigma$ and $c$) fixed.
Hence, the function 
$\rbfn$ is only parameterized by $w$.
For more information on regression with radial basis functions, 
please refer to \cite{Stulp2015} and~\appref{app_func_approx}.

\subsection{Pose Optimization}\label{sec_po}
The pose optimizer finds the robot's body pose~$u$ 
that maximizes the number of safe footholds for all the legs. 
This is casted as a non-linear optimization problem. 
The notion of safe footholds is provided by the function~$\rbfn(z_h, w)$ that
maps a hip height $z_h$ to a number of safe foothold~\gls{nsf}, 
and is parameterized by $w$.
Since the pose optimizer is solving for the body pose $u$, 
the function~$\rbfn(z_h)$ should be encoded using the body pose rather than the 
hip heights (${\rbfn = \rbfn(z_h(u))}$).
This is done by estimating the hip height as a function of the body pose 
(${z_h = z_h(u)}$) as shown in~\appref{po_appendix}.

\subsection{Single-Horizon Pose Optimization}
The pose optimization problem is formulated as
\begin{eqnarray}
\underset{u=[z_b,\beta, \gamma]}{\text{maximize~}}   &~& 
\mathcal{C} (\rbfn_l( z_{h_l}(z_b,\beta, \gamma) )) ~~ \forall l \in N_l
\label{po_cost_1} \\
\text{subject to} 
&~& u_{\min} \leq u \leq u_{\max} \label{po_u_1} \\
&~& \Delta u_{\min} \leq \Delta u \leq \Delta u_{\max} \label{po_delta_u_1}
\end{eqnarray}
where 
$u=[z_b,\beta, \gamma]\in\Rnum^3$ are the decision variables (robot body pose)
consisting of the robot height, roll and pitch, respectively,
$\mathcal{C}$ is the cost function,
$\rbfn_l$ is $\rbfn$ for every leg $l$ 
where $N_l=4$ is the number of legs, 
$z_{h_l}\in\Rnum$ is the hip height of the leg~$l$,
and
$u_{\min}$ and $u_{\max}$ are the lower and upper bounds
of the decision variables, respectively.
$\Delta u = u - u_{k-1}$ 
is the numerical difference of~$u$ 
where~$u_{k-1}$ is the output of~$u$ at the previous instant,
and
$\Delta u_{\min}$ and $\Delta u_{\max}$ are the lower and upper bounds
of $\Delta u$, respectively.
We can re-write~\eref{po_delta_u_1}~as
\begin{equation}
\Delta u_{\min} + u_{k-1} \leq u \leq \Delta u_{\max} + u_{k-1}. \label{po_delta_u_2}
\end{equation}

The cost function in~\eref{po_cost_1} maximizes~$\rbfn$ for all of the legs.
We designed several types of cost functions as detailed next.
The constraints in~\eref{po_u_1} and~\eref{po_delta_u_1}
ensure that the decision variables and their variations are bounded.

\subsection{Cost Functions}\label{sec_cost_options}
A standard cost function can be the sum of the squares of~$\rbfn_l$ for all 
of the legs
\begin{equation}
\mathcal{C}_{\mathrm{sum}} = \sum_{l=1}^{N_l=4} \Vert \rbfn_l (z_{h_l}) \Vert^2_Q 
\label{eq_cost_option_1}
\end{equation}
where another option could be the product of the squares of~$\rbfn_l$ for all of the legs
\begin{equation}
\mathcal{C}_{\mathrm{prod}} = \prod_{l=1}^{N_l=4} \Vert \rbfn_l (z_{h_l}) \Vert^2_Q .
\label{eq_cost_option_2}
\end{equation}

The key difference between 
an additive cost~$\mathcal{C}_{\mathrm{sum}}$ 
and a multiplicative cost~$\mathcal{C}_{\mathrm{prod}}$  
is that the latter puts equal weighting for each~$\rbfn_l$. 
This is important since we do not want the optimizer
to find a pose that maximizes~$\rbfn$ for one leg 
while compromising the other leg(s).
One can also define the cost 
\begin{equation}
\mathcal{C}_{\mathrm{int}} = \sum_{l=1}^{N_l=4} \Vert 
\int_{z_{h_l}-m}^{z_{h_l}+m} \rbfn_l (z_{h_l}) ~ d z_{h_l} ~
\Vert^2_Q 
\label{eq_cost_option_3}
\end{equation}
which is the \textit{sum of squared integrals}
that can be numerically approximated as
\begin{equation}
\int_{z_{h_l}-m}^{z_{h_l}+m} \rbfn_l (z_{h_l}) ~ d z_{h_l}
\approx
m \cdot ( \rbfn_l (z_{h_l}-m) + \rbfn_l (z_{h_l}+m)) 
\end{equation}
yielding
\begin{equation}
\mathcal{C}_{\mathrm{int}} = \sum_{l=1}^{N_l=4} \Vert    
m \cdot ( \rbfn_l (z_{h_l}-m) + \rbfn_l (z_{h_l}+m) ) 
\Vert^2_Q .
\label{eq_cost_option_4}
\end{equation}

In this cost option, 
we do not find the pose that maximizes~$\rbfn$.
Instead, we want to find the pose that maximizes 
the area around $\rbfn$ that is defined by the margin $m$.
Using $\mathcal{C}_{\mathrm{int}}$ is important since it adds 
robustness in case there is any error in the pose tracking during execution. 
Because of possible tracking errors during execution, 
the robot might end up in the pose~$u^*\pm m$ instead of~$u^*$.
If we use~$\mathcal{C}_{\mathrm{int}}$ as a cost function, 
the optimizer will find poses that maximizes the number of safe footholds
not just for~$u^*$ but within a vicinity of~$m$.
More details on the use of $\mathcal{C}_{\mathrm{int}}$ as a cost function
in the pose optimization of the~\gls{vpa} can be found in~\appref{cint_appendix}.

\subsection{Receding-Horizon Pose Optimization}\label{sec_receding_PO}
Adapting the robot's pose during dynamic locomotion 
requires reasoning about what is ahead of the robot:
the robot should not just consider its current state but also future ones. 
For that, we extend the pose optimizer to consider
the current and future states of the robot in a receding horizon manner. 
To formulate the receding horizon pose optimizer, 
instead of considering~$\rbfn_l~\forall~{l\in N_l}$ in the single horizon case, 
the pose optimizer will consider~$\rbfn_{l,j}~\forall~{l\in N_l},~ {j \in N_h}$
where~$N_h$ is the receding horizon number. 
We compute~$\rbfn_{l,j}$ in the same way explained in the pose evaluation stage. 
More details on computing~$\rbfn_{l,j}$ can be found in~\appref{receding_def}.

The receding horizon pose optimization problem is 
\begin{eqnarray}
\underset{u=[u_1^T, \cdots, u_{N_h}^T]}{\text{maximize}}  &~&
\sum_{j=1}^{N_h} \mathcal{C}_j (\rbfn_{l,j}( z_{h_{l,j}}(u_j) )) \nonumber\\
&+& \sum_{j=1}^{N_h-1} \Vert u_j - u_{j+1} \Vert \nonumber \\
&~&\forall l \in N_l,~ j \in N_h \label{receding_po_cost_1}\\
\text{subject to} &~&
u_{\min} \leq u \leq u_{\max} \\
&~& \Delta u_{\min} \leq \Delta u \leq \Delta u_{\max} 
\label{eq_receding_po}
\end{eqnarray}
where $u=[u_1^T, \cdots, u_j^T, \cdots, u_{N_h}^T]\in\Rnum^{3N_h}$
are the decision variables during the entire receding horizon $N_h$.
Each variable~$u_j = [z_{b,j},\beta_j, \gamma_j]\in\Rnum^3$ is 
the optimal pose of the horizon~$j$.

The first term in~\eref{receding_po_cost_1} 
is the sum of the cost functions~$\mathcal{C}_j$ during the entire horizon~\text{($\forall j \in N_h$)}.
The cost~$\mathcal{C}_j$ can be any of the aforementioned cost functions.
The second term in~\eref{receding_po_cost_1} 
penalizes the deviation between two consecutive  
optimal poses within the receding horizon
($u_j$ and $u_{j+1}$). 
The second term is added so that each optimal pose $u_j$
is also taking into account the optimal pose 
of the upcoming sequence~$u_{j+1}$ (to connect the solutions in a smooth way).
Similar to the single horizon pose optimizer, 
$u_{\min}$ and $u_{\max}$ are the lower and upper bounds
of the decision variables, respectively.
Furthermore,~$\Delta u$ denotes the numerical difference of $u$,
while~$\Delta u_{\min}$ and $\Delta u_{\max}$ are the lower and upper bounds
of $\Delta u$, respectively.
Note that the constraints of the single horizon and the receding horizon 
are of different dimensions.

\section{System Overview}\label{sec_sys_over}
Our locomotion framework that is shown in~\fref{fig_2}(D) is based on the \gls{rcf}~\cite{Barasuol2013}.
\gls{vital} complements the \gls{rcf} with an exteroceptive terrain-aware layer composed of the \gls{vfa} and the \gls{vpa}. 
\gls{vital} takes the robot states, the terrain map and user commands as inputs, 
and sends out the selected footholds and body pose to the \gls{rcf} (perceptive) layer.
The \gls{rcf} takes the robot states and the references from \gls{vital}, 
and uses them inside a motion generation and a motion control block. 
The motion generation block generates the trajectories of the leg and the body, 
and adjusts them with the reflexes from~\cite{Barasuol2013,Focchi2013}. 
The legs and body references from the motion generation block are sent to the motion control block.
The motion control block consists of a \gls{wbc}~\cite{Fahmi2019} 
that generates desired torques that are tracked via a low-level torque
controller~\cite{Boaventura2015}, and sent to the robot's joints.
The framework also includes a state estimation block that feeds 
back the robot states to each of the aforementioned layers~\cite{Camurri2017}. 
More implementation details on~\acrshort{vital} and the entire framework is in~\appref{misc_details}.

We demonstrate \gls{vital} on 
the \unit[90]{kg} \acrshort{hyq} and the \unit[140]{kg} \acrshort{hyqreal} quadruped robots.
Each leg of the two robots has 3~degrees of freedom (3~actuated joints). 
The torques and angles of the 12~joints of both robots are directly measured. 
The bodies of \acrshort{hyq} and \acrshort{hyqreal} have a tactical-grade~\gls{imu} (KVH 1775). 
More information on \acrshort{hyq} and \acrshort{hyqreal}
can be found in~\cite{Semini2011}, and~\cite{Semini2019} respectively.

We noticed a significant drift in the states of the robots in experiment. 
To tackle this issue, the state estimator fused the data from a motion capture system and the \gls{imu}.
This reduced the drift in the base states of the robots albeit not eliminating it completely. 
Improving the state estimation is an ongoing work and is out of the scope of this article.
We used the grid map interface~\cite{Fankhauser2016} to get the terrain map in simulation. 
Due to the issues with state estimation on the real robots, 
we constructed the grid map before the experiments, and used the motion capture system to locate
the map with respect to the robot.

\section{Results}\label{sec_results_vital}
We evaluate~\gls{vital} on~\acrshort{hyq} and~\acrshort{hyqreal}.
We consider all the \criteria mentioned earlier for the~\gls{vfa} and the~\gls{vpa}.
We use the receding horizon pose optimizer of~\eref{eq_receding_po}
and the sum of squared integral of~\eref{eq_cost_option_4}.
We choose stair climbing as an application for~\gls{vital}. 
Climbing stairs is challenging for~\acrshort{hyq} 
due to its limited leg workspace in the sagittal plane.
Videos associated with the upcoming results can be found in the supplementary materials and~\cite{Video}. 
Finally, an analysis of the accuracy of the~\glspl{cnn} and the computational time of~\gls{vital}
can be found in~\appref{est_acc} and~\appref{comp_anal}, respectively.

\subsection{Climbing Stairs (Simulation)}\label{sec_sim1}
\begin{figure}
\centering
\includegraphics[width=0.9\columnwidth]{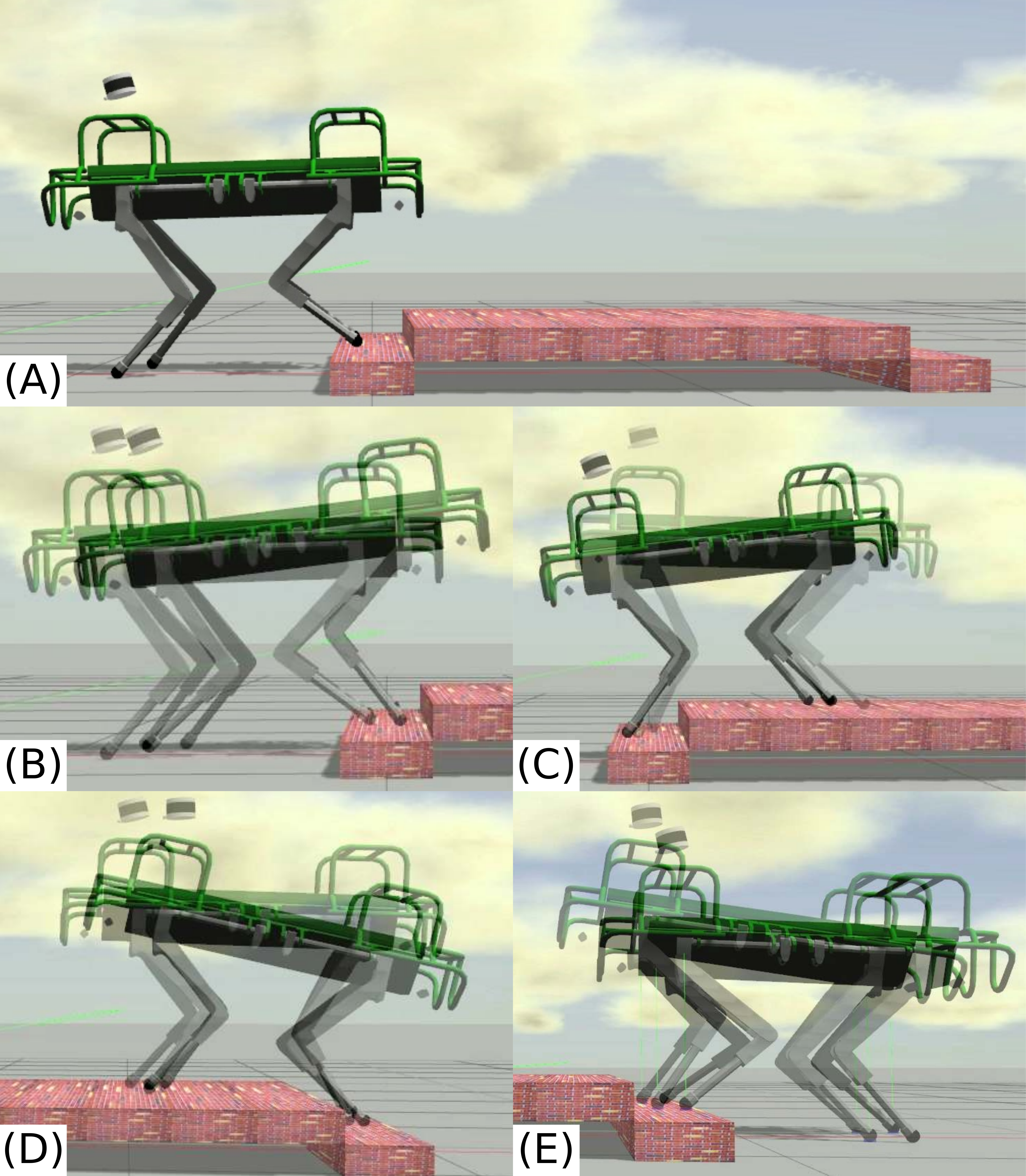}		
\caption
[\acrshort{hyq} climbing stairs in simulation.]
{
\acrshort{hyq} climbing stairs in simulation.
(A)~The full scenario.
(B)~The~robot~pitches up to allow for safe footholds for the front legs.
(C)~The~robot~lifts up the hind hips to avoid hind leg collisions with the step. 
(D)~The~robot~pitches down to allow for safe footholds for the front legs. 
(E)~The~robot~lowers the hind hips to allow for safe footholds for the legs when stepping down.
}
\label{fig_3}
\end{figure}

\begin{figure}
\centering
\includegraphics[width=\columnwidth]{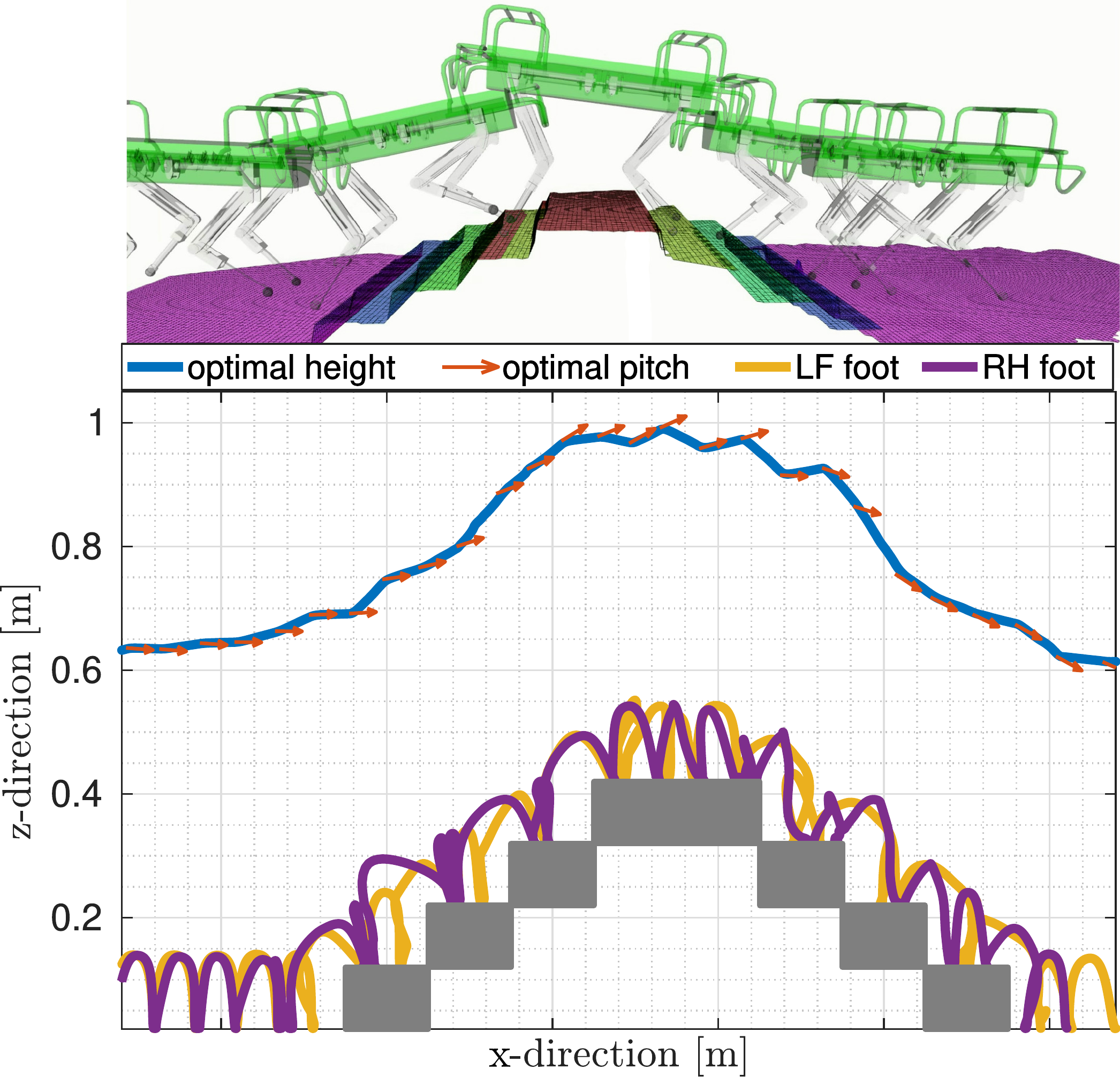}		
\caption[Climbing Stairs: A More Complex Scenario.]{Climbing Stairs: A More Complex Scenario. 	
Top: Overlayed screenshots of \acrshort{hyq} climbing stairs.
Bottom: the optimal height and corresponding pitch (presented by the arrows) 
and the foot trajectories of \acrshort{lf} and \acrshort{rh} legs.}
\label{fig_4}
\end{figure}

We carried out multiple simulations where~\acrshort{hyq} 
is climbing the stairs shown in~\fref{fig_3}.
Each step has a rise of~\unit[10]{cm}, and a go of~\unit[25]{cm}.
\acrshort{hyq} is commanded to trot with a desired forward velocity 
of~\unit[0.2]{m/s} using the~\gls{vpa} and the~\gls{vfa}.
Figure~\ref{fig_3} shows screenshots of one simulation run,
and~\vref{1} shows three simulation runs.

Figure~\ref{fig_3} shows 
the ability of the \gls{vpa} in adapting the robot pose to increase the chances
of the legs to succeed in finding a safe foothold. 
In~\fref{fig_3}(B), 
\acrshort{hyq} raised its body and pitched upwards
so that the front hips are raised 
to increase the workspace of the front legs when stepping up. 
In~\fref{fig_3}(C), 
\acrshort{hyq} raised its body and pitched downwards 
so that the hind hips are raised.
This is done for two reasons. 
First, 
to have a larger clearance between the hind legs and the obstacle, 
and thus avoiding leg collision with the edge of the stairs.
Second,
to increase the workspace of the hind legs when stepping up, 
and thus avoiding reaching the workspace limits and
collisions along the foot swing trajectory.
In~\fref{fig_3}(D), 
\acrshort{hyq} lowered its body and pitched downwards
so that the front hips are lowered. 
This is done for two reasons:
First, 
to increase the workspace of the front legs when stepping down, 
and thus avoiding reaching the workspace limits. 
Second, 
to have a larger clearance between the front legs and the obstacle, 
and thus avoiding leg collisions. 
In~\fref{fig_3}(E), 
\acrshort{hyq} lowered its hind hips
to increase the hind legs' workspace when stepping down.

Throughout these simulations, the robot continuously adapted its body pose and its feet
to find the best trade-off between increasing the kinematic feasibility, 
and avoiding trajectory and leg collision. 
This can be seen in~\vref{1}
where the robot's legs and the corresponding feet trajectories
never collided with the terrain. 
The robot took multiple steps around the same foot location before stepping over an obstacle.
The reason behind this is that the robot 
waited for the \gls{vpa} to change the pose and allow for safe footholds,
and then the \gls{vfa} took the decision of stepping over the obstacle.

We carried out another scenario where \acrshort{hyq} is climbing the stairs setup in~\fref{fig_4}
where each step has a rise of~\unit[10]{cm}, and a go of~\unit[25]{cm}.
\acrshort{hyq} is commanded to trot with a desired forward velocity of $0.2$\unit{m/s} using~\acrshort{vital}.
The results are reported in~\fref{fig_4} and~\vref{2}. 
Figure~\ref{fig_4} shows the robot's height and pitch based on the \acrshort{vpa}, 
and the corresponding feet trajectories of the~\acrshort{lf} leg and the~\acrshort{rh} leg based on the \acrshort{vfa}.
\acrshort{hyq}'s behavior was similar to the previous section:
it accomplished the task without collisions or reaching workspace limits.

\subsection{Climbing Stairs (Experiments)}\label{sec_fig_7}
\begin{figure*}
\centering
\includegraphics[width=\textwidth]{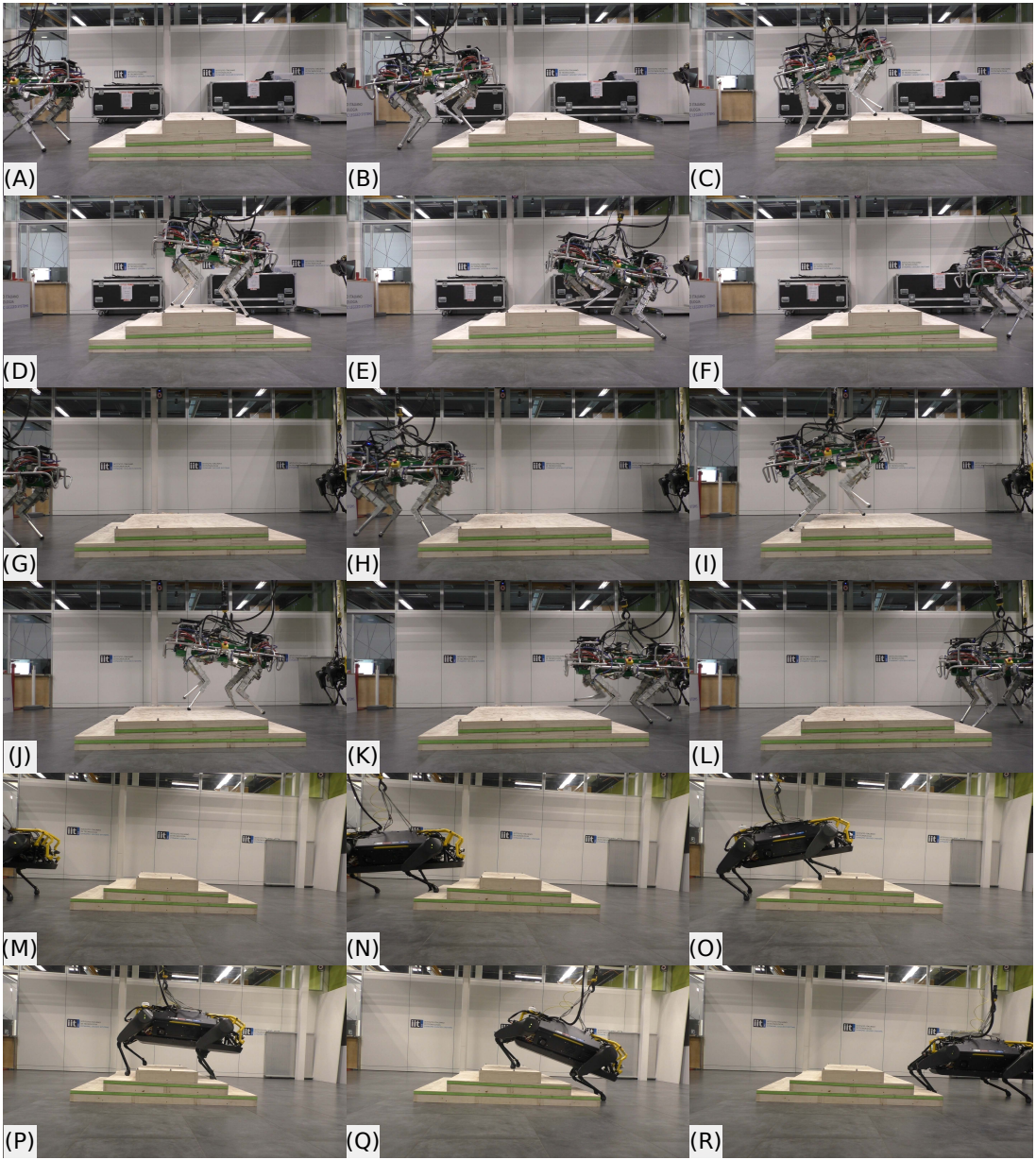}		
\caption
[\acrshort{hyq} and \acrshort{hyqreal} climbing stairs in experiment.]
{\acrshort{hyq} and \acrshort{hyqreal} climbing stairs in experiment.
(A-F)~\acrshort{hyq} crawling over with~\unit[0.1]{m/s} commanded forward velocity.
(G-L)~\acrshort{hyq} trotting over with~\unit[0.25]{m/s} commanded forward velocity.
(M-R)~\acrshort{hyqreal} crawling over with~\unit[0.2]{m/s} commanded forward velocity.
}
\label{fig_6}
\end{figure*} 
\begin{figure}[t!]
\centering
\includegraphics[width=\columnwidth]{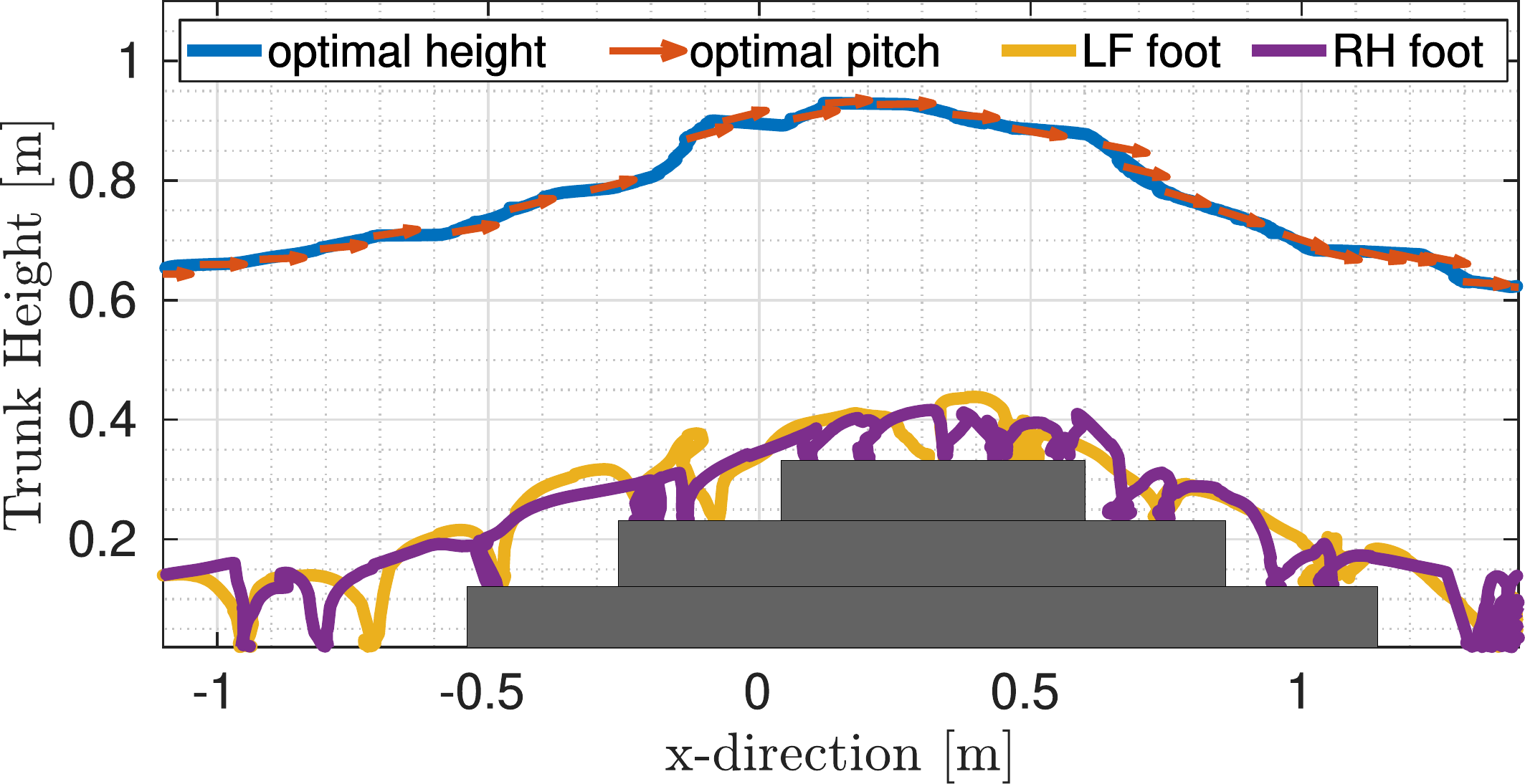}		
\caption
[\acrshort{hyq} climbing stairs in experiment.]
{
\acrshort{hyq} climbing stairs in experiment.
The figure shows the optimal height and corresponding pitch (presented by the arrows) 
based on the \acrshort{vpa}, 
and the foot trajectories of the~\acrshort{lf} and~\acrshort{rh} legs
based on the \acrshort{vfa}.
}
\label{fig_7}
\end{figure} 

To validate \gls{vital} in experiments, 
we created three sets of experiments using
the setups shown in~\fref{fig_1}. 
Each step has a rise of~\unit[10]{cm}, and a go of~\unit[28]{cm}.

In the first set of experiments, 
\acrshort{hyq} is commanded to crawl over the setups in~\fref{fig_1}(A,B)
with a desired forward velocity of~\unit[0.1]{m/s} using the \gls{vpa} and the \gls{vfa}.
\text{Figures~\ref{fig_6}(A-F)} show screenshots of one trial.
\vref{3} shows \acrshort{hyq} climbing back and forth the setup in~\fref{fig_1}(B) five times. 
\vref{4} shows \acrshort{hyq} climbing the~\fref{fig_1}(A) setup, which is reported in~\fref{fig_7}. 
Figure~\ref{fig_7} shows the robot's height and pitch based on the \gls{vpa}, 
and the corresponding feet trajectories of the \gls{lf} leg and the \gls{rh} leg based on the \gls{vfa}.
This set of experiments confirms that~\gls{vital} is effective on the real platform.
The robot managed to accomplish the task without collisions or reaching workspace limits. 

In the second set of experiments, \acrshort{hyq} is commanded to trot 
over the setup in~\fref{fig_1}(B) with a desired forward velocity of~\unit[0.25]{m/s} using the \gls{vpa} and the \gls{vfa}.
Figures~\ref{fig_6}(\text{G-L}) show separate screenshots of this trial.
\vref{5} shows three trials of \acrshort{hyq} climbing the same setup. 
This set of experiments shows that~\gls{vital} can handle different gaits.

Finally, in the third set of experiments,
\acrshort{hyqreal} is commanded to crawl 
over the setup in~\fref{fig_1}(C) with a desired forward velocity of~\unit[0.2]{m/s} 
using the~\gls{vpa} and the~\gls{vfa}.
The results are reported in~\fref{fig_6}(M-R) that show screenshots of this trial.
\vref{6} shows \acrshort{hyqreal} climbing this stair setup (\fref{fig_1}(C)). 
This set of experiments shows that~\gls{vital} can work on different legged platforms.

\begin{table}[t!]
\centering 
\caption{Mean Absolute Tracking Errors of the Body Pitch~$\beta$ and Height~$z_b$
of \acrshort{hyq} \& \acrshort{hyqreal} using \acrshort{vital} in the Experiments 
detailed in \sref{sec_fig_7} and in~\fref{fig_1}.\label{tab_mae_exp}}
\renewcommand{\arraystretch}{1.25}
\begin{tabular}{lcc}
\hline \hline
Description                                              &  $\beta$~[deg] & $z_b$~[cm] \\ \hline
Exp. (A): \acrshort{hyq} crawling at \unit[0.1]{m/s}     &  1.0           & 1.5 \\
Exp. (B): \acrshort{hyq} trotting at \unit[0.2]{m/s}     &  1.0           & 2.3 \\
Exp. (C): \acrshort{hyqreal} crawling at \unit[0.2]{m/s} &  1.0           & 4.2 \\
\hline \hline
\end{tabular}
\end{table}

The tracking performance of these three sets of experiments are shown in~\tref{tab_mae_exp}.
The table shows the mean absolute tracking errors of 
the body pitch~$\beta$ and height~$z_b$ of~\acrshort{hyq} and~\acrshort{hyqreal}.

\subsection{Climbing Stairs with Different Forward Velocities}\label{sec_sim_4}
\begin{figure}[t!]
\centering
\includegraphics[width=0.9\columnwidth]{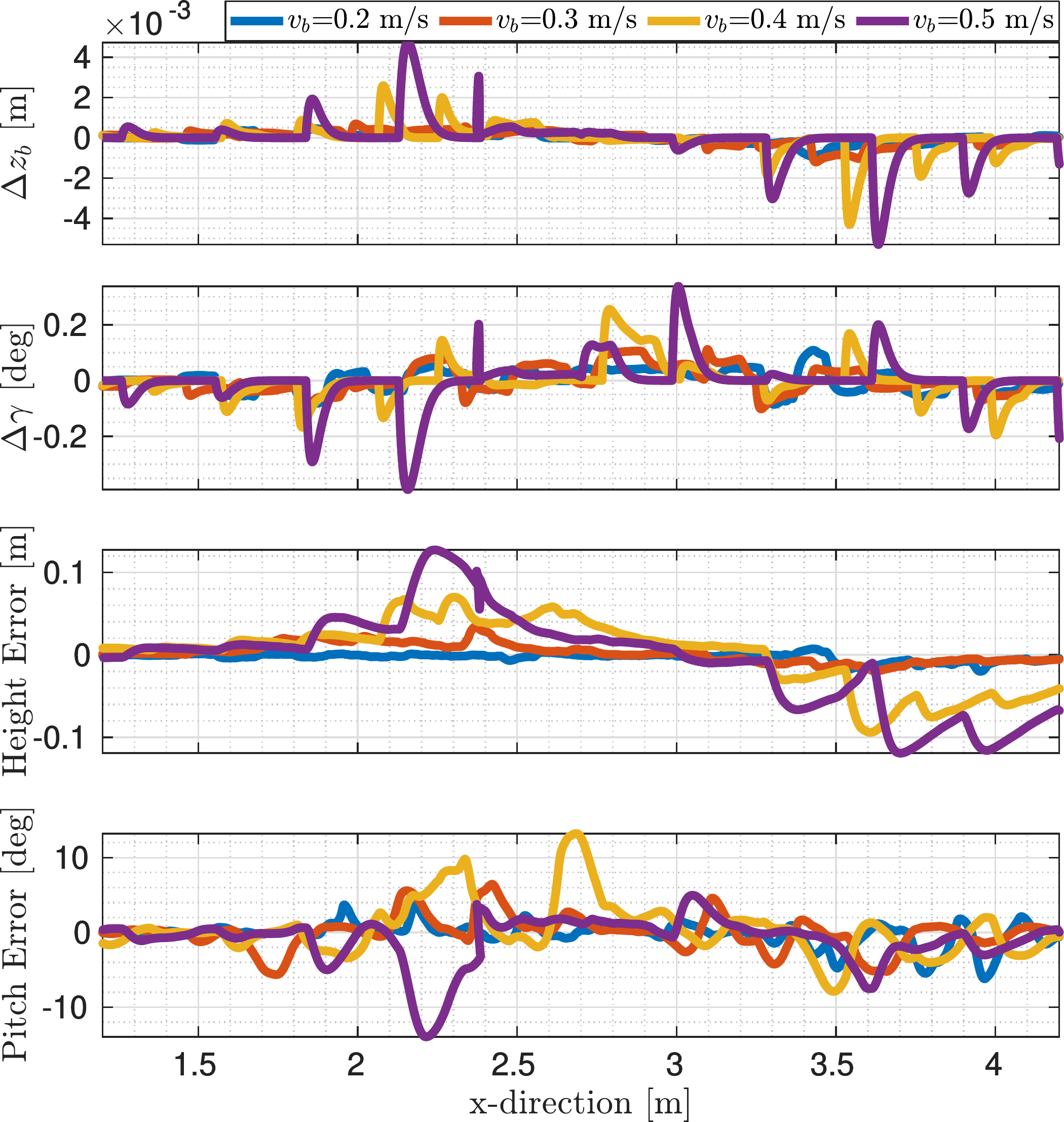}		
\caption
[\acrshort{hyq}'s Performance under Different Commanded Velocities]
{\acrshort{hyq}'s performance using~\gls{vital} under different commanded velocities.
The top two plots show the numerical difference of the body height and pitch ($\Delta z_b$ and $\Delta \gamma$), 
and the bottom two plots show the tracking errors of the body height and pitch.}
\label{fig_8}
\end{figure} 

\begin{table}[t!]
\centering 
\caption{Mean Absolute Tracking Errors of the Body Pitch~$\beta$ and Height~$z_b$
of  \acrshort{hyq} using \acrshort{vital} 
in Simulation with Different Forward Velocities
as Detailed in~\sref{sec_sim_4} and~\fref{fig_8}.\label{tab_mae_sim}}
\renewcommand{\arraystretch}{1.25}
\begin{tabular}{lcc}
\hline \hline
Description & $\beta$~[deg] & $z_b$~[cm]   \\ \hline
Sim. (A): \acrshort{hyq} trotting at \unit[0.2]{m/s} &  0.7 & 0.3 \\
Sim. (B): \acrshort{hyq} trotting at \unit[0.3]{m/s} &  1.1 & 0.7 \\
Sim. (C): \acrshort{hyq} trotting at \unit[0.4]{m/s} &  2.3 & 2.6 \\
Sim. (D): \acrshort{hyq} trotting at \unit[0.5]{m/s} &  1.6 & 3.8 \\
\hline \hline
\end{tabular}
\end{table}

We evaluate the performance of \acrshort{hyq}
under different commanded velocities using~\gls{vital}. 
We carried out a series of simulations using the stairs setup shown in~\fref{fig_4}. 
\acrshort{hyq} is commanded to trot at four different forward velocities:
\unit[0.2]{m/s}, \unit[0.3]{m/s}, \unit[0.4]{m/s}, and~\unit[0.5]{m/s}.
The results are reported in~\fref{fig_8} and in~\vref{7}.
Additionally, 
the tracking performance of these series of simulations
at the four different forward velocities are shown in~\tref{tab_mae_sim}.
The table shows the mean absolute tracking errors of  
the body pitch~$\beta$ and height~$z_b$
of~\acrshort{hyq} at the corresponding commanded forward velocity.
Figure~\ref{fig_8} shows the numerical differences~$\Delta z_b$ and~$\Delta\gamma$, 
and the tracking errors of the body height and pitch, respectively. 

\acrshort{hyq} was able to climb the stairs terrain under different commanded velocities.
However, as the commanded velocity increases, 
\acrshort{hyq} started having faster (abrupt) changes in the body pose
as shown in the top two plots of~\fref{fig_8}. 
As a result, 
the height and pitch tracking errors
increase proportionally to the commanded speed
as shown in the bottom two plots of~\fref{fig_8}.  
This can also be seen in~\tref{tab_mae_sim} where the mean absolute tracking errors
of the pitch and height increase proportionally to the commanded speed.

Similarly to~\acrshort{hyq}, we evaluate \gls{vital} on \acrshort{hyqreal} 
and commanded it to trot with five different forward velocities:
\unit[0.2]{m/s}, \unit[0.3]{m/s}, \unit[0.4]{m/s}, \unit[0.5]{m/s}, and~\unit[0.75]{m/s}.
We report this simulation in~\vref{8}
where we show that \gls{vital} is robot independent. 
Yet, since the workspace of \acrshort{hyqreal} is larger than \acrshort{hyq}, 
this scenario was more feasible to traverse for \acrshort{hyqreal}.
Thus, \acrshort{hyqreal} was able to reach a higher commanded velocity than 
the ones reported for \acrshort{hyq}.

\subsection{Comparing the \acrshort{vpa} with a Baseline (Experiments)}\label{tbr_result}
\begin{figure}[t!]
\centering
\includegraphics[width=0.9\columnwidth]{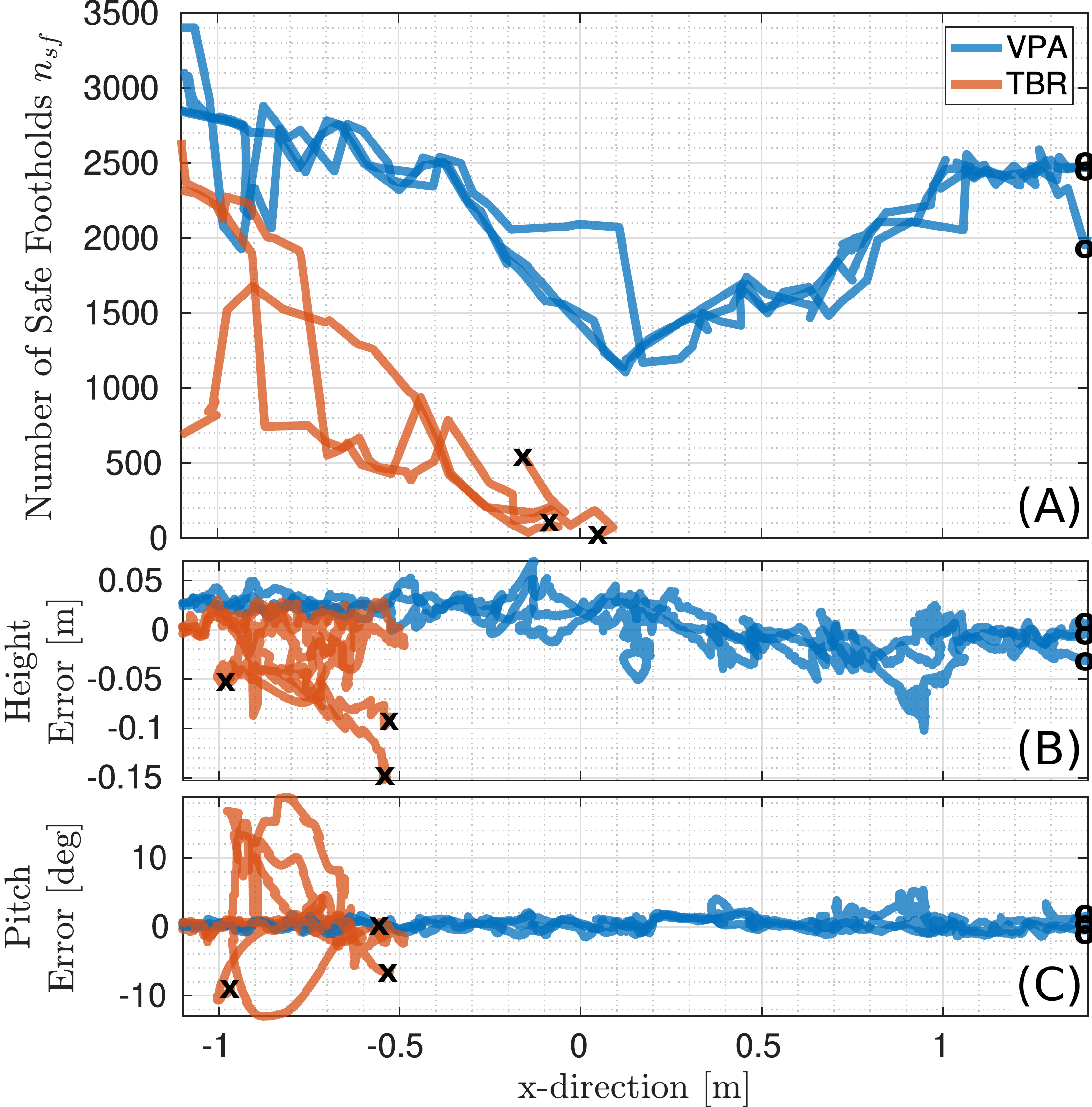}		
\caption
[The difference between the \acrshort{vpa} and the \acrshort{tbr}.]
{ 
The difference between the \acrshort{vpa} and the \acrshort{tbr} in six experiments (3 each).
(A)~The number of safe footholds corresponding to the robot pose.
(B,C)~The body height and pitch tracking errors, respectively. 
Circles~(\textbf{o})  and crosses (\textbf{x})  are successful and failed trials, respectively.
Unlike the~\acrshort{vpa}, the~\acrshort{tbr} failed to climb the stairs
because the~\acrshort{tbr} resulted in almost no safe footholds for the four legs to reach.
}
\label{fig_9}
\end{figure} 
We compare the \gls{vpa} with another vision-based pose adaptation strategy:
the \gls{tbr}~\cite{Villarreal2019}.
The \gls{tbr} generates pose references based on the footholds selected by the \gls{vfa}. 
The~\gls{tbr} fits a plane that passes through the given selected footholds, 
and sets the orientation of this plane as a body orientation reference to the robot.
The elevation reference of the \gls{tbr} is a constant distance
from the center of the approximated plane that passes through 
the selected footholds.
We chose the~\gls{tbr} instead of an optimization-based strategy 
since the latter does not provide references that are fast enough with respect to the~\gls{vpa}.

Using the stairs setup in~\fref{fig_1}(A), we conducted six experimental trials:
three with the~\gls{vpa} and three with the~\gls{tbr}.  
All trials were with the \gls{vfa}. 
In all trials, 
\acrshort{hyq} is commanded to crawl with a desired forward velocity of~\unit[0.1]{m/s}.
The results are reported in~\fref{fig_9} and~\vref{9}. 
Figure~\ref{fig_9}(A) shows the number of safe footholds corresponding to the robot pose
from the~\gls{vpa} and the~\gls{tbr}. 
The robot height and pitch tracking errors are shown in \fref{fig_9}(B,C), respectively.

As shown in~\vref{9}, \acrshort{hyq} failed to climb the stairs with the~\gls{tbr}, 
while it succeeded with the~\gls{vpa}. 
This is because, unlike the~\gls{vpa},
the~\gls{tbr} does not aim to 
put the robot in a pose that maximizes the chances of the legs to succeed in finding safe footholds. 
As shown in~\fref{fig_9}(A), the number of safe footholds from using the \gls{tbr} was below the ones from using the \gls{vpa}. 
During critical periods when the robot was around \unit[0]{m} in the x-direction,
the number of safe footholds from using the~\acrshort{tbr} almost reached zero.
The low number of safe footholds for the~\gls{tbr} compared to the~\gls{vpa}
is reflected in the tracking of the robot height and pitch as shown in 
{\fref{fig_9}(B,C)} 
where the tracking errors from the~\gls{tbr} were higher than the~\gls{vpa}.

The difference between the \gls{vpa} and the \gls{tbr} can be further explained in~\vref{9}.  
When the \gls{tbr} is used, the robot is adapting its pose \textit{given} the selected foothold. 
But, if the selected foothold is not reached, or if there is a high tracking error, 
the robot reaches a body pose that results in a smaller number of safe footholds. 
Thus, the feet end up colliding with the terrain and hence the robot falls. 
On the other hand, the~\gls{vpa} is able to put the robot 
in a pose that maximizes the number of safe footholds. 
As a result, the feet found alternative safe footholds to select from, 
which resulted in no collision, and succeeded in climbing the stairs.
The~\gls{vpa} optimizes for the number of safe footholds.
Thus, if there is a variation around the optimal pose (tracking error),
the~\gls{vfa} still finds more footholds to step on, which is not the case~with~the~\gls{tbr}.

\subsection{Comparing the \acrshort{vpa} with a Baseline (Simulation)}\label{tbr_result_sim}
\begin{figure}[h!]
	\centering
	\includegraphics[width=0.9\columnwidth]{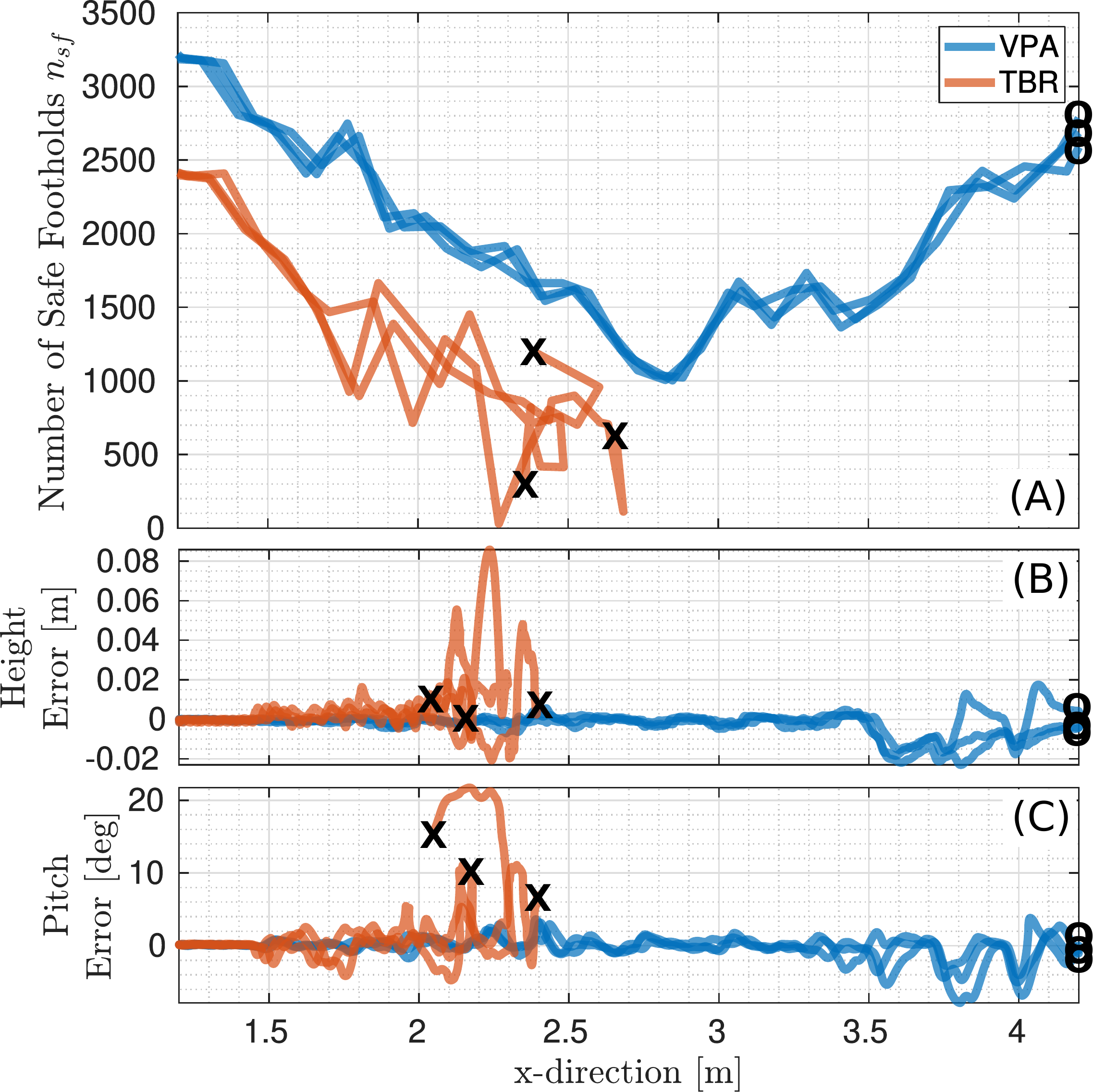}		
	\caption[The difference between the \acrshort{vpa} and the \acrshort{tbr} in six simulations (3 each).]
	{The difference between the \acrshort{vpa} and the \acrshort{tbr} in six simulations (3 each).
		(A)~The number of safe footholds corresponding to the robot pose.
		(B,C)~The body height and pitch tracking errors, respectively. 
		Circles~(\textbf{o})  and crosses (\textbf{x})  are successful and failed trials, respectively.
		Unlike the~\acrshort{vpa}, the~\acrshort{tbr} failed to climb the stairs
		because the~\acrshort{tbr} resulted in almost no safe footholds for the four legs to reach.}
	\label{fig_res_3_1}
\end{figure}

Similar to experiments, and using the stairs setup in~\fref{fig_4}, we compare  the~\acrshort{vpa} with  the~\acrshort{tbr}.
We conducted six simulations: three with the~\acrshort{vpa} and three with the~\acrshort{tbr}.  
All trials were with the \acrshort{vfa}. 
In all trials, \acrshort{hyq} is commanded to trot with a $0.2$\unit{m/s} desired forward velocity.
The results are reported in~\fref{fig_res_3_1} and~\vref{10}. 
Figure~\ref{fig_res_3_1}(A) shows the number of safe footholds corresponding to the robot pose from the~\acrshort{vpa}, and the~\acrshort{tbr}. 
The tracking errors of the robot height and pitch are shown in~\fref{fig_res_3_1}(B,C), respectively.
These trials show that~\acrshort{hyq} failed to climb the stairs using the~\acrshort{tbr}, while it succeeded using the~\acrshort{vpa}.

\subsection{Climbing Stairs with Gaps}\label{sec_sim_7}
\begin{figure}[h!]
\centering
\includegraphics[width=0.7\columnwidth]{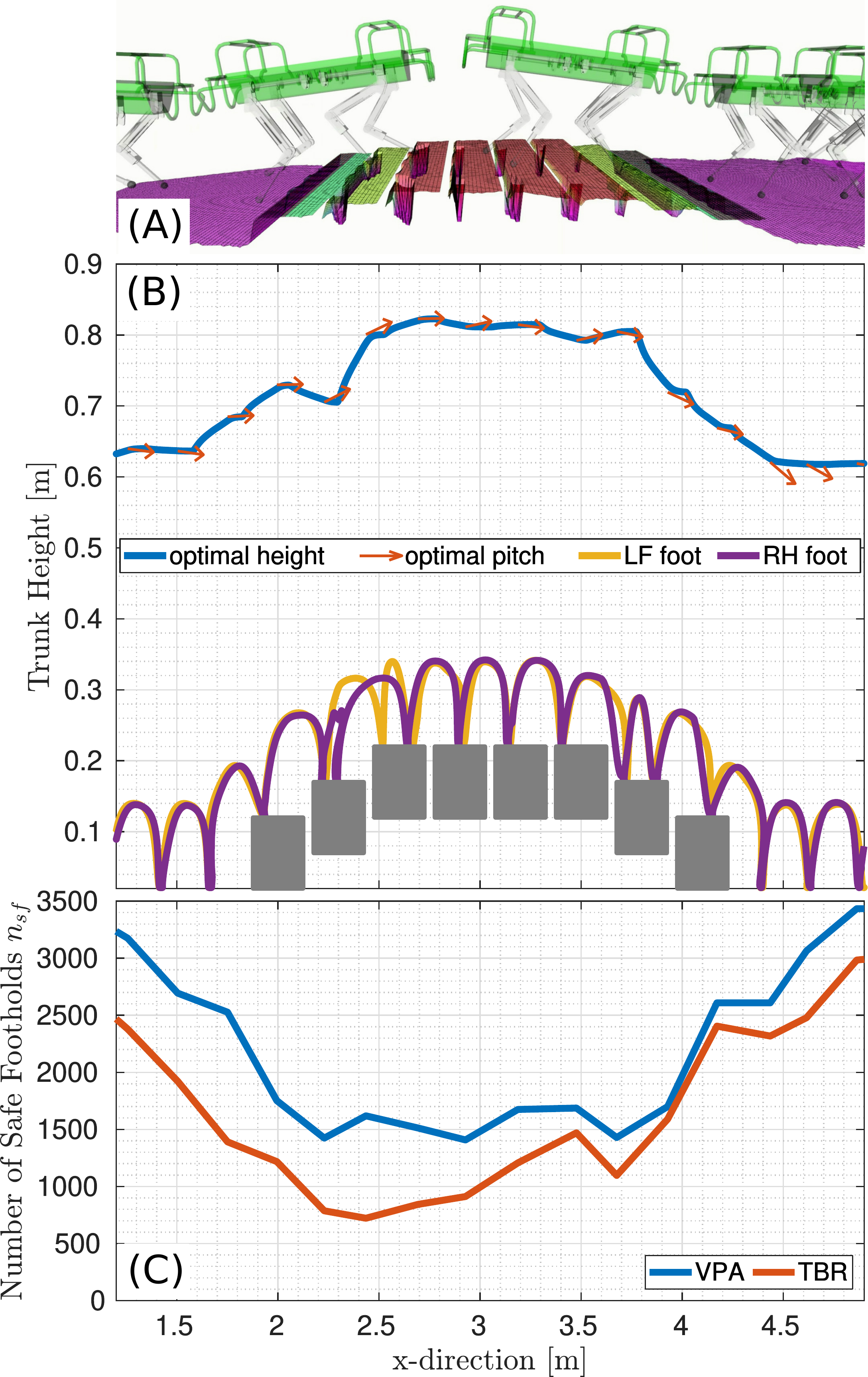}
\caption[\acrshort{hyq} climbing gapped stairs.]{\acrshort{hyq} climbing gapped stairs.
(A)~Screenshots of~\acrshort{hyq} climbing the setup.
(B)~The robot's height and pitch based on the~\acrshort{vpa}, and the corresponding feet trajectories of the~\acrshort{lf} and~\acrshort{rh} legs based on the~\acrshort{vfa}.
(C)~The number of safe footholds using the~\acrshort{vpa} and the~\acrshort{tbr}.}
\label{fig_5}
\end{figure}

We show \acrshort{hyq}'s capabilities of climbing stairs with gaps using \acrshort{vital}, and we compare the~\acrshort{vpa} with the~\acrshort{tbr}.
In this scenario, \acrshort{hyq} is commanded to trot at~\unit[0.4]{m/s}.
Figure~\ref{fig_5}(A) shows overlayed screenshots of the simulation and the used setup.
Figure~\ref{fig_5}(B) shows the robot's height and pitch based on the~\acrshort{vpa}, 
and the corresponding feet trajectories of the~\acrshort{lf} leg and the~\acrshort{rh} leg based on the~\acrshort{vfa},
and Fig.~\ref{fig_5}(C) shows the number of safe footholds using the~\acrshort{vpa} and the~\acrshort{tbr}.
Because of~\acrshort{vital}, \acrshort{hyq} was able to climb the stairs with gaps 
while continuously adapting its pose and feet. 
Furthermore, the number of safe footholds from using the~\acrshort{tbr} is always lower than from using the~\acrshort{vpa},
which shows that indeed the~\acrshort{vpa} outperforms the~\acrshort{tbr}.
\vref{11} shows the output of this simulation using~\gls{vital}.

\begin{figure}[h!]
\centering
\includegraphics[width=0.9\columnwidth]{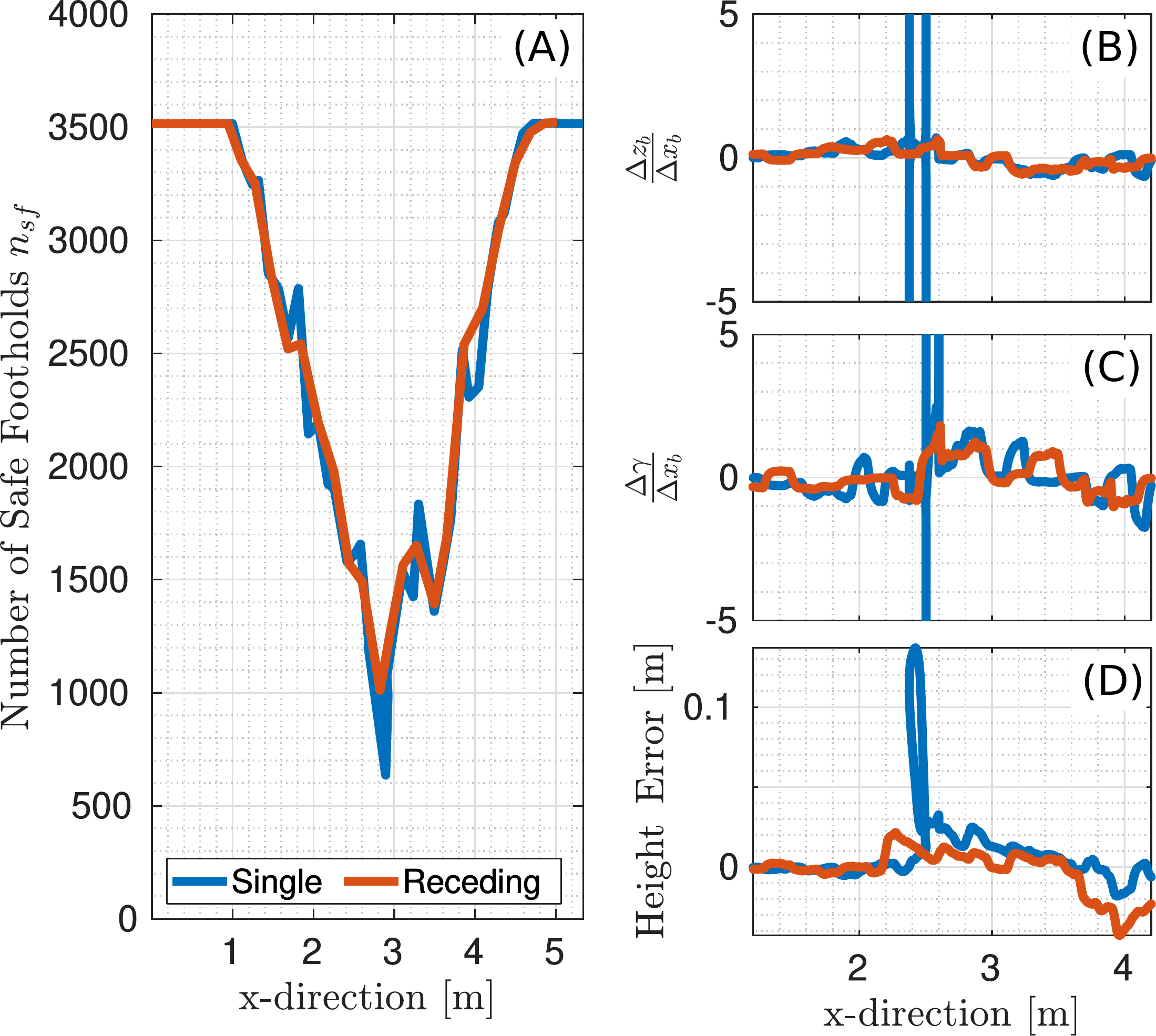}
\caption[Pose Optimization: Single vs. Receding Horizons.]
{Pose Optimization: Single vs. Receding Horizons.
(A)~The number of safe footholds.
(B)~Variation (numerical difference) of the robot's height. 
(C)~Variation (numerical difference) of the robot's pitch. 
(D)~The tracking error of the robot's height.}
\label{res_5_1}
\end{figure} 
\subsection{Pose Optimization: Single vs. Receding Horizons}\label{sec_sim_5}
To analyze the differences between the receding horizon and the single horizon 
in pose optimization, we use the stairs setup in~\fref{fig_4}
with a commanded forward velocity of~\unit[0.4]{m/s}, 
and report the outcome in~\fref{res_5_1} and~\vref{12}. 
The main advantage of using a receding horizon instead of a single horizon 
is that the pose optimization can consider future decisions. 
Thus, if the robot is trotting at higher velocities,
the pose optimizer can adapt the robot's pose before hand. 
This can result in a better adaptation strategy 
with less variations in the generated optimal pose.
Thus, we analyze the two approaches by taking a look at the variations in the body pose
\begin{equation}
\acute{z}_b = \frac{\Delta z_b}{\Delta x_b} 
~,~~~\text{and}~~~  
\acute{\gamma} = \frac{\Delta \gamma}{\Delta x_b}
\end{equation}
where $\acute{z}_b$ and $\acute{\gamma}$
are the numerical differences (variations) of 
the robot height $z_b$ and pitch $\gamma$ 
with respect to the robot forward position $x_b$, respectively.

Figure~\ref{res_5_1} reports the differences between the two cases.
The number of safe footholds is shown in~\fref{res_5_1}(A).
The variations in $\acute{z}_b$ and $\acute{\gamma}$ 
are shown in~\fref{res_5_1}(B,C), respectively. 
Finally, the tracking error of the robot's height is shown in~\fref{res_5_1}(D).
As shown in Figure~\ref{res_5_1} the receding horizon 
resulted in less variations in the body pose compared to 
the single horizon. 
This resulted in a smaller tracking error for the receding horizon
in the body height, which resulted in slightly larger number of safe footholds. 
All in all, the receding horizon reduces variations in the desired trajectories 
which improves the trajectory tracking response.

The differences between the receding and single horizon in the pose optimization
can also be noticed in~\vref{12}. 
In the case of a single horizon, 
the robot was struggling while climbing up the stairs but 
was able to recover and accomplish the task.
However, using the receding horizon, the robot was able 
to adapt its pose in time, 
and thus resulting in safer footholds
that allowed the robot to accomplish the task.

\begin{figure}[h!]
\centering
\includegraphics[width=0.9\columnwidth]{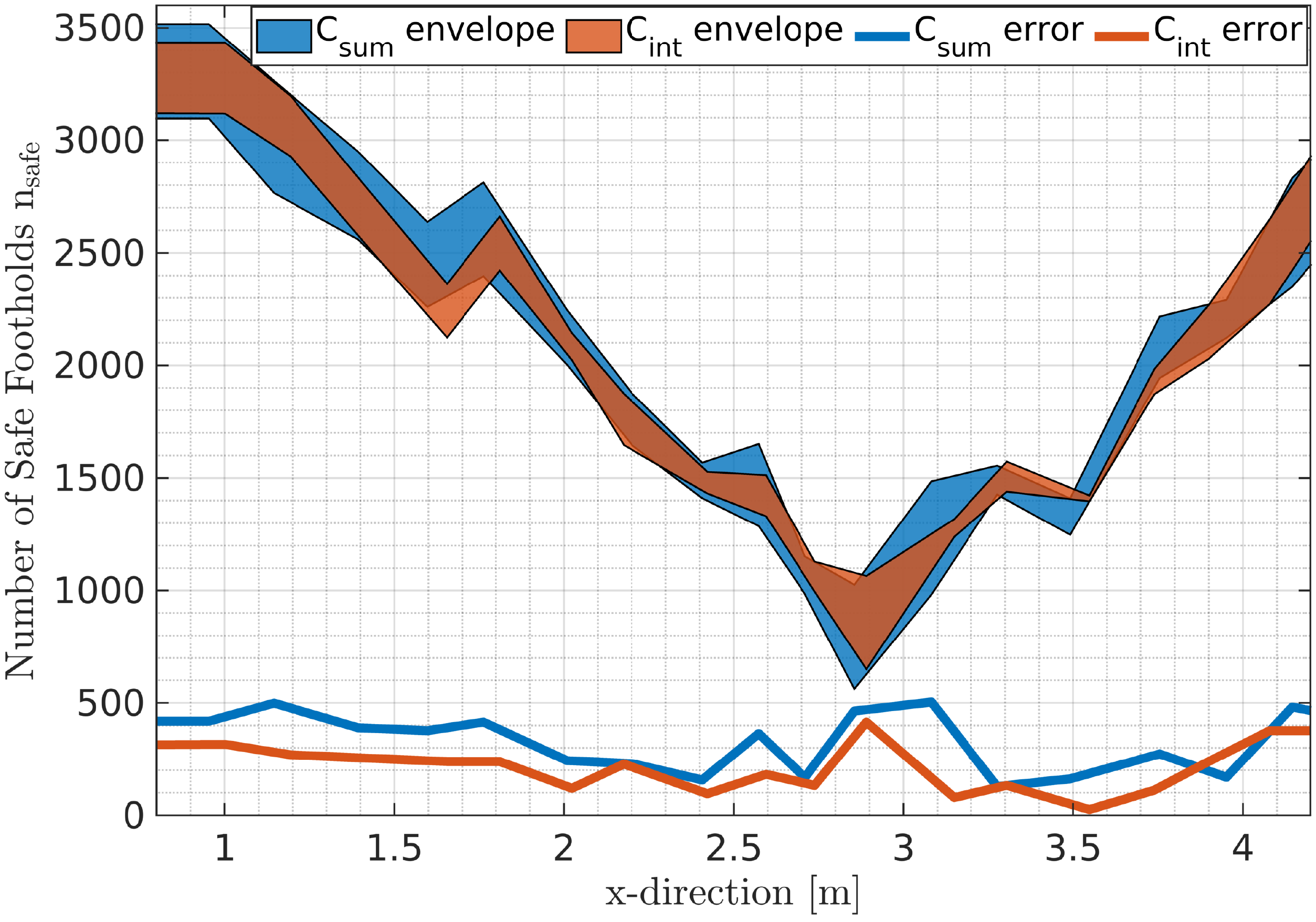}		
\caption[Pose Optimization: $\mathcal{C}_{\mathrm{sum}}$ vs. $\mathcal{C}_{\mathrm{int}}$.]
{Pose Optimization: $\mathcal{C}_{\mathrm{sum}}$ vs. $\mathcal{C}_{\mathrm{int}}$.
The shaded areas are the envelopes of the number of safe footholds. 
The lines are the thicknesses (errors) between these envelopes.}
\label{res_6_1}
\end{figure} 

\subsection{Pose Optimization: $\mathcal{C}_{\mathrm{sum}}$ vs. $\mathcal{C}_{\mathrm{int}}$}
\label{sec_sim_6}

To analyze the differences between $\mathcal{C}_{\mathrm{sum}}$ and $\mathcal{C}_{\mathrm{int}}$
in the pose optimization, 
we use the stairs setup in~\fref{fig_4} with a commanded forward velocity of~\unit[0.4]{m/s}, 
and report the outcome in~\fref{res_6_1}.
The main advantage of using $\mathcal{C}_{\mathrm{int}}$
over $\mathcal{C}_{\mathrm{sum}}$ is that $\mathcal{C}_{\mathrm{int}}$
will result in a pose that does not just maximize the number of safe footholds for all of the legs, 
but also ensures that the 
number of safe footholds of the poses around the optimal pose is still high.
To compare the two cost functions, 
we take a look at the number of safe footholds. 
In particular, we evaluate the number of safe footholds 
corresponding to the optimal pose, and the poses around it with a margin of $m=$~\unit[0.025]{m}.
Thus, in~\fref{res_6_1}, we plot the envelope (shaded area) between 
$\rbfn(u^* + m)$ and $\rbfn(u^* - m)$ for both cases, and the thickness between these envelopes 
which we refer to as error 
\begin{equation}
\textbf{error}~=\vert \rbfn(u^* + m)-\rbfn(u^* - m)\vert .
\end{equation}

As shown in~\fref{res_6_1}
the envelope of the number of safe footholds 
resulting from $\mathcal{C}_{\mathrm{int}}$
is almost always encapsulated by $\mathcal{C}_{\mathrm{sum}}$.
The thickness (error) of
the number of safe footholds resulting from using $\mathcal{C}_{\mathrm{int}}$ 
is always smaller than $\mathcal{C}_{\mathrm{sum}}$.
This means that any variation of $m$ in the optimal pose
will be less critical if  $\mathcal{C}_{\mathrm{int}}$ is used compared to 
$\mathcal{C}_{\mathrm{sum}}$.

\begin{figure}[h!]
\centering
\includegraphics[width=\columnwidth]{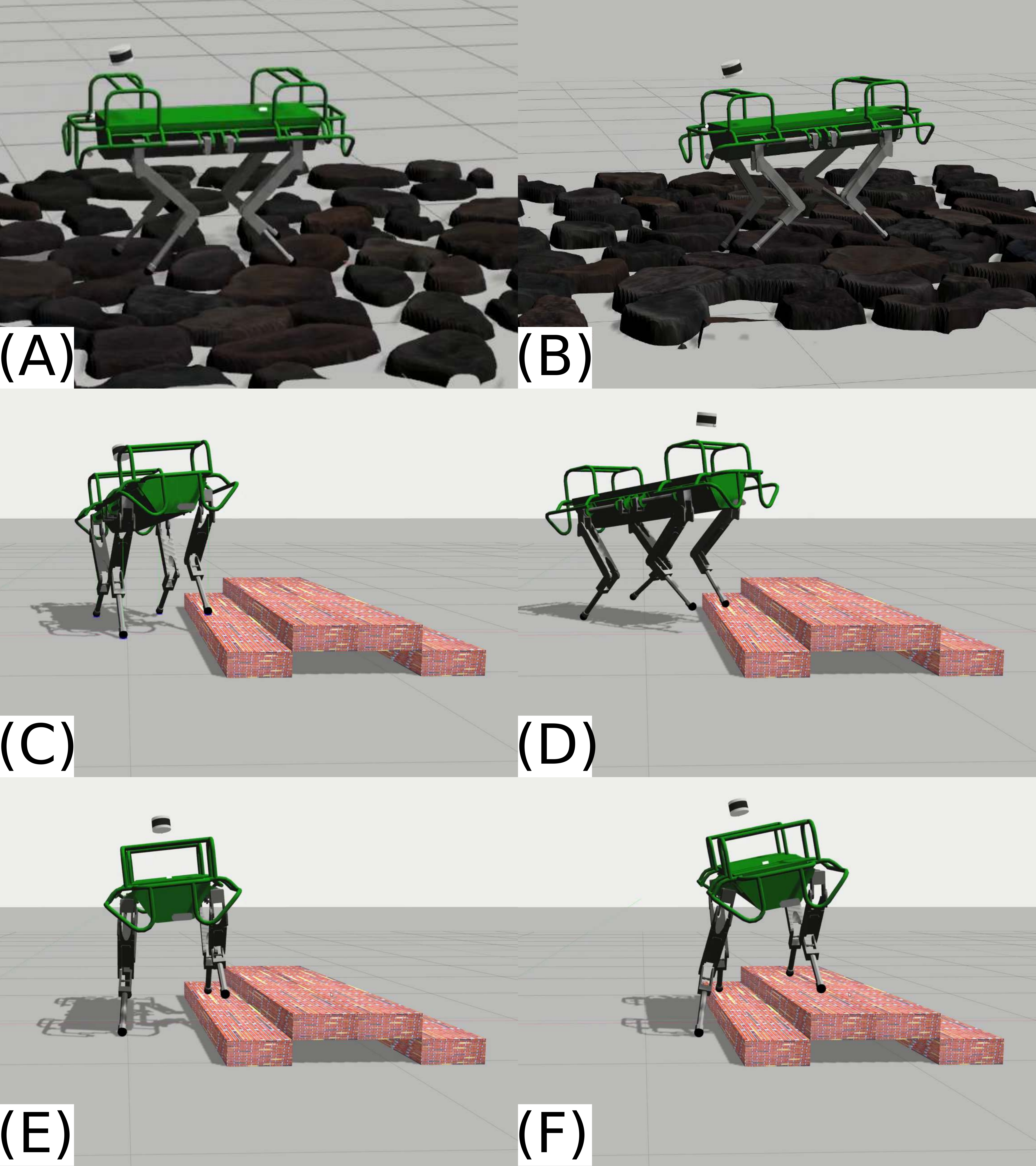}		
\caption[\acrshort{hyq} traversing rough terrain and climbing stairs sideways using~\gls{vital}.]
{\acrshort{hyq} traversing rough terrain and climbing stairs sideways.
(A,B)~\acrshort{hyq} traversing rough terrain with and without~\gls{vital}, respectively.
(C,D)~\acrshort{hyq} climbing stairs while yawing (commanding the yaw rate) using~\gls{vital}.
(E,F)~\acrshort{hyq} climbing stairs sideways using~\gls{vital}.}
\label{res_15}
\end{figure} 

\subsection{Locomotion over Rough Terrain}
We evaluate the performance of~\acrshort{hyq} in traversing 
rough terrain as shown in~\fref{res_15}(A,B) and in~\vref{13}. 
We conducted two simulations:
one with~\gls{vital} and thus with exteroceptive and proprioceptive reactions as shown in \fref{res_15}(A),
and another without~\gls{vital} and thus only with proprioceptive reactions as shown in~\fref{res_15}(B). 
\acrshort{hyq} was commanded to traverse the rough terrain with a forward velocity of~\unit[0.2]{m/s}.
No hyper parameters re-tuning, or \glspl{cnn} re-training were needed.

As shown in~\vref{13},~\acrshort{hyq} was able to successfully traverse the terrain in both cases. 
With~\gls{vital},~\acrshort{hyq} collided less with the terrain and continuously adapted its footholds over the small cobblestones. 
Without~\gls{vital},~\acrshort{hyq} traversed the rough terrain, yet, with significantly more effort. 
Additionally, without~\gls{vital},~\acrshort{hyq} continuously collided with the terrain, and in some incidents, the feet got stuck. 
For this reason, we had to re-tune the gait parameters, and increase the step height to reduce these incidents. 
The robot's feet also kept slipping since the feet were always close to edges and corners.

\subsection{Climbing Stairs with Different Commands}
Instead of commanding only forward velocities as in the previous sections, 
we command~\acrshort{hyq} to climb the stairs with~\gls{vital}
while yawing (commanding the yaw rate) as shown in~\vref{14} and~\fref{res_15}(C,D), and 
to climb stairs laterally as shown in~\vref{15} and~\fref{res_15}(E,F). 
Climbing stairs sideways is more challenging than facing the stairs 
since the range of motion of the robot's roll orientation is more restricted versus the pitch orientation. 
That said, because of~\gls{vital}, \acrshort{hyq} was still able to climb these stairs in both cases as shown in~\vref{14} and~\vref{15}.

\section{Conclusion}\label{sec_conclusion_vital}
We presented \gls{vital} which is an online vision-based locomotion planning strategy.
\gls{vital} consists of the~\gls{vpa} for pose adaptation, 
and the~\gls{vfa} for foothold selection. 
The~\gls{vpa} introduces a different paradigm 
to current state-of-the-art pose adaptation strategies. 
The~\gls{vpa} finds body poses that maximize the chances 
of the legs to succeed in reaching safe footholds. 
This notion of success emerges from the robot's skills. 
These skills are encapsulated in the \gls{fec}
that include (but are not limited to)
the terrain roughness, kinematic feasibility, leg collision, and foot trajectory collision. 
The~\gls{vfa} is a foothold selection algorithm
that continuously adapts the robot's trajectory based on the~\criteria.
The~\gls{vfa} algorithm of this work extends 
our previous work in~\cite{Esteban2020, Villarreal2019} as well as the state of the art~\cite{Fankhauser2018,Jenelten2020}. 
Since the computation of the~\criteria is usually expensive, 
we rely on approximating these criteria with~\glspl{cnn}.

The robot's skills and the notion of success provided 
by the \gls{fec} allowed the~\gls{vpa}
to generate body poses that maximize the chances of success in reaching safe footholds. 
This resulted in body poses that are aware of the terrain
and aware of what the robot and its legs can do. 
For that reason, 
the~\gls{vpa} was able to generate body poses that give a better chance
for the~\gls{vfa} to select safe footholds. 
As a result, because of~\gls{vital}, 
\acrshort{hyq} and \acrshort{hyqreal} were able to traverse
multiple terrains with various forward velocities and different gaits
without colliding or reaching workspace limits. 
The terrains included stairs, gaps, and rough terrains, 
and the commanded velocities varied from \unit[0.2]{m/s} to~\unit[0.75]{m/s}.
The~\gls{vpa} outperformed other strategies for pose adaptation. 
We compared~\gls{vpa}
with the~\gls{tbr} which is another vision based pose adaptation strategy,
and showed that indeed the~\gls{vpa} puts the robot
in a pose that provides the feet with higher number of safe footholds. 
Because of this, 
the~\gls{vpa} made our robots succeed in various scenarios where the~\gls{tbr} failed.

\section{Limitations and Future Work}
One issue that we faced during experiment was in tracking the motion of the robot, especially for~\acrshort{hyqreal}. 
We were using a~\gls{wbc} for motion tracking.  
We believe that the motion tracking and our strategy can be improved by using a~\gls{mpc} alongside the~\gls{wbc}.
Similarly, instead of using a model-based controller (\gls{mpc} or \gls{wbc}), 
we hypothesize that an~\gls{drl}-based controller can also improve the robustness and reliability of the overall 
robot behavior. 

As explained in~\sref{sec_sys_over}, 
one other key limitation was regarding the perception system. 
State estimation introduced a significant drift that caused a major noise and drift in the terrain map. 
Albeit not being a limitation to the suggested approach, 
we plan on improving the state estimation and perception system of~\acrshort{hyq} and~\acrshort{hyqreal}
to allow us to test~\gls{vital} in the wild.

The pose optimization problem of the~\gls{vpa} does not reason about the robot's dynamics. 
This did not prevent~\acrshort{hyq} and~\acrshort{hyqreal}
from achieving dynamic locomotion while traversing challenging terrains at high speeds. 
However, we believe that 
incorporating the robot's dynamics into~\gls{vital} may result in a better overall performance. 
That said, we believe that in the future, the~\gls{vpa} should also reason about the robot's dynamics. 
For instance, 
one can augment the \criteria with another criterion that ensures that the selected footholds are dynamically feasible by the robot.

Additionally, in the future,
we plan to extend the~\gls{vpa} of~\gls{vital} to not only send pose references, 
but also reason about the robot's body twist. 
We also plan to augment the robot skills to not only consider foothold evaluation criteria, 
but also skills that are tailored to the robot pose. 
Finally, in this work, \gls{vital} considered heightmaps which are 2.5D maps. 
In the future, we plan to consider full 3D maps that will enable~\gls{vital} 
to reason about navigating in confined space (inspired by \cite{Buchanan2020b}).

\chapter{ViTAL - Implementation Details}\label{chap_6}
\vitalPaper
\section{CNN approximation in the VFA and the VPA}\label{cnn_approx}
In the~\gls{vfa}, the foothold evaluation stage is approximated with a \acrshort{cnn}~\cite{Lecun1989} as explained in~\remref{remark_cnn_exact}.
The \acrshort{cnn} approximates the mapping between~$\heurtuple$ and~$\optimal$.
The heightmap~$\hmapvfa$~in~$\heurtuple$ passes through three convolutional layers 
with~$5 \times 5$ kernels, $2 \times 2$ padding, Leaky ReLU activation~\cite{Maas2013}, and $2 \times 2$ max-pooling operation.
The resulted one-dimensional feature vector is concatenated with 
the rest of the variables in the tuple~$\heurtuple$,
namely, $\hheight, \bodyvel, \gaitparams,$ and $\nominal$. 
This new vector passes through two fully-connected layers with Leaky ReLU and softmax activations.
The parameters of the \acrshort{cnn} are optimized to minimize the cross-entropy loss~\cite{Bishop2006} of classifying a candidate foothold location as optimal~$\optimal$.

In the~\gls{vpa}, the pose evaluation and the function approximation is approximated with a \acrshort{cnn} as explained in~\remref{remark_cnn_exact_vpa}.
The \acrshort{cnn} infers the weights~$w$ of~$\rbfn$ given~$\heurtuplevpa$ (the mapping between~$\heurtuplevpa$ and~$w$).
The heightmap $H_{\mathrm{vpa}}\in\Rnum^{33\times33}$ passes through three convolutional layers 
with $5 \times 5$ kernels, $2 \times 2$ padding, Leaky ReLU activation, and $2 \times 2$ max-pooling operation.
The body velocities~$\bodyvel$ pass through a fully-connected layer with Leaky ReLU activation 
that is then concatenated with the one-dimensional feature vector obtained from the heightmap.
This new vector passes through two fully-connected layers with Leaky ReLU and linear activations.
The parameters of this \acrshort{cnn} are optimized 
to minimize the mean squared error loss
between the number of safe footholds~\acrshort{nsf} predicted by $\rbfn(z_{h},w)$ 
and $\rbfn(z_{h},\hat{w})$ where $\hat{w}$ 
are the function parameters approximated by the~\acrshort{cnn}.

For both \glspl{cnn}, we used the Adam optimizer~\cite{Kingma2015} with a learning rate of $0.001$,
and  we used a validation-based early-stopping using a $9$-to-$1$~proportion to reduce overfitting.
The datasets required for training this \glspl{cnn} are collected 
by running simulated terrain scenarios that consist of bars, gaps, stairs, and rocks. 
In this work, we considered a $33\times33$ heightmap with a resolution of \unit[0.02]{m} 
($H_{\mathrm{vfa}}, H_{\mathrm{vpa}}\in\Rnum^{33\times33}$).

\begin{figure}[t!]
\centering
\includegraphics[width=\columnwidth]{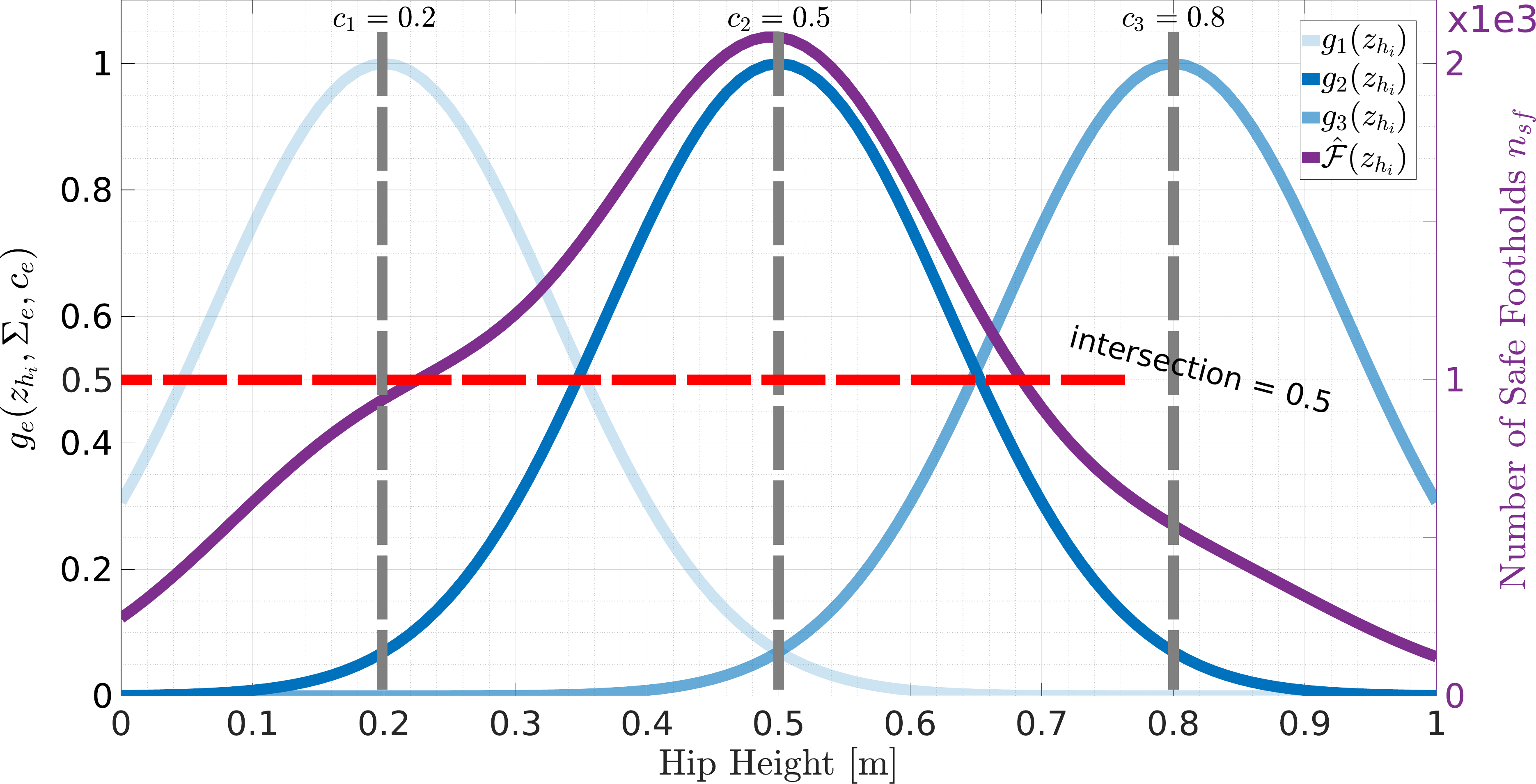}		
\caption{An illustration of the Function Approximation of the \acrshort{vpa}.}
\label{fig_func_approx}
\end{figure} 

\section{On the Function Approximation of the \acrshort{vpa}}\label{app_func_approx}
As explained in~\sref{sec_func_approx}, the function~$\rbfn(z_{h_i}, w)$ 
\begin{equation}
\rbfn(z_{h_i}, w) = \sum_{e=1}^{E} w_e \cdot g(z_{h_i}, \Sigma_e, c_e)
\end{equation}
is defined as the weighted sum of Gaussian basis functions
\begin{equation}
g(z_{h_i}, \Sigma_e, c_e) = \text{exp}(-0.5 (z_{h_i}-c_e)^T \Sigma_e^{-1} (z_{h_i} -c_e)).
\end{equation}
The parameters $\Sigma_e$ and $c_e$ are the widths and centers of the Gaussian function~$g_e$~(see Section~3.1 in~\cite{Stulp2015}).
In the literature,~$c_e$ 
is usually referred to as the mean or the expected value, and $\Sigma_e$ as the standard deviation.
The regression algorithm should predict the weights~$w_e$, and the parameters~$\Sigma_e$ and~$c_e$.
To reduce the dimensionality of the problem, as explained in~\sref{sec_func_approx}, and in~Section~4.1 in~\cite{Stulp2015},
we decided to fix the values of the parameters of the Gaussian functions~$\Sigma_e$ and~$c_e$.
In detail, 
the centers~$c_e$ are spaced equidistantly within the bounds of the hip heights~$\elementheights$,
and the widths are determined by the value at which the Gaussian functions intersect.
That way, the regression algorithm only outputs the weights~$w_e$.
Figure~\ref{fig_func_approx} shows an example of the function approximation. 
In this example, the bounds of the  hip heights~$\elementheights$ are \unit[0.2]{m} and \unit[0.8]{m}.
Assuming a number of basis functions~$E=3$, the centers~$c_e$ are then chosen to be equidistant within the bounds, 
and thus, the centers~$c_e$ are \unit[0.2]{m}, \unit[0.3]{m} and \unit[0.8]{m}. 
By choosing the Gaussian functions to intersect at 0.5, the widths~$\Sigma_e$ are~0.13.

\section{Representing the Hip Heights in terms of the Body Pose}\label{po_appendix}
To represent the hip heights in terms of the body pose, 
we first write the forward kinematics of the robot's hips
\begin{equation}p_{h_i}^W = p_{b}^W ~+~ R_b^W p_{h_i}^b\label{hh_respresentation_1}\end{equation}
where $p_{h_i}^W\in\Rnum^3$~is the position of the hip of the $i$th leg in the world frame, 
$p_{b}^W\in\Rnum^3$~is the position of the robot's base in the world frame,
$R_b^W\in SO(3)$~is the rotation matrix mapping vectors from the base frame to the world frame,
and $p_{h_i}^b\in\Rnum^3$~is the position of the hip of the $i$th leg in the base frame.
The rotation matrix $R_b^W$ is a representation of the Euler angles of the robot's base
with sequence of roll~$\beta$, pitch~$\gamma$, and yaw~$\psi$ (Cardan angles)~\cite{Diebel2006}.
The variable $p_{h_i}^b$ is obtained from the CAD of the robot. 
Expanding~\eref{hh_respresentation_1} yields
\begin{eqnarray}
\begin{bmatrix} x_{h_i}^W \\ y_{h_i}^W \\ z_{h_i}^W \end{bmatrix}
&\hspace{-5pt}=\hspace{-5pt}& 
\begin{bmatrix} x_{b}^W \\ y_{b}^W \\ z_{b}^W \end{bmatrix} + R_b^w (\beta, \gamma, \psi) \begin{bmatrix} x_{h_i}^b \\ y_{h_i}^b \\ z_{h_i}^b \end{bmatrix}\\
&\hspace{-5pt}=\hspace{-5pt}& 
\begin{bmatrix} x_{b}^W \\ y_{b}^W \\ z_{b}^W \end{bmatrix} 
+ \begin{bmatrix} \cdots  & \cdots & \cdots \\ \cdots  & \cdots & \cdots \\ -s\gamma & c\gamma~s\beta &  c\gamma~c\beta\end{bmatrix}
\begin{bmatrix} x_{h_i}^b \\ y_{h_i}^b \\ z_{h_i}^b \end{bmatrix} 
\label{hh_respresentation_2}
\end{eqnarray}
where $s$ and $c$ are sine and cosine of the angles, respectively. 
Since we are interested only in the hip heights, 
the z-component (third row) of~\eref{hh_respresentation_2} yields
\begin{equation}
z_{h_i}^W = z_{b}^W - x_{h_i}^b s\gamma + y_{h_i}^b c\gamma s\beta + z_{h_i}^b c\gamma c\beta.
\end{equation}

\section{Using the Sum of Squared Integrals in the Pose Optimization of the~\gls{vpa}}\label{cint_appendix}
\begin{figure}[t!]
\centering
\includegraphics[width=\columnwidth]{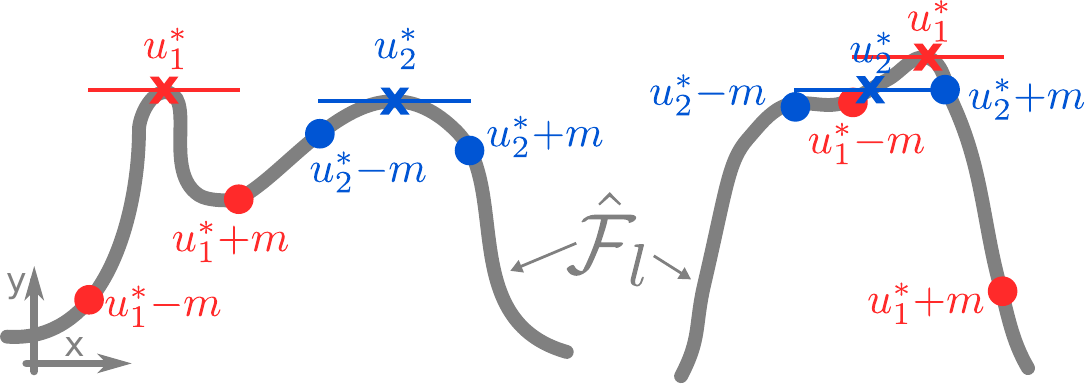}		
\caption[Using the sum of squared integrals as a cost function in the pose optimization of the~\gls{vpa}]
{Using the sum of squared integrals as a cost function in the pose optimization of the~\gls{vpa}.
The two curves represent~$\rbfn_l$. The x-axis is the hip height and the y-axis is~\acrshort{nsf}.
The figure shows two optimal poses: $u_1^*$~which is from 
using~$\mathcal{C}_{\mathrm{sum}}$ or~$\mathcal{C}_{\mathrm{prod}}$,
and $u_2^*$~which is from using~$\mathcal{C}_{\mathrm{int}}$.}
\label{fig_po}
\end{figure}
Using~$\mathcal{C}_{\mathrm{int}}$ as a cost function 
can be motivated by taking~\fref{fig_po} as an example. 
In this figure, 
there are two curves that represent~$\rbfn_l$
where the horizontal axis is the hip height and the vertical axis represent~\acrshort{nsf}.
The figure shows two optimal poses where $u_1^*$~is the optimal pose
using the cost functions~$\mathcal{C}_{\mathrm{sum}}$  or $\mathcal{C}_{\mathrm{prod}}$ that only maximize for~$\rbfn_l$,
and $u_2^*$~is the optimal pose using the cost function~$\mathcal{C}_{\mathrm{int}}$.
As shown in the figure,
if $\mathcal{C}_{\mathrm{sum}}$ or $\mathcal{C}_{\mathrm{prod}}$
is used, the optimal pose will be $u_1^*$
which is indeed the one that results in the maximum~$\rbfn_l$.
However, if there is a tracking error of $m$ 
(thus the robot reaches $u_1^* \pm m$), 
the robot might end up in the pose $u_1^* \pm m$
that results in a small number of safe footholds.
Using 
$\mathcal{C}_{\mathrm{int}}$ will take into account the safe footholds within a margin $m$.
This might result in a pose that does not yield the maximum number of safe footholds, 
but it will result in a safer foothold in case the robot pose has any tracking errors. 
Note that using
the sum of squared integrals~$\mathcal{C}_{\mathrm{int}}$
as a cost function
is similar to smoothing~$\rbfn_l$ with respect to the hip height 
(a moving average smoothing).
Using a smoothing function~$\mathcal{C}_{s}$ may take the form
\begin{equation}
\mathcal{C}_{s} = \sum_{l=1}^{N_l=4} \Vert  
\frac{1}{2\epsilon}  
\sum_{i=-\epsilon}^{\epsilon}  \rbfn_l (z_{h_l}+i) \
\Vert^2_Q.
\end{equation}

\section{Defining the Receding Horizon}\label{receding_def}
In the receding horizon there is a tuple~$\heurtuplevpaR$ for every $j$th horizon
that is defined~as
\begin{equation}
\heurtuplevpaR = (\hmapvpaR,\bodyvel, \gaitparams)
\label{vpa_tuble_receding}
\end{equation}
hence sharing the same body twist~$\bodyvel$ and gait parameters~$\gaitparams$ but a different heightmap~$\hmapvpaR$.
For every leg, a heightmap of horizon~$j+1$ is overlapping with  the previous horizon's~$j$ heightmap.
This overlap has a magnitude of~$\Delta h$ taking the same direction as the body velocity~$\dot{x}_b$.
Without loss of generality, we chose the magnitude of the overlap to be half of the diagonal size of the heightmap in this work. 
To sum up, we first gather~$\heurtuplevpaR$ that share the same~$\bodyvel$ and~$\gaitparams$, but a different~$\hmapvpaR$.
Then, we evaluate~$\heurtuplevpaR$ and approximate the output using the function approximation 
yielding~$\rbfn_{j}$ that is sent to the optimizer for all of the legs.

\section{Miscellaneous Settings}\label{misc_details}
In this work, all simulations were conducted on an Intel Core i7 quad-core CPU, 
and all experiments were running on an onboard Intel Core i7 quad-core CPU where state estimation, mapping, and controls were running. 
The~\gls{rcf} (including the~\gls{wbc}) runs at \unit[250]{Hz}, 
the low-level controller runs at \unit[1000]{Hz},
the state estimator runs at \unit[333]{Hz}, 
and the mapping algorithm runs at \unit[20]{Hz}.
The~\gls{vpa} and the~\gls{vfa} run asynchronously at the
maximum possible update rate.

\acrshort{vital} is implemented in Python. 
The~\acrshortpl{cnn} are implemented in PyTorch~\cite{Adam2019}.
As explained in~\appref{cnn_approx}, in this work, we considered a $33\times33$ heightmap with a resolution of \unit[0.02]{m} 
($H_{\mathrm{vfa}}, H_{\mathrm{vpa}}\in\Rnum^{33\times33}$).
The finite set~$\finitesetheights$
consisted of a hip height range between~\unit[0.2]{m} and~\unit[0.8]{m}
with a resolution of~\unit[0.02]{m} yielding~$\hipheightsamples=31$ samples. 
The number of radial basis functions used in the function approximation was $E=30$.
The pose optimization problem is 
solved with a 
trust-region interior point method~\cite{Wright1996,Byrd1999} which is a non-linear optimization problem solver that we solved using
SciPy~\cite{Virtanen2020}.
The bounds of the pose optimization problem~$u_{\min}$
and~$u_{\max}$ are~$[\unit[0.2]{m},\unit[-0.35]{rad},\unit[-0.35]{rad}]$, and~$[\unit[0.8]{m},\unit[0.35]{rad},\unit[0.35]{rad}]$, respectively.
We used a receding horizon of~$N_h=2$ with a map overlap of half the size of the heightmap.
For a heightmap of a size of~$33\times33$ and a resolution of~\unit[0.02]{m}, the map overlap~$\Delta h$ is~\unit[0.33]{m}.
We used Gazebo~\cite{Koenig2004} for the simulations, and ROS for communication.

\section{Estimation Accuracy}\label{est_acc}
We compare the estimation accuracy of the~\gls{vfa}
by comparing the output of the foothold evaluation stage (explained in \sref{vfaa}) given the same input tuple $\heurtuple$.
That is to say, we compare the estimation accuracy of the~\gls{vfa} by comparing $\ghat$ versus $\g$ (see \remref{remark_cnn_exact}).
To do so, once trained, we generated a dataset of \unit[4401]{samples} from randomly sampled heightmaps for every leg. 
This analysis was done on~\acrshort{hyq}.

As explained in~\sref{vfaa}, 
from all of the safe candidate footholds in~$\fecout$, 
the~\gls{vfa} chooses the optimal foothold to be the one closest to the nominal foothold. 
Thus, to fairly analyse the estimation accuracy of the~\gls{vfa}, we present three main measures: 
\textit{perfect match} being the amount of samples where $\ghat$ outputted the exact value of $\g$,
\textit{safe footholds}, being the amount of samples where  $\ghat$ did not output the exact value of $\g$, but rather a foothold that is safe but not closest to the nominal foothold, 
and \textit{mean distance}, being the average distance of the estimated optimal foothold from $\ghat$ relative to the exact foothold from $\g$. 
These measures are presented as the mean of all legs. 

Based on that, 
the perfect match measure is~$74.0\%$. 
Thus, $74\%$ of $\ghat$ perfectly matched $\g$. 
The safe footholds measure is~$93.7\%$. 
Thus, $93.7\%$ of $\ghat$ were deemed safe.
Finally, the mean distance of the estimated optimal foothold from $\ghat$ relative to the exact foothold from $\g$ is~\unit[0.02]{m}.
This means that, on average, $\ghat$ yielded optimal footholds that are~\unit[0.02]{m} far from the optimal foothold from $\g$.
Note that, the radius of~\acrshort{hyq}'s foot, and the resolution of the heightmap is also~\unit[0.02]{m}, 
which means that the average distance measure is still acceptable especially since we account for this value in the uncertainty margin as explained 
in \remref{remark_uncertainty_margin}.

Similar to the~\gls{vfa}, we compare the accuracy of~\gls{vpa}
by comparing  the output of the pose evaluation stage (explained in \sref{sec_vpa_pipeline}) given the same input tuple $\heurtuple$.
That is to say, we compare the estimation accuracy of the~\gls{vpa} by comparing $\finitesetfeasibles$ versus $\rbfn$ 
(see \remref{rem3} and \remref{remark_cnn_exact_vpa}).
To do so, we ran one simulation using the stairs setup shown in~\fref{fig_4} on~\acrshort{hyq}, and gathered the input tuple $\heurtuple$. 
Then, we ran the~\gls{vpa} offline, once with the exact evaluation (yielding~$\finitesetfeasibles$) 
and once with the approximate one (yielding~$\rbfn$).

Based on this simulation run, 
the mean values of the exact and the approximate evaluations are 
$\mathrm{mean}(\finitesetfeasibles) = 1370$ and
$\mathrm{mean}(\rbfn) = 1322$, respectively. 
This yields an estimation accuracy 
$\mathrm{mean}(\rbfn)/\mathrm{mean}(\finitesetfeasibles)$ 
of~$96.5\%$.

\section{Computational Analysis}\label{comp_anal}
To analyze the computational time of the~\gls{vfa} and the~\gls{vpa}, 
we ran one simulation using the stairs setup shown in~\fref{fig_4} on~\acrshort{hyq} 
to gather the input tuples of the~\gls{vfa} and the~\gls{vpa}, $\heurtuple$ and $\heurtuplevpa$, respectively. 
Then, we ran both algorithms offline, once with the exact evaluation and once using the~\glspl{cnn}, 
and collected the time it took to run both algorithms (all stages included).
The mean and standard deviation of the time taken to compute the exact and the~\gls{cnn}-approximated \gls{vfa} (per leg) algorithms are
$\unit[7.5]{ms} \pm \unit[1]{ms}$, and
$\unit[3.5]{ms} \pm \unit[1]{ms}$, respectively. 
The mean and standard deviation of the time taken to compute the exact and the~\gls{cnn}-approximated \gls{vpa} algorithms are
$\unit[720]{ms} \pm  \unit[68]{ms}$, and
$\unit[180]{ms} \pm \unit[60]{ms}$, respectively. 
Hence, the~\gls{vfa} and the~\gls{vpa} can run at roughly~\unit[280]{Hz} and~\unit[5]{Hz}, respectively.
This also shows that the~\glspl{cnn} can speed up the evaluation of the \gls{vfa} and the \gls{vpa} 
up to 4 times and 2 times, respectively.

Note that it takes longer to compute the~\gls{vfa} of this work versus our previous work~\cite{Villarreal2019}.
This is because the~\gls{vfa} of this work considers more inputs than in our previous work, 
and thus, the size of the~\gls{cnn} is larger. 
As can be seen, the~\gls{vpa} runs at a relatively lower update rate compared to the~\gls{vfa}.
We believe that this is not an issue since 
the~\gls{vfa} runs at the legs-level while the~\gls{vpa} runs as the body-level
which means that the legs experience faster dynamics than the body.

During simulations and experiments, 
the~\glspl{cnn} were running on a CPU.
A significant amount of computational time can be reduced if we run the~\glspl{cnn} of the~\gls{vfa} and the~\gls{vpa} on a GPU.
Likewise, a significant amount of computational time can be reduced if a different pose optimization solver is used. 
However, both suggestions are beyond the scope of this work, and are left as a future work.

\clearpage
\newpage

\glsresetall \chapter{Conclusion}\label{chap_conc}
\section{Summary}
\gls{tal} is an essential element to achieve \gls{ati} for legged robots. 
For that to happen, legged robots should be able 
to perceive, understand, and adapt to their surrounding terrain
using  their proprioceptive and exteroceptive (visual) information. 
This thesis presented \gls{tal} strategies, 
both at the proprioceptive and vision-based level. 
The first part
(Chapters~\ref{chap_pwbc}-\ref{chap_lsens}) was on \gls{ptal} strategies and 
the second part (\chapref{chap_vital}) was on \gls{etal} strategies. 

In \chapref{chap_pwbc}, 
we presented a \gls{ptal} strategy that made legged robots adapt 
to the terrain inclination and frictional properties. 
We presented a \gls{pwbc} framework for quadruped robots
where the locomotion control problem was casted as  
a \gls{qp} that took into account
the full robot rigid body dynamics, the actuation limits, the joint kinematic limits 
and the contact interaction.
The contact interaction included
the terrain's inclination (normals), 
frictional and unilaterality properties, and the rigid contact interaction,
and were encoded in the~\gls{qp} formulation.
We encoded the terrain inclination, frictional properties, 
and the rigid contact interaction 
in the \gls{qp} formulation. 
As a result, 
the quadruped robot was able to reliably traverse various challenging terrains
with different friction coefficients, and under different gaits. 

In \chapref{chap_stance}, 
we presented a \gls{ptal} strategy that made legged robots adapt to soft terrain. 
We introduced the \gls{stance} approach 
that extended capabilities of  the previously presented \gls{pwbc} in \chapref{chap_pwbc}.
\gls{stance} consisted of a \gls{awbc} and a \gls{ste}.
The \gls{ste} provided the \gls{awbc} with the current terrain impedance
that the \gls{awbc} then used to adapt the robot's motion accordingly. 
As a result, 
the quadruped robot was able to 
%
differentiate between compliances under each foot, 
and to adapt online 
to multiple terrains with different compliances (rigid and soft) without pre-tuning.

In \chapref{chap_lsens}, 
we looked into one of the remaining limitations of locomotion over soft terrain. 
We were able to 
investigate how and why does 
soft terrain affect state estimation for legged robots. 
As a result, we 
showed that soft terrain negatively affects state estimation for legged robots, 
and that the state estimates have a noticeable drift over soft terrain compared to rigid terrain.

In \chapref{chap_vital}, 
we presented a \gls{vital} strategy
consisting of a 
\gls{vpa} algorithm that 
introduced a paradigm shift for pose adaptation strategies,
and a \gls{vfa} algorithm that 
extended state-of-the-art foothold selection strategies. 
Instead of the commonly used pose adaptation techniques that 
optimizes body poses given the selected footholds, 
we proposed to find body poses 
that maximizes the chances of reaching safe footholds.
This was done by relying on a set of robots skills
that represented the capabilities of the robot and its legs.
The skills 
were then learned via self-supervised learning using \glspl{cnn}. 
\gls{vital} allowed our robots to 
select the footholds based on their capabilities, 
and simultaneously find poses that maximize the chances
of reaching safe footholds. 
As a result, 
our quadruped robots were able to
traverse stairs, gaps, and various other terrains at different speeds. 

We believe that the contributions of this thesis 
allows legged robots
to traverse a wider range of terrains
with different geometries and physical properties. 
We believe that this would not have been possible 
without exploiting the robot's 
proprioceptive and exteroceptive (visual) information.

\section{Future Directions}
The contributions of this thesis 
allowed us to advance in many research directions,
yet, there remains many further improvements 
in these research directions. 
Below, we provide 
pointers to where we believe these improvements should be going. 

\newpage
\subsection*{Whole-Body Control}
\begin{itemize}
\item 
\gls{wbc} is still an active topic in research. 
Perhaps one promising direction is the use of control Lyapunov function based \gls{qp}.
This approach has shown a rapid convergence compared to the standard \gls{wbc} 
formulations~\cite{Reher2020}. 
Yet, further validation of the approach should be done 
to show its capabilities in more challenging terrains and different gaits. 

\item 
\gls{wbc}s are model-based optimization approaches 
that rely on inverse dynamics. 
Hence, one possible direction is to improve the robot dynamics model 
using model learning techniques. 
The hybrid nature of floating based systems 
makes it harder to learn its inverse dynamics. 
One approach is to learn the errors between the dynamics model and the actual dynamics 
(the residual in the dynamics) of the legged robot \cite{Grandia2018}. 
%

\item 
There are other challenges in \gls{wbc} in specific scenarios. 
One challenge is when the robot loses contact or if the robot 
gets lifted up. 
The current \gls{wbc} formulations cannot handle this case. 
Thus, some implementations still rely on a joint-level PD loop
albeit not being used in our work. 
Perhaps one way to improve this is to
add constraints regarding the stance feet. 
These constraints would keep the stance feet close to the robot base, 
thus, if the robot gets lifted, the feet do not move away from the base. 
\end{itemize}

\subsection*{Soft Terrain Adaptation}
\begin{itemize}
\item As mentioned in \chapref{chap_stance}, 
there are various research directions in locomotion over soft terrain
mainly in state estimation and low-level control.
The former was tackled in \chapref{chap_lsens} while the latter 
is not yet tackled. 
With this in mind, we believe
there is  great potential improvements of low-level control 
to generalize beyond rigid terrain. 
As a first step, 
perhaps there should be a formal discussion on the effects of soft terrain 
on the low-level control, 
and possible ideas on how to improve the performance of the low-level control
to adapt to terrains with different impedances. 

\item 
The \gls{awbc} presented in \chapref{chap_stance} was superior to 
the \gls{swbc}. 
As we mentioned in~\chapref{chap_stance}, 
the differences between these two controllers were more evident under 
dynamic motions. 
Perhaps other \glspl{wbc} 
could perform better than the \gls{swbc} and the \gls{awbc} under slower motions. 
Hence, it might be of great potential to compare the \gls{awbc}
with the controllers mentioned in \cite{Flayols2020}.

\item 
The \gls{ste} presented in \chapref{chap_stance} 
relied on the \gls{kv} model
which is linear springs and dampers parallel and perpendicular to the contact point. 
As we reported in \chapref{chap_stance}, we chose this model because 
it was simple, and it is computationally inexpensive. 
Based on that, we believe there is a lot of work to do in that aspect. 
First, it is perhaps important to analyze the trade-off between 
using a more complicated model (non-linear) that is slower to estimate its parameters
versus a simpler model that is faster to compute. 


\item \gls{stance} 
is implemented at the \gls{wbc}  level
which resulted in the controller being \gls{c3}.
One possible extension is to exploit the work of \gls{stance}
to also make \gls{mpc} \gls{c3}. 
For that, the \gls{mpc} optimization problem 
should be re-formulated to take into account the terrain impedance parameters. 

\end{itemize}

\subsection*{State Estimation}
Although this thesis was not fundamentally
focusing on state estimation, 
we believe that state estimation is crucial in \gls{tal}.
\gls{tal} mandates an accurate estimate of 
the robot states as well as the map of the environment (terrain).

As we mentioned in \chapref{chap_lsens}, 
the performance of state estimation degrades on soft terrain.
This is because state estimators still rely on rigid body assumptions
mainly in leg odometry. 
One way to deal with that is to have a
velocity bias in leg odometry. 
This velocity bias should be adaptive and not constant since
it depends on the type of terrain and on the gait used. 
Perhaps another way to deal with this
is to reformulate the leg odometry module
and augment it with the terrain impedance knowledge.

\subsection*{Vision-based Terrain Aware Locomotion}
We believe that our work on \gls{vital} and specifically the \gls{vpa}
algorithm introduced a different way of thinking and of approaching 
pose adaptation problems. 
This opened more research questions and a lot of possible improvements. 
\gls{vital}'s core idea is to teach the robot a set of skills. 
These skills were encapsulated by the~\criteria.  
Perhaps the most important future work in that aspect
is to augment the robot with skills that are tailored to the robot's pose
and not just the legs. 
For example, the skills can include the robot's ability to traverse 
confined spaces (inspired from~\cite{Buchanan2020b}). 
Furthermore, in~\chapref{chap_vital}, we reported 
that we faced some issues during experiment. 
These issues were mainly in tracking the motion of the robot especially for~\acrshort{hyqreal}. 
Thus, a possible way to improve this is to use the~\gls{mpc} controller in~\cite{Villarreal2020}.

\clearpage~\newpage
\newpage
\thispagestyle{plain}
\addcontentsline{toc}{chapter}{Bibliography}
\bibliographystyle{./includes/IEEEtran}
\bibliography{./includes/references.bib}

\chapter*{Curriculum Vitae}
\addcontentsline{toc}{chapter}{Curriculum Vitae}
\begin{wrapfigure}{r}{0.4\textwidth}
\begin{center}
\includegraphics[width=0.34\textwidth]{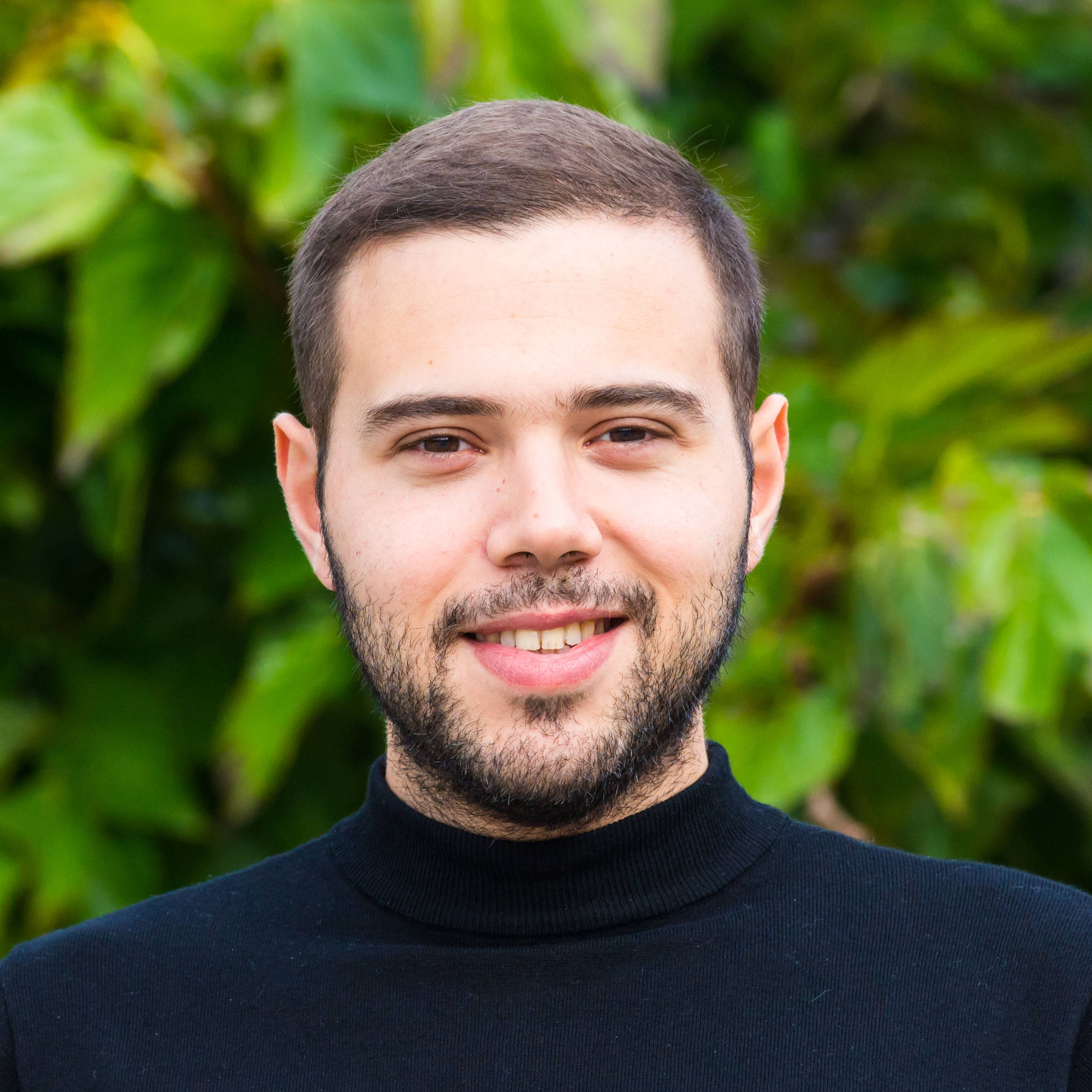}
Shamel Fahmi \\ \url{www.shamelfahmi.com}
\end{center}
\end{wrapfigure}
Shamel Fahmi (S'19) was born in 1993 in Cairo, Egypt.
He received the B.Sc. degree in mechatronics from 
the German University in Cairo (GUC), Cairo, Egypt, in 2015, 
and did his bachelor thesis at the
Institute of Automatic Control (IRT) at RWTH Aachen University, Aachen, Germany. 
During his bachelor, he was a researcher at the 
Institute of Automotive Engineering (IKA) 
at RWTH Aachen University, Aachen, Germany. 
He received
the M.Sc. degree in systems and control from 
the University of Twente (UT), Enschede, the Netherlands, in 2017. 
During his masters, he was a researcher at the 
Robotics and Mechatronics Center (RMC) at the German Aerospace Center (DLR), 
Oberpfaffenhofen, Germany.
From December 2017 to April 2021, he joined the Dynamic Legged Systems (DLS) Lab
at the Italian Institute of Technology (IIT), Genoa, 
Italy for the Ph.D. degree in advanced and humanoid robotics.
His research interests include robotics, controls, optimization, and learning for dynamical systems.
\end{document}